# DetectRL-X: Towards Reliable Multilingual and Real-World LLM-Generated Text Detection

**Junchao Wu**[1,*] **Yefeng Liu**[2] **Chenyu Zhu**[2] **Hao Zhang**[2] **Zeyu Wu**[1] **Tianqi Shi**[2]
**Yichao Du**[2] **Longyue Wang**[2] **Weihua Luo**[2] **Jinsong Su**[3] **Derek F. Wong**[1,†]
[1]NLP[2]CT Lab, Department of Computer and Information Science, University of Macau
[2]Alibaba Group [3]Xiamen University

## Abstract

The effective detection and governance of Large Language Model (LLM) generated content has become increasingly critical due to the growing risk of misuse. Despite the impressive performance of existing detectors, their reliability and potential in multilingual, real-world scenarios remain largely underexplored. In this study, we introduce DetectRL-X, a comprehensive multilingual benchmark designed to evaluate advanced detectors across 8 dimensions. The benchmark encompasses 8 languages commonly used in commercial contexts and collects human-written texts from 6 domains highly susceptible to LLM misuse. To better aligned with real-world applications, We create LLM-generated texts using 4 popular commercial LLMs, and include typical AI-assisted writing operations such as polishing, expanding, and condensing to capture authentic usage patterns. Furthermore, we develop a multilingual framework for paraphrasing and perturbation attacks to simulate diverse human modifications and writing noise, enabling stress testing of detectors across languages. Experimental results on DetectRL-X reveal the strengths and limitations of current state-of-the-art detectors when applied to diverse linguistic resources. We further analyze how domains, generators, attack strategies, text length, and refinement operations influence performance in different languages, underscoring DetectRL-X as an effective benchmark for strengthening multilingual and language-specific detectors.[1]

## 1 Introduction

The rapid growth of LLMs has intensified concerns about the misuse of LLM-generated texts (LGT) (Wu et al., 2025a). While hallucinations may introduce false or misleading information (Zhang et al., 2023; Yang et al., 2026), deliberate abuse can facilitate public opinion manipulation (Zhang et al., 2025), plagiarism, and unfairness in education (Lee et al., 2023), underscoring the importance of reliable detection methods. Existing detection benchmarks generally evaluate detectors across domains and generators by constructing idealized test data (e.g., TuringBench (Uchendu et al., 2021)). Despite efforts to mirror real-world settings, most still focus on English (e.g., DetectRL (Wu et al., 2024) and RAID (Dugan et al., 2024)) and overlook the impact of text refinement operations on detection. While some benchmarks include multilingual data, their domain distributions and evaluation settings often remain unbalanced and narrow (e.g., HC3 (Guo et al., 2023), MULTITuDE (Macko et al., 2023), M4 (Wang et al., 2024b)), making it hard to fairly evaluate detector robustness across languages and real-world scenarios, as summarized in Table 1.

To bridge these gaps, we propose DetectRL-X, the most large-scale and challenging multilingual benchmark for LGT detection. It contains 3.46 million samples covering 8 languages, 6 domains, 4 generators, 8 attack strategies, 4 text-length granularities, and 3 types of refinement operations, with 8 evaluation dimensions comparing 12 detectors. We collect human-written texts (HWT) from domains with extensive LLM adoption, including Academic, News, Novel, Wiki, SEO, and Webtext. The corresponding LGTs are produced using Deepseek-V3 (DeepSeek-AI et al., 2024), Gemini-2.5-flash (Comanici et al., 2025), GPT-4o (OpenAI, 2023), and Qwen-Max (Yang et al., 2024). We define three text refinement operations frequently observed in everyday LLM usage. Among them, *human-written & LLM-refined text* (HLT) is distinguished from HWT and LGT, allowing us to extend the task from Binary to Ternary classification and evaluate the detectors' ability to identify such hybrid text. In contrast, *LLM-generated & LLM-*

---

*Work was conducted during the internships of Junchao Wu at Alibaba Group.
†Corresponding to derekfw@um.edu.mo.
[1]Code and data: `https://github.com/AIDC-AI/Marco-LLM/tree/main/DetectRL-X`.

| **Benchmark ↓ Eval→** | Size | Task | | Real-world | | | | | | Multilingual | | | |
|---|---|---|---|---|---|---|---|---|---|---|---|---|---|
| | | BINARY | TERNARY | Multi-Domain | Multi-Generator | Paraphrase Attack | Perturbation Attack | Multi-Length | Multi-Operation | Language Num | Domain Balance? | Generator Balance? | Multilingual Attacks? |
| TuringBench (Uchendu et al., 2021) | 168K | ✓ | ✗ | ✓ | ✓ | ✗ | ✗ | ✗ | ✗ | 1 | ✗ | ✗ | ✗ |
| HC3 (Uchendu et al., 2021) | 125K | ✓ | ✗ | ✓ | ✓ | ✗ | ✗ | ✗ | ✗ | 2 | ✗ | ✗ | ✗ |
| MGTBench (He et al., 2024) | 21K | ✓ | ✗ | ✗ | ✓ | ✓ | ✓ | ✓ | ✗ | 1 | ✗ | ✗ | ✗ |
| MULTITuDE (Macko et al., 2023) | 74K | ✓ | ✗ | ✗ | ✓ | ✗ | ✓ | ✗ | ✗ | 11 | ✗ | ✓ | ✗ |
| M4 (Wang et al., 2024b) | 245K | ✓ | ✗ | ✓ | ✓ | ✗ | ✗ | ✓ | ✗ | 7 | ✗ | ✗ | ✗ |
| MAGE (Li et al., 2024) | 448K | ✓ | ✗ | ✓ | ✓ | ✓ | ✗ | ✓ | ✗ | 1 | ✗ | ✗ | ✗ |
| RAID (Dugan et al., 2024) | 6,287K | ✓ | ✗ | ✓ | ✓ | ✓ | ✓ | ✗ | ✗ | 1 | ✗ | ✗ | ✗ |
| Stumbling Blocks (Wang et al., 2024a) | 10K | ✓ | ✗ | ✗ | ✓ | ✓ | ✓ | ✗ | ✗ | 1 | ✗ | ✓ | ✗ |
| DetectRL (Wu et al., 2024) | 235K | ✓ | ✗ | ✓ | ✓ | ✓ | ✓ | ✓ | ✗ | 1 | ✗ | ✗ | ✗ |
| **DETECTRL-X (Ours)** | 3,456K | ✓ | ✓ | ✓ | ✓ | ✓ | ✓ | ✓ | ✓ | 8 | ✓ | ✓ | ✓ |

Table 1: Comparison of DETECTRL-X with existing benchmarks.

*refined text* remains categorized as LGT, and we further analyze how additional refinement operations influence its detectability.

In this paper, we study the following questions: **(1) How do detectors perform in multilingual and cross-lingual settings? (2) Which real-world factors influence detector performance within and across languages, and to what extent? (3) Can detectors identify HLT and withstand diverse text refinement operations?**

The proposed benchmark effectively addresses the research objectives while posing significant challenges to existing detection methods. Experimental results reveal that statistical detectors generally struggle with multilingual LGT in the BINARY task, whereas neural-based models achieve superior performance. Interestingly, in cross-lingual settings, statistical methods exhibit relative stability, while neural approaches experience a more acute performance drop, particularly with the introduction of the HLT category. Furthermore, both detector types deteriorate sharply under real-world perturbations. The TERNARY task proves substantially more demanding: most statistical methods lose their efficacy, whereas neural architectures demonstrate greater resilience, highlighting their potential for future hybrid-text detection research.

In summary, our benchmark offers a comprehensive evaluation of LGT detection across diverse languages and scenarios. By exposing the limitations of current detectors and the unique challenges of HLT, this work paves the way for more robust and equitable detection systems.

## 2 Related Work

### 2.1 Progress of LGT Detection

Current detection methods can generally be divided into two main categories: neural-based methods and statistical-based methods (Wu et al., 2025a). Neural-based methods typically employ supervised approaches, such as fine-tuning XLM-RoBERTa-Classifier and mDeBERTa classifiers (Liu et al., 2019; He et al., 2021). While these methods can achieve relatively superior detection performance, they heavily rely on large amounts of training data and are prone to overfitting, which limits their ability to generalize effectively to out-of-distribution (OOD) data (Li et al., 2024; Wu et al., 2025b). In contrast, statistical-based methods focus on identifying statistical features that can distinguish between two types of text and use distribution thresholds for classification, such as Log-Likelihood (Solaiman et al., 2019), Log-Rank (Gehrmann et al., 2019), DetectLLM-LRR (Su et al., 2023), GECScore (Wu et al., 2025c), ReviseDetect. (Zhu et al., 2023), Fast-DetectGPT (Bao et al., 2024), Lastde++ (Xu et al., 2025) Binoculars (Hans et al., 2024) and Repreguard (Chen et al., 2025b). Although these methods may require impractical white-box access to the generator, their higher interpretability makes them more convincing and trustworthy to end-users.

### 2.2 LGT Detection Benchmark

While significant strides have been made in LGT detection benchmarks, existing efforts often rely on idealized test data to evaluate performance across domains and generators (Wu et al., 2025a). For example, although TuringBench (Uchendu et al., 2021) and MAGE (Li et al., 2024) offer large-scale datasets spanning diverse domains, they remain primarily English-centric and provide insufficient simulation of challenging stress-testing scenarios. Recent benchmarks like Stumbling Blocks (Wang et al., 2024a), RAID (Dugan et al., 2024), and DetectRL (Wu et al., 2024), better align with real-world conditions by introducing attack strategies to simulate text rewriting and noise perturbations; however, their scope is similarly confined to English. Multilingual efforts, such as M4 (Wang et al., 2024b), broaden language coverage but suffer from imbalanced and limited distributions across languages and domains. For example, although M4 includes seven languages, most non-English samples are restricted to a single domain, and generator coverage is uneven, leading to unfair cross-lingual comparisons. Similar multilingual benchmarks, including HC3 (Guo et al., 2023) and MULTITuDE (Macko et al., 2023), also focus primarily on idealized detection settings and provide limited exploration of real-world stress scenarios and their impacts across languages. Moreover, existing benchmarks generally overlook the influence of text refinement operations such as polishing, expansing, and condensing applied to both HWT and LGT, which can significantly obscure detection boundaries. To address these gaps, this study introduces a general, scalable, and large-scale multilingual benchmark framework aimed at constructing more real-world evaluation environments and promoting the development of robust and equitable detection systems through systematic stress testing.

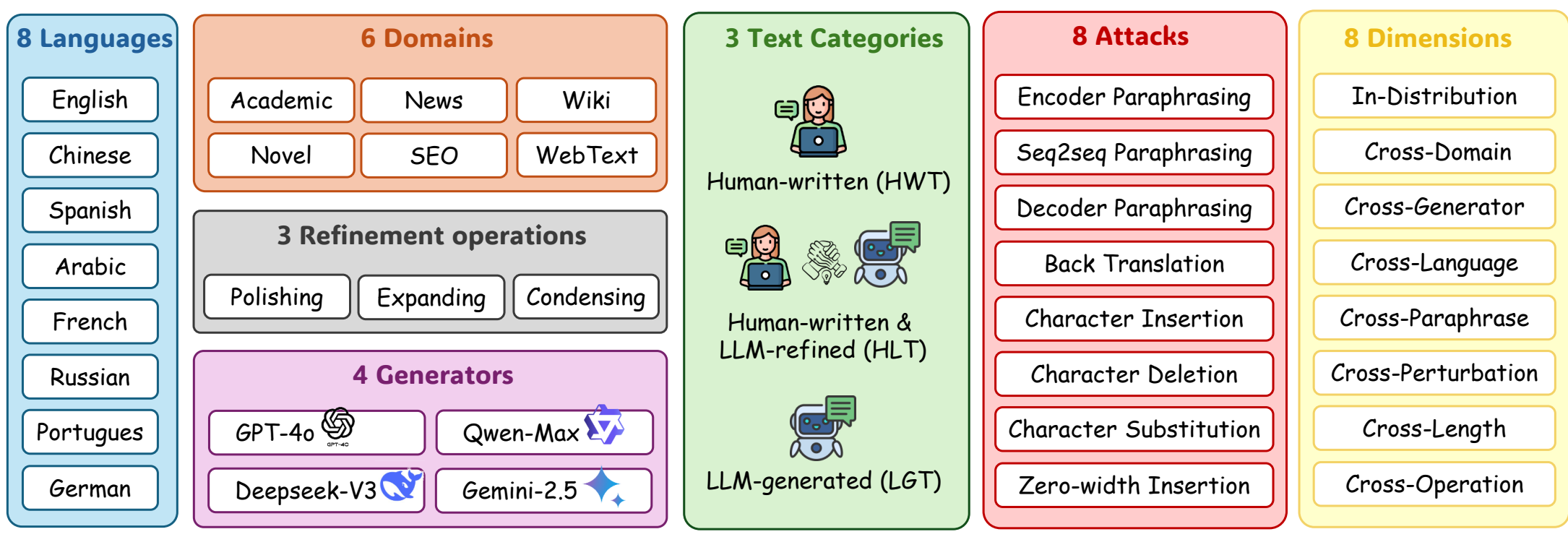

Figure 1: An overview of the structure of the DETECTRL-X. The benchmark comprises 3.46 million samples, making it the largest known multilingual LGT detection dataset. It includes 8 languages, 6 domains, 4 generators, 8 attack scenarios, 4 text-length granularities, and 3 types of refinement operations. The benchmark supports BINARY and TERNARY classification tasks, and includes 8 evaluation dimensions comparing 12 representative detectors.

# 3 DETECTRL-X

## 3.1 Benchamrking Data Design

### 3.1.1 Task Definition

In this study, we extend and formalize the LGT detection task. Traditionally, this task has been formulated as a BINARY classification problem distinguishing between HWT and LGT. However, with the growing use of LLMs for text refinement, many real-world texts now often result from human–LLM collaboration (Zhang et al., 2024). In practice, HWT refined by an LLM is generally acceptable, whereas purely LGT is typically restricted (Wu et al., 2025a). To better reflect this real-world usage, we extend the detection task from BINARY to a TERNARY setting that can also identify HLT. Specifically, given a text $t \in \mathcal{T}$, the traditional BINARY and TERNARY detection functions are defined as:

$$f_{\text{BINARY}} : \mathcal{T} \to \{\text{HWT}, \text{LGT}\},$$
$$f_{\text{TERNARY}} : \mathcal{T} \to \{\text{HWT}, \text{HLT}, \text{LGT}\}.$$

The detector aims to learn an optimal function $f^*$ maximizing accuracy over the data $\mathcal{D}$:

$$f^* = \arg\max_{f} \mathbb{E}_{(x,y)\sim\mathcal{D}}[\,\mathbf{1}\{f(x) = y\}\,]$$

where $x$ denotes a input text and $y \in \{\text{HWT}, \text{HLT}, \text{LGT}\}$ is its ground-truth label. To capture text refinement in human–LLM collaboration, we define a set of representative operations reflecting common LLM-assisted writing practices:

$$\mathcal{R} = \{r_p, r_e, r_c\},$$

where $r_p$, $r_e$, and $r_c$ denote *Polishing*, *Expanding*, and *Condensing*, respectively. Based on text source and refinement process, the refined text categories can be formally defined as:

$$\text{HLT} = r(\text{HWT}), \quad \text{LGT} = r(\text{LGT}), \quad r \in \mathcal{R}.$$

This formulation enables systematic evaluation of how refinement operations influence detection performance and allows detectors to distinguish HWT, HLT, and multi-step LGT more effectively, clarifying the boundary between *LLM-assisted writing* and *fully LLM-automated writing*.

### 3.1.2 Languages

To evaluate the robustness and cross-lingual generalization of LGT detection methods, we collect a dataset covering 8 languages from 5 language families: English and German (Germanic),

Spanish, French, and Portuguese (Romance), Russian (Slavic), Arabic (Semitic), and Chinese (Sino-Tibetan). These languages differ significantly in grammar, lexicon, and writing system, representing diverse linguistic structures and playing crucial roles in global communication and business, thus providing high practical value for detection research. Based on language morphological richness and typological distance from English (Arnett and Bergen, 2025; Kuribayashi et al., 2020), the selected languages are grouped into three complexity levels: high (Arabic, Russian, Chinese), medium (German, French, Spanish, Portuguese), and low (English). The classification criteria and linguistic rationale are detailed in Appendix B.1.

#### 3.1.3 Data Sources

We curated HWTs from 6 domains where LLMs are widely used and prone to misuse: Academic writing, News reporting, Novel creation, Search Engine Optimization (SEO) content, Wikipedia entries (Wiki), and general Web text (WebText) (Wang et al., 2024b; Chen and Wang, 2025; Venkatraman et al., 2025; Quaremba et al., 2025). To ensure authenticity and avoid contamination from LGTs, we collected samples exclusively from publicly available internet sources published prior to 2022 years. We implemented multi-stage filtering strategies to remove malformed texts, encoding errors, and ensure balanced representation across writing styles and complexity levels within each domain. Detailed descriptions of the specific data sources for each domain are provided in Appendix B.2.

#### 3.1.4 Generators

To ensure the diversity and representativeness of the generated texts, we selected 4 mainstream multilingual LLMs as generators, including Deepseek-V3 (DeepSeek-AI et al., 2024), Gemini-2.5-flash (Comanici et al., 2025), GPT-4o (OpenAI, 2023), and Qwen-Max (Yang et al., 2024). These generators collectively support all 8 languages of interest and represent the latest and most popular technological advancements in the world, widely used in various daily scenarios (OpenRouter, 2025). Using them as text generators allows our evaluation to more accurately reflect the performance and robustness of detectors in real-world applications. Detailed generator configurations, prompt templates, and parameter settings are provided in Appendix B.3.

#### 3.1.5 Revision Operations

To reflect real-world human-LLM collaboration, we define three revision operations based on established writing taxonomies (Faigley and Witte, 1981; Coenen et al., 2021): (1) **Polishing** (improving fluency and style); (2) **Expanding** (adding details and elaboration); and (3) **Condensing** (removing redundancies). These operations are applied to HWT to create the HLT category and to LGT to test detector robustness against post-processing. We consistently use Qwen-Max as the revision assistant to ensure high-quality, uniform data construction across all languages. Detailed prompt templates and implementation settings are provided in Appendix B.7.

#### 3.1.6 Attacks and Data A ugmentation

To comprehensively evaluate detector robustness in practical scenarios, we propose a multilingual data augmentation and attack framework that aligns closely with real-world detection challenges. While data augmentation is traditionally used to improve model generalization (Wei and Zou, 2019; Chen et al., 2023; Liang et al., 2026), we employ these techniques as adversarial transformations to probe the vulnerabilities of detection systems (Wu et al., 2025a). The framework incorporates diverse paraphrasing and perturbation strategies to enhance the coverage and realism of existing benchmarks. Unlike prior work, it ensures that all attack methods are uniformly applicable across different languages, providing a solid basis for multilingual LGT detection research. The attack suite includes two categories: (1) **Paraphrase Attacks:** These focus on semantic-preserving transformations (Krishna et al., 2023; Sennrich et al., 2016), including Encoder Paraphrasing (EP), Seq2seq Paraphrasing (SP), Decoder Paraphrasing (DP), and Back-Translation (BT); and (2) **Perturbation Attacks:** These introduce fine-grained noise to the text (Wang et al., 2024a), including Character Insertion (CI), Character Substitution (CS), Character Deletion (CD), and Zero-width Insertion (ZI). Detailed descriptions and implementation details for attack strategies are provided in Appendix B.4 and B.5.

**Multiple Text Length Granularities** To assess length sensitivity, we generated semantically complete sub-samples of 64, 128, 256, and 512 tokens through spaCy-based sentence segmentation and merging, representing diverse text-length scenarios. Detailed procedures are provided in Appendix B.6.

### 3.2 Data Statistics and Distribusion

The resulting DETECTRL-X dataset comprises 3,456,000 samples, with a 2:1 train-test split that maintains a balanced distribution across all categories. Detailed statistics are described in Table 5 and Appendix C.

### 3.3 Evaluation Module Design

Our evaluation framework systematically assesses the performance of detectors across 8 key dimensions to comprehensively measure their robustness and generalization capabilities. Based on the definition of the detection task (§ 3.1.1), we have established separate leaderboards for BINARY and TERNARY classification tasks. The specific descriptions of the 8 evaluation dimensions are as follows:

- **In-Distribution Performance**: Evaluates detector within a mixed distribution across 8 languages, 6 domains, and 4 generators.
- **Cross-Domain Performance**: Evaluates generalization to unseen domains, across all generators and languages.
- **Cross-Generator Performance**: Evaluates generalization to unseen generators, across all domains and languages.
- **Cross-Language Performance**: Evaluates generalization to unseen languages, across all domains and generators.
- **Cross-Paraphrase Performance**: Evaluates robustness against 4 paraphrase attacks, across all domains, generators, and languages.
- **Cross-Perturbation Performance**: Evaluates robustness against 4 perturbation attacks, across all domains, generators, and languages.
- **Cross-Length Performance**: Evaluates robustness to 4 text length granularities, across all domains, generators and languages.
- **Cross-Operations Performance**: Evaluates robustness to 3 text refinement operations, across all domains, generators and languages.

**Evaluation Metrics** We adopt two primary evaluation metrics: Best $F_1$ Score ($F_1^B$) and $F_1$ at False Positive Rate = 0.01 ($F_1^F$), both calculated using macro-averaging. The $F_1^B$ metric captures the optimal balance between precision and recall, the optimal threshold is determined by maximizing Youden's $J$ statistic on the training set, reflecting the detector's overall performance. The $F_1^F$ metric measures the $F_1$ score at a fixed low false positive rate (FPR = 0.01), highlighting the detector's reliability in scenarios with minimal tolerance for false positives. The threshold $\tau$ is calibrated on the training set and evaluated on the test set. Alongside our multi-dimensional framework, this metric ensures a evaluation of both effectiveness and practical usability.

## 4 Experiment

### 4.1 Detection Methods

To comprehensively evaluate the performance of various detection methods across diverse application scenarios, we select a representative set of detectors covering both statistical and neural-based state-of-the-art (SOTA) approaches. The evaluated methods include Log-Likelihood (Solaiman et al., 2019), Log-Rank (Gehrmann et al., 2019), DetectLLM-LRR (Su et al., 2023), GEC-Score (Wu et al., 2025c), ReviseDetect (Zhu et al., 2023), Fast-DetectGPT (Bao et al., 2024), Binoculars (Hans et al., 2024), Lastde++ (Xu et al., 2025), RepreGuard (Chen et al., 2025b), XLM-RoBERTa-Classifier (X-Rob-Classifier, Liu et al. (2019)), mDeBERTa-Classifier (He et al., 2021), and Biscope (Guo et al., 2024). Since LLMs in real-world applications typically operate as black-boxes and are inaccessible, watermarking methods are excluded from our evaluation scopes. Detailed descriptions and implementation details of each detector are provided in Appendix E.1.

### 4.2 LeaderBoard and Results

#### 4.2.1 Main Results

We evaluate existing detectors on DETECTRL-X, the results are shown in Table 2. Higher average scores indicate stronger generalization and practical performance across diverse detection scenarios.

Overall, neural-based detectors consistently outperform statistical-based detectors on both the BINARY and TERNARY leaderboards, showing consistent ranking trends across the two classification tasks. X-Rob-Classifier and mDeBERTa-Classifier achieve the best results: X-Rob-Classifier ranks first on the BINARY leaderboard with average $F_1^B$ and $F_1^F$ scores of 95.58% and 91.31%, while mDeBERTa-Classifier leads on the TERNARY leaderboard with 87.68% and 81.10%, and ranks second on the BINARY leaderboard with 95.48% and 93.20%, respectively. Even the weaker neural-based detector, Biscope, performs well, achieving average $F_1^B$ and $F_1^F$ scores of 80.06% and 63.62% (Binary) and 59.69% and 37.91% (Ternary), surpassing the best statistical-based detector, GECScore, by 9.96% and 11.84% in $F_1^B$, and by 6.26% and 1.81% in $F_1^F$. These results show that neural-based detectors are significantly more robust and adaptable in multilingual and real-world conditions, while statistical-based detectors experience notable performance degradation in complex mixed distributions and distribution shifts.

| Binary | In-Distribution | | Cross-Domain | | Cross-Generator | | Cross-Language | | Cross-Paraphrase | | Cross-Perturbation | | Cross-Length | | Cross-Operation | | Avg. | | **Rank** |
|---|---|---|---|---|---|---|---|---|---|---|---|---|---|---|---|---|---|---|---|
| | $F_1^B$ | $F_1^F$ | $F_1^B$ | $F_1^F$ | $F_1^B$ | $F_1^F$ | $F_1^B$ | $F_1^F$ | $F_1^B$ | $F_1^F$ | $F_1^B$ | $F_1^F$ | $F_1^B$ | $F_1^F$ | $F_1^B$ | $F_1^F$ | $F_1^B$ | $F_1^F$ | |
| Log-Likelihood | 66.89 | 38.09 | 64.76 | 41.17 | 65.96 | 38.08 | 62.68 | 46.89 | 32.95 | 35.74 | 26.84 | 33.12 | 58.82 | 35.33 | 63.07 | 34.10 | 55.25 | 37.81 | 10 |
| Log-Rank | 67.21 | 39.94 | 65.21 | 43.03 | 66.12 | 39.94 | 63.55 | 47.41 | 39.62 | 35.38 | 28.05 | 33.12 | 58.37 | 35.99 | 63.07 | 34.32 | 56.40 | 38.64 | 9 |
| DetectLLM-LRR | 69.05 | 17.83 | 67.26 | 20.32 | 66.86 | 17.77 | 68.14 | 22.16 | 40.45 | 35.89 | 36.37 | 33.48 | 56.25 | 38.47 | 59.71 | 37.14 | 58.01 | 27.88 | 8 |
| Fast-DetectGPT | 54.22 | 42.43 | 53.69 | 43.06 | 54.10 | 42.45 | 53.18 | 43.67 | 42.89 | 36.73 | 30.88 | 33.31 | 56.43 | 47.37 | 40.76 | 34.99 | 48.27 | 40.50 | 11 |
| Lastde++ | 38.98 | 33.54 | 37.98 | 33.62 | 38.86 | 33.54 | 38.17 | 33.58 | 53.12 | 35.08 | 39.76 | 33.77 | 36.23 | 33.39 | 36.65 | 33.44 | 39.97 | 33.74 | 12 |
| Binoculars | 75.87 | 59.75 | 75.54 | 58.37 | 75.87 | 59.75 | 73.65 | 60.51 | 53.48 | 41.11 | 45.62 | 36.17 | 74.89 | 66.18 | 70.48 | 51.62 | 68.17 | 54.18 | 6 |
| Revise-Detect | 82.40 | 0.00 | 81.95 | 22.05 | 82.37 | 0.00 | 80.03 | 0.00 | 18.28 | 0.00 | 0.33 | 0.00 | 41.33 | 0.00 | 79.61 | 0.00 | 58.29 | 2.76 | 7 |
| GECScore | 83.22 | 61.74 | 81.69 | 66.17 | 83.28 | 61.66 | 78.40 | 65.59 | 38.01 | 33.61 | 31.02 | 33.41 | 73.83 | 59.05 | 91.36 | 77.66 | 70.10 | 57.36 | 4 |
| RepreGuard | 73.21 | 56.95 | 78.07 | 59.54 | 74.09 | 53.90 | 54.24 | 47.68 | 66.37 | 46.86 | 66.17 | 50.11 | 71.01 | 50.68 | 75.65 | 59.88 | 69.85 | 53.20 | 5 |
| X-Rob-Classifier | 99.95 | 99.50 | 94.76 | 84.99 | 99.19 | 99.10 | 96.62 | 88.35 | 77.32 | 64.13 | 97.92 | 96.34 | 98.90 | 98.42 | 99.94 | 99.65 | 95.58 | 91.31 | 1 |
| mDeBERTa-Classifier | 99.96 | 99.50 | 96.04 | 90.87 | 99.33 | 99.15 | 97.63 | 95.70 | 73.43 | 63.88 | 98.26 | 97.93 | 99.20 | 98.86 | 99.97 | 99.72 | 95.48 | 93.20 | 2 |
| Biscope | 92.85 | 83.66 | 89.44 | 74.04 | 91.39 | 79.07 | 79.94 | 62.09 | 57.76 | 37.62 | 57.39 | 33.38 | 81.09 | 56.95 | 90.62 | 82.17 | 80.06 | 63.62 | 3 |

| Ternary | In-Distribution | | Cross-Domain | | Cross-Generator | | Cross-Language | | Cross-Paraphrase | | Cross-Perturbation | | Cross-Length | | Cross-Operation | | Avg. | | **Rank** |
|---|---|---|---|---|---|---|---|---|---|---|---|---|---|---|---|---|---|---|---|
| | $F_1^B$ | $F_1^F$ | $F_1^B$ | $F_1^F$ | $F_1^B$ | $F_1^F$ | $F_1^B$ | $F_1^F$ | $F_1^B$ | $F_1^F$ | $F_1^B$ | $F_1^F$ | $F_1^B$ | $F_1^F$ | $F_1^B$ | $F_1^F$ | $F_1^B$ | $F_1^F$ | |
| Log-Likelihood | 42.63 | 35.90 | 39.26 | 35.26 | 41.05 | 35.18 | 38.95 | 33.26 | 27.86 | 21.27 | 21.25 | 15.03 | 36.86 | 31.10 | 42.21 | 35.63 | 36.26 | 30.33 | 8 |
| Log-Rank | 43.03 | 36.15 | 39.33 | 35.38 | 41.36 | 35.29 | 39.74 | 33.97 | 28.58 | 21.23 | 22.62 | 15.34 | 39.59 | 31.02 | 41.51 | 21.23 | 36.97 | 28.70 | 7 |
| DetectLLM-LRR | 38.88 | 35.33 | 37.37 | 34.71 | 39.31 | 34.73 | 38.22 | 34.48 | 26.13 | 21.56 | 23.09 | 19.48 | 36.39 | 30.34 | 36.62 | 35.05 | 34.50 | 30.71 | 9 |
| Fast-DetectGPT | 32.14 | 27.81 | 29.41 | 26.79 | 27.53 | 25.74 | 30.62 | 26.86 | 26.36 | 21.96 | 19.00 | 17.08 | 30.45 | 26.88 | 29.41 | 27.88 | 28.12 | 25.12 | 10 |
| Lastde++ | 16.67 | 16.67 | 20.36 | 17.01 | 17.38 | 16.87 | 19.14 | 16.94 | 16.67 | 16.67 | 19.00 | 17.08 | 16.69 | 16.69 | 16.67 | 16.67 | 17.82 | 16.83 | 12 |
| Binoculars | 44.06 | 32.97 | 46.57 | 34.98 | 45.66 | 35.99 | 46.05 | 36.22 | 32.63 | 21.75 | 29.67 | 18.69 | 42.31 | 32.22 | 45.98 | 35.50 | 41.62 | 31.04 | 6 |
| Revise-Detect | 56.05 | 46.12 | 50.77 | 41.44 | 56.15 | 46.35 | 52.39 | 44.65 | 31.22 | 19.26 | 28.98 | 16.78 | 44.43 | 34.69 | 54.03 | 46.45 | 46.75 | 36.97 | 5 |
| GECScore | 56.60 | 42.61 | 53.26 | 50.30 | 56.04 | 48.44 | 53.29 | 48.77 | 31.03 | 15.93 | 30.07 | 14.54 | 47.22 | 28.80 | 55.29 | 39.41 | 47.85 | 36.10 | 4 |
| RepreGuard | 23.33 | 17.06 | 35.04 | 27.90 | 30.11 | 20.47 | 28.80 | 28.51 | 25.19 | 17.85 | 20.39 | 16.91 | 26.71 | 20.15 | 30.28 | 22.12 | 27.48 | 21.37 | 11 |
| X-Rob-Classifier | 94.40 | 93.04 | 73.01 | 50.51 | 88.85 | 80.22 | 69.12 | 48.82 | 77.03 | 70.06 | 95.25 | 94.67 | 85.32 | 79.92 | 79.27 | 81.77 | 82.78 | 74.88 | 2 |
| mDeBERTa-Classifier | 96.81 | 95.81 | 79.17 | 61.48 | 92.07 | 88.06 | 76.26 | 59.17 | 85.25 | 79.11 | 97.22 | 96.18 | 92.14 | 88.53 | 82.52 | 80.47 | 87.68 | 81.10 | 1 |
| Biscope | 70.10 | 49.60 | 63.82 | 41.56 | 67.03 | 45.44 | 53.46 | 34.78 | 48.75 | 26.19 | 55.94 | 29.00 | 55.17 | 28.15 | 63.25 | 48.56 | 59.69 | 37.91 | 3 |

Table 2: DetectRL-X Leaderboard. This leaderboard evaluates the robustness performance of different detectors in multilingual real-world settings and is divided into two sub-leaderboards: Binary and Ternary classification. It provides insights for developers with varying detection objectives.

### 4.2.2 Challenges and Findings

**Limitations of Statistical-Based Detectors.** Even in the In-Distribution setting, statistical-based detectors struggle with data spanning multiple domains, generators, and languages, achieving an average $F_1^B$ score of 67.89%. The best-performing system, GECScore, reaches only 83.22%, indicating a lack of robustness to real-world heterogeneity. The main reason is that most existing studies evaluate detectors within single domains or generators, leading to an overestimation of their effectiveness. This highlights the need for comprehensive and complex evaluation settings and benchmarks.

**Cross-Lingual Generalization Challenge.** a Neural-based detectors experience substantial degradation in cross-lingual settings, revealing limited transferability across languages. In Binary, the average $F_1^B$ drops from 95.3% to 91.4% (↓ 3.9%), while in Ternary, the decline is more pronounced, falling from 87.10% to 66.28% (↓ 20.82%), with strong mDeBERTa-Classifier dropping 20.55%. In contrast, statistical-based detectors show smaller average decreases (around 4.8% for Binary and 0.7% for Ternary) but exhibit higher result variance, highlighting the need for language-invariant feature modeling and stronger cross-lingual robustness.

**Focus More on Cross-Domain than Generator.** Our findings reveal that Cross-Domain is more challenging than Cross-Generator. In Binary, the average $F_1^B$ of neural-based detectors drops by 2.95%, compared to only 0.78% in Cross-Generator. In Ternary, the gap widens further, with the performance of the powerful mDeBERTa-Classifier decreasing by 18.2%, far exceeding its 4.9% decline in Cross-Generator. These results show that while detectors remain relatively stable across generators, they experience sharp performance degradation under domain shifts, making Cross-Domain generalization a critical performance bottleneck.

**Vulnerability to Paraphrase and Perturbation.** Both paraphrase and perturbation substantially degrade detector performance. In Binary, neural-based detectors' $F_1^B$ scores drop by 28.1% and 13.1% in Cross-Paraphrase and Cross-Perturbation, respectively, while in TERNARY, the declines are 16.8% and 4.3%, respectively. Paraphrase consistently exerts a stronger effect than perturbation in both settings, with a more pronounced impact in BINARY than in TERNARY. Statistical-based detectors suffer greater performance losses (25–40% for BINARY and 13–27% for TERNARY) due to feature distribution drift. Despite better performance, neural-based detectors still experience drops of up to 35.5% under such perturbations, revealing limited robustness to semantic and surface variations in multilingual contexts.

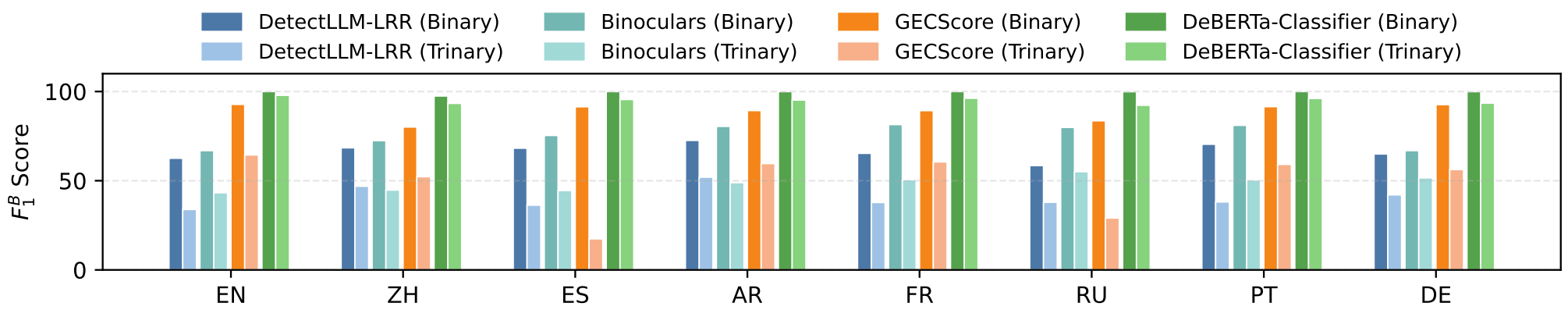

Figure 2: In-Distribution Performance of Different Detectors on Different Languages.

**Sensitivity to Length and Operation Variations.** Variations in text length and operations reduce detection robustness. In BINARY, neural-based detectors' $F_1^B$ scores drop by around 4.5% and 1% under Cross-Length and Cross-Operation, respectively. In TERNARY, the averages decline by 11.9% and 13.4%, with the best mDeBERTa-Classifier dropping by 4.96% and 14.29%. Greater variation in text length and operations weakens detector discriminability, particularly in TERNARY. Statistical-based detectors exhibit larger decreases (around 35–40% for both BINARY and TERNARY), indicating limited adaptability to text length and operation variations.

**Intrinsic Difficulty of TERNARY Task.** Across all experiments, TERNARY is more challenging than BINARY detection. Under In-Distribution, statistical-based detectors' $F_1^B$ drops from 67.9% to 39.3% (↓ 28.6%), while neural-based detectors decrease from 97.6% to 87.1% (↓ 10.5%). For the overall average results, statistical-based show a 23.0% decrease (from 58.3% to 35.3%) and neural-based detectors a 13.7% decrease (from 90.4% to 76.7%). In Cross-Domain, Language, Paraphrase, Length, and Operation challenges, this performance gap further widens due to blurred boundaries and feature overlap among LGT, HWT, and HLT. Although the mDeBERTa-Classifier achieves an $F_1^B$ score of 87.68% in TERNARY, it shows strong potential but still requires further optimization for practical deployment. Overall, TERNARY better reflects hybrid authorship and reveals the performance limitations of current methods.

**Usability Gaps Under Low False Positive Rate.** We find that many detectors achieve high $F_1^B$ but drop sharply in $F_1^F$, revealing difficulty in maintaining accuracy under strict reliability constraints. For instance, GECScore and Binoculars show declines of 12.7% and 14.0% from $F_1^B$ to $F_1^F$. Since real-world applications such as academic integrity checks and content safety require minimal false positives, both metrics should be considered to ensure detectors are not only accurate but also reliable.

## 5 Analysis across Different Languages

This section analyzes representative detectors to reveal their strengths and weaknesses across different languages. The selected detectors include DetectLLM-LRR (gray-box statistical-based method using simple logit features), Binoculars (gray-box-based statistical method using contrastive logit features), GECScore (black-box statistical method), and mDeBERTa-Classifier (neural-based method). These detectors achieve the best performance in their respective categories, making them suitable candidates for detailed comparative analysis.

### 5.1 ID Performance on Different Languages

In the In-Distribution (ID) setting, the average $F_1^B$ results by language complexity are shown in Figure 2. High, medium, and low-complexity languages achieve $F_1^B$ scores of 81.9%, 83.5%, and 80.5% in BINARY, and 58.9%, 57.7%, and 59.9% in TERNARY, respectively. The variations in performance across language complexity levels remain within ±3%, indicating that language complexity has minimal impact on detection performance. mDeBERTa-Classifier performs best across all languages, exceeding 99% in BINARY and about 95% in TERNARY. GECScore, Binoculars, and DetectLLM-LRR show slightly lower $F_1^B$ on high-complexity languages than on other languages, with performance gaps of only 1% to 5%. At the single-language level, the detectors perform best in Arabic, achieving $F_1^B$ of 85.6% and 63.4%, while they perform worst in Russian, with 80.5% and 53.6%. Despite both being high-complexity languages, this supports that there is no clear correlation between linguistic complexity and detection performance.

### 5.2 Cross-Language Generalization on Different Languages

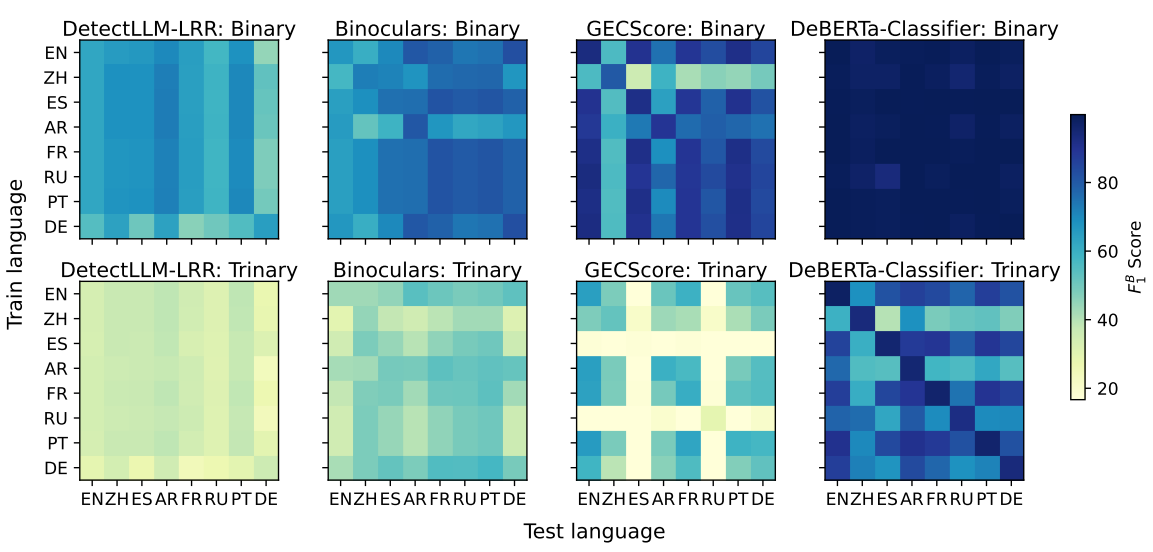

Figure 3: Generalization Performance of Different Detectors Across Languages.

We evaluated detectors' generalization across languages differing in linguistic complexity and family affiliation (Table 15). The overall trend shows that as linguistic complexity increases, generalization performance declines, and this degradation becomes even more pronounced when moving BINARY to TERNARY. Among all detectors, mDeBERTa-Classifier exhibited the highest stability in BINARY, with an average $F_1^B$ of 97.22%, only 2.42% lower than its In-distribution performance. In contrast, DetectLLM-LRR, Binoculars, and GECScore showed larger declines of 6.70%, 8.41%, and 15.78%, respectively. Generalization between low and medium-complexity languages is relatively stable. In contrast, for high-complexity languages (e.g. Arabic, Russian, Chinese), statistical-based methods drop by about 17%, whereas mDeBERTa-Classifier remains above 95%. Cross-language family transfer introduced an additional challenge. When transferring from Indo-European to Semitic or Sino-Tibetan languages, the average $F_1^B$ of all detectors reduced from 65–75% to 40–55%, corresponding to an average loss of 19.15%. In TERNARY, the best detector (mDeBERTa-Classifier) achieves 76.26% overall, with a 21.37% drop from BINARY and a 20.55% drop from In-distribution of TERNARY, further underscores the significant cross-lingual challenges introduced by HLT, which render detectors highly robust in BINARY nearly ineffective in TERNARY.

### 5.3 Cross-Domain on Different Languages

We evaluated four detectors in both BINARY and TERNARY settings across six domains (Figure 4). In the BINARY task, domain variation exhibited minimal impact, with the performance gap between formal domains (e.g., Academic) and informal ones (e.g., SEO, WebText) remaining below 2%. Among the evaluated models, mDeBERTa-Classifier demonstrated the highest stability, achieving an average $F_1^B$ of 97% in BINARY and 78% in TERNARY classification. The 19% performance degradation suggests that the introduction of HLT significantly amplifies domain sensitivity.In contrast, statistical-based detectors yielded $F_1^B$ scores ranging from 60% to 85%, with marginal domain-induced fluctuations. While their performance dropped by $\sim$5% across domains upon incorporating HLT, the mDeBERTa-Classifier fluctuated by only approximately 2%, further underscoring its robustness against domain shifts. Linguistic analysis revealed a similar trend: while Indo-European languages yielded consistent results, languages such as Chinese, Arabic, and Russian saw a 3–7% decline for statistical methods, likely due to their heavier reliance on domain-specific features and limited cross-lingual generalization.

### 5.4 Cross Attacks on Different Languages

We evaluated detector robustness across 8 attack strategies in different languages (Figure 5a). Across all detectors and languages, average $F_1^B$ scores ranged from 39% under Seq2Seq paraphrasing to 58% under Back-Translation, indicating that attack strategy was the primary factor driving performance variation. In BINARY, the performance gap between categories was narrow: perturbation averaged 53.6%, while paraphrasing averaged 50.0%. In TERNARY, paraphrasing attacks proved more disruptive, reducing the aggregate $F_1$ to 38.7%, compared to 41.8% for perturbations. This shift suggests that the HLT category specifically complicates the decision boundary for rewritten text, moving the primary threat from surface-level noise to deep semantic consistency. Across languages, mean $F_1^B$ values were broadly similar, ranging from 41.1% in Russian to 46.8% in Chinese, a modest 5% gap indicating general cross-lingual stability. However, Chinese and Arabic exhibited higher variability across attack types ($\sigma \approx 24.1\%$) than Indo-European languages ($\sigma \approx 19.8\%$), indicating that higher linguistic complexity increases attack sensitivity.

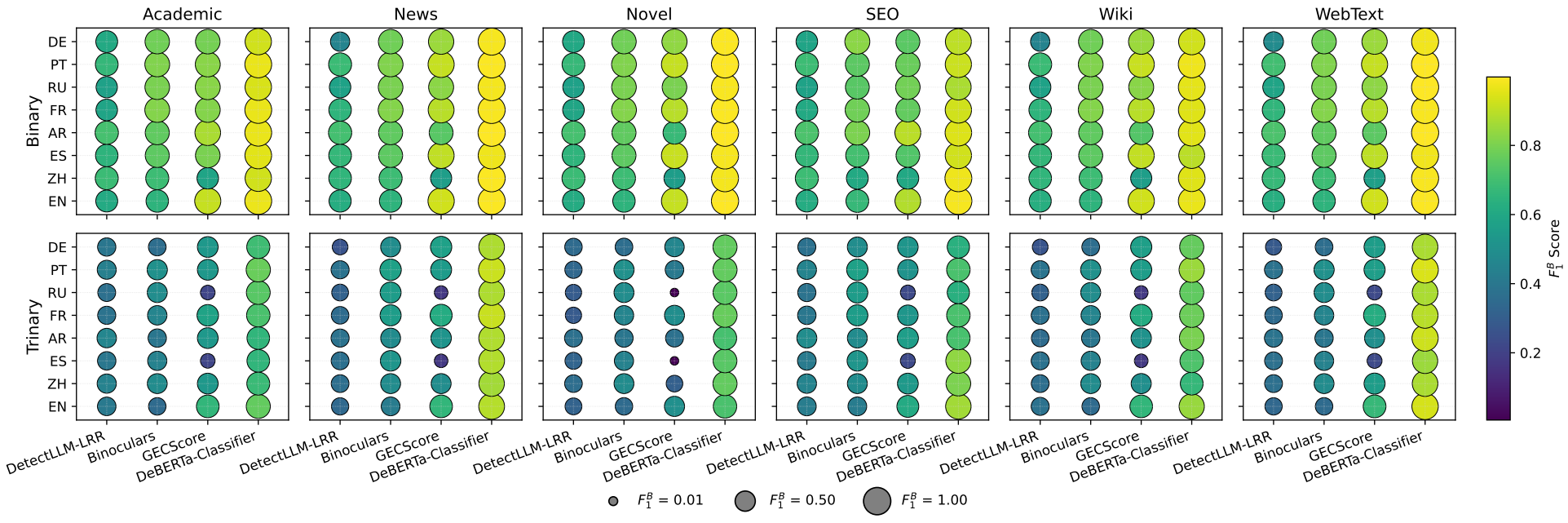

Figure 4: Multilingual Performance of Detectors Across Training Domains.

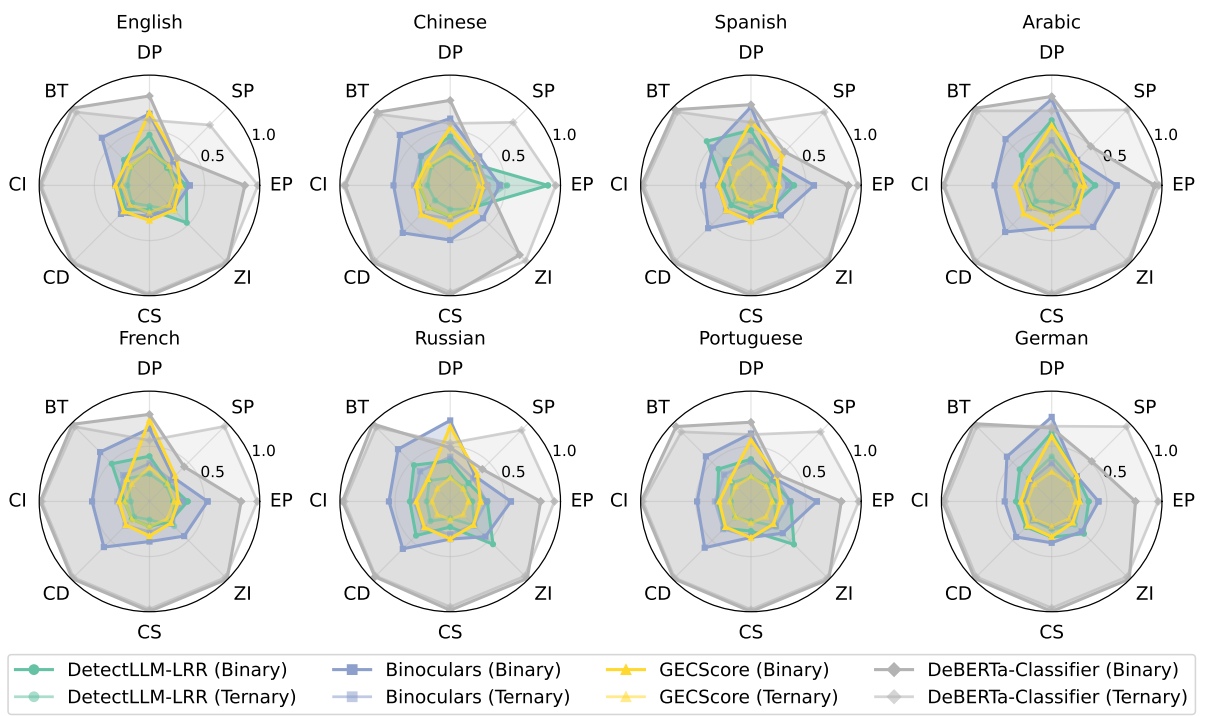
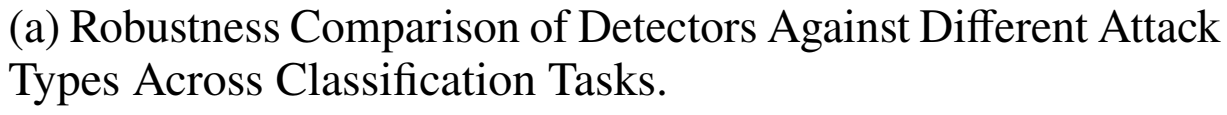

(a) Robustness Comparison of Detectors Against Different Attack Types Across Classification Tasks.

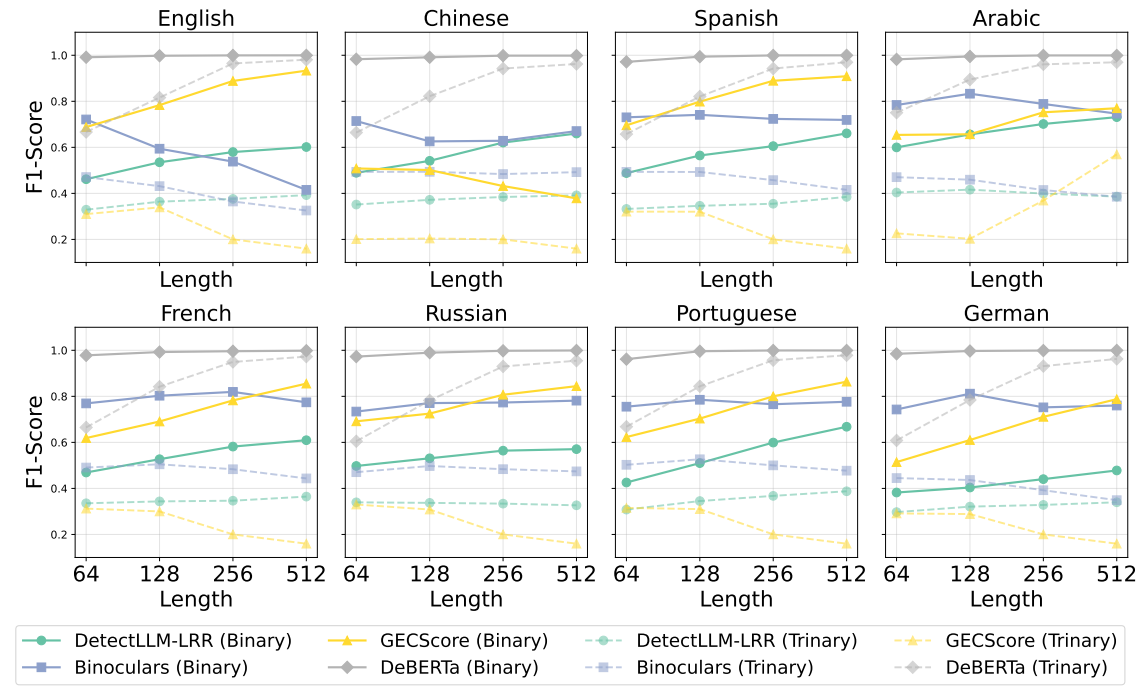

(b) Robustness Comparison of Detectors Against Different Text Lengths Across Classification Tasks.

## 5.5 Cross Length and Operation on Different Languages.

We evaluated detector robustness across 4 length granularities and 3 text refinement operations.

**Text Length.** As illustrated in Figure 5b, detection performance correlates positively with text length, confirming that expanded context improves reliability. The mDeBERTa-Classifier consistently yields the highest and most resilient $F_1^B$ scores, scaling from 96% to nearly 100% in BINARY and from 66% to 97% in TERNARY tasks. Conversely, statistical detectors exhibit lower and more volatile performance. Notably, linguistic complexity influences initial accuracy: complex languages (e.g., Arabic, Russian, and Chinese) yield lower scores for short texts (45-55%) compared to others (55–65%). However, this gap narrows as length increases, with scores converging to 65-73% in BINARY. This suggests that sufficient context can mitigate cross-linguistic performance disparities.

**Refinement Operations.** Cross-operation analysis reveals significant language-dependent variances. The mDeBERTa-Classifier maintains near-perfect $F_1^B$ scores (99.8%-100%) across all languages, demonstrating superior cross-lingual consistency. In contrast, statistical detectors remain sensitive to linguistic variation; for instance, Binoculars scores ∼60% for Arabic and Chinese but ∼75% for English and German. Among operations, Expanding induces the most substantial performance degradation (15-25% lower than other operations), particularly in complex languages. Condensing shows moderate impact, while Polishing remains the least disruptive, likely because it preserves essential stylistic cues while introducing minimal character perturbations.

## 6 Conclusion

In this paper, we present DETECTRL-X, a multilingual benchmark for evaluating the effectiveness and robustness of LGT detectors under real-world scenarios. It extends the traditional BINARY task to a TERNARY classification that distinguishing HLT cases. Results reveal that current detectors exhibit major weaknesses in multilingual and real-world settings. We further analyze the actual factors and extent that affect performance, and provide insights into multilingual robustness through cross-language comparisons. Additionally, DETECTRL-X provides a multilingual data-augmentation framework for building adversarial benchmarks, advancing research toward more fair, robust, and real-world-oriented detection.

## Limitations

Considering the rapid innovation in LLMs, we acknowledge that the time relevance of our benchmarking may be a potential limitation. Due to the swift advancements in LLMs, the text generated by these models may become increasingly indistinguishable from human-written text, posing greater challenges. Although we have used the most cutting-edge models available to mitigate the impact of model updates, this remains a concern.

Another related limitation is the diversity of languages. Our current benchmark includes only eight commonly used languages and does not cover a broader range of regional languages. This decision was made due to the difficulties in data collection and the practical value of the languages included. In future work, we plan to incorporate a more diverse set of language resources to guide the development of more globally applicable detectors.

Furthermore, our benchmark primarily evaluates LLMs as standalone generators, overlooking complex integrated scenarios like Retrieval-Augmented and Multi-Agent Generation. Such dynamic workflows can produce hybrid statistical signatures (Chen et al., 2025a; Jiang et al., 2025; Zhang et al., 2026) that lie beyond the scope of this study and are left for future work.

## Ethical Considerations

The release of DetectRL-X is intended to advance research on the robustness of LLM-generated text detection, particularly within multilingual and real-world scenarios. While we acknowledge the potential risks of misuse associated with open-sourcing dataset construction frameworks and attack strategies, such as circumventing existing detection systems or developing adversarial mechanisms, we maintain that providing transparent stress-testing benchmarks and attack paths is crucial for establishing more resilient defense architectures.

Regarding data ethics, although we have employed a rigorous cleaning process involving automated filtering combined with LLM-as-a-judge and human review to mitigate Personally Identifiable Information and offensive content, residual risks may still exist. Consequently, the use of this dataset and technical framework is strictly restricted to non-commercial academic research purposes. We urge researchers to exercise caution during utilization and encourage users to promptly report any problematic content identified.

## Acknowledgments

This work was supported in part by the Science and Technology Development Fund of Macau SAR (Grant Nos. FDCT/0007/2024/AKP, EF2024-00185-FST), the UM and UMDF (Grant Nos. MYRG-GRG2024-00165-FST-UMDF, MYRG-GRG2025-00236-FST, EF2023-00151-FST), the Stanley Ho Medical Development Foundation (Grant No. SHMDF-AI/2026/001), and the National Natural Science Foundation of China (Grant No. 62266013).

## References

Catherine Arnett and Benjamin Bergen. 2025. Why do language models perform worse for morphologically complex languages? In *Proceedings of the 31st International Conference on Computational Linguistics, COLING 2025, Abu Dhabi, UAE, January 19-24, 2025*, pages 6607–6623. Association for Computational Linguistics.

Guangsheng Bao, Yanbin Zhao, Zhiyang Teng, Linyi Yang, and Yue Zhang. 2024. Fast-detectgpt: Efficient zero-shot detection of machine-generated text via conditional probability curvature. In *The Twelfth International Conference on Learning Representations, ICLR 2024, Vienna, Austria, May 7-11, 2024*. OpenReview.net.

Can Chen and Jun-Kun Wang. 2025. Online detection of llm-generated texts via sequential hypothesis testing by betting. In *Forty-second International Conference on Machine Learning, ICML 2025, Vancouver, BC, Canada, July 13-19, 2025*, Proceedings of Machine Learning Research. PMLR / OpenReview.net.

Guanhua Chen, Yutong Yao, Lidia S. Chao, Xuebo Liu, and Derek F. Wong. 2025a. SGIC: A self-guided iterative calibration framework for RAG. In *Proceedings of the 63rd Annual Meeting of the Association for Computational Linguistics (Volume 1: Long Papers), ACL 2025, Vienna, Austria, July 27 - August 1, 2025*, pages 28357–28370. Association for Computational Linguistics.

Jiaao Chen, Derek Tam, Colin Raffel, Mohit Bansal, and Diyi Yang. 2023. An empirical survey of data augmentation for limited data learning in NLP. *Trans. Assoc. Comput. Linguistics*, 11:191–211.

Xin Chen, Junchao Wu, Shu Yang, Runzhe Zhan, Zeyu Wu, Ziyang Luo, Di Wang, Min Yang, Lidia S. Chao, and Derek F. Wong. 2025b. Repreguard: Detecting llm-generated text by revealing hidden representation patterns. *CoRR*, abs/2508.13152.

Andy Coenen, Luke Davis, Daphne Ippolito, Emily Reif, and Ann Yuan. 2021. Wordcraft: a human-ai collaborative editor for story writing. *CoRR*, abs/2107.07430.

Gheorghe Comanici, Eric Bieber, Mike Schaekermann, Ice Pasupat, Noveen Sachdeva, Inderjit S. Dhillon, Marcel Blistein, Ori Ram, Dan Zhang, Evan Rosen, Luke Marris, Sam Petulla, Colin Gaffney, Asaf Aharoni, Nathan Lintz, Tiago Cardal Pais, Henrik Jacobsson, Idan Szpektor, Nan-Jiang Jiang, and 81 others. 2025. Gemini 2.5: Pushing the frontier with advanced reasoning, multimodality, long context, and next generation agentic capabilities. *CoRR*, abs/2507.06261.

DeepSeek-AI, Aixin Liu, Bei Feng, Bing Xue, Bingxuan Wang, Bochao Wu, Chengda Lu, Chenggang Zhao, Chengqi Deng, Chenyu Zhang, Chong Ruan, Damai Dai, Daya Guo, Dejian Yang, Deli Chen, Dongjie Ji, Erhang Li, Fangyun Lin, Fucong Dai, and 80 others. 2024. Deepseek-v3 technical report. *CoRR*, abs/2412.19437.

Liam Dugan, Alyssa Hwang, Filip Trhlík, Andrew Zhu, Josh Magnus Ludan, Hainiu Xu, Daphne Ippolito, and Chris Callison-Burch. 2024. RAID: A shared benchmark for robust evaluation of machine-generated text detectors. In *Proceedings of the 62nd Annual Meeting of the Association for Computational Linguistics (Volume 1: Long Papers), ACL 2024, Bangkok, Thailand, August 11-16, 2024*, pages 12463–12492. Association for Computational Linguistics.

Lester Faigley and Stephen Witte. 1981. Analyzing revision. *College Composition & Communication*, 32(4):400–414.

Rudolph Flesch. 1948. A new readability yardstick. *Journal of applied psychology*, 32(3):221.

Sebastian Gehrmann, Hendrik Strobelt, and Alexander M. Rush. 2019. GLTR: statistical detection and visualization of generated text. In *Proceedings of the 57th Conference of the Association for Computational Linguistics, ACL 2019, Florence, Italy, July 28 - August 2, 2019, Volume 3: System Demonstrations*, pages 111–116. Association for Computational Linguistics.

Biyang Guo, Xin Zhang, Ziyuan Wang, Minqi Jiang, Jinran Nie, Yuxuan Ding, Jianwei Yue, and Yupeng Wu. 2023. How close is chatgpt to human experts? comparison corpus, evaluation, and detection. *CoRR*, abs/2301.07597.

Hanxi Guo, Siyuan Cheng, Xiaolong Jin, Zhuo Zhang, Kaiyuan Zhang, Guanhong Tao, Guangyu Shen, and Xiangyu Zhang. 2024. Biscope: Ai-generated text detection by checking memorization of preceding tokens. In *Advances in Neural Information Processing Systems 38: Annual Conference on Neural Information Processing Systems 2024, NeurIPS 2024, Vancouver, BC, Canada, December 10 - 15, 2024*.

Abhimanyu Hans, Avi Schwarzschild, Valeriia Cherepanova, Hamid Kazemi, Aniruddha Saha, Micah Goldblum, Jonas Geiping, and Tom Goldstein. 2024. Spotting llms with binoculars: Zero-shot detection of machine-generated text. In *Forty-first International Conference on Machine Learning, ICML 2024, Vienna, Austria, July 21-27, 2024*. OpenReview.net.

Pengcheng He, Xiaodong Liu, Jianfeng Gao, and Weizhu Chen. 2021. Deberta: decoding-enhanced bert with disentangled attention. In *9th International Conference on Learning Representations, ICLR 2021, Virtual Event, Austria, May 3-7, 2021*. OpenReview.net.

Xinlei He, Xinyue Shen, Zeyuan Chen, Michael Backes, and Yang Zhang. 2024. Mgtbench: Benchmarking machine-generated text detection. In *Proceedings of the 2024 on ACM SIGSAC Conference on Computer and Communications Security, CCS 2024, Salt Lake City, UT, USA, October 14-18, 2024*, pages 2251–2265. ACM.

Xinke Jiang, Yue Fang, Rihong Qiu, Haoyu Zhang, Yongxin Xu, Hao Chen, Wentao Zhang, Ruizhe Zhang, Yuchen Fang, Xu Chu, and 1 others. 2025. Tc-rag: Turing-complete rag's case study on medical llm systems. *ACL 2025*.

Kalpesh Krishna, Yixiao Song, Marzena Karpinska, John Wieting, and Mohit Iyyer. 2023. Paraphrasing evades detectors of ai-generated text, but retrieval is an effective defense. In *Advances in Neural Information Processing Systems 36: Annual Conference on Neural Information Processing Systems 2023, NeurIPS 2023, New Orleans, LA, USA, December 10 - 16, 2023*.

Tatsuki Kuribayashi, Takumi Ito, Jun Suzuki, and Kentaro Inui. 2020. Language models as an alternative evaluator of word order hypotheses: A case study in japanese. In *Proceedings of the 58th Annual Meeting of the Association for Computational Linguistics, ACL 2020, Online, July 5-10, 2020*, pages 488–504. Association for Computational Linguistics.

Jooyoung Lee, Thai Le, Jinghui Chen, and Dongwon Lee. 2023. Do language models plagiarize? In *Proceedings of the ACM Web Conference 2023, WWW 2023, Austin, TX, USA, 30 April 2023 - 4 May 2023*, pages 3637–3647. ACM.

Yafu Li, Qintong Li, Leyang Cui, Wei Bi, Zhilin Wang, Longyue Wang, Linyi Yang, Shuming Shi, and Yue Zhang. 2024. MAGE: machine-generated text detection in the wild. In *Proceedings of the 62nd Annual Meeting of the Association for Computational Linguistics (Volume 1: Long Papers), ACL 2024, Bangkok, Thailand, August 11-16, 2024*, pages 36–53. Association for Computational Linguistics.

Yuqing Liang, Jiancheng Xiao, Wensheng Gan, and Philip S Yu. 2026. Watermarking techniques for large language models: A survey. *Artificial Intelligence Review*, 59(2):74.

Yinhan Liu, Myle Ott, Naman Goyal, Jingfei Du, Mandar Joshi, Danqi Chen, Omer Levy, Mike Lewis, Luke Zettlemoyer, and Veselin Stoyanov. 2019. Roberta: A robustly optimized BERT pretraining approach. *CoRR*, abs/1907.11692.

Dominik Macko, Róbert Móro, Adaku Uchendu, Jason Samuel Lucas, Michiharu Yamashita, Matús Pikuliak, Ivan Srba, Thai Le, Dongwon Lee, Jakub Simko, and Mária Bieliková. 2023. Multitude: Large-scale multilingual machine-generated text detection benchmark. In *Proceedings of the 2023 Conference on Empirical Methods in Natural Language Processing, EMNLP 2023, Singapore, December 6-10, 2023*, pages 9960–9987. Association for Computational Linguistics.

OpenAI. 2023. GPT-4 technical report. *CoRR*, abs/2303.08774.

OpenRouter. 2025. AI Model Rankings.

Gerrit Quaremba, Elizabeth Black, Denny Vrandecic, and Elena Simperl. 2025. Wetbench: A benchmark for detecting task-specific machine-generated text on wikipedia. *CoRR*, abs/2507.03373.

Rico Sennrich, Barry Haddow, and Alexandra Birch. 2016. Improving neural machine translation models with monolingual data. In *Proceedings of the 54th Annual Meeting of the Association for Computational Linguistics, ACL 2016, August 7-12, 2016, Berlin, Germany, Volume 1: Long Papers*. The Association for Computer Linguistics.

Irene Solaiman, Miles Brundage, Jack Clark, Amanda Askell, Ariel Herbert-Voss, Jeff Wu, Alec Radford, and Jasmine Wang. 2019. Release strategies and the social impacts of language models. *CoRR*, abs/1908.09203.

Jinyan Su, Terry Yue Zhuo, Di Wang, and Preslav Nakov. 2023. Detectllm: Leveraging log rank information for zero-shot detection of machine-generated text. In *Findings of the Association for Computational Linguistics: EMNLP 2023, Singapore, December 6-10, 2023*, pages 12395–12412. Association for Computational Linguistics.

Adaku Uchendu, Zeyu Ma, Thai Le, Rui Zhang, and Dongwon Lee. 2021. TURINGBENCH: A benchmark environment for turing test in the age of neural text generation. In *Findings of the Association for Computational Linguistics: EMNLP 2021, Virtual Event / Punta Cana, Dominican Republic, 16-20 November, 2021*, pages 2001–2016. Association for Computational Linguistics.

Saranya Venkatraman, Nafis Irtiza Tripto, and Dongwon Lee. 2025. Collabstory: Multi-llm collaborative story generation and authorship analysis. In *Findings of the Association for Computational Linguistics: NAACL 2025, Albuquerque, New Mexico, USA, April 29 - May 4, 2025*, Findings of ACL, pages 3665–3679. Association for Computational Linguistics.

Yichen Wang, Shangbin Feng, Abe Bohan Hou, Xiao Pu, Chao Shen, Xiaoming Liu, Yulia Tsvetkov, and Tianxing He. 2024a. Stumbling blocks: Stress testing the robustness of machine-generated text detectors under attacks. In *Proceedings of the 62nd Annual Meeting of the Association for Computational Linguistics (Volume 1: Long Papers), ACL 2024, Bangkok, Thailand, August 11-16, 2024*, pages 2894–2925. Association for Computational Linguistics.

Yuxia Wang, Jonibek Mansurov, Petar Ivanov, Jinyan Su, Artem Shelmanov, Akim Tsvigun, Chenxi Whitehouse, Osama Mohammed Afzal, Tarek Mahmoud, Toru Sasaki, Thomas Arnold, Alham Fikri Aji, Nizar Habash, Iryna Gurevych, and Preslav Nakov. 2024b. M4: multi-generator, multi-domain, and multi-lingual black-box machine-generated text detection. In *Proceedings of the 18th Conference of the European Chapter of the Association for Computational Linguistics, EACL 2024 - Volume 1: Long Papers, St. Julian's, Malta, March 17-22, 2024*, pages 1369–1407. Association for Computational Linguistics.

Jason W. Wei and Kai Zou. 2019. EDA: easy data augmentation techniques for boosting performance on text classification tasks. In *Proceedings of the 2019 Conference on Empirical Methods in Natural Language Processing and the 9th International Joint Conference on Natural Language Processing, EMNLP-IJCNLP 2019, Hong Kong, China, November 3-7, 2019*, pages 6381–6387. Association for Computational Linguistics.

Junchao Wu, Shu Yang, Runzhe Zhan, Yulin Yuan, Lidia S. Chao, and Derek Fai Wong. 2025a. A survey on llm-generated text detection: Necessity, methods, and future directions. *Comput. Linguistics*, 51(1):275–338.

Junchao Wu, Runzhe Zhan, Qianli Wang, Yulin Yuan, Lidia S. Chao, and Derek F. Wong. 2025b. Overview of the NLPCC 2025 shared task 1: Llm-generated text detection. In *Natural Language Processing and Chinese Computing - 14th National CCF Conference, NLPCC 2025, Urumqi, China, August 7-9, 2025, Proceedings, Part IV*, Lecture Notes in Computer Science, pages 263–274. Springer.

Junchao Wu, Runzhe Zhan, Derek F. Wong, Shu Yang, Xuebo Liu, Lidia S. Chao, and Min Zhang. 2025c. Who wrote this? the key to zero-shot llm-generated text detection is gecscore. In *Proceedings of the 31st International Conference on Computational Linguistics, COLING 2025, Abu Dhabi, UAE, January 19-24, 2025*, pages 10275–10292. Association for Computational Linguistics.

Junchao Wu, Runzhe Zhan, Derek F. Wong, Shu Yang, Xinyi Yang, Yulin Yuan, and Lidia S. Chao. 2024. Detectrl: Benchmarking llm-generated text detection in real-world scenarios. In *Advances in Neural Information Processing Systems 38: Annual Conference on Neural Information Processing Systems 2024, NeurIPS 2024, Vancouver, BC, Canada, December 10 - 15, 2024*.

Yihuai Xu, Yongwei Wang, Yifei Bi, Huangsen Cao, Zhouhan Lin, Yu Zhao, and Fei Wu. 2025. Training-free llm-generated text detection by mining token probability sequences. In *The Thirteenth International Conference on Learning Representations, ICLR 2025, Singapore, April 24-28, 2025*. OpenReview.net.

An Yang, Baosong Yang, Beichen Zhang, Binyuan Hui, Bo Zheng, Bowen Yu, Chengyuan Li, Dayiheng Liu, Fei Huang, Haoran Wei, Huan Lin, Jian Yang, Jianhong Tu, Jianwei Zhang, Jianxin Yang, Jiaxi Yang, Jingren Zhou, Junyang Lin, Kai Dang, and 22 others. 2024. Qwen2.5 technical report. *CoRR*, abs/2412.15115.

Shu Yang, Jingyu Hu, Tong Li, Hanqi Yan, Wenxuan Wang, and Di Wang. 2026. Automonitor-bench: Evaluating the reliability of llm-based misbehavior monitor. *CoRR*, abs/2601.05752.

Jiayi Zhang, Shu Yang, Junchao Wu, Derek F. Wong, and Di Wang. 2025. Understanding and mitigating political stance cross-topic generalization in large language models. *Preprint*, arXiv:2508.02360.

Qihui Zhang, Chujie Gao, Dongping Chen, Yue Huang, Yixin Huang, Zhenyang Sun, Shilin Zhang, Weiye Li, Zhengyan Fu, Yao Wan, and Lichao Sun. 2024. Llm-as-a-coauthor: Can mixed human-written and machine-generated text be detected? In *Findings of the Association for Computational Linguistics: NAACL 2024, Mexico City, Mexico, June 16-21, 2024*, Findings of ACL, pages 409–436. Association for Computational Linguistics.

Ruizhe Zhang, Xinke Jiang, Zhibang Yang, Zhixin Zhang, Jiaran Gao, Yuzhen Xiao, Hongbin Lai, Xu Chu, Junfeng Zhao, and Yasha Wang. 2026. Stackplanner: A centralized hierarchical multi-agent system with task-experience memory management. *ACL 2026*.

Yue Zhang, Yafu Li, Leyang Cui, Deng Cai, Lemao Liu, Tingchen Fu, Xinting Huang, Enbo Zhao, Yu Zhang, Yulong Chen, Longyue Wang, Anh Tuan Luu, Wei Bi, Freda Shi, and Shuming Shi. 2023. Siren's song in the AI ocean: A survey on hallucination in large language models. *CoRR*, abs/2309.01219.

Biru Zhu, Lifan Yuan, Ganqu Cui, Yangyi Chen, Chong Fu, Bingxiang He, Yangdong Deng, Zhiyuan Liu, Maosong Sun, and Ming Gu. 2023. Beat llms at their own game: Zero-shot llm-generated text detection via querying chatgpt. In *Proceedings of the 2023 Conference on Empirical Methods in Natural Language Processing, EMNLP 2023, Singapore, December 6-10, 2023*, pages 7470–7483. Association for Computational Linguistics.

# A The Use of LLMs

We used LLMs responsibly and within specific limits for this work. Experimentally, LLMs were used to generate data for training and evaluation. For the writing process, they served solely to assist with grammar checking and linguistic refinement. It is important to clarify that the research design and the original draft were written entirely by humans.

# B Data collection

## B.1 Language Complexity Classification

To assess how linguistic differences influence model representation and detection behavior (Arnett and Bergen, 2025; Kuribayashi et al., 2020), We categorize the eight examined languages into three levels of complexity: *high*, *medium*, and *low*, based on their morphological richness and typological distance from English .

### Classification Criteria

- **Morphological Richness:** The degree of inflectional and derivational variation. Highly morphological languages (eg, Arabic, Russian) increase token diversity and complicate subword segmentation, posing challenges for detector robustness compared to analytic languages.

- **Typological Distance from English:** This metric accounts for differences in syntax (word order), script, and language family. Languages more distant from the English-centric training data of most LLMs often exhibit distinct representation patterns in latent space.

### Language Grouping

- **High-complexity languages:** *Arabic, Russian, Chinese.* These languages exhibit extensive morphology (e.g., non-concatenative morphological systems in Arabic), flexible or non-linear syntax (e.g., free word order in Russian), and unique writing systems (e.g., logographic characters in Chinese). Such characteristics increase representation difficulty and tokenization cost.

- **Medium-complexity languages:** *German, French, Spanish, Portuguese.* These languages display moderate inflectional variation and relatively regular syntactic structures.

Morphological agreement and verb conjugation create some modeling challenges but generally follow predictable grammatical patterns.

- **Low-complexity language:** *English.* English features simplified morphology, fixed word order, and high encoding efficiency, imposing minimal modeling overhead and serving as a baseline for cross-lingual comparison.

**Relevance to Detection Tasks** This classification provides a structured basis for analyzing how linguistic variability affects model performance and cross-lingual transferability. High-complexity languages generally pose greater challenges for tokenization and feature extraction, whereas low-complexity languages often yield more stable detection and stronger transfer performance.

### B.2 Human-Written Data Resources

We collected diverse human-authored texts from multiple representative and discriminative LLM application scenarios to construct a benchmark corpus that encompasses a wide range of linguistic features and writing styles, aiming to comprehensively evaluate detectors' ability to identify authentic human creations with different writing styles and content characteristics.

The selected datasets cover six major writing domains: academic writing (Academic), news reporting (News), novel creation (Novel), search engine optimization texts (SEO), encyclopedia entries (Wiki), and web texts (WebText). This diversified data selection strategy holds significant importance for AI detection tasks: academic texts exhibit rigorous logical structures and domain-specific knowledge expression; news reporting reflects objective and standardized journalistic writing styles; novel works present creative narrative techniques and rich emotional expression; SEO texts demonstrate specific writing techniques optimized for search engine algorithms, including keyword placement, meta description optimization, and other technical writing features; encyclopedia entries embody accurate and concise knowledge-based writing characteristics; web texts represent the diverse language usage patterns in daily communication.

By integrating these different types of textual resources, we constructed a comprehensive benchmark corpus that covers various linguistic features and writing styles, providing a solid data foundation for training robust LLM-generated text detectors. The specific statistical information, collection methods, and technical details of each dataset are as follows:

**Academic Writing** We collected academic papers from the Directory of Open Access Journals (DOAJ)[2], a community-curated directory containing high-quality, peer-reviewed open access journals across all disciplines. We extracted and processed PDF documents to obtain clean textual content.

**News** News articles were primarily sourced from reputable international news media outlets, including BBC[3] and other major news organizations, ensuring professional journalistic standards and Operationorial quality across multiple languages.

**Novel** Creative fiction texts were extracted from Common Crawl,[4] a publicly available web archive containing crawled web data. We identified and extracted novel excerpts, short stories, and other creative writing content.

**SEO** Search engine optimization content was obtained from two sources: publicly available web content optimized for search engines, and proprietary high-quality SEO materials from the authors' affiliated organization.

**Wiki** Encyclopedia entries were collected from Wikipedia,[5] ensuring balanced representation across various topics and languages while maintaining Wikipedia's characteristic encyclopedic writing style.

**WebText** General web content was also sourced from Common Crawl, representing diverse writing styles found in blogs, forums, and other online platforms.

All human-written data were collected from sources published before 2022 years to avoid potential contamination from LLM-generated content. We implemented a systematic three-stage filtering pipeline to ensure data quality. First, we applied language identification to filter out texts not matching our target eight languages, ensuring all collected samples conform to our designated language set. Second, we performed length-based filtering to remove excessively short and long texts, retaining only samples within a practical and usable length range. Third, we computed the perplexity (PPL) for each text using pre-trained language models and filtered out samples with excessively high PPL values as well as a small portion of samples with extremely low PPL values, which may indicate anomalous or formulaic content.

[2] https://doaj.org
[3] https://www.bbc.com
[4] https://commoncrawl.org
[5] https://www.wikipedia.org

### B.3 Generators Used and Generation Settings

We employed four state-of-the-art language models as text generators: GPT-4o, Gemini, DeepSeek-V3, and Qwen-Max. These models represent leading commercial and open-source LLMs with diverse architectural designs and training methodologies, reflecting the full spectrum of synthetic text characteristics that detection systems might encounter in real-world scenarios, while also being the most commonly selected models in people's daily usage. This section details the selected generator models, their architectural characteristics, and the specific generation parameters and engineering strategies used to produce synthetic texts for benchmark evaluation. The models we sampled are shown in Table 3.

| Generator | API Service | Version |
|---|---|---|
| GPT-4o | OpenAI | gpt-4o-2024-11-20 |
| Gemini-2.5 | Google | gemini-2.5-flash |
| DeepSeek-V3 | DeepSeek | deepseek-chat |
| Qwen-Max | Alibaba | qwen-max-2025-01-25 |

Table 3: Details of the generative models that is used to produce LLM-generated text.

Throughout all text generation tasks, we set the temperature parameter to 1 to promote more diverse and rich text generation. For different domain data, we designed distinct prompts to guide model synthesis, with each language having its own version to better activate the model's linguistic capabilities. To construct more challenging detection texts that closely resemble human daily usage scenarios, we employed a two-stage "summarize-generate" framework. Specifically, we first invoked the LLM to summarize human-written texts into a single sentence, then prompted the LLM to regenerate content based on this summary, thereby significantly enhancing the prior probability of consistency with human sample distributions. Our results reveal that this data construction method presents significant challenges to existing detectors. The following are the "summarize-generate" instructions adopted for different domain texts:

**Prompt for Summarizing in English**

**System:** You are an expert in creating accurate and concise summaries in English. Your task is to create summaries that capture the main points and essence of the original text, while maintaining clear and coherent style.
**User:** Please create a concise summary of the following text in English. The summary should: - Capture the most important points and significant events - Preserve narrative coherence if it's a story - Be considerably shorter than the original text - Use clear and fluent language
Text to summarize: {original_text}
Summary:

**Prompt for Summarizing in Arabic**

**System:** أنت خبير في إنشاء ملخصات دقيقة ومختصرة باللغة العربية. مهمتك هي إنشاء ملخصات تلتقط النقاط الرئيسية وجوهر النص الأصلي، مع الحفاظ على أسلوب واضح ومتماسك.
**User:** يرجى إنشاء ملخص مختصر للنص التالي باللغة العربية. يجب أن يشمل الملخص: - التقاط أهم النقاط والأحداث الأكثر أهمية - الحفاظ على التماسك السردي إذا كان النص قصة - أن يكون أقصر بكثير من النص الأصلي - استخدام لغة واضحة ومتدفقة
النص المراد تلخيصه:{original_text}
الملخص:

**Prompt for Summarizing in Chinese**

**System:** 你是一个专业的中文文本摘要专家。 你的任务是创建准确、简洁的中文摘要，能够捕捉原文的主要观点和核心内容， 同时保持清晰连贯的表达风格。
**User:** 请为以下文本创建一个简洁的中文摘要。摘要要求： - 准确捕捉最重要的观点和最关键的事件 - 如果是故事类文本，请保持叙述的连贯性 - 长度要明显短于原文 - 使用清晰流畅的中文表达
需要摘要的文本：{original_text}
摘要：

**Prompt for Summarizing in French**

**System:** Vous êtes un expert dans la création de résumés précis et concis en français. Votre tâche est de créer des résumés qui capturent les points principaux et l'essence du texte original, tout en maintenant un style clair et cohérent.
**User:** Veuillez créer un résumé concis du texte suivant en français. Le résumé doit : - Capturer les points les plus importants et les événements les plus significatifs - Préserver la cohérence narrative s'il s'agit d'une histoire - Être considérablement plus court que le texte original - Utiliser un langage clair et fluide
Texte à résumer :{original_text}
Résumé :

| Language | News | Webtext | Wiki | Novel | SEO | Academic |
|---|---|---|---|---|---|---|
| **English** | news article | web content | Wikipedia article | narrative or fiction text | SEO-optimized content | academic writing |
| 中文 **(Chinese)** | 新闻报道 | 网页内容 | 百科条目 | 小说或叙事文本 | 搜索引擎优化内容 | 学术写作 |
| **العربية (Arabic)** | مقال إخباري | محتوى ويب | مقال ويكيبيديا | نص سردي أو خيالي | محتوى محسن لمحركات البحث | كتابة أكاديمية |
| **Français (French)** | article de presse | contenu web | article Wikipédia | texte narratif ou de fiction | contenu optimisé pour le référencement | rédaction académique |
| **Deutsch (German)** | Nachricht-enartikel | Web-Inhalt | Wikipedia-Artikel | Erzähltext oder Fiktion | SEO-optimierter Inhalt | wis-senschaftliches Schreiben |
| **Português (Portuguese)** | artigo de notícias | conteúdo web | artigo da Wikipédia | texto narrativo ou ficção | conteúdo otimizado para SEO | escrita acadêmica |
| **Русский (Russian)** | новост-ная статья | веб-контент | статья Википе-дии | Художествен-ный текст | SEO-контент | научная статья |
| **Español (Spanish)** | artículo de noticias | contenido web | artículo de Wikipedia | texto narrativo o de ficción | contenido optimizado para SEO | Escritura académica |

Table 4: Text Category Descriptions.

## Prompt for Summarizing in German

**System:** Sie sind ein Experte für die Erstellung präziser und prägnanter Zusammenfassungen auf Deutsch. Ihre Aufgabe ist es, Zusammenfassungen zu erstellen, die die Hauptpunkte und das Wesentliche des ursprünglichen Textes erfassen und dabei einen klaren und kohärenten Stil beibehalten.
**User:** Bitte erstellen Sie eine prägnante Zusammenfassung des folgenden Textes auf Deutsch. Die Zusammenfassung sollte: - Die wichtigsten Punkte und bedeutendsten Ereignisse erfassen - Die narrative Kohärenz bewahren, falls es sich um eine Geschichte handelt - Deutlich kürzer als der ursprüngliche Text sein - Eine klare und fließende Sprache verwenden
Zu zusammenfassender Text:{original_text}
Zusammenfassung:

## Prompt for Summarizing in Portuguese

**System:** Você é um especialista na criação de resumos precisos e concisos em português. Sua tarefa é criar resumos que capturem os pontos principais e a essência do texto original, mantendo um estilo claro e coerente.
**User:** Por favor, crie um resumo conciso do seguinte texto em português. O resumo deve: - Capturar os pontos mais importantes e os eventos mais significativos - Preservar a coerência narrativa se for uma história - Ser consideravelmente mais curto que o texto original - Usar linguagem clara e fluida
Texto para resumir:{original_text}
Resumo:

## Prompt for Summarizing in Russian

**System:** Вы эксперт по созданию точных и кратких резюме на русском языке. Ваша задача - создавать резюме, которые отражают основные моменты и суть исходного текста, сохраняя при этом ясный и последовательный стиль.
**User:** Пожалуйста, создайте краткое резюме следующего текста на русском языке. Резюме должно: - Отражать наиболее важные моменты и значимые события - Сохранять повествовательную связность, если это история - Быть значительно короче исходного текста - Использовать ясный и плавный язык
Текст для реферирования:{original_text}
Резюме:

## Prompt for Summarizing in Spanish

**System:** Eres un experto en crear resúmenes concisos y precisos en español. Tu tarea es generar resúmenes que capturen los puntos principales y la esencia del texto original, manteniendo un estilo claro y coherente.
**User:** Por favor, crea un resumen conciso del siguiente texto en español. El resumen debe: - Capturar los puntos principales y eventos más importantes - Mantener la coherencia narrativa si es una historia - Ser significativamente más breve que el texto original - Usar un lenguaje claro y fluido
Texto a resumir:{original_text}
Resumen:

### Prompt for Generating in English

**System:** You are an expert English writer with great creativity. Your task is to compose original texts based on a summary, with the freedom to develop and enrich ideas. You should aim for a target length and use your own writing style, always ensuring impeccable grammar.
**User:** Create an original text based on the following summary. The resulting text should be a {category}. You have total freedom to develop and enrich the presented ideas, without strictly limiting yourself to the explicit information in the summary. The goal is to achieve an an approximate length of {target_length} words, using your personal writing style and always maintaining impeccable grammar.
Summary:{summary}
Final Text:

### Prompt for Generating in Arabic

**System:** أنت كاتب خبير باللغة العربية يتمتع بإبداع كبير. مهمتك هي تأليف نصوص أصلية من خلاص التلخيص، مع حرية تطوير وإثراء الأفكار. يجب أن تهدف إلى طول مستهدف وأن تستخدم أسلوبك الخاص في الكتابة، مع ضمان قواعد نحوية لا تشوبها شائبة دائماً.
**User:** اكتب نصاً أصلياً بناءً على التلخيص التالي. يجب أن يكون النص الناتج {category}. لك الحرية الكاملة في تطوير وإثراء الأفكار المقدمة، دون أن تقتصر بشكل صارم على المعلومات الصريحة في التلخيص.
الهدف هو الوصول إلى طول تقريبي يبلغ {target_length} كلمة، باستخدام أسلوبك الشخصي في الكتابة مع الحفاظ على قواعد نحوية مثالية دائماً.
التلخيص:{summary}
النص النهائي:

### Prompt for Generating in Chinese

**System:** 你是一位富有创造力的中文写作专家。你的任务是根据摘要创作原创文本，可以自由发挥和丰富想法。你需要达到指定的目标长度，运用自己独特的写作风格，并始终保持完美的语法和表达。
**User:** 请根据以下摘要创作一篇原创文本。生成的文本应该是{category}类型。你可以完全自由地发展和丰富摘要中的想法，不必严格限制在摘要的明确信息范围内。 目标是达到大约{target_length}个字符的长度，使用你个人的写作风格，始终保持完美的语法和表达。
摘要：{summary}
最终文本：

### Prompt for Generating in French

**System:** Vous êtes un écrivain expert en français doté d'une grande créativité. Votre tâche est de composer des textes originaux à partir d'un résumé, avec la liberté de développer et d'enrichir les idées. Vous devez viser une longueur cible et utiliser votre propre style d'écriture, en garantissant toujours une grammaire impeccable.
**User:** Créez un texte original à partir du résumé suivant. Le texte résultant doit être un {category}. Vous avez une totale liberté pour développer et enrichir les idées présentées, sans vous limiter strictement aux informations explicites du résumé. L'objectif est d'atteindre une longueur approximative de {target_length} mots, en utilisant votre style d'écriture personnel et en maintenant toujours une grammaire impeccable.
Résumé:{summary}
Texte final:

### Prompt for Generating in German

**System:** Sie sind ein erfahrener deutscher Schriftsteller mit großer Kreativität. Ihre Aufgabe ist es, originelle Texte auf der Grundlage einer Zusammenfassung zu verfassen, mit der Freiheit, Ideen zu entwickeln und zu bereichern. Sie sollten eine Ziellänge anstreben und Ihren eigenen Schreibstil verwenden, wobei Sie stets eine einwandfreie Grammatik gewährleisten müssen.
**User:** Erstellen Sie einen originellen Text basierend auf der folgenden Zusammenfassung. Der resultierende Text sollte ein {category} sein. Sie haben völlige Freiheit, die dargestellten Ideen zu entwickeln und zu bereichern, ohne sich streng auf die expliziten Informationen der Zusammenfassung zu beschränken. Das Ziel ist es, eine ungefähre Länge von {target_length} Wörtern zu erreichen, wobei Sie Ihren persönlichen Schreibstil verwenden und stets eine einwandfreie Grammatik beibehalten.
Zusammenfassung:{summary}
Endgültiger Text:

### Prompt for Generating in Portuguese

**System:** Você é um escritor experiente em português com grande criatividade. Sua tarefa é compor textos originais a partir de um resumo, com a liberdade de desenvolver e enriquecer as ideias. Você deve almejar um comprimento alvo e empregar seu próprio estilo de escrita, garantindo sempre uma gramática impecável.
**User:** Crie um texto original a partir do seguinte resumo. O texto resultante deve ser um {category}. Você tem total liberdade para desenvolver e enriquecer as ideias apresentadas, sem se limitar estritamente às informações explícitas do resumo. O objetivo é atingir um comprimento aproximado de {target_length} palavras, utilizando seu estilo pessoal de escrita e mantendo sempre uma gramática impecável.
Resumo:{summary}
Texto final:

### Prompt for Generating in Russian

**System:** Вы — опытный русскоязычный автор с большой креативностью. Ваша задача — создавать оригинальные тексты на основе резюме, имея свободу развивать и обогащать идеи. Вы должны стремиться к достижению целевой длины и использовать свой собственный стиль письма, всегда обеспечивая безупречную грамматику.
**User:** Создайте оригинальный текст на основе следующего резюме. Полученный текст должен быть {category}. У вас есть полная свобода развивать и обогащать представленные идеи, не ограничиваясь строго явной информацией в резюме. Цель — достичь приблизительно {target_length} слов** по длине, используя ваш личный стиль письма и поддерживая при этом идеальную грамматику.
Резюме:{summary}
Итоговый текст:

**Prompt for Generating in Spanish**

**System:** Eres un experto escritor en español con gran creatividad. Tu tarea es componer textos originales a partir de un resumen, con la libertad de desarrollar y enriquecer las ideas. Debes aspirar a una longitud objetivo y emplear tu propio estilo de escritura, garantizando siempre una gramática impecable.
**User:** Crea un texto original a partir del siguiente resumen. El texto resultante debe ser un {category}. Tienes total libertad para desarrollar y enriquecer las ideas presentadas, sin limitarte estrictamente a la información explícita del resumen. El objetivo es alcanzar una longitud aproximada de {target_length} palabras, utilizando tu estilo personal de escritura y manteniendo siempre una gramática impecable.
Resumen:{summary}
Texto final:

### B.4 Paraphrase Attack

Paraphrasing attack is a technique that systematically rewrites text content to generate semantically equivalent yet expressionally diverse textual variants, while preserving the core meaning. In this study, we have carefully designed and implemented four complementary paraphrasing strategies to construct a comprehensive, diverse, and multilingual attack framework, as detailed below:

**Encoder Paraphrsing** This approach employs advanced encoder architectures capable of intelligently identifying and masking key textual segments, thereby generating alternative content that is strictly semantically equivalent but significantly varied at the lexical level, while ensuring the linguistic fluency of the rewritten text. Specifically, building upon an Encoder-only model, we randomly mask 15% of tokens in the input samples and then resample to generate new tokens that are semantically consistent. To accommodate multiple language environments, we adopt the multilingual encoder mBERT[6] to perform the masked prediction task.

**Seq2seq Paraphrasing** Leveraging powerful sequence-to-sequence architectures, this technique enables fine-grained rewriting of complex structures and long passages, effectively simulating the diverse patterns found in human-authored text. To this end, we constructed a dedicated LGT-HWT pair dataset comprising 192k samples across 8 languages, 6 domains, and 4 model types, ensuring no overlap with existing benchmarks. On this basis, we fine-tuned the multilingual mT5[7] model using this corpus to build a high-quality paraphrase generator.

**Decoder Paraphrasing** Capitalizing on large-scale pretrained decoder-only language models, this approach demonstrates exceptional capability in creative paraphrasing, generating grammatically robust and semantically diverse variants. Adopting the same LGT-HWT dataset established in the Seq2seq Paraphrasing section, we applied it to Direct Preference Optimization (DPO). Specifically, we fine-tuned the multilingual Qwen-2.5-7B[8] model via DPO to construct an efficient paraphrasing system.

**Back-Translation** As a classic paraphrasing technique, Back-Translation generates semantically faithful yet expressively novel textual variants by exploiting the linguistic transformations inherent in a "source → target → source" bidirectional pipeline. In this study, we employed the multilingual mBART[9] model to perform this task. To ensure optimal generation quality, we selected English as the unified pivot language for all non-English inputs, while utilizing Chinese as the pivot for English text.

### B.5 Perturbation Attack

To systematically evaluate the security and generalization of detectors within complex, real-world globalized deployment scenarios, we propose a language-agnostic and universally applicable adversarial perturbation framework. This framework is designed to simulate pervasive perturbations that transcend specific languages and writing systems. To ensure consistency in evaluation, all strategies adhere to a unified "low-level intervention" principle: independent of specific linguistic grammars, we tokenize the input and randomly sample 15% of the words for physical-level modification. This design ensures seamless transferability across arbitrary linguistic scenarios while preserving the core semantic intent of the original text.

[6] https://huggingface.co/google-bert/bert-base-multilingual-cased

[7] https://huggingface.co/google/mt5-base

[8] https://huggingface.co/Qwen/Qwen-2.5-7B

[9] https://huggingface.co/facebook/mbart-large-50-many-to-many-mmt

**Character Insertion** This strategy involves systematically repeating the preceding character at arbitrary positions within selected words, simulating common "repetitive keystroke" errors found across various input devices. Leveraging the physical sequential nature of writing systems, this method is entirely independent of specific lexical or syntactic rules. Whether applied to Latin, Cyrillic, or other alphabetic systems, this technique disrupts sub-word morphological structures with a unified logic. Its inherent universality allows for direct application to multilingual datasets without the need for language-specific customization.

**Character Substitution** By exploiting the vast Unicode standard, this strategy replaces characters with visually identical (or highly similar) but uniquely encoded counterparts (e.g., substituting the Latin "a" with the Cyrillic "а"). We construct a dictionary of visually similar characters for random character substitution of selected words. This dictionary is derived from authoritative cross-textual obfuscation dictionaries, including Unicode Confusables,[10] Arabic Shaping,[11] and open-source homoglyph corpora.[12] It transcends single-language character set limitations, effectively implementing low-level encoding obfuscation on any Unicode-based text, including code-mixed data, thereby rigorously testing model robustness in multilingual environments.

**Character Deletion** This strategy randomly removes characters from selected words, mimicking spelling omissions that occur during rapid typing. Character omission is a fundamental noise form present in all writing systems. As a non-parametric attack method, it requires no external knowledge bases or dictionaries. By introducing pure structural entropy, it forces the model to rely on strong context recovery and error-correction capabilities across different languages, thus verifying the detector's generalization boundaries against low-quality, noisy multilingual inputs.

**Zero-width Insertion** We inject non-printing or zero-width Unicode control characters (including Zero Width Space `U+200B`, Zero Width Non-Joiner `U+200C`, Zero Width Joiner `U+200D`, and Word Joiner `U+2060`) into the text stream. These control characters are integral to the core Unicode standard and are supported by virtually all global text processing systems. Consequently, this attack effectively penetrates language barriers, remaining latent within digital text of any language. It exploits the dissociation between underlying encoding and surface rendering, creating a universal attack vector that is visually "invisible" to human readers of any language but significantly disrupts tokenizer logic at the machine processing layer.

### B.6 Various Text Length

To comprehensively evaluate the robustness and adaptability of our proposed detection framework in handling texts of varying lengths, we designed a systematic experimental study covering four representative input length categories: 64 tokens, 128 tokens, 256 tokens, and 512 tokens. This multi-dimensional experimental design reveals how detector performance varies with input scale and provides crucial insights into the framework's practical value across diverse application scenarios. It should be noted that we set the upper limit at 512 tokens, as this length covers the majority of real-world text processing requirements; for ultra-long texts exceeding this threshold, we can employ strategies such as sliding windows or semantic segmentation to divide them into multiple interconnected chunks for separate detection, which is more flexible and efficient in practical deployment. Specifically, the 64-token and 128-token settings simulate short-text scenarios (such as reviews and comments), while the 256-token and 512-token configurations align with long-text applications requiring deep comprehension (such as news reports).

In our experimental protocol, we adopted a fine-grained construction method based on sentence semantic units, leveraging spaCy toolkit[13] to achieve precise and natural text length control. The implementation process is as follows: first, we utilize spaCy's pre-trained sentence boundary detection component to perform accurate sentence segmentation on original documents; subsequently, we employ a progressive merging strategy, starting from the document beginning and sequentially merging adjacent sentences according to their natural order until the total token count reaches the level closest to the preset target length without exceeding this threshold. This methodology ensures strict variable control while maintaining the original text's in-herent discourse structure and semantic coherence, making the experimental results more persuasive and practically instructive.

[10]https://www.unicode.org/Public/security/latest/confusables.txt

[11]https://www.unicode.org/Public/UCD/latest/ucd/ArabicShaping.txt

[12]https://github.com/contr4l/SimilarCharacter

[13]`www.spacy.io`

### B.7 Various Operation Behavior

To better reflect real-world scenarios and capture the complexity of text evolution in human-LLM collaborative environments, we classify Operationing behaviors that can be applied to both HWT and LGT. For HWT, these Operationing operations represent human authors utilizing LLM capabilities to refine their original content, serving as a third category of samples ”human writing augmented by AI” (HLT). For LGT, these identical behaviors correspond to post-processing operations designed to explore how different generation workflows impact detector robustness. We categorize these behaviors into three types that reflect the primary taxonomy of human-LLM collaboration in real-world writing assistance scenarios (Faigley and Witte, 1981; Coenen et al., 2021):

- (1) **Polishing**, which involves restructuring or rephrasing text to improve fluency, correct grammatical issues, and better conform to specific domain styles or audience preferences;
- (2) **Expansing**, where additional details, examples, or elaborations are incorporated to make the content more comprehensive and substantial;
- (3) **Condensing**, which focuses on removing redundant, repetitive, or unnecessary information to achieve more concise and focused expression.

To construct diverse Operationing behavior examples and ensure high-quality, authentic data, we consistently use Qwen-Max as the Operationing assistant throughout the data construction process. This consistent approach ensures that all Operationed samples maintain comparable quality standards while preserving the unique characteristics of each Operationing type.

This unified framework for Operationing behaviors is crucial for understanding the full scope of text modification processes in modern writing workflows. Whether applied to originally human-created content or machine-generated text, these Operationing operations create distinct text patterns that pose unique challenges to more fine-grained detection systems. By systematically analyzing the performance of different categories of detectors across these various Operationing types, we can better understand the evolution of text characteristics throughout the entire human-LLM collaboration process and develop more robust and generalizable detectors.

The prompts we used for different Operationing behavior types are as follows:

#### B.7.1 Prompt for Polishing

**Prompt for Polishing in English**

**System:** You are an English writing expert. Your task is to rewrite the given text, maintaining exactly the same length and meaning, but using your own writing style.
**User:** Rewrite the following text in your own style, maintaining the same length. Do not modify words in ALL CAPS and avoid grammatical errors:
Input: {original_text}
Output:

**Prompt for Polishing in Arabic**

**System:** أنت كاتب خبير في اللغة العربية. مهمتك هي إعادة كتابة النصوص مع الحفاظ على نفس الطول والمعنى بالضبط، ولكن باستخدام أسلوبك الخاص في الكتابة.
**User:** أعد كتابة النص التالي بأسلوبك الخاص، مع الحفاظ على نفس الطول الأصلي. لا تغير الكلمات المكتوبة بالأحرف الكبيرة وتجنب الأخطاء النحوية:
{original_text}
المخرجات:

**Prompt for Polishing in Chinese**

**System:** 你是一位中文写作专家。你的任务是重写文本，保持完全相同的长度和含义，但使用你自己的写作风格。
**User:** 请用你自己的风格重写以下文本，保持与原文相同的长度。不要修改大写的词语，避免语法错误：
{original_text}
输出：

**Prompt for Polishing in French**

**System:** Vous êtes un expert écrivain français. Votre tâche est de réécrire des textes en conservant exactement la même longueur et le même sens, mais en utilisant votre propre style d’écriture.
**User:** Réécrivez le texte suivant dans votre propre style, en conservant la même longueur que l’original. Ne modifiez pas les mots en MAJUSCULES et évitez les erreurs grammaticales :
{original_text}
Sortie :

**Prompt for Polishing in German**

**System:** Sie sind ein erfahrener deutscher Schriftsteller. Ihre Aufgabe ist es, Texte umzuschreiben, dabei exakt die gleiche Länge und Bedeutung beizubehalten, aber Ihren eigenen Schreibstil zu verwenden.
**User:** Schreiben Sie den folgenden Text in Ihrem eigenen Stil um, behalten Sie dabei die gleiche Länge wie das Original bei. Ändern Sie keine Wörter in GROSSBUCHSTABEN und vermeiden Sie Grammatikfehler:
{original_text}
Ausgabe:

**Prompt for Polishing in Portuguese**

**System:** Você é um escritor especialista em português. Sua tarefa é reescrever textos mantendo exatamente o mesmo comprimento e significado, mas usando seu próprio estilo de escrita.
**User:** Reescreva o seguinte texto no seu próprio estilo, mantendo o mesmo comprimento do original. Não modifique as palavras em MAIÚSCULAS e evite erros gramaticais:
{original_text}
Saída:

**Prompt for Polishing in Russian**

**System:** Вы являетесь экспертом-писателем русского языка. Ваша задача - переписывать тексты, сохраняя точно такую же длину и смысл, но используя свой собственный стиль письма.
**User:** Перепишите следующий текст в своем собственном стиле, сохранив ту же длину, что и оригинал. Не изменяйте слова, написанные ЗАГЛАВНЫМИ БУКВАМИ, и избегайте грамматических ошибок:
{original_text}
Вывод:

**Prompt for Polishing in Spanish**

**System:** Eres un experto escritor en español. Tu tarea es reescribir textos manteniendo exactamente la misma longitud y significado, pero usando tu propio estilo de escritura.
**User:** Reescribe el siguiente texto en tu propio estilo, manteniendo la misma longitud original. No modifiques las palabras en MAYÚSCULAS y evita errores gramaticales:
{original_text}
Salida:

### B.7.2 Prompt for Expanding

**Prompt for Expanding in English**

**System:** You are an expert in English writing. Your mission is to expand upon a given text, making it richer and more detailed. While preserving the core meaning of the original content, you should employ your own distinctive and engaging writing style.
**User:** Please expand the following text in your own style to make it richer and more detailed. Do not modify any capitalized words, and ensure the output is free of grammatical errors:
{original_text}
Output:

**Prompt for Expanding in Arabic**

**System:** أنت خبير في الكتابة باللغة العربية. مهمتك هي التوسع في النص وإثراؤه، مع الحفاظ على جوهر المعنى الأصلي، وتقديم محتوى أكثر تفصيلاً وغنى، مع استخدام أسلوبك الكتابي الخاص والمميز.
**User:** يرجى توسيع النص التالي بأسلوبك الخاص لجعله أكثر ثراءً وتفصيلاً. لا تقم بتعديل الكلمات المكتوبة بأحرف لاتينية كبيرة ، وتجنب الأخطاء النحوية:
{original_text}
المخرجات:

**Prompt for Expanding in Chinese**

**System:** 你是一位中文写作专家。你的任务是对文本进行扩写，在保留原文核心语义的基础上，使其内容更加丰富详细，并使用你自己的写作风格。
**User:** 请用你自己的风格扩写以下文本，使其内容更加丰富详细。不要修改大写的词语，避免语法错误：
{original_text}
输出：

**Prompt for Expanding in French**

**System:** Vous êtes un expert en rédaction française. Votre mission est de développer un texte donné pour l'enrichir et le détailler. Tout en préservant le sens fondamental du contenu original, vous devez utiliser votre propre style d'écriture, unique et captivant.
**User:** Veuillez développer le texte suivant avec votre propre style pour le rendre plus riche et plus détaillé. Ne modifiez pas les sigles ou les mots écrits entièrement en majuscules, et assurez-vous que le résultat est exempt d'erreurs grammaticales :
{original_text}
Sortie :

**Prompt for Expanding in German**

**System:** Sie sind ein Experte für das Schreiben auf Deutsch. Ihre Mission ist es, einen vorgegebenen Text zu erweitern, um ihn reichhaltiger und detaillierter zu machen. Behalten Sie dabei die Kernaussage des Originaltextes bei und verwenden Sie Ihren eigenen, unverkennbaren Schreibstil.
**User:** Bitte erweitern Sie den folgenden Text in Ihrem eigenen Stil, um ihn reichhaltiger und detaillierter zu gestalten. Verändern Sie keine Wörter oder Abkürzungen, die vollständig in Großbuchstaben geschrieben sind, und vermeiden Sie grammatikalische Fehler:
{original_text}
Ausgabe:

### Prompt for Expanding in Portuguese

**System:** Você é um especialista em redação em língua portuguesa. Sua missão é expandir um determinado texto para enriquecê-lo e detalhá-lo. Preservando o núcleo do significado original, você deve utilizar seu próprio estilo de escrita, que seja único e cativante.
**User:** Por favor, expanda o texto a seguir com seu próprio estilo para torná-lo mais rico e detalhado. Não modifique siglas ou palavras escritas inteiramente em letras maiúsculas e garanta que o resultado não contenha erros gramaticais:
{original_text}
Saída:

### Prompt for Expanding in Russian

**System:** Вы — эксперт по написанию текстов на русском языке. Ваша задача — расширить и дополнить предложенный текст, сделав его более насыщенным и подробным. При этом необходимо сохранить основную суть оригинального содержания и использовать свой собственный, уникальный авторский стиль.
**User:** Пожалуйста, расширьте следующий текст в своём собственном стиле, чтобы сделать его более насыщенным и подробным. Не изменяйте аббревиатуры или слова, написанные полностью заглавными буквами, и избегайте грамматических ошибок:

{original_text}

Вывод:

### Prompt for Expanding in Spanish

**System:** Usted es un experto en redacción en español. Su misión es desarrollar un texto dado para enriquecerlo y detallarlo. Preservando el sentido fundamental del contenido original, debe emplear su propio estilo de redacción, que sea único y atractivo.
**User:** Por favor, desarrolle el siguiente texto con su propio estilo para hacerlo más rico y detallado. No modifique las siglas o palabras escritas completamente en mayúsculas y asegúrese de que el resultado no contenga errores gramaticales:
{original_text}
Salida:

### B.7.3 Prompt for Condensing

### Prompt for Condensing in English

**System:** YYou are an expert English writer. Your task is to condense a text, preserving its core meaning by making the content more concise and eliminating redundancy. You should use your own writing style.
**User:** Please condense the following text in your own style, making it more concise and eliminating redundancy. Do not modify capitalized words and avoid grammatical errors:
{original_text}
Output:

### Prompt for Condensing in Arabic

**System:** أنت خبير في الكتابة باللغة العربية. مهمتك هي تكثيف النص، مع الحفاظ على معناه الجوهري، وجعله أكثر إيجازًا وخلوًا من الحشو والتكرار. يجب أن تستخدم أسلوبك الكتابي الخاص.
**User:** يُرجى تكثيف الن التالي بأسلوبك الخاص، لجعله أكثر إيجازًا وخاليًا من الحشو. لا تعدّل الكلمات المكتوبة بأحرف كبيرة، وتجنب الأخطاء النحوية:
{original_text}
المخرجات:

### Prompt for Condensing in Chinese

**System:** 你是一位中文写作专家。你的任务是对文本进行缩写，在保留原文核心语义的基础上，使其内容更加精炼而不冗余，并使用你自己的写作风格。
**User:** 请用你自己的风格缩写以下文本，使其内容更加精炼而不冗余。不要修改大写的词语，避免语法错误：
{original_text}
输出：

### Prompt for Condensing in French

**System:** Vous êtes un expert en rédaction française. Votre mission est de condenser un texte pour le rendre plus concis et éliminer les redondances, tout en préservant son sens essentiel. Vous devez appliquer votre propre style d'écriture.
**User:** Veuillez condenser le texte suivant dans votre propre style, en le rendant plus concis et en éliminant les redondances. Ne modifiez pas les mots en majuscules et évitez les erreurs grammaticales :
{original_text}
Sortie :

### Prompt for Condensing in German

**System:** Sie sind ein Experte für das Verfassen deutscher Texte. Ihre Aufgabe ist es, einen Text zu kürzen. Gestalten Sie ihn prägnanter, beseitigen Sie Redundanzen und bewahren Sie dabei die Kernaussage. Verwenden Sie Ihren eigenen Schreibstil.
**User:** Bitte kürzen Sie den folgenden Text in Ihrem eigenen Stil, sodass er prägnanter und frei von Redundanzen ist. Ändern Sie keine Wörter in Großbuchstaben und vermeiden Sie grammatikalische Fehler:
{original_text}
Ausgabe:

### Prompt for Condensing in Portuguese

**System:** Você é um especialista em redação em português. Sua tarefa é condensar um texto, tornando-o mais conciso e eliminando redundâncias, ao mesmo tempo que preserva o significado central. Use seu próprio estilo de escrita.
**User:** Por favor, condense o texto a seguir em seu próprio estilo, tornando-o mais conciso e livre de redundâncias. Não modifique as palavras em maiúsculas e evite erros gramaticais:

{original_text}
Saída:

**Prompt for Condensing in Russian**

**System:** Вы — эксперт по написанию текстов на русском языке. Ваша задача — сократить текст, сделав его более сжатым и лаконичным, устранив избыточность и сохранив его основной смысл. Используйте свой авторский стиль.
**User:** Пожалуйста, сократите следующий текст в своём собственном стиле, сделав его более сжатым и свободным от избыточности. Не изменяйте слова, написанные заглавными буквами, и избегайте грамматических ошибок:
{original_text}
Вывод:

**Prompt for Condensing in Spanish**

**System:** Eres un escritor experto en español. Tu tarea es condensar un texto, preservando su significado central para hacerlo más conciso y eliminar redundancias. Debes usar tu propio estilo de redacción.
**User:** Por favor, condensa el siguiente texto con tu propio estilo, haciéndolo más conciso y eliminando las redundancias. No modifiques las palabras en mayúsculas y evita los errores gramaticales:
{original_text}
Salida:

## C Dataset Statistics

| Category | Subcategory | Sample Count |
|---|---|---|
| Original Sample | Human-written | 144,000 |
| | LLM-generated | 144,000 |
| LLM-refined Sample (Various Operation Behaviour) | Human + Rewritting | 144,000 |
| | Human + Expanding | 144,000 |
| | Human + Condesing | 144,000 |
| | LLM + Rewritting | 144,000 |
| | LLM + Expanding | 144,000 |
| | LLM + Condesing | 144,000 |
| Paraphrasing Attacks | LLM + Encoder Paraphrasing | 144,000 |
| | LLM + Seq2seq Paraphrasing | 144,000 |
| | LLM + Decoder Paraphrasing | 144,000 |
| | LLM + Back-Translation | 144,000 |
| Perturbation Attacks | LLM + Character Insertion | 144,000 |
| | LLM + Character Substitution | 144,000 |
| | LLM + Character Deletion | 144,000 |
| | LLM + Zero-width Insertion | 144,000 |
| Multi-length Samples | Human with 64 Tokens | 144,000 |
| | Human with 128 Tokens | 144,000 |
| | Human with 256 Tokens | 144,000 |
| | Human with 512 Tokens | 144,000 |
| | LLM with 64 Tokens | 144,000 |
| | LLM with 128 Tokens | 144,000 |
| | LLM with 256 Tokens | 144,000 |
| | LLM with 512 Tokens | 144,000 |
| **Total** | | **3,456,000** |

Table 5: Dataset Statistics.

Detailed statistics of our dataset are summarized in Table 5. The dataset comprises 3,456,000 samples spanning several critical dimensions, including human-written content, LLM-generated text, augmented operations, paraphrasing/perturbation attacks, and multi-length evaluations. This large-scale benchmark offers an unprecedented platform to rigorously evaluate the performance, robustness, and generalization of text generation models.

The core of the dataset consists of 144,000 human-written samples paired with 144,000 LLM-generated counterparts, establishing a balanced human-AI comparison group (Table 6). To enrich data diversity across operation types, we curated 432,000 samples of human-originated text augmented by LLMs, alongside 432,000 LLM-generated samples refined through polishing, expanding, and condensing.

To assess model robustness, we implemented a dual-attack framework. For paraphrasing attacks, four advanced strategies including Encoder Paraphrasing, Seq2seq Paraphrasing, Decoder Paraphrasing, and Back-Translation, were utilized to generate 576,000 high-quality samples. Similarly, for perturbation attacks, we employed four techniques (Character Insertion, Character Substitution, Character Deletion, and Zero-width Insertion) to produce an additional 576,000 samples. Finally, to investigate length generalization, we compiled 1,152,000 samples across four length categories (64, 128, 256, and 512 tokens), providing a benchmark for performance across varying text complexities.

**Dataset Split** The dataset is divided into training and testing sets with a 2:1 ratio. To ensure fair and representative evaluation, the split is balanced across all domains, generators, and languages, maintaining consistent distributions in each subset. This design allows reliable cross-lingual and cross-domain performance comparison while preventing bias toward any specific source.

## D Textual features analysis of DetectRL-X

In this section, we analyze the textual features of DetectRL-X samples to provide further insights.

**Text length** We performed a statistical analysis of text length distribution in DetectRL-X, as shown in Figure 6. Our analysis of text length, measured as the number of tokens computed by the xlm-X-Rob-base tokenizer, reveals significant variations across different categories. Specifically, texts from the academic writing and novel domains exhibit considerably greater lengths when compared to those from the news and webtext categories. Furthermore, linguistic characteristics appear to influence text length; Western languages, including English and Spanish, tend to produce shorter texts, whereas languages such as Chinese, Arabic, and Russian are associated with longer outputs. Despite these variations across category and language, the length distributions prove to be remarkably consistent, both across the different models and between LLM-generated and LLM-refined texts.

| Language | Domain | HWT | LGT | | | | HLT (E-HWT) | | | E-LGT | | | |
|---|---|---|---|---|---|---|---|---|---|---|---|---|---|
| | | Original | GPT-4o | Gemini | DeepSeek | Qwen | Polishing | Expanding | Condensing | Polishing | Expanding | Condensing | |
| EN | Academic | 3,000 | 750 | 750 | 750 | 750 | 3,000 | 3,000 | 3,000 | 3,000 | 3,000 | 3,000 | 24,000 |
| | News | 3,000 | 750 | 750 | 750 | 750 | 3,000 | 3,000 | 3,000 | 3,000 | 3,000 | 3,000 | 24,000 |
| | Novel | 3,000 | 750 | 750 | 750 | 750 | 3,000 | 3,000 | 3,000 | 3,000 | 3,000 | 3,000 | 24,000 |
| | SEO | 3,000 | 750 | 750 | 750 | 750 | 3,000 | 3,000 | 3,000 | 3,000 | 3,000 | 3,000 | 24,000 |
| | Wiki | 3,000 | 750 | 750 | 750 | 750 | 3,000 | 3,000 | 3,000 | 3,000 | 3,000 | 3,000 | 24,000 |
| | WebText | 3,000 | 750 | 750 | 750 | 750 | 3,000 | 3,000 | 3,000 | 3,000 | 3,000 | 3,000 | 24,000 |
| ZH | Academic | 3,000 | 750 | 750 | 750 | 750 | 3,000 | 3,000 | 3,000 | 3,000 | 3,000 | 3,000 | 24,000 |
| | News | 3,000 | 750 | 750 | 750 | 750 | 3,000 | 3,000 | 3,000 | 3,000 | 3,000 | 3,000 | 24,000 |
| | Novel | 3,000 | 750 | 750 | 750 | 750 | 3,000 | 3,000 | 3,000 | 3,000 | 3,000 | 3,000 | 24,000 |
| | SEO | 3,000 | 750 | 750 | 750 | 750 | 3,000 | 3,000 | 3,000 | 3,000 | 3,000 | 3,000 | 24,000 |
| | Wiki | 3,000 | 750 | 750 | 750 | 750 | 3,000 | 3,000 | 3,000 | 3,000 | 3,000 | 3,000 | 24,000 |
| | WebText | 3,000 | 750 | 750 | 750 | 750 | 3,000 | 3,000 | 3,000 | 3,000 | 3,000 | 3,000 | 24,000 |
| ES | Academic | 3,000 | 750 | 750 | 750 | 750 | 3,000 | 3,000 | 3,000 | 3,000 | 3,000 | 3,000 | 24,000 |
| | News | 3,000 | 750 | 750 | 750 | 750 | 3,000 | 3,000 | 3,000 | 3,000 | 3,000 | 3,000 | 24,000 |
| | Novel | 3,000 | 750 | 750 | 750 | 750 | 3,000 | 3,000 | 3,000 | 3,000 | 3,000 | 3,000 | 24,000 |
| | SEO | 3,000 | 750 | 750 | 750 | 750 | 3,000 | 3,000 | 3,000 | 3,000 | 3,000 | 3,000 | 24,000 |
| | Wiki | 3,000 | 750 | 750 | 750 | 750 | 3,000 | 3,000 | 3,000 | 3,000 | 3,000 | 3,000 | 24,000 |
| | WebText | 3,000 | 750 | 750 | 750 | 750 | 3,000 | 3,000 | 3,000 | 3,000 | 3,000 | 3,000 | 24,000 |
| AR | Academic | 3,000 | 750 | 750 | 750 | 750 | 3,000 | 3,000 | 3,000 | 3,000 | 3,000 | 3,000 | 24,000 |
| | News | 3,000 | 750 | 750 | 750 | 750 | 3,000 | 3,000 | 3,000 | 3,000 | 3,000 | 3,000 | 24,000 |
| | Novel | 3,000 | 750 | 750 | 750 | 750 | 3,000 | 3,000 | 3,000 | 3,000 | 3,000 | 3,000 | 24,000 |
| | SEO | 3,000 | 750 | 750 | 750 | 750 | 3,000 | 3,000 | 3,000 | 3,000 | 3,000 | 3,000 | 24,000 |
| | Wiki | 3,000 | 750 | 750 | 750 | 750 | 3,000 | 3,000 | 3,000 | 3,000 | 3,000 | 3,000 | 24,000 |
| | WebText | 3,000 | 750 | 750 | 750 | 750 | 3,000 | 3,000 | 3,000 | 3,000 | 3,000 | 3,000 | 24,000 |
| FR | Academic | 3,000 | 750 | 750 | 750 | 750 | 3,000 | 3,000 | 3,000 | 3,000 | 3,000 | 3,000 | 24,000 |
| | News | 3,000 | 750 | 750 | 750 | 750 | 3,000 | 3,000 | 3,000 | 3,000 | 3,000 | 3,000 | 24,000 |
| | Novel | 3,000 | 750 | 750 | 750 | 750 | 3,000 | 3,000 | 3,000 | 3,000 | 3,000 | 3,000 | 24,000 |
| | SEO | 3,000 | 750 | 750 | 750 | 750 | 3,000 | 3,000 | 3,000 | 3,000 | 3,000 | 3,000 | 24,000 |
| | Wiki | 3,000 | 750 | 750 | 750 | 750 | 3,000 | 3,000 | 3,000 | 3,000 | 3,000 | 3,000 | 24,000 |
| | WebText | 3,000 | 750 | 750 | 750 | 750 | 3,000 | 3,000 | 3,000 | 3,000 | 3,000 | 3,000 | 24,000 |
| RU | Academic | 3,000 | 750 | 750 | 750 | 750 | 3,000 | 3,000 | 3,000 | 3,000 | 3,000 | 3,000 | 24,000 |
| | News | 3,000 | 750 | 750 | 750 | 750 | 3,000 | 3,000 | 3,000 | 3,000 | 3,000 | 3,000 | 24,000 |
| | Novel | 3,000 | 750 | 750 | 750 | 750 | 3,000 | 3,000 | 3,000 | 3,000 | 3,000 | 3,000 | 24,000 |
| | SEO | 3,000 | 750 | 750 | 750 | 750 | 3,000 | 3,000 | 3,000 | 3,000 | 3,000 | 3,000 | 24,000 |
| | Wiki | 3,000 | 750 | 750 | 750 | 750 | 3,000 | 3,000 | 3,000 | 3,000 | 3,000 | 3,000 | 24,000 |
| | WebText | 3,000 | 750 | 750 | 750 | 750 | 3,000 | 3,000 | 3,000 | 3,000 | 3,000 | 3,000 | 24,000 |
| PT | Academic | 3,000 | 750 | 750 | 750 | 750 | 3,000 | 3,000 | 3,000 | 3,000 | 3,000 | 3,000 | 24,000 |
| | News | 3,000 | 750 | 750 | 750 | 750 | 3,000 | 3,000 | 3,000 | 3,000 | 3,000 | 3,000 | 24,000 |
| | Novel | 3,000 | 750 | 750 | 750 | 750 | 3,000 | 3,000 | 3,000 | 3,000 | 3,000 | 3,000 | 24,000 |
| | SEO | 3,000 | 750 | 750 | 750 | 750 | 3,000 | 3,000 | 3,000 | 3,000 | 3,000 | 3,000 | 24,000 |
| | Wiki | 3,000 | 750 | 750 | 750 | 750 | 3,000 | 3,000 | 3,000 | 3,000 | 3,000 | 3,000 | 24,000 |
| | WebText | 3,000 | 750 | 750 | 750 | 750 | 3,000 | 3,000 | 3,000 | 3,000 | 3,000 | 3,000 | 24,000 |
| DE | Academic | 3,000 | 750 | 750 | 750 | 750 | 3,000 | 3,000 | 3,000 | 3,000 | 3,000 | 3,000 | 24,000 |
| | News | 3,000 | 750 | 750 | 750 | 750 | 3,000 | 3,000 | 3,000 | 3,000 | 3,000 | 3,000 | 24,000 |
| | Novel | 3,000 | 750 | 750 | 750 | 750 | 3,000 | 3,000 | 3,000 | 3,000 | 3,000 | 3,000 | 24,000 |
| | SEO | 3,000 | 750 | 750 | 750 | 750 | 3,000 | 3,000 | 3,000 | 3,000 | 3,000 | 3,000 | 24,000 |
| | Wiki | 3,000 | 750 | 750 | 750 | 750 | 3,000 | 3,000 | 3,000 | 3,000 | 3,000 | 3,000 | 24,000 |
| | WebText | 3,000 | 750 | 750 | 750 | 750 | 3,000 | 3,000 | 3,000 | 3,000 | 3,000 | 3,000 | 24,000 |
| **Total** | | **144,000** | **36,000** | **36,000** | **36,000** | **36,000** | **144,000** | **144,000** | **144,000** | **144,000** | **144,000** | **144,000** | **1,152,000** |

Table 6: Dataset Statistics of Original Samples and LLM-refined Samples under Different Operation Behaviour. "HLT (E-HWT)" denotes "Human-written & LLM refined" where original human samples are polished, expanded, or condensed by LLMs. "E-LGT" denotes "LLM-generated & LLM refined" where original LLM samples are polished, expanded, or condensed by LLMs.

**N-grams** We conducted a statistical analysis of the n-gram distribution in DetectRL-X, focusing on unigrams, bigrams, and trigrams. The results are presented in Figure 7. Our analysis of n-gram distributions highlights significant diversity variations. The Novel category shows the most extensive vocabulary, directly opposing the SEO category, which has the least. Model-wise, gemini-2.5-flash generates the most diverse n-grams, distinctly outperforming gpt-40, deepseek-v3, and qwen-max, all of which exhibit more constrained lexical patterns. Linguistically, languages like French, German, and Portuguese display high n-gram richness, whereas Chinese is a notable outlier with exceptionally low diversity. Across all these comparisons, however, the distributions for LLM-refined and LLM-generated texts remain remarkably consistent and nearly identical.

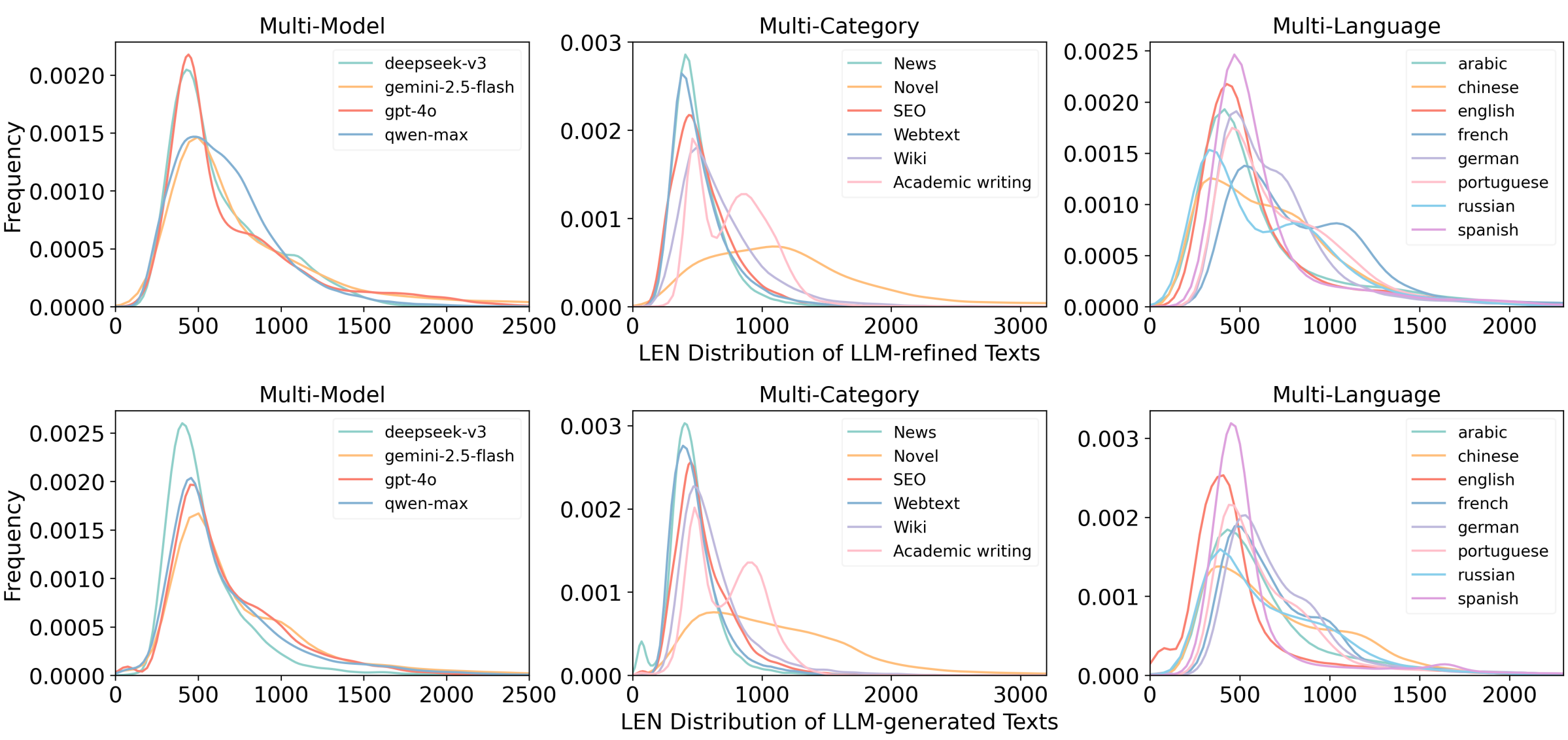

Figure 6: Text length distribution of DETECTRL-X.

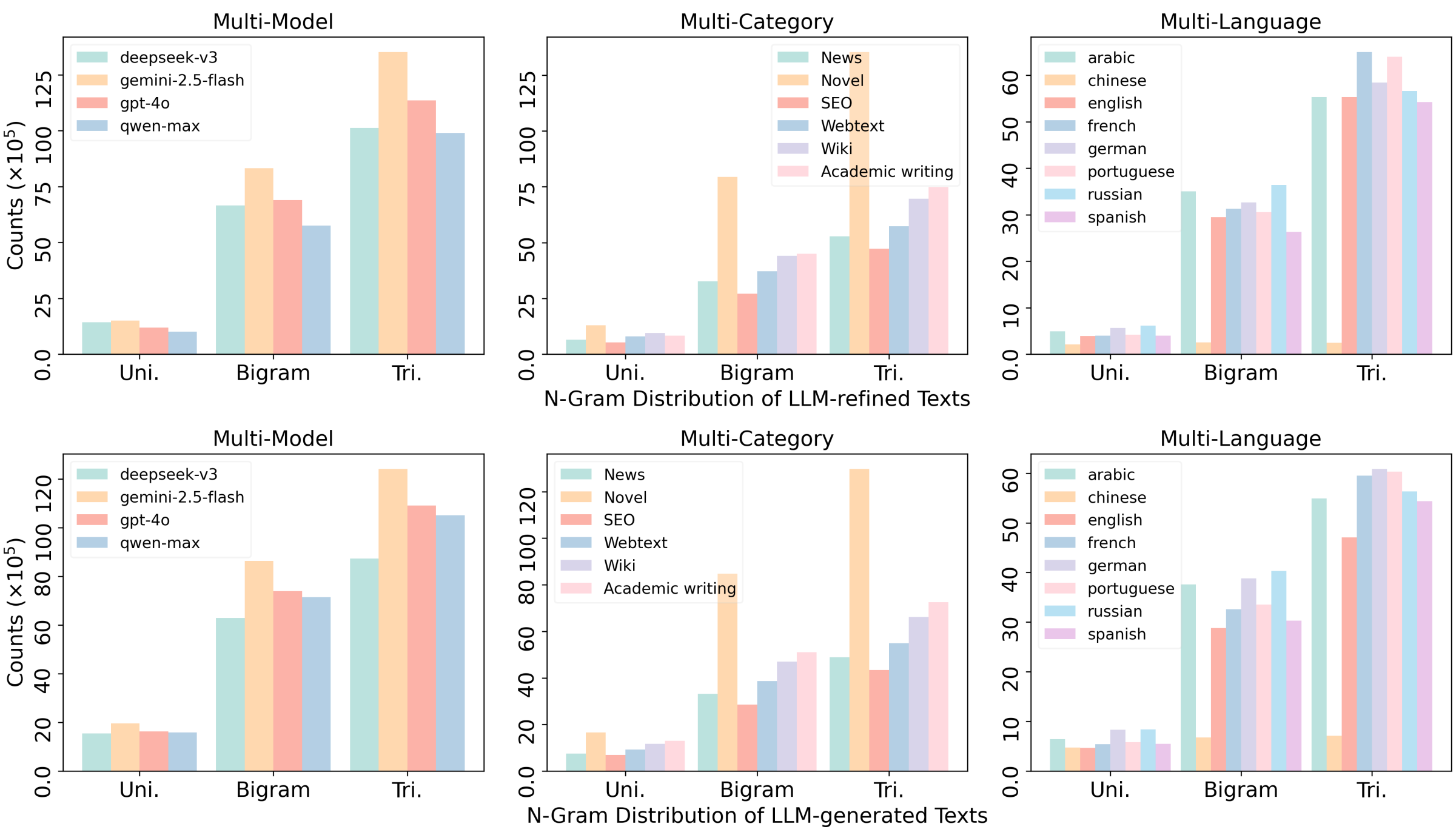

Figure 7: N-gram distribution of DETECTRL-X.

**Readability** The readability distribution of DetectRL-X samples was analyzed using the Flesch Reading Ease Score (FRES). The FRES (Flesch, 1948) assesses reading difficulty based on word and sentence length and is computed as follows:

$$\text{FRES} = 206.835 - 1.015 \times \left(\frac{\text{Total Words}}{\text{Total Sentences}}\right) - 84.6 \times \left(\frac{\text{Total Syllables}}{\text{Total Words}}\right)$$

Higher scores indicating better readability. The results of our readability assessment in Figure 8 indicate that while LLM refinement promotes cross-model consistency, significant variations persist across content categories and languages. We observed a distinct hierarchy in readability among the categories, with Novel texts being the most accessible and Academic writing proving to be the least. Furthermore, the analysis by language revealed a substantial gap: both Arabic and Russian achieved unexpectedly high readability scores, whereas German and Portuguese were situated at the lower end of the readability spectrum. A crucial observation, however, is that the distributions for LLM-refined texts show a much higher degree of uniformity across the models compared to the more divergent scores of LLM-generated texts.

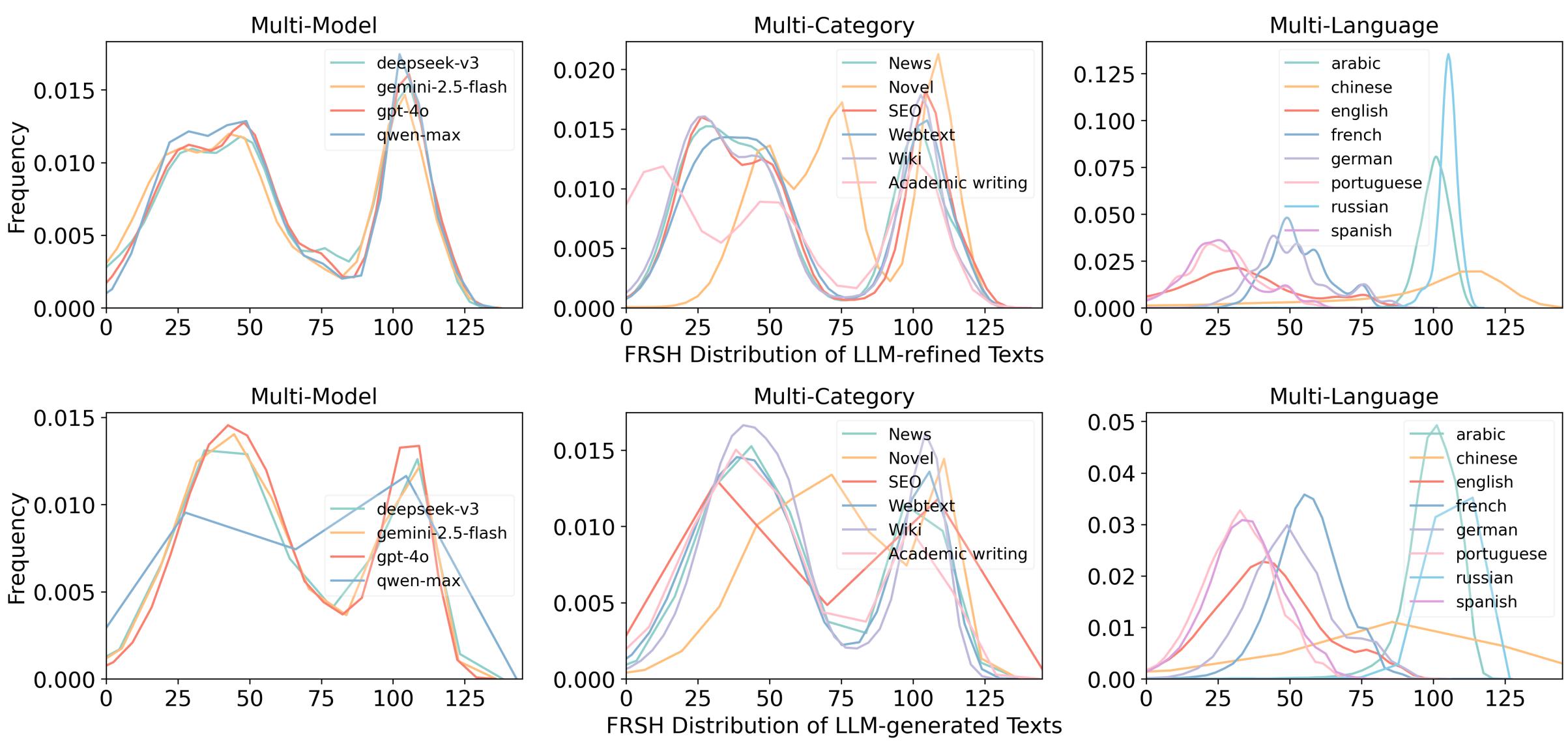

Figure 8: Readability distribution of DETECTRL-X.

**Lexical diversity** We statistically analyzed the Lexical Diversity Score (LDS) of DetectRL-X, which is defined as follows:

$$\text{LDS} = \frac{\text{Number of Unique Word Types}}{\text{Total Number of Words}}$$

In our examination of lexical diversity, a notable pattern of consistency emerged across several dimensions as show in Figure 9. Both LLM-generated and LLM-refined texts exhibit nearly identical distributions, and a similar uniformity is observed across the different models, with scores generally clustering around a lexical diversity of approximately 0.6. However, this consistency does not extend to content categories or languages. Counter-intuitively, the Novel domain shows a comparatively low lexical richness, in contrast to the higher diversity found in SEO, News, and Webtext samples. Furthermore, a clear divergence is noted among languages, where French texts tend to have lower diversity, while Russian texts demonstrate a higher score.

**Comparative Analysis of HWT, LGT, and HLT.** To further characterize the linguistic properties of our dataset, we conduct a comparative analysis across the three primary data types: HWT, LGT, and HLT. This analysis aims to quantify the stylistic shifts that occur during the revision process and highlight the distinct patterns that detectors must navigate.Table 7 summarizes the mean and standard deviation of four key textual metrics across the entire corpus.

Our analysis reveals that HLT exhibits a unique "hybrid" characteristic that distinguishes it from both pure human and pure machine-generated content:

- **Lexical Richness vs. Conciseness:** HLT inherits the high lexical diversity characteristic of LGT (TTR: 0.57), yet it results in the most concise average text length. This suggests that the revision operations (Polishing, Condensing) effectively retain semantic density while removing redundancy.
- **Readability and Flow:** While LGT often features more complex or formulaic sentence structures (indicated by a lower Flesch score of 48.82), the human-AI collaborative process in HLT restores readability (Flesch: 53.29) to a level closer to that of HWT (55.52).
- **Distributional Coverage:** The substantial standard deviations observed across all met-

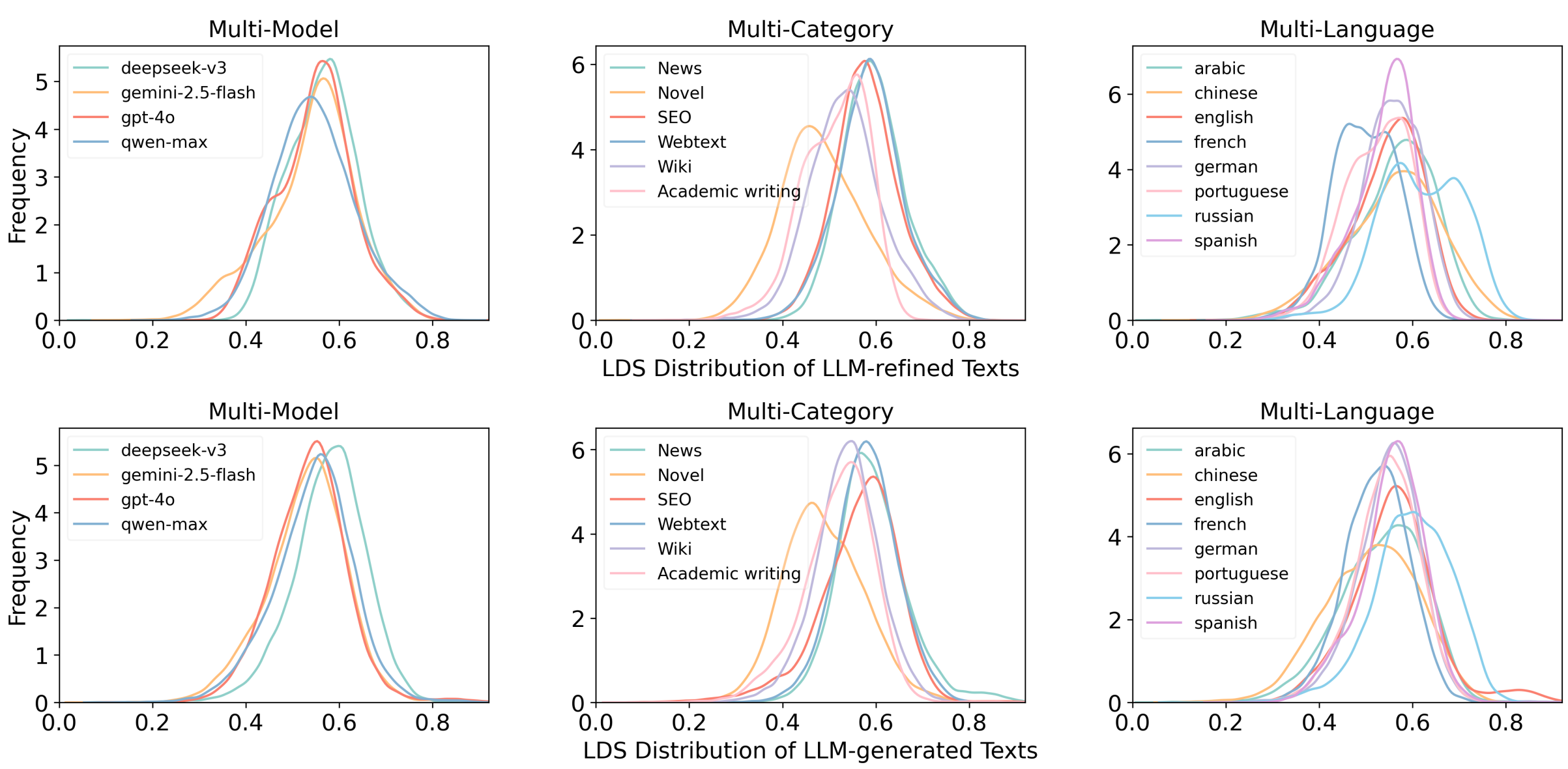

Figure 9: Lexical diversity distribution of DETECTRL-X.

Table 7: Comparative analysis of textual features across HWT, LGT, and HLT. Results report mean ± standard deviation.

| Metric | HWT | LGT | HLT |
|---|---|---|---|
| Length (Words) ↓ | $567.29 \pm 432.68$ | $534.49 \pm 406.68$ | $\mathbf{474.79 \pm 307.43}$ |
| Lexical Diversity (TTR) ↑ | $0.53 \pm 0.10$ | $\mathbf{0.57 \pm 0.10}$ | $\mathbf{0.57 \pm 0.10}$ |
| N-gram Diversity (Dist-3) ↑ | $0.95 \pm 0.05$ | $\mathbf{0.97 \pm 0.03}$ | $0.96 \pm 0.05$ |
| Readability (Flesch) ↑ | $\mathbf{55.52 \pm 30.77}$ | $48.82 \pm 30.46$ | $53.29 \pm 30.66$ |

rics confirm that DetectRL-X encompasses a wide and representative distribution of textual features, avoiding the narrow stylistic biases often found in synthetic datasets.

These findings underscore the challenge HLT poses to current detectors: it combines the structural fluency of human writing with the lexical patterns of LLMs, creating a subtle adversarial boundary in the feature space.

# E Evaluated Detectors

## E.1 Descriptions of Evaluated Detectors

We provide detailed descriptions of each detector evaluated, explaining their working principles and key characteristics. These detectors excel in different application scenarios, and by using a combination of these methods, we can achieve a more comprehensive evaluation and detection of LLM-generated text.

- **Log-Likelihood** (Solaiman et al., 2019): This method computes the log-likelihood of the input text using a pre-trained language model. Lower likelihoods indicate that the text is more likely to be machine-generated. It is a widely adopted baseline in text detection tasks. Specifically, Log-Likelihood leverages the language model's understanding of natural language by calculating the probability of each word in its context to determine the overall likelihood of the text being generated.

- **Log-Rank** (Gehrmann et al., 2019): This method ranks the predicted probability of each token and uses the average token rank as a distinguishing feature, capturing distributional differences between human-written and machine-generated text. Log-Rank assumes that human-written text will have a more uniform ranking in the language model compared to machine-generated text, thus using the average rank to differentiate between the two.

- **DetectLLM-LRR** (Su et al., 2023): By combining log-likelihood and token rank information, this method leverages dual statistical signals to enhance detection performance. LRR improves detection accuracy by simulta-neously considering both log-likelihood and token rank, thereby capturing more nuanced features of the generated text.

- **Fast-DetectGPT** (Bao et al., 2024): As an efficient variant of DetectGPT, this method distinguishes generated text through local perturbation and curvature analysis, achieving a balance between detection accuracy and computational efficiency. Fast-DetectGPT introduces local perturbations and analyzes their impact on model output, maintaining high accuracy while reducing computational costs.

- **Binoculars** (Hans et al., 2024): This approach adopts a dual-perspective mechanism, comparing the model outputs on both original and perturbed samples to improve robustness against paraphrasing and adversarial attacks. Binoculars enhances resistance to rephrasing and adversarial attacks by contrasting the model's behavior on original and perturbed versions of the text.

- **ReviseDetect** (Zhu et al., 2023): Based on a rewriting model, this method measures the extent to which the input text has been rewritten by the larger model. Larger rewriting typically indicates that the text is machine-generated. ReviseDetect effectively identifies text generated by large language models by analyzing the degree of rewriting.

- **GECScore** (Wu et al., 2025c): Based on a grammatical error correction model, this method measures the amount of modification made by the correction system. Larger changes typically indicate that the text is machine-generated. GECScore assesses the text's machine-generated nature by quantifying the corrections made by a grammatical error correction model.

- **Lastde++** (Xu et al., 2025): By improving the architecture and training strategy of the detector, this approach enhances generalization and robustness across multiple domains and languages. Lastde++ optimizes the model structure and training process to improve the detector's adaptability in various languages and domains.

- **RepreGuard** (Chen et al., 2025b): This method captures the differences in neuronal activation patterns when the model processes human-written text and LLM-generated text, extracting high-dimensional features from LLM-generated text to distinguish between the two types of text. RepreGuard analyzes the internal activation patterns of the neural network to accurately identify LLM-generated text.

- **X-Rob-Classifier** (Liu et al., 2019): This discriminative method fine-tunes the X-Rob pre-trained model for BINARY or multi-class classification, representing a mainstream approach in neural-based detection. X-Rob-Classifier leverages the powerful pre-trained X-Rob model to perform text classification tasks after fine-tuning.

- **mDeBERTa-Classifier** (He et al., 2021): Utilizing the multilingual version of mDeBERTa as the classifier, this method improves cross-lingual detection capability and is well-suited for multilingual scenarios. mDeBERTa-Classifier enhances cross-lingual detection by using a multilingual pre-trained model.

- **Biscope** (Guo et al., 2024): As a state-of-the-art approach, Biscope combines generation-based and discrimination-based signals to improve both accuracy and robustness, representing the latest advancements in LLM-generated text detection. Biscope integrates the advantages of both generative and discriminative methods to achieve better detection accuracy and robustness.

**Implementation details of the statistics-based approaches** For statistical-based gray-box detection methods, we uniformly employs Qwen-2.5-7B[14] as the scoring model to accommodate multilingual detection requirements. Among these, FastDetectGPT, Binoculars, and Lastde++ require additional reference models, for which we uniformly adopt Qwen-2.5-7B-Instruct.[15] For black-box statistical detection methods (such as ReviseDetect and GECScore), Qwen-Max is used to perform text revision, error correction, and related operations. For neural-based detectors, all classifiers are trained with identical parameter configurations. Detailed training parameter settings are provided in Appendix E.2. Following Chen et al. (2025b) and Hans et al. (2024),

[14] https://huggingface.co/Qwen/Qwen2.5-7B
[15] https://huggingface.co/Qwen/Qwen2.5-7B-Instruct

all detectors are retrained and fairly compared on the same training and test sets.

**Extending BINARY Statistical-Based Detector to TERNARY Classification** The original detector is a BINARY classifier that distinguishes between HWT and LGT based on a statistical feature $x$ and a single threshold $T$:

$$\text{label} = \begin{cases} \text{HWT}, & x < T, \\ \text{LGT}, & x \geq T. \end{cases}$$

To include the intermediate category HLT, the BINARY decision rule is extended to a TERNARY formulation by introducing two thresholds $T_1$ and $T_2$ ($T_1 < T_2$):

$$\text{label} = \begin{cases} \text{HWT}, & x < T_1, \\ \text{HLT}, & T_1 \leq x < T_2, \\ \text{LGT}, & x \geq T_2. \end{cases}$$

The thresholds $T_1$ and $T_2$ are determined through an exhaustive search on the validation set to maximize the overall F1-Score of the three-class classification. This dual-threshold formulation generalizes the original BINARY detector and enables more precise separation among HWT, HLT, and LGT.

### E.2 Parameters of Neural-Based Detectors

**X-Rob-Classifier** We use the `FacebookAI/xlm-X-Rob-base` model as the pre-trained backbone. The model is fine-tuned for `3 epochs` using a batch size of `32`. To ensure a fair and stable comparison across all experiments, we utilize a linear learning rate decay (from $2 \times 10^{-5}$ to 0) without a warmup phase. We deliberately omit early stopping and consistently evaluate the final checkpoint. All experiments are conducted on an NVIDIA® A100 Tensor Core GPU (80 GB).

| Parameters | Settings |
|---|---|
| Model Name | FacebookAI/xlm-roberta-base |
| Learning Rate | 2e-5 |
| Batch Size | 32 |
| Epochs | 3 |
| Seed | 2025 |
| GPU Envs | NVIDIA® A100 Tensor Core GPU 80GB |

Table 8: Parameters for X-Rob-Classifier Training.

**mDeBERTa-Classifier** We use the `microsoft/mdeberta-v3-base` model as the pre-trained backbone. The model is fine-tuned for `3 epochs` using a batch size of `32`. To ensure a fair and stable comparison across all experiments, we utilize a linear learning rate decay (from $2 \times 10^{-5}$ to 0) without a warmup phase. We deliberately omit early stopping and consistently evaluate the final checkpoint. All experiments are conducted on an NVIDIA® A100 Tensor Core GPU (80 GB).

| Parameters | Settings |
|---|---|
| Model Name | microsoft/mdeberta-v3-base |
| Learning Rate | 2e-5 |
| Batch Size | 32 |
| Epochs | 3 |
| Seed | 2025 |
| GPU Envs | NVIDIA® A100 Tensor Core GPU 80GB |

Table 9: Parameters for mDeBERTa-Classifier Training.

**BiScope-Classifier** We employ `Qwen2.5-7B-Instruct` for feature extraction and a `RandomForestClassifier` for classification. For neural components, we use a batch size of 128 and a linear learning rate decay ($2 \times 10^{-5}$ to 0) over 3 epochs without warmup. We consistently evaluate the final checkpoint. All experiments are conducted on an NVIDIA® A100 (80GB) GPU.

| Parameters | Settings |
|---|---|
| Classifier Model | RandomForestClassifier |
| Detect Model | Qwen/Qwen2.5-7B-Instruct |
| Learning Rate | 2e-5 |
| Batch Size | 128 |
| Epochs | 3 |
| Seed | 2025 |
| GPU Envs | NVIDIA® A100 Tensor Core GPU 80GB |

Table 10: Parameters for Biscope-Classifier Training.

## F Error Analysis and Case Studies

This section focuses on analyzing the causes of detection failures, systematically investigating their underlying mechanisms through a combination of statistical metric analysis and representative case studies. Manual inspection of misclassified samples reveals that failures are predominantly driven by *implicit distributional overlaps* between text categories, rather than explicit linguistic errors in the text itself. To quantify and interpret these overlaps, we adopt **Likelihood** as the core analytical metric (an intuitive and interpretable indicator of text entropy), and structure our analysis around three key dimensions: **Boundary Ambiguity of HLT**, **Ad-**

Table 11: Likelihood Distributions of Different Text Categories.

| Category | Source | Mean ± Std | Best F1 Threshold |
|---|---|---|---|
| Baselines | HWT | $-2.2381 \pm 0.4240$ | $-1.995830$ |
| | LGT | $-1.8309 \pm 0.3431$ | – |
| HLT Variants | Polishing | $-1.8275 \pm 0.3435$ | – |
| | Expanding | $-1.8606 \pm 0.3500$ | – |
| | Condensing | $-1.8265 \pm 0.3431$ | – |

**versarial Attack Mechanics**, and **Generalization Limitations**.

### F.1 Boundary Ambiguity in HLT

The fundamental challenge in TERNARY classification lies in the "fuzzy boundary" of HLT, which causes its feature distribution to overlap extensively with both HWT and LGT. Table 11 presents the likelihood distribution statistics of different text categories, where HLT variants (Polishing, Expanding, Condensing) exhibit mean likelihood values (-1.8275, -1.8606, -1.8265) extremely close to LGT (-1.8309), directly explaining the classification difficulty.

To further reveal the overlap mechanisms, we analyze three representative failure cases (Table 12) and summarize three typical failure modes:

- **"Formulaic Trap" (HWT→LGT Misclassification)**: Extremely standardized human writing (e.g., formal acknowledgments) exhibits low entropy (likelihood $= -1.7890 >$ LGT average of $-1.8309$). The machine-like regularity of such texts confuses the model, leading to misclassification as LGT.
- **"Creative Mimicry" (LGT→HWT Misclassification)**: When LLMs simulate specific literary styles using low-frequency vocabulary, the likelihood value drops significantly ($-2.1910$), creating a statistical overlap with the HWT distribution (mean $= -2.2381$). This mimicry of human creative writing characteristics results in misclassification as HWT.
- **"Entity Anchoring" (HLT→HWT Misclassification)**: HLT retains a large number of named entities (e.g., proper nouns, specific events) from the original human text, which act as high-perplexity anchors. These entities pull the overall likelihood of HLT down to $-2.1511$, effectively masking the smoothed syntactic patterns unique to AI generation, thus leading to misclassification as HWT.

### F.2 Attack Mechanics: Distributional Shift vs. Fragmentation

Adversarial attacks cause LGT to deviate from the in-distribution range, leading to detection failures through two core mechanisms: semantic distributional shift and sub-word fragmentation. Table 13 presents the likelihood changes of LGT under different attack types.

- **Semantic Distributional Shift**: Paraphrasing attacks (e.g., Seq2Seq models) increase lexical variance by rephrasing LGT while preserving core meaning. This disrupts the "smooth" statistical signature of LLM generation, reducing the likelihood from $-1.8309$ to $-4.1994$, which effectively mimics the high-entropy feature distribution of HWT.
- **Sub-word Fragmentation**: Character substitution attacks physically disrupt the integrity of words (e.g., replacing 'a' with '$\alpha$'). This forces the tokenizer to split words into meaningless sub-word fragments, introducing artificial noise into the feature space. As a result, the likelihood of LGT drops drastically to $-5.7146$, making it indistinguishable from high-perplexity text.

### F.3 Generalization Failures: Statistical Mismatch & Variance Explosion.

The poor generalization of detectors in cross-domain, cross-language, and cross-length scenarios is rooted in the statistical mismatch between training and testing distributions, as well as variance explosion caused by information sparsity. Table 14 summarizes the likelihood distribution characteristics under different generalization scenarios.

- **Domain Shift**: Semantic distributions vary significantly across domains. For example, the likelihood of HWT in WebText ($-2.0764$) is notably higher than that in Academic domain ($-2.4426$). This shift moves the HWT distribution closer to the LGT feature range ($\sim -1.90$), increasing the overlap between the two and leading to misclassification.
- **Cross-Language Mismatch**: The likelihood distributions of HWT and LGT differ between English and Chinese. Chinese text exhibits higher variance (HWT: 0.5137, LGT: 0.4339) compared to English (HWT: 0.4190, LGT: 0.3756), which is attributed to the structural differences between logographic and alphabetic languages. This mismatch reduces the detector's performance on unseen languages.

Table 12: Representative Failure Cases and Feature Analysis.

| Cases | Text Snippet | Likelihood | Prediction |
|---|---|---|---|
| HWT | I would like to thank the staff and management of Biltmore Estate for their support of Serafina and the Black Cloak and their commitment to preserving an ... | $-1.7890$ | LGT |
| LGT | August 15, 1892 My Dearest Friend, As I sit here, pen in hand, the sun is setting over the distant hills, casting a golden glow that seems to set the world ablaze ... | $-2.1910$ | HWT |
| HLT | Chapter 26 The hospital room was cold and sterile, its walls painted a pale institutional green that seems to drain the light from the air. Mrs. Parker lay on the bed ... | $-2.1511$ | HWT |

Table 13: Likelihood Distributions of LGT Under Adversarial Attacks

| Attack Type | Category | Mean ± Std |
|---|---|---|
| Normal | HWT | $-2.2381 \pm 0.4240$ |
| | LGT | $-1.8309 \pm 0.3431$ |
| Seq2Seq (Paraphrasing) | LGT | $-4.1994 \pm 0.6535$ |
| Char-Substitution (Perturbation) | LGT | $-5.7146 \pm 0.5845$ |

Table 14: Likelihood Distributions Under Generalization Scenarios.

| Dimension | Category | Type | Mean ± Std | Best F1 Threshold |
|---|---|---|---|---|
| Domain | Academic | HWT | $-2.4426 \pm 0.4324$ | $-2.2066$ |
| | | LGT | $-1.8721 \pm 0.3226$ | |
| | WebText | HWT | $-2.0764 \pm 0.3382$ | |
| | | LGT | $-1.9020 \pm 0.3206$ | |
| Language | English | HWT | $-2.4837 \pm 0.4190$ | $-2.3163$ |
| | | LGT | $-2.1121 \pm 0.3756$ | |
| | Chinese | HWT | $-2.4659 \pm 0.5137$ | |
| | | LGT | $-2.2604 \pm 0.4339$ | |
| Length | 512 Tokens | HWT | $-2.2381 \pm 0.4240$ | $-1.9958$ |
| | | LGT | $-1.8309 \pm 0.3431$ | |
| | 64 Tokens | HWT | $-2.9355 \pm 0.8295$ | |
| | | LGT | $-2.3738 \pm 0.5330$ | |

- **Variance Explosion (Shorter Texts)**: Information sparsity in short texts leads to severe variance explosion. For 64-token texts, the standard deviation of HWT likelihood ($\sigma = 0.8295$) is nearly double that of 512-token texts ($\sigma = 0.4240$). The expanded distribution range causes massive overlap between HWT and LGT, resulting in significant performance degradation for short text detection.

# G Detailed Results

## G.1 Generalization Results for Languages.

The detailed results of generalization for languages in BINARY task and TRINARY task are presented in Table 15 and Table 16.

## G.2 Generalization Results for Domains.

The detailed results of generalization for domains in BINARY task and TRINARY task are presented in Table 17 and Table 18.

## G.3 Generalization Results for Generators.

The detailed results of generalization for generators in BINARY task and TRINARY task are presented in Table 19 and Table 20.

## G.4 Robustness Results for Cross-Domain on Different Languages.

The detailed results for cross-domain on different languages in BINARY task and TRINARY task are presented in Table 21 and Table 22.

## G.5 Robustness Results for Cross-Generator on Different Languages.

The detailed results for cross-generator on different languages in BINARY task and TRINARY task are presented in Table 23 and Table 24.

## G.6 Robustness Results for Cross-Paraphrase on Different Languages.

The detailed results for cross-paraphrase on different languages in BINARY task and TRINARY task are presented in Table 25 and Table 26.

## G.7 Robustness Results for Cross-Perturbation on Different Languages.

The detailed results for cross-perturbation on different languages in BINARY task and TRINARY task are presented in Table 27 and Table 28.

## G.8 Robustness Results for Cross-Length on Different Languages.

The detailed results for cross-length on different languages in BINARY task and TRINARY task are presented in Table 29 and Table 30.

| Detectors↓ | Train↓ Test→ | English | | Chinese | | Spanish | | Arabic | | French | | Russian | | Portuguese | | German | | Average | |
|---|---|---|---|---|---|---|---|---|---|---|---|---|---|---|---|---|---|---|---|
| | Metrics | $F_1^B$ | $F_1^F$ | $F_1^B$ | $F_1^F$ | $F_1^B$ | $F_1^F$ | $F_1^B$ | $F_1^F$ | $F_1^B$ | $F_1^F$ | $F_1^B$ | $F_1^F$ | $F_1^B$ | $F_1^F$ | $F_1^B$ | $F_1^F$ | $F_1^B$ | $F_1^F$ |
| Log-Likelihood | English | 64.88 | 46.48 | 55.73 | 39.48 | 58.54 | 43.12 | 67.84 | 49.15 | 56.44 | 41.49 | 46.48 | 33.59 | 55.08 | 40.60 | 64.89 | 46.81 | 63.64 | 54.44 |
| | Chinese | 65.45 | 37.17 | 58.09 | 37.00 | 48.36 | 38.40 | 61.62 | 35.18 | 45.89 | 37.97 | 39.48 | 34.97 | 44.85 | 37.72 | 55.71 | 39.68 | 55.60 | 40.70 |
| | Spanish | 73.97 | 50.26 | 48.36 | 38.40 | 74.44 | 53.65 | 70.07 | 42.86 | 72.05 | 55.09 | 60.61 | 33.85 | 73.17 | 48.07 | 62.87 | 37.47 | 66.68 | 49.49 |
| | Arabic | 59.96 | 42.68 | 49.68 | 38.18 | 74.06 | 52.27 | 70.83 | 41.57 | 71.27 | 54.42 | 56.91 | 60.42 | 74.54 | 53.98 | 65.57 | 37.00 | 66.91 | 48.61 |
| | French | 56.44 | 41.49 | 45.89 | 37.97 | 73.97 | 50.26 | 68.20 | 39.89 | 72.02 | 52.30 | 61.68 | 59.04 | 75.28 | 51.54 | 59.05 | 36.26 | 65.66 | 47.03 |
| | Russian | 46.48 | 33.59 | 39.48 | 34.97 | 60.61 | 33.85 | 49.18 | 33.39 | 60.88 | 34.57 | 64.36 | 36.48 | 62.16 | 34.19 | 41.66 | 33.35 | 54.46 | 34.31 |
| | Portuguese | 55.08 | 40.60 | 44.85 | 37.72 | 73.17 | 48.07 | 66.75 | 38.41 | 71.60 | 50.48 | 62.84 | 57.77 | 74.62 | 49.34 | 56.67 | 35.29 | 64.81 | 45.56 |
| | German | 64.89 | 46.81 | 55.71 | 39.68 | 64.60 | 61.24 | 67.84 | 50.05 | 57.34 | 61.40 | 47.97 | 64.57 | 63.98 | 62.12 | 73.55 | 42.28 | 63.65 | 54.99 |
| Log-Rank | English | 65.12 | 47.78 | 58.68 | 40.96 | 61.34 | 45.96 | 70.02 | 48.87 | 58.14 | 44.60 | 50.52 | 34.92 | 58.90 | 43.95 | 65.71 | 51.91 | 65.46 | 52.05 |
| | Chinese | 64.72 | 41.12 | 60.57 | 38.23 | 51.58 | 43.80 | 64.44 | 37.59 | 45.71 | 47.28 | 45.16 | 50.29 | 53.44 | 44.04 | 70.91 | 34.13 | 58.70 | 42.40 |
| | Spanish | 68.51 | 56.92 | 53.33 | 39.96 | 73.71 | 53.86 | 71.37 | 45.24 | 70.91 | 55.69 | 59.96 | 57.92 | 74.96 | 54.39 | 60.84 | 36.69 | 67.00 | 49.46 |
| | Arabic | 61.83 | 45.51 | 53.77 | 39.71 | 73.69 | 52.64 | 71.69 | 43.96 | 70.76 | 54.58 | 59.48 | 57.13 | 74.75 | 53.06 | 61.52 | 36.29 | 67.14 | 48.56 |
| | French | 58.14 | 44.60 | 49.89 | 39.38 | 72.57 | 50.73 | 68.99 | 42.51 | 70.84 | 53.21 | 62.70 | 36.34 | 74.35 | 51.28 | 54.61 | 35.60 | 65.40 | 47.25 |
| | Russian | 50.52 | 34.92 | 42.75 | 35.88 | 62.27 | 34.65 | 54.67 | 33.51 | 62.70 | 36.34 | 62.94 | 37.33 | 63.74 | 34.90 | 41.52 | 33.35 | 56.20 | 35.13 |
| | Portuguese | 58.90 | 43.95 | 50.96 | 39.15 | 73.27 | 49.20 | 69.92 | 41.23 | 71.26 | 52.16 | 62.35 | 54.75 | 74.92 | 49.88 | 56.53 | 35.30 | 66.03 | 46.26 |
| | German | 65.71 | 51.91 | 60.02 | 43.80 | 60.32 | 64.79 | 67.38 | 57.35 | 52.83 | 64.74 | 47.85 | 63.80 | 61.31 | 66.53 | 72.20 | 43.41 | 62.45 | 58.19 |
| DetectLLM-LRR | English | 62.58 | 44.70 | 65.58 | 41.78 | 66.45 | 37.97 | 69.97 | 38.00 | 65.00 | 41.58 | 57.67 | 36.78 | 68.19 | 36.92 | 44.88 | 33.54 | 63.32 | 39.01 |
| | Chinese | 62.71 | 48.59 | 68.42 | 44.99 | 68.03 | 41.97 | 72.49 | 42.10 | 64.42 | 45.61 | 58.91 | 39.26 | 70.40 | 40.13 | 52.77 | 33.75 | 65.54 | 42.22 |
| | Spanish | 62.73 | 51.25 | 68.30 | 48.23 | 68.22 | 45.40 | 72.65 | 46.56 | 64.45 | 49.43 | 58.84 | 41.55 | 70.43 | 43.69 | 51.97 | 34.12 | 65.47 | 45.27 |
| | Arabic | 62.72 | 51.31 | 68.19 | 48.31 | 68.14 | 45.48 | 72.58 | 46.65 | 64.84 | 49.49 | 58.94 | 41.60 | 70.47 | 43.75 | 51.39 | 34.14 | 65.43 | 45.33 |
| | French | 62.60 | 49.87 | 67.11 | 46.59 | 67.61 | 43.53 | 71.57 | 43.96 | 65.34 | 47.36 | 58.50 | 40.12 | 69.69 | 41.52 | 47.88 | 33.88 | 64.58 | 43.55 |
| | Russian | 62.61 | 46.39 | 67.10 | 43.05 | 67.61 | 39.41 | 71.59 | 39.47 | 65.33 | 43.01 | 58.48 | 37.76 | 69.73 | 38.04 | 48.00 | 33.59 | 64.60 | 40.22 |
| | Portuguese | 62.69 | 52.15 | 67.59 | 49.53 | 68.09 | 46.75 | 72.18 | 48.23 | 64.98 | 50.51 | 58.59 | 42.33 | 70.43 | 45.05 | 49.49 | 34.44 | 65.04 | 46.39 |
| | German | 54.89 | 61.62 | 63.87 | 63.80 | 50.58 | 64.15 | 63.93 | 67.75 | 46.36 | 63.87 | 50.13 | 55.63 | 55.61 | 65.32 | 64.98 | 41.63 | 57.34 | 61.22 |
| Fast-DetectGPT | English | 49.30 | 37.70 | 46.12 | 36.45 | 48.77 | 36.15 | 54.59 | 56.18 | 48.90 | 38.77 | 58.81 | 49.99 | 49.56 | 36.04 | 56.97 | 37.43 | 52.94 | 41.61 |
| | Chinese | 49.02 | 40.38 | 46.03 | 38.48 | 48.65 | 39.36 | 57.04 | 61.32 | 49.77 | 43.55 | 61.16 | 57.50 | 49.87 | 39.21 | 56.74 | 42.64 | 53.58 | 46.23 |
| | Spanish | 49.31 | 39.13 | 46.04 | 37.36 | 48.73 | 37.79 | 54.68 | 59.53 | 49.08 | 41.27 | 59.26 | 54.14 | 49.61 | 37.68 | 56.76 | 40.03 | 53.00 | 44.11 |
| | Arabic | 44.59 | 35.22 | 42.10 | 34.99 | 44.56 | 34.48 | 64.44 | 49.15 | 49.39 | 35.57 | 63.87 | 43.06 | 45.28 | 34.38 | 50.69 | 35.10 | 51.88 | 37.98 |
| | French | 47.51 | 39.13 | 44.79 | 37.37 | 47.93 | 37.79 | 62.14 | 59.54 | 50.88 | 41.26 | 65.10 | 54.18 | 48.54 | 37.68 | 55.69 | 40.03 | 54.15 | 44.12 |
| | Russian | 47.31 | 38.76 | 44.39 | 37.08 | 47.48 | 37.25 | 62.90 | 59.00 | 50.67 | 40.43 | 65.19 | 53.12 | 48.30 | 37.10 | 55.27 | 39.36 | 54.00 | 43.47 |
| | Portuguese | 49.32 | 39.84 | 46.03 | 38.03 | 48.72 | 38.79 | 54.70 | 60.85 | 49.04 | 42.73 | 59.29 | 56.42 | 49.54 | 38.73 | 56.71 | 41.80 | 52.98 | 45.51 |
| | German | 49.29 | 40.54 | 46.06 | 38.58 | 48.66 | 39.46 | 54.60 | 61.41 | 48.68 | 43.62 | 58.73 | 57.69 | 49.52 | 39.43 | 57.00 | 42.83 | 52.87 | 46.37 |
| Binoculars | English | 66.85 | 42.03 | 60.47 | 55.52 | 70.23 | 49.88 | 80.37 | 51.92 | 78.09 | 53.98 | 74.11 | 55.98 | 74.97 | 55.28 | 82.49 | 45.80 | 73.88 | 51.47 |
| | Chinese | 57.27 | 42.03 | 72.43 | 55.52 | 71.37 | 49.88 | 67.45 | 51.92 | 75.96 | 53.98 | 76.85 | 55.98 | 77.39 | 55.28 | 67.21 | 45.80 | 71.07 | 51.47 |
| | Spanish | 64.54 | 48.17 | 68.66 | 65.76 | 75.35 | 58.38 | 75.61 | 58.01 | 81.45 | 62.33 | 79.87 | 65.39 | 81.03 | 64.31 | 78.35 | 53.48 | 75.87 | 59.75 |
| | Arabic | 65.45 | 57.27 | 52.27 | 72.43 | 58.95 | 71.37 | 80.45 | 67.45 | 67.31 | 75.96 | 62.21 | 76.85 | 63.45 | 77.39 | 78.74 | 67.21 | 66.92 | 71.07 |
| | French | 64.54 | 48.17 | 68.66 | 65.76 | 75.35 | 58.38 | 75.61 | 58.01 | 81.45 | 62.33 | 79.87 | 65.39 | 81.03 | 64.31 | 78.35 | 53.48 | 75.87 | 59.75 |
| | Russian | 64.54 | 48.17 | 68.66 | 65.76 | 75.35 | 58.38 | 75.61 | 58.01 | 81.45 | 62.33 | 79.87 | 65.39 | 81.03 | 64.31 | 78.35 | 53.48 | 75.87 | 59.75 |
| | Portuguese | 64.54 | 48.17 | 68.66 | 65.76 | 75.35 | 58.38 | 75.61 | 58.01 | 81.45 | 62.33 | 79.87 | 65.39 | 81.03 | 64.31 | 78.35 | 53.48 | 75.87 | 59.75 |
| | German | 66.85 | 57.27 | 60.47 | 72.43 | 70.23 | 71.37 | 80.37 | 67.45 | 78.09 | 75.96 | 74.11 | 76.85 | 74.97 | 77.39 | 82.49 | 67.21 | 73.88 | 71.07 |
| ReviseDetect | English | 89.30 | 0.00 | 43.52 | 0.00 | 87.30 | 0.00 | 81.61 | 0.00 | 86.21 | 0.00 | 83.96 | 0.00 | 87.04 | 0.00 | 85.23 | 0.00 | 82.31 | 0.00 |
| | Chinese | 76.32 | 0.00 | 80.46 | 0.00 | 71.38 | 0.00 | 72.91 | 0.00 | 72.01 | 0.00 | 71.99 | 0.00 | 70.97 | 0.00 | 73.22 | 0.00 | 73.38 | 0.00 |
| | Spanish | 85.12 | 0.00 | 40.81 | 0.00 | 88.31 | 0.00 | 68.54 | 0.00 | 83.64 | 0.00 | 79.46 | 0.00 | 88.19 | 0.00 | 79.57 | 0.00 | 78.54 | 0.00 |
| | Arabic | 89.80 | 0.00 | 47.50 | 0.00 | 83.48 | 0.00 | 86.93 | 0.00 | 84.97 | 0.00 | 83.98 | 0.00 | 83.33 | 0.00 | 85.44 | 0.00 | 82.25 | 0.00 |
| | French | 87.83 | 0.00 | 41.97 | 0.00 | 88.32 | 0.00 | 75.39 | 0.00 | 85.59 | 0.00 | 82.46 | 0.00 | 88.02 | 0.00 | 82.90 | 0.00 | 80.88 | 0.00 |
| | Russian | 89.17 | 0.00 | 43.33 | 0.00 | 87.47 | 0.00 | 80.87 | 0.00 | 86.38 | 0.00 | 83.92 | 0.00 | 87.23 | 0.00 | 84.82 | 0.00 | 82.19 | 0.00 |
| | Portuguese | 84.83 | 0.00 | 40.80 | 0.00 | 88.25 | 0.00 | 67.78 | 0.00 | 83.49 | 0.00 | 79.12 | 0.00 | 88.07 | 0.00 | 79.15 | 0.00 | 78.26 | 0.00 |
| | German | 89.61 | 0.00 | 44.50 | 0.00 | 86.38 | 0.00 | 83.18 | 0.00 | 86.14 | 0.00 | 84.12 | 0.00 | 86.04 | 0.00 | 85.68 | 0.00 | 82.45 | 0.00 |
| GECScore | English | 92.78 | 71.34 | 56.01 | 54.89 | 89.82 | 72.16 | 75.04 | 42.50 | 88.90 | 66.40 | 83.03 | 58.61 | 91.05 | 65.30 | 84.94 | 56.44 | 83.23 | 61.36 |
| | Chinese | 56.29 | 89.43 | 80.09 | 59.59 | 35.62 | 75.15 | 59.07 | 89.12 | 41.90 | 77.71 | 46.27 | 79.83 | 44.78 | 78.67 | 49.23 | 76.22 | 54.32 | 79.14 |
| | Spanish | 89.77 | 69.93 | 55.24 | 54.89 | 91.44 | 70.51 | 64.31 | 42.43 | 88.64 | 64.68 | 78.07 | 58.38 | 90.36 | 63.46 | 81.86 | 55.57 | 80.55 | 60.33 |
| | Arabic | 88.31 | 91.25 | 59.81 | 55.49 | 73.26 | 91.24 | 89.31 | 67.97 | 76.30 | 89.23 | 78.71 | 80.04 | 77.14 | 91.18 | 74.83 | 83.50 | 78.22 | 81.80 |
| | French | 91.50 | 68.31 | 55.54 | 54.89 | 91.25 | 68.96 | 68.80 | 42.25 | 89.28 | 63.62 | 80.40 | 58.25 | 91.38 | 62.06 | 83.78 | 54.48 | 82.05 | 59.42 |
| | Russian | 92.92 | 71.24 | 56.17 | 54.89 | 89.13 | 71.94 | 77.19 | 42.50 | 88.54 | 66.19 | 83.58 | 58.60 | 90.46 | 65.19 | 85.05 | 56.34 | 83.42 | 61.26 |
| | Portuguese | 91.71 | 72.12 | 55.62 | 54.89 | 91.04 | 73.35 | 69.76 | 42.62 | 89.31 | 67.32 | 80.70 | 58.82 | 91.50 | 66.60 | 84.00 | 57.04 | 82.26 | 62.04 |
| | German | 92.64 | 68.25 | 55.93 | 54.89 | 90.13 | 68.85 | 74.13 | 42.23 | 89.18 | 63.62 | 82.56 | 58.22 | 91.17 | 61.88 | 84.89 | 54.47 | 83.11 | 59.36 |
| Lastde++ | English | 37.52 | 33.39 | 34.62 | 33.50 | 35.52 | 33.45 | 44.92 | 34.15 | 37.53 | 33.60 | 45.40 | 33.92 | 33.36 | 33.60 | 39.45 | 34.06 | 39.14 | 33.71 |
| | Chinese | 33.69 | 33.30 | 33.88 | 33.41 | 33.59 | 33.38 | 35.78 | 33.77 | 34.31 | 33.53 | 35.23 | 33.75 | 33.97 | 33.44 | 34.99 | 33.75 | 34.45 | 33.54 |
| | Spanish | 35.51 | 33.30 | 35.04 | 33.41 | 35.31 | 33.37 | 43.13 | 33.78 | 37.54 | 33.54 | 42.65 | 33.75 | 34.62 | 33.45 | 39.97 | 33.71 | 38.27 | 33.54 |
| | Arabic | 36.72 | 33.32 | 34.38 | 33.38 | 35.79 | 33.37 | 44.53 | 33.69 | 37.99 | 33.48 | 45.01 | 33.52 | 34.22 | 33.43 | 40.01 | 33.51 | 39.07 | 33.46 |
| | French | 36.52 | 33.33 | 34.57 | 33.47 | 35.70 | 33.43 | 44.44 | 33.99 | 38.17 | 33.57 | 44.80 | 33.87 | 34.40 | 33.55 | 40.05 | 33.93 | 39.04 | 33.65 |
| | Russian | 37.54 | 33.35 | 34.56 | 33.49 | 35.55 | 33.44 | 44.87 | 34.01 | 37.72 | 33.58 | 45.25 | 33.88 | 33.48 | 33.54 | 39.90 | 33.98 | 39.18 | 33.66 |
| | Portuguese | 34.83 | 33.30 | 34.43 | 33.41 | 34.80 | 33.37 | 41.33 | 33.78 | 36.68 | 33.52 | 40.38 | 33.74 | 34.79 | 33.46 | 38.87 | 33.72 | 37.20 | 33.54 |
| | German | 36.43 | 33.30 | 34.74 | 33.41 | 35.67 | 33.36 | 44.20 | 33.77 | 38.10 | 33.53 | 44.55 | 33.75 | 34.48 | 33.44 | 40.19 | 33.75 | 38.99 | 33.54 |
| RepreGuard | English | 96.97 | 95.62 | 95.89 | 96.63 | 96.90 | 96.03 | 96.91 | 86.18 | 96.41 | 98.15 | 91.82 | 93.34 | 97.97 | 98.14 | 94.68 | 93.84 | 95.91 | 94.87 |
| | Chinese | 81.96 | 92.48 | 94.92 | 91.63 | 66.69 | 70.33 | 66.70 | 70.05 | 66.68 | 68.54 | 66.69 | 68.04 | 66.68 | 68.99 | 66.67 | 66.69 | 70.88 | 72.99 |
| | Spanish | 6.17 | 3.83 | 0.00 | 0.00 | 97.05 | 97.15 | 0.03 | 0.03 | 36.40 | 25.91 | 0.00 | 0.00 | 91.62 | 87.22 | 8.15 | 6.48 | 41.64 | 38.87 |
| | Arabic | 15.81 | 2.86 | 44.99 | 18.51 | 76.62 | 42.91 | 82.84 | 71.06 | 81.19 | 68.30 | 74.29 | 48.86 | 84.88 | 62.02 | 66.49 | 71.66 | 70.03 | 53.74 |
| | French | 67.76 | 67.76 | 67.36 | 67.37 | 71.25 | 71.33 | 81.90 | 82.08 | 96.46 | 96.49 | 77.18 | 77.26 | 74.00 | 74.07 | 67.93 | 67.97 | 74.47 | 74.53 |
| | Russian | 0.00 | 0.00 | 0.00 | 0.00 | 0.03 | 0.00 | 17.01 | 0.27 | 0.00 | 0.00 | 67.15 | 9.23 | 0.00 | 0.00 | 0.17 | 0.00 | 18.42 | 1.24 |
| | Portuguese | 0.10 | 0.03 | 0.00 | 0.00 | 65.22 | 39.93 | 0.00 | 0.00 | 2.96 | 0.37 | 0.00 | 0.00 | 84.55 | 80.34 | 0.07 | 0.00 | 27.97 | 21.03 |
| | German | 47.03 | 28.79 | 0.10 | 0.00 | 36.97 | 17.35 | 1.32 | 0.76 | 6.92 | 3.28 | 1.00 | 0.27 | 30.20 | 8.03 | 89.71 | 86.40 | 34.58 | 24.20 |
| X-Rob-Classifier | English | 99.96 | 99.49 | 96.38 | 96.21 | 96.63 | 81.36 | 94.81 | 93.21 | 98.92 | 98.07 | 97.52 | 96.26 | 99.34 | 99.29 | 98.67 | 98.21 | 95.73 | 92.19 |
| | Chinese | 99.02 | 98.88 | 97.32 | 94.72 | 94.71 | 70.72 | 97.75 | 93.41 | 97.35 | 83.95 | 93.74 | 72.61 | 96.27 | 81.85 | 97.09 | 85.26 | 97.32 | 94.72 |
| | Spanish | 98.85 | 98.79 | 96.90 | 95.93 | 99.88 | 99.73 | 98.80 | 98.70 | 99.13 | 99.10 | 97.25 | 95.68 | 99.67 | 99.50 | 99.32 | 99.26 | 98.26 | 96.87 |
| | Arabic | 99.38 | 99.36 | 89.76 | 62.50 | 95.17 | 79.98 | 99.75 | 99.49 | 97.79 | 91.42 | 90.73 | 66.78 | 95.70 | 73.21 | 95.06 | 74.92 | 92.65 | 64.16 |
| | French | 99.36 | 99.31 | 96.67 | 87.06 | 97.58 | 84.56 | 98.77 | 97.91 | 99.90 | 99.53 | 96.72 | 84.89 | 98.73 | 95.75 | 98.69 | 89.57 | 97.68 | 90.99 |
| | Russian | 98.97 | 95.40 | 96.27 | 96.21 | 93.78 | 59.61 | 98.78 | 98.05 | 99.04 | 97.09 | 99.65 | 99.47 | 99.66 | 99.47 | 97.93 | 84.00 | 96.54 | 80.05 |
| | Portuguese | 98.84 | 95.95 | 97.25 | 97.04 | 98.55 | 86.74 | 98.52 | 98.37 | 99.54 | 99.45 | 98.38 | 98.10 | 99.91 | 99.62 | 98.52 | 91.74 | 97.87 | 93.74 |
| | German | 99.58 | 99.42 | 97.39 | 97.21 | 99.07 | 98.99 | 97.51 | 91.88 | 99.52 | 99.44 | 97.32 | 94.97 | 99.49 | 99.40 | 99.75 | 99.52 | 96.91 | 94.11 |
| mDeBERTa-Classifier | English | 99.96 | 99.49 | 97.07 | 96.94 | 98.62 | 98.11 | 98.99 | 98.92 | 99.30 | 99.26 | 98.01 | 97.94 | 99.62 | 99.44 | 98.82 | 98.67 | 97.60 | 97.01 |
| | Chinese | 99.09 | 99.09 | 97.47 | 95.81 | 97.26 | 95.06 | 99.19 | 99.13 | 98.22 | 97.73 | 95.57 | 94.35 | 99.05 | 99.01 | 97.82 | 97.03 | 97.47 | 95.81 |
| | Spanish | 99.34 | 99.32 | 98.52 | 98.20 | 99.95 | 99.51 | 99.40 | 99.24 | 99.77 | 99.49 | 99.17 | 99.13 | 99.89 | 99.50 | 99.67 | 99.41 | 98.91 | 98.90 |
| | Arabic | 99.66 | 99.43 | 98.16 | 97.94 | 98.77 | 98.56 | 99.94 | 99.64 | 99.17 | 99.14 | 97.09 | 95.91 | 99.22 | 99.22 | 97.97 | 97.07 | 96.28 | 91.84 |
| | French | 99.67 | 99.48 | 98.39 | 98.27 | 99.54 | 99.41 | 99.83 | 99.46 | 99.97 | 99.36 | 98.99 | 98.86 | 99.87 | 99.49 | 99.81 | 99.36 | 98.56 | 98.24 |
| | Russian | 99.09 | 98.64 | 97.20 | 96.77 | 93.44 | 59.12 | 99.57 | 99.42 | 98.72 | 97.36 | 99.89 | 99.48 | 99.58 | 99.45 | 97.92 | 86.07 | 97.04 | 93.60 |
| | Portuguese | 99.70 | 99.48 | 98.92 | 98.87 | 98.68 | 96.20 | 99.58 | 99.42 | 99.69 | 99.47 | 99.47 | 99.35 | 99.97 | 99.42 | 98.73 | 98.31 | 97.60 | 95.59 |
| | German | 99.65 | 99.47 | 98.87 | 98.72 | 98.67 | 98.13 | 99.35 | 99.27 | 99.55 | 99.44 | 98.04 | 97.90 | 99.38 | 99.32 | 99.92 | 99.51 | 97.54 | 94.59 |
| Biscope | English | 96.71 | 95.00 | 56.46 | 37.18 | 87.56 | 46.43 | 86.19 | 66.38 | 84.98 | 43.92 | 84.54 | 47.26 | 83.43 | 39.53 | 81.03 | 39.41 | 80.77 | 52.50 |
| | Chinese | 68.92 | 33.33 | 95.43 | 91.88 | 47.45 | 34.93 | 75.40 | 33.33 | 55.03 | 34.96 | 55.21 | 33.33 | 43.71 | 34.60 | 43.91 | 34.32 | 61.65 | 41.46 |
| | Spanish | 85.85 | 79.23 | 63.59 | 39.02 | 96.90 | 93.92 | 77.29 | 57.23 | 96.42 | 91.27 | 86.09 | 63.41 | 96.56 | 92.67 | 93.12 | 81.08 | 84.03 | 72.24 |
| | Arabic | 92.24 | 78.68 | 82.16 | 46.47 | 82.17 | 53.12 | 91.23 | 73.00 | 80.36 | 48.44 | 85.72 | 33.33 | 80.55 | 48.74 | 76.32 | 44.61 | 78.95 | 50.57 |
| | French | 71.86 | 59.89 | 78.55 | 48.81 | 95.86 | 92.47 | 71.73 | 39.33 | 96.65 | 92.58 | 82.86 | 58.30 | 96.61 | 92.27 | 92.60 | 81.14 | 83.62 | 73.05 |
| | Russian | 91.18 | 78.14 | 78.76 | 57.77 | 93.86 | 78.68 | 89.08 | 76.07 | 91.49 | 68.11 | 90.47 | 71.16 | 93.36 | 76.45 | 88.76 | 62.99 | 83.32 | 65.19 |
| | Portuguese | 72.19 | 45.89 | 76.48 | 44.18 | 95.89 | 92.25 | 75.51 | 40.08 | 96.33 | 92.07 | 83.68 | 58.46 | 96.84 | 92.76 | 92.31 | 80.34 | 83.42 | 70.56 |
| | German | 86.31 | 77.82 | 74.78 | 38.58 | 96.32 | 91.32 | 80.10 | 56.41 | 95.86 | 89.45 | 83.18 | 57.44 | 96.07 | 91.25 | 93.28 | 83.66 | 83.79 | 71.12 |

Table 15: Performance Comparison of Different Detectors on In-Distribution and Language Generalization in BINARY Task. Detector performance is visualized using teal color gradients, with darker intensities indicating superior detection capabilities.

| Detectors↓ | Train↓ Test→ | English | | Chinese | | Spanish | | Arabic | | French | | Russian | | Portuguese | | German | | Average | |
|---|---|---|---|---|---|---|---|---|---|---|---|---|---|---|---|---|---|---|---|
| | Metrics | $F_1^B$ | $F_1^F$ | $F_1^B$ | $F_1^F$ | $F_1^B$ | $F_1^F$ | $F_1^B$ | $F_1^F$ | $F_1^B$ | $F_1^F$ | $F_1^B$ | $F_1^F$ | $F_1^B$ | $F_1^F$ | $F_1^B$ | $F_1^F$ | $F_1^B$ | $F_1^F$ |
| Log-Likelihood | English | 35.68 | 35.68 | 30.77 | 30.77 | 28.81 | 28.81 | 35.31 | 35.31 | 25.43 | 25.43 | 23.96 | 23.96 | 29.38 | 29.38 | 38.92 | 38.92 | 31.95 | 31.95 |
| | Chinese | 43.83 | 35.61 | 39.15 | 30.91 | 30.60 | 27.13 | 33.34 | 34.80 | 25.29 | 24.29 | 22.17 | 23.47 | 27.11 | 28.00 | 43.05 | 38.26 | 34.52 | 31.29 |
| | Spanish | 32.58 | 32.58 | 26.20 | 26.20 | 40.28 | 40.28 | 37.48 | 37.48 | 38.46 | 38.46 | 31.15 | 31.15 | 40.18 | 40.18 | 35.65 | 35.65 | 35.92 | 35.92 |
| | Arabic | 36.21 | 30.41 | 32.64 | 24.17 | 46.01 | 40.28 | 50.27 | 35.99 | 44.96 | 39.03 | 35.19 | 33.35 | 49.05 | 40.61 | 45.40 | 32.03 | 44.49 | 35.28 |
| | French | 37.71 | 30.69 | 34.06 | 24.28 | 43.08 | 40.38 | 48.56 | 36.19 | 42.28 | 38.99 | 33.45 | 33.18 | 44.89 | 40.71 | 46.80 | 32.43 | 43.63 | 35.38 |
| | Russian | 25.82 | 22.85 | 24.98 | 19.96 | 36.71 | 29.38 | 36.59 | 22.76 | 39.32 | 30.27 | 44.89 | 33.07 | 41.07 | 30.24 | 24.97 | 19.42 | 35.94 | 26.76 |
| | Portuguese | 35.11 | 30.00 | 31.63 | 23.79 | 46.70 | 39.96 | 50.12 | 35.66 | 45.77 | 38.80 | 36.73 | 33.74 | 50.54 | 40.45 | 43.12 | 31.05 | 44.36 | 35.00 |
| | German | 41.13 | 35.07 | 37.24 | 29.46 | 39.57 | 35.73 | 43.24 | 36.93 | 35.85 | 32.17 | 26.13 | 26.34 | 37.22 | 35.27 | 52.32 | 39.78 | 40.79 | 34.49 |
| Log-Rank | English | 35.50 | 35.50 | 31.90 | 31.90 | 33.83 | 33.83 | 36.87 | 36.87 | 30.04 | 30.04 | 26.42 | 26.42 | 34.08 | 34.08 | 39.24 | 39.24 | 34.01 | 34.01 |
| | Chinese | 44.09 | 35.47 | 41.28 | 32.12 | 36.25 | 32.39 | 37.56 | 36.52 | 31.78 | 28.56 | 25.17 | 25.77 | 33.70 | 33.04 | 49.10 | 39.29 | 38.80 | 33.49 |
| | Spanish | 34.33 | 34.33 | 29.72 | 29.72 | 39.56 | 39.56 | 38.28 | 38.28 | 37.43 | 37.43 | 31.11 | 31.11 | 39.64 | 39.64 | 35.21 | 35.21 | 36.10 | 36.10 |
| | Arabic | 36.24 | 31.85 | 33.42 | 26.54 | 46.38 | 39.58 | 50.53 | 36.68 | 46.06 | 38.43 | 39.57 | 34.13 | 50.86 | 40.37 | 38.50 | 29.80 | 44.04 | 35.32 |
| | French | 39.32 | 32.16 | 37.62 | 27.12 | 41.52 | 39.92 | 48.68 | 37.12 | 41.22 | 38.53 | 35.49 | 33.64 | 44.20 | 40.51 | 44.09 | 30.73 | 43.35 | 35.58 |
| | Russian | 28.73 | 25.26 | 27.19 | 21.26 | 37.39 | 30.35 | 39.92 | 25.11 | 39.99 | 31.30 | 43.29 | 31.79 | 40.97 | 30.75 | 24.53 | 19.29 | 36.62 | 27.43 |
| | Portuguese | 37.51 | 32.13 | 35.49 | 27.07 | 45.33 | 39.85 | 51.34 | 37.05 | 44.33 | 38.50 | 37.63 | 33.77 | 49.77 | 40.53 | 41.81 | 30.59 | 44.47 | 35.56 |
| | German | 43.29 | 35.49 | 41.27 | 31.87 | 38.03 | 34.52 | 41.16 | 37.03 | 34.46 | 30.67 | 26.51 | 26.80 | 36.27 | 34.67 | 50.87 | 39.01 | 40.49 | 34.26 |
| DetectLLM-LRR | English | 33.87 | 33.87 | 36.95 | 36.95 | 37.10 | 37.10 | 38.70 | 38.70 | 34.97 | 34.97 | 31.48 | 31.48 | 38.45 | 38.45 | 28.02 | 28.02 | 35.34 | 35.34 |
| | Chinese | 42.74 | 33.88 | 46.90 | 36.93 | 41.16 | 37.09 | 49.94 | 38.74 | 40.54 | 34.83 | 38.35 | 31.52 | 45.69 | 38.37 | 37.09 | 28.45 | 43.72 | 35.36 |
| | Spanish | 33.44 | 33.44 | 36.88 | 36.88 | 36.28 | 36.28 | 38.39 | 38.39 | 33.15 | 33.15 | 31.14 | 31.14 | 37.46 | 37.46 | 31.78 | 31.78 | 35.11 | 35.11 |
| | Arabic | 42.25 | 34.07 | 45.52 | 35.66 | 40.71 | 36.19 | 51.98 | 37.37 | 41.15 | 35.34 | 39.46 | 30.82 | 44.70 | 37.26 | 30.59 | 23.83 | 43.01 | 34.29 |
| | French | 38.94 | 33.98 | 37.97 | 36.69 | 39.12 | 37.08 | 35.12 | 38.48 | 37.79 | 35.18 | 32.15 | 31.37 | 40.03 | 38.39 | 30.22 | 26.92 | 36.88 | 35.19 |
| | Russian | 42.19 | 34.09 | 46.62 | 35.83 | 39.33 | 36.47 | 50.10 | 37.61 | 39.94 | 35.52 | 37.91 | 31.04 | 43.58 | 37.56 | 33.85 | 24.42 | 42.78 | 34.54 |
| | Portuguese | 33.74 | 33.74 | 37.06 | 37.06 | 36.86 | 36.86 | 38.70 | 38.70 | 34.27 | 34.27 | 31.27 | 31.27 | 38.10 | 38.10 | 29.88 | 29.88 | 35.33 | 35.33 |
| | German | 35.31 | 29.28 | 35.45 | 34.08 | 35.23 | 27.11 | 25.62 | 34.92 | 29.96 | 24.78 | 27.43 | 26.85 | 33.65 | 29.69 | 42.06 | 35.54 | 33.56 | 30.71 |
| Fast-DetectGPT | English | 26.33 | 26.33 | 24.59 | 24.59 | 25.88 | 25.88 | 30.78 | 30.78 | 26.36 | 26.36 | 32.51 | 32.51 | 26.54 | 26.54 | 30.90 | 30.90 | 28.59 | 28.59 |
| | Chinese | 26.00 | 20.47 | 30.70 | 19.84 | 25.80 | 19.85 | 42.01 | 36.73 | 27.63 | 21.36 | 40.49 | 30.56 | 27.48 | 19.57 | 30.06 | 20.46 | 32.23 | 24.20 |
| | Spanish | 26.42 | 26.42 | 24.58 | 24.58 | 26.06 | 26.06 | 29.61 | 29.61 | 26.23 | 26.23 | 31.58 | 31.58 | 26.48 | 26.48 | 30.87 | 30.87 | 28.34 | 28.34 |
| | Arabic | 22.15 | 19.12 | 24.22 | 18.43 | 22.61 | 18.09 | 43.45 | 29.42 | 23.47 | 19.51 | 35.63 | 25.78 | 22.86 | 18.06 | 23.70 | 18.87 | 28.20 | 21.27 |
| | French | 24.52 | 24.52 | 22.91 | 22.91 | 24.48 | 24.48 | 34.48 | 34.48 | 27.15 | 27.15 | 34.92 | 34.92 | 25.05 | 25.05 | 28.63 | 28.63 | 28.55 | 28.55 |
| | Russian | 26.76 | 22.18 | 30.33 | 21.02 | 28.12 | 22.08 | 43.14 | 33.98 | 30.90 | 24.78 | 44.35 | 32.97 | 29.83 | 22.16 | 33.11 | 24.61 | 34.42 | 26.27 |
| | Portuguese | 24.57 | 24.57 | 23.04 | 23.04 | 24.67 | 24.67 | 34.39 | 34.39 | 27.21 | 27.21 | 34.94 | 34.94 | 25.18 | 25.18 | 28.84 | 28.84 | 28.63 | 28.63 |
| | German | 31.14 | 25.60 | 33.06 | 23.96 | 31.17 | 25.72 | 35.04 | 33.32 | 34.46 | 27.34 | 41.86 | 34.68 | 33.80 | 25.96 | 38.73 | 30.29 | 35.97 | 29.05 |
| Binoculars | English | 43.24 | 31.14 | 43.25 | 39.12 | 45.02 | 39.11 | 54.02 | 37.33 | 51.09 | 42.02 | 48.58 | 42.08 | 49.87 | 42.32 | 53.08 | 37.45 | 49.69 | 39.00 |
| | Chinese | 29.94 | 25.97 | 44.75 | 36.27 | 38.03 | 32.16 | 35.01 | 31.95 | 39.31 | 34.60 | 42.95 | 36.35 | 42.78 | 35.67 | 31.42 | 29.26 | 38.36 | 32.97 |
| | Spanish | 34.92 | 25.97 | 48.33 | 36.27 | 44.45 | 32.16 | 39.45 | 31.95 | 45.56 | 34.60 | 48.86 | 36.35 | 50.37 | 35.67 | 35.94 | 29.26 | 44.06 | 32.97 |
| | Arabic | 42.09 | 34.80 | 42.85 | 36.63 | 49.19 | 40.39 | 48.90 | 41.39 | 53.40 | 43.97 | 51.56 | 42.82 | 52.89 | 43.35 | 51.80 | 43.23 | 49.36 | 40.83 |
| | French | 37.93 | 31.14 | 47.61 | 39.12 | 48.61 | 39.11 | 42.73 | 37.33 | 50.51 | 42.02 | 51.05 | 42.08 | 53.54 | 42.32 | 42.82 | 37.45 | 47.26 | 39.00 |
| | Russian | 34.92 | 25.97 | 48.33 | 36.27 | 44.45 | 32.16 | 39.45 | 31.95 | 45.56 | 34.60 | 48.86 | 36.35 | 50.37 | 35.67 | 35.94 | 29.26 | 44.06 | 32.97 |
| | Portuguese | 34.92 | 25.97 | 48.33 | 36.27 | 44.45 | 32.16 | 39.45 | 31.95 | 45.56 | 34.60 | 48.86 | 36.35 | 50.37 | 35.67 | 35.94 | 29.26 | 44.06 | 32.97 |
| | German | 41.84 | 31.14 | 48.49 | 39.12 | 52.17 | 39.11 | 48.22 | 37.33 | 55.56 | 42.02 | 55.05 | 42.08 | 57.52 | 42.32 | 48.85 | 37.45 | 51.60 | 39.00 |
| Revise-Detect | English | 62.48 | 50.46 | 35.91 | 33.22 | 57.01 | 50.34 | 58.14 | 44.73 | 58.86 | 49.19 | 56.95 | 47.21 | 56.91 | 51.26 | 57.60 | 47.36 | 56.46 | 47.20 |
| | Chinese | 32.73 | 42.31 | 52.64 | 44.77 | 22.53 | 24.01 | 30.47 | 42.76 | 25.36 | 31.76 | 24.14 | 30.96 | 22.45 | 28.87 | 29.40 | 34.30 | 32.48 | 36.44 |
| | Spanish | 56.77 | 46.50 | 33.76 | 32.58 | 58.91 | 49.55 | 49.14 | 37.56 | 56.09 | 45.46 | 53.81 | 42.88 | 59.46 | 49.76 | 53.72 | 42.73 | 53.66 | 44.04 |
| | Arabic | 62.55 | 50.55 | 36.02 | 33.24 | 56.85 | 50.32 | 58.21 | 44.90 | 58.78 | 49.19 | 57.01 | 47.26 | 56.74 | 51.29 | 57.55 | 47.42 | 56.43 | 47.24 |
| | French | 58.92 | 48.91 | 34.40 | 32.88 | 58.92 | 50.33 | 53.71 | 41.96 | 58.30 | 47.97 | 56.02 | 45.79 | 58.97 | 51.03 | 55.91 | 45.32 | 55.27 | 46.08 |
| | Russian | 59.99 | 49.07 | 34.64 | 32.90 | 59.10 | 50.35 | 54.68 | 42.20 | 58.75 | 48.11 | 56.37 | 45.90 | 58.85 | 51.11 | 56.46 | 45.54 | 55.77 | 46.20 |
| | Portuguese | 55.09 | 44.54 | 33.52 | 32.47 | 58.45 | 48.54 | 46.57 | 34.68 | 54.85 | 43.97 | 52.64 | 41.58 | 59.35 | 48.66 | 52.24 | 40.89 | 52.61 | 42.65 |
| | German | 62.56 | 50.68 | 36.04 | 33.27 | 56.71 | 50.29 | 58.38 | 45.14 | 58.77 | 49.25 | 57.02 | 47.33 | 56.66 | 51.26 | 57.44 | 47.44 | 56.40 | 47.29 |
| GECScore | English | 64.44 | 47.96 | 48.34 | 29.96 | 16.84 | 0.00 | 51.36 | 33.50 | 59.24 | 47.54 | 16.84 | 0.00 | 51.61 | 39.30 | 54.66 | 43.70 | 55.13 | 48.89 |
| | Chinese | 48.08 | 41.17 | 52.24 | 40.77 | 20.58 | 0.00 | 43.50 | 47.65 | 41.73 | 33.19 | 20.58 | 0.00 | 41.57 | 29.49 | 48.77 | 34.53 | 41.23 | 44.62 |
| | Spanish | 17.41 | 0.00 | 16.67 | 16.67 | 17.41 | 0.00 | 16.67 | 16.67 | 17.41 | 0.00 | 16.68 | 16.68 | 16.99 | 16.99 | 16.67 | 0.00 | 29.01 | 29.01 |
| | Arabic | 64.55 | 49.60 | 48.62 | 30.18 | 16.80 | 0.00 | 59.61 | 42.15 | 56.69 | 47.13 | 16.80 | 0.00 | 49.04 | 37.86 | 54.96 | 45.46 | 57.24 | 51.95 |
| | French | 63.92 | 45.32 | 48.38 | 28.95 | 17.41 | 0.00 | 49.42 | 30.68 | 60.54 | 45.05 | 17.41 | 0.00 | 53.09 | 36.63 | 55.21 | 40.49 | 55.94 | 47.05 |
| | Russian | 16.67 | 16.67 | 16.67 | 16.67 | 16.68 | 16.67 | 19.49 | 16.73 | 16.67 | 16.67 | 29.07 | 17.44 | 16.67 | 16.67 | 16.69 | 16.67 | 20.36 | 16.77 |
| | Portuguese | 65.60 | 53.45 | 48.53 | 33.74 | 14.56 | 0.00 | 48.72 | 36.78 | 62.85 | 53.11 | 14.56 | 0.00 | 59.13 | 47.99 | 57.41 | 48.38 | 53.94 | 49.48 |
| | German | 58.26 | 48.71 | 39.18 | 29.67 | 8.62 | 0.00 | 46.51 | 36.65 | 55.82 | 47.60 | 8.62 | 0.00 | 47.21 | 38.78 | 52.82 | 44.79 | 56.24 | 50.66 |
| Lastde++ | English | 16.70 | 16.70 | 16.76 | 16.76 | 16.66 | 16.66 | 16.67 | 16.67 | 16.67 | 16.67 | 16.67 | 16.67 | 16.67 | 16.67 | 16.67 | 16.67 | 16.68 | 16.68 |
| | Chinese | 23.63 | 16.87 | 25.55 | 16.94 | 26.26 | 16.98 | 25.91 | 17.11 | 24.30 | 16.84 | 24.30 | 16.83 | 25.81 | 17.14 | 24.79 | 17.12 | 25.24 | 16.98 |
| | Spanish | 18.63 | 16.99 | 19.15 | 17.00 | 20.24 | 17.12 | 20.78 | 17.19 | 19.38 | 16.84 | 19.10 | 16.86 | 20.35 | 17.24 | 20.05 | 17.14 | 19.75 | 17.05 |
| | Arabic | 17.68 | 17.68 | 16.19 | 16.19 | 16.39 | 16.39 | 16.88 | 16.88 | 16.07 | 16.07 | 16.89 | 16.89 | 16.24 | 16.24 | 16.45 | 16.45 | 16.65 | 16.65 |
| | French | 17.32 | 17.17 | 17.40 | 17.20 | 17.80 | 17.48 | 18.07 | 17.53 | 17.36 | 17.04 | 17.29 | 17.07 | 17.89 | 17.53 | 17.80 | 17.40 | 17.62 | 17.30 |
| | Russian | 24.29 | 16.78 | 25.42 | 16.87 | 21.83 | 16.91 | 21.93 | 17.62 | 21.75 | 17.05 | 24.73 | 17.52 | 21.20 | 16.97 | 21.39 | 17.45 | 23.11 | 17.15 |
| | Portuguese | 18.57 | 17.00 | 18.40 | 17.00 | 16.90 | 17.12 | 16.89 | 17.19 | 16.95 | 16.85 | 17.19 | 16.86 | 16.89 | 17.24 | 16.84 | 17.14 | 17.37 | 17.05 |
| | German | 16.70 | 16.70 | 16.81 | 16.81 | 16.66 | 16.66 | 16.67 | 16.67 | 16.66 | 16.66 | 16.66 | 16.66 | 16.66 | 16.66 | 16.67 | 16.67 | 16.69 | 16.69 |
| RepreGuard | English | 49.69 | 47.53 | 51.31 | 46.56 | 47.86 | 51.75 | 52.53 | 46.07 | 41.12 | 51.30 | 44.64 | 49.21 | 47.35 | 52.55 | 44.22 | 48.72 | 46.12 | 45.11 |
| | Chinese | 25.58 | 42.84 | 46.89 | 37.17 | 17.10 | 17.13 | 16.77 | 17.68 | 16.97 | 16.99 | 16.89 | 17.03 | 17.19 | 17.24 | 17.47 | 17.49 | 24.37 | 27.36 |
| | Spanish | 18.73 | 18.73 | 16.67 | 16.67 | 18.01 | 18.01 | 16.67 | 16.67 | 41.25 | 41.25 | 16.68 | 16.68 | 34.61 | 34.61 | 16.99 | 16.99 | 29.01 | 29.01 |
| | Arabic | 27.33 | 21.43 | 34.83 | 27.80 | 17.29 | 23.35 | 22.88 | 20.19 | 16.70 | 18.29 | 20.23 | 24.90 | 17.09 | 21.10 | 17.44 | 17.41 | 27.82 | 27.15 |
| | French | 30.77 | 30.77 | 16.67 | 16.67 | 45.14 | 45.14 | 16.74 | 16.74 | 16.88 | 16.88 | 16.68 | 16.68 | 46.75 | 46.75 | 19.65 | 19.65 | 31.92 | 31.92 |
| | Russian | 16.67 | 16.67 | 16.67 | 16.67 | 16.68 | 16.67 | 19.49 | 16.73 | 16.67 | 16.67 | 29.07 | 17.44 | 16.67 | 16.67 | 16.69 | 16.67 | 20.36 | 16.77 |
| | Portuguese | 27.17 | 27.17 | 16.67 | 16.67 | 25.37 | 25.37 | 16.67 | 16.67 | 41.87 | 41.87 | 16.68 | 16.68 | 17.17 | 17.17 | 17.09 | 17.09 | 28.98 | 28.98 |
| | German | 16.67 | 16.67 | 16.67 | 16.67 | 16.67 | 16.67 | 16.67 | 16.67 | 16.68 | 16.68 | 16.67 | 16.67 | 16.67 | 16.67 | 22.31 | 22.31 | 21.79 | 21.79 |
| X-Rob-Classifier | English | 96.66 | 95.77 | 57.26 | 56.60 | 74.12 | 59.58 | 76.28 | 56.48 | 83.32 | 70.89 | 64.13 | 48.60 | 82.63 | 65.98 | 77.25 | 62.97 | 77.58 | 59.22 |
| | Chinese | 52.16 | 46.22 | 88.12 | 87.91 | 47.02 | 40.47 | 76.58 | 59.34 | 56.61 | 54.12 | 56.19 | 51.98 | 53.40 | 54.10 | 58.30 | 51.94 | 61.42 | 37.08 |
| | Spanish | 69.44 | 53.54 | 57.48 | 53.45 | 91.81 | 87.77 | 75.35 | 64.55 | 80.96 | 58.44 | 66.23 | 58.34 | 72.74 | 54.74 | 69.66 | 47.45 | 68.10 | 52.70 |
| | Arabic | 79.79 | 62.73 | 52.33 | 53.77 | 55.83 | 40.64 | 92.06 | 87.56 | 60.85 | 52.60 | 57.73 | 49.54 | 61.81 | 25.37 | 60.70 | 55.10 | 64.52 | 29.05 |
| | French | 76.91 | 79.11 | 58.37 | 61.85 | 79.32 | 58.01 | 73.25 | 70.21 | 93.05 | 92.48 | 65.48 | 62.09 | 81.57 | 76.24 | 76.63 | 62.62 | 75.38 | 66.85 |
| | Russian | 72.16 | 53.87 | 62.77 | 53.44 | 67.66 | 44.07 | 75.77 | 45.54 | 74.67 | 53.71 | 89.56 | 86.91 | 73.57 | 54.41 | 70.88 | 48.19 | 68.04 | 39.19 |
| | Portuguese | 74.43 | 63.97 | 56.74 | 66.26 | 74.88 | 53.32 | 74.40 | 74.04 | 80.23 | 66.30 | 64.06 | 64.06 | 92.99 | 92.83 | 67.53 | 51.87 | 75.38 | 61.07 |
| | German | 72.13 | 70.79 | 56.99 | 58.52 | 63.80 | 45.13 | 70.36 | 65.12 | 70.16 | 54.44 | 58.65 | 56.71 | 64.47 | 54.07 | 88.33 | 85.65 | 62.52 | 45.37 |
| mDeBERTa-Classifier | English | 97.85 | 97.56 | 68.58 | 57.30 | 81.66 | 70.35 | 86.30 | 73.38 | 83.91 | 71.93 | 77.60 | 62.44 | 86.53 | 78.97 | 81.82 | 66.73 | 83.08 | 70.39 |
| | Chinese | 59.38 | 41.73 | 93.33 | 84.57 | 40.02 | 28.75 | 68.48 | 54.84 | 48.78 | 34.05 | 51.67 | 36.81 | 52.79 | 42.38 | 47.54 | 31.13 | 59.50 | 41.03 |
| | Spanish | 85.47 | 82.59 | 59.90 | 57.67 | 95.52 | 92.81 | 87.53 | 79.85 | 88.80 | 82.91 | 79.69 | 58.64 | 88.92 | 84.94 | 84.63 | 74.28 | 84.37 | 74.07 |
| | Arabic | 76.91 | 57.22 | 55.35 | 53.27 | 54.44 | 20.63 | 95.20 | 94.47 | 58.20 | 48.40 | 56.23 | 45.74 | 62.14 | 51.03 | 54.48 | 44.75 | 63.84 | 47.98 |
| | French | 86.59 | 82.69 | 60.82 | 63.22 | 84.84 | 23.89 | 89.11 | 81.76 | 96.21 | 95.50 | 74.77 | 52.55 | 89.52 | 84.43 | 86.32 | 77.82 | 84.35 | 68.28 |
| | Russian | 77.89 | 60.75 | 75.78 | 67.59 | 62.52 | 23.07 | 80.04 | 71.68 | 69.53 | 32.98 | 92.26 | 89.28 | 69.83 | 55.76 | 70.19 | 48.67 | 72.63 | 40.26 |
| | Portuguese | 90.32 | 89.13 | 70.14 | 64.44 | 84.03 | 32.06 | 90.16 | 84.28 | 87.97 | 83.36 | 82.21 | 61.32 | 96.08 | 94.93 | 82.01 | 66.05 | 85.18 | 68.66 |
| | German | 86.80 | 85.61 | 72.00 | 69.10 | 67.35 | 36.84 | 83.85 | 80.91 | 77.12 | 58.07 | 68.19 | 65.47 | 71.37 | 60.78 | 93.50 | 90.80 | 77.15 | 62.67 |
| Biscope | English | 80.52 | 68.88 | 24.22 | 18.92 | 50.73 | 27.70 | 59.05 | 33.04 | 45.40 | 31.35 | 50.42 | 26.94 | 45.54 | 26.92 | 43.44 | 22.29 | 53.38 | 31.59 |
| | Chinese | 36.40 | 16.67 | 71.48 | 51.96 | 28.89 | 17.34 | 35.49 | 16.67 | 35.08 | 17.49 | 29.06 | 16.67 | 27.58 | 17.05 | 22.91 | 17.33 | 39.67 | 22.61 |
| | Spanish | 56.85 | 45.71 | 25.53 | 17.67 | 74.84 | 56.43 | 26.74 | 20.48 | 73.40 | 51.80 | 47.24 | 33.66 | 73.66 | 52.90 | 63.26 | 45.43 | 58.45 | 41.60 |
| | Arabic | 61.52 | 39.79 | 29.34 | 20.33 | 38.02 | 29.43 | 67.83 | 40.54 | 41.14 | 16.67 | 54.34 | 16.67 | 37.78 | 30.26 | 33.43 | 21.69 | 49.54 | 27.64 |
| | French | 53.23 | 43.41 | 27.32 | 18.70 | 71.09 | 51.34 | 23.01 | 18.98 | 74.71 | 53.61 | 40.34 | 30.52 | 74.08 | 54.37 | 60.70 | 42.29 | 56.30 | 40.38 |
| | Russian | 58.44 | 40.52 | 24.07 | 17.88 | 60.70 | 33.16 | 51.24 | 34.93 | 56.64 | 33.36 | 63.45 | 40.02 | 58.50 | 38.14 | 59.63 | 29.30 | 57.58 | 32.17 |
| | Portuguese | 48.45 | 32.52 | 25.08 | 17.68 | 71.40 | 51.35 | 27.34 | 20.75 | 72.94 | 53.18 | 40.22 | 28.84 | 75.56 | 56.26 | 61.42 | 41.83 | 55.97 | 40.81 |
| | German | 47.79 | 39.32 | 24.15 | 17.17 | 70.97 | 52.24 | 28.36 | 22.64 | 70.76 | 51.00 | 48.64 | 33.73 | 71.18 | 51.02 | 70.03 | 48.56 | 56.79 | 41.45 |

Table 16: Performance Comparison of Different Detectors on In-Distribution and Language Generalization in TERNARY Task. Detector performance is visualized using teal color gradients, with darker intensities indicating superior detection capabilities.

| Detectors↓ | Train↓ Test→ | Academic | | News | | Novel | | SEO | | Wiki | | WebText | | Average | |
|---|---|---|---|---|---|---|---|---|---|---|---|---|---|---|---|
| | Metrics | $F_1^B$ | $F_1^F$ | $F_1^B$ | $F_1^F$ | $F_1^B$ | $F_1^F$ | $F_1^B$ | $F_1^F$ | $F_1^B$ | $F_1^F$ | $F_1^B$ | $F_1^F$ | $F_1^B$ | $F_1^F$ |
| Log-Likelihood | Academic | 77.94 | 46.73 | 57.83 | 46.43 | 55.01 | 37.00 | 70.57 | 44.66 | 56.06 | 52.55 | 54.05 | 42.23 | 60.25 | 44.04 |
| | News | 68.07 | 37.87 | 66.51 | 35.25 | 49.21 | 33.47 | 67.82 | 35.79 | 72.33 | 38.39 | 60.76 | 34.62 | 62.45 | 35.46 |
| | Novel | 77.51 | 50.71 | 61.09 | 51.73 | 56.44 | 39.51 | 71.85 | 49.25 | 60.74 | 58.45 | 56.86 | 46.53 | 64.09 | 48.18 |
| | SEO | 76.80 | 44.27 | 63.44 | 42.54 | 56.51 | 35.65 | 72.90 | 41.73 | 64.42 | 48.54 | 58.79 | 39.49 | 63.81 | 41.98 |
| | Wiki | 67.15 | 37.87 | 66.22 | 35.25 | 48.62 | 33.47 | 67.04 | 35.79 | 72.23 | 38.37 | 60.31 | 34.59 | 61.93 | 35.37 |
| | WebText | 70.24 | 40.27 | 67.01 | 37.61 | 50.63 | 34.06 | 69.69 | 37.77 | 72.25 | 42.31 | 61.37 | 36.12 | 65.20 | 37.23 |
| Log-Rank | Academic | 77.66 | 47.95 | 58.78 | 47.42 | 55.64 | 38.52 | 72.50 | 45.34 | 59.02 | 53.23 | 54.59 | 43.30 | 61.12 | 45.16 |
| | News | 69.09 | 40.32 | 67.42 | 36.73 | 50.44 | 34.50 | 69.73 | 37.08 | 73.43 | 40.44 | 60.76 | 35.98 | 65.15 | 37.40 |
| | Novel | 77.70 | 52.48 | 60.44 | 54.00 | 56.02 | 41.38 | 73.02 | 50.63 | 61.20 | 60.13 | 55.86 | 48.08 | 64.06 | 50.46 |
| | SEO | 76.65 | 47.90 | 63.63 | 47.24 | 56.17 | 38.49 | 73.82 | 45.24 | 66.05 | 53.05 | 58.35 | 43.17 | 64.62 | 45.02 |
| | Wiki | 65.66 | 39.87 | 66.63 | 36.28 | 48.64 | 34.33 | 66.35 | 36.72 | 72.86 | 39.90 | 59.18 | 35.67 | 61.61 | 36.38 |
| | WebText | 70.71 | 42.49 | 67.65 | 39.21 | 51.28 | 35.34 | 70.89 | 39.02 | 73.13 | 44.22 | 61.06 | 37.85 | 66.32 | 39.16 |
| DetectLLM-LRR | Academic | 72.69 | 48.41 | 60.08 | 47.14 | 54.24 | 39.56 | 75.80 | 43.12 | 64.53 | 48.95 | 54.06 | 43.43 | 60.93 | 44.27 |
| | News | 66.71 | 45.38 | 64.79 | 43.39 | 50.35 | 38.01 | 67.76 | 40.47 | 71.21 | 44.66 | 57.57 | 40.62 | 62.06 | 41.81 |
| | Novel | 72.59 | 49.73 | 60.39 | 48.71 | 54.29 | 40.21 | 75.80 | 44.70 | 64.86 | 50.88 | 54.18 | 44.85 | 62.95 | 46.86 |
| | SEO | 72.53 | 53.38 | 61.72 | 53.51 | 54.54 | 42.10 | 75.59 | 49.06 | 67.17 | 56.02 | 55.35 | 48.47 | 62.75 | 49.94 |
| | Wiki | 66.62 | 44.96 | 64.64 | 42.84 | 50.23 | 37.89 | 67.37 | 40.01 | 71.22 | 44.09 | 57.50 | 40.20 | 61.93 | 41.39 |
| | WebText | 68.67 | 43.93 | 65.07 | 41.52 | 51.65 | 37.43 | 70.42 | 38.73 | 71.64 | 42.57 | 57.77 | 39.02 | 62.46 | 40.21 |
| Fast-DetectGPT | Academic | 65.07 | 50.21 | 55.27 | 44.99 | 52.96 | 51.87 | 42.19 | 41.31 | 53.34 | 40.85 | 48.07 | 46.38 | 50.92 | 45.98 |
| | News | 65.05 | 47.66 | 56.12 | 43.55 | 55.62 | 49.74 | 43.05 | 40.48 | 53.34 | 39.15 | 50.27 | 45.00 | 51.80 | 44.41 |
| | Novel | 62.42 | 43.54 | 54.08 | 40.73 | 58.08 | 45.88 | 44.60 | 38.74 | 51.20 | 36.94 | 51.21 | 42.28 | 52.40 | 41.22 |
| | SEO | 64.32 | 44.96 | 55.35 | 41.52 | 56.97 | 47.38 | 44.09 | 39.46 | 52.74 | 37.73 | 51.23 | 43.28 | 52.70 | 42.18 |
| | Wiki | 65.07 | 47.65 | 55.26 | 43.41 | 52.92 | 49.71 | 42.19 | 40.41 | 53.33 | 39.09 | 48.05 | 44.93 | 51.20 | 44.35 |
| | WebText | 63.71 | 41.02 | 55.01 | 39.41 | 57.50 | 43.21 | 44.27 | 37.56 | 52.32 | 35.75 | 51.24 | 40.79 | 52.78 | 39.63 |
| Binoculars | Academic | 77.37 | 61.19 | 80.48 | 64.33 | 75.83 | 59.42 | 69.71 | 52.18 | 76.67 | 59.06 | 74.70 | 61.68 | 75.87 | 59.64 |
| | News | 77.37 | 61.19 | 80.48 | 64.33 | 75.83 | 59.42 | 69.71 | 52.18 | 76.67 | 59.06 | 74.70 | 61.68 | 75.87 | 59.64 |
| | Novel | 77.37 | 53.46 | 80.48 | 55.35 | 75.83 | 50.16 | 69.71 | 46.26 | 76.67 | 50.03 | 74.70 | 53.18 | 75.87 | 51.87 |
| | SEO | 74.27 | 61.19 | 80.02 | 64.33 | 71.67 | 59.42 | 71.98 | 52.18 | 73.43 | 59.06 | 71.36 | 61.68 | 73.72 | 59.64 |
| | Wiki | 77.37 | 61.19 | 80.48 | 64.33 | 75.83 | 59.42 | 69.71 | 52.18 | 76.67 | 59.06 | 74.70 | 61.68 | 75.87 | 59.64 |
| | WebText | 77.37 | 61.19 | 80.48 | 64.33 | 75.83 | 59.42 | 69.71 | 52.18 | 76.67 | 59.06 | 74.70 | 61.68 | 75.87 | 59.64 |
| ReviseDetect | Academic | 89.33 | 80.28 | 80.61 | 74.59 | 73.23 | 69.98 | 85.70 | 74.07 | 86.10 | 81.95 | 82.22 | 74.71 | 82.89 | 75.82 |
| | News | 87.54 | 0.00 | 80.27 | 0.00 | 73.54 | 0.00 | 82.12 | 0.00 | 86.65 | 0.00 | 81.08 | 0.00 | 81.87 | 0.00 |
| | Novel | 89.28 | 0.00 | 81.07 | 0.00 | 73.34 | 0.00 | 84.67 | 0.00 | 86.70 | 0.00 | 82.15 | 0.00 | 82.87 | 0.00 |
| | SEO | 88.47 | 0.00 | 78.89 | 0.00 | 73.09 | 0.00 | 86.62 | 0.00 | 83.68 | 0.00 | 80.90 | 0.00 | 81.94 | 0.00 |
| | Wiki | 87.07 | 58.74 | 80.10 | 56.71 | 73.45 | 52.82 | 81.72 | 54.29 | 86.51 | 62.42 | 80.92 | 54.26 | 81.63 | 55.58 |
| | WebText | 88.99 | 0.00 | 81.13 | 0.00 | 73.55 | 0.00 | 83.94 | 0.00 | 87.00 | 0.00 | 81.98 | 0.00 | 82.77 | 0.00 |
| GECScore | Academic | 90.77 | 87.72 | 76.72 | 81.46 | 65.15 | 72.10 | 91.86 | 86.30 | 85.47 | 85.65 | 80.07 | 83.98 | 82.68 | 83.00 |
| | News | 88.90 | 67.74 | 82.04 | 64.19 | 71.55 | 57.38 | 87.94 | 63.17 | 86.56 | 67.22 | 84.77 | 65.32 | 83.97 | 64.50 |
| | Novel | 86.66 | 29.74 | 80.93 | 31.61 | 72.24 | 30.85 | 85.03 | 30.69 | 84.81 | 31.15 | 83.25 | 30.40 | 82.31 | 30.81 |
| | SEO | 89.81 | 88.03 | 71.71 | 81.62 | 61.81 | 71.87 | 92.05 | 86.74 | 82.46 | 85.93 | 75.24 | 84.31 | 78.02 | 82.83 |
| | Wiki | 88.60 | 74.22 | 81.94 | 68.77 | 71.88 | 62.75 | 87.56 | 68.41 | 86.38 | 72.57 | 84.58 | 70.13 | 83.55 | 69.26 |
| | WebText | 89.13 | 73.28 | 82.04 | 68.04 | 71.41 | 61.92 | 88.27 | 67.31 | 86.83 | 71.77 | 84.77 | 69.16 | 83.77 | 68.55 |
| Lastde++ | Academic | 39.84 | 33.63 | 39.32 | 33.36 | 46.91 | 34.05 | 28.38 | 33.20 | 35.99 | 33.34 | 39.75 | 33.51 | 37.74 | 33.44 |
| | News | 38.58 | 34.01 | 41.00 | 33.42 | 44.51 | 35.01 | 26.41 | 33.02 | 36.87 | 33.31 | 40.47 | 33.67 | 36.31 | 33.48 |
| | Novel | 39.33 | 33.66 | 36.68 | 33.36 | 46.94 | 34.00 | 30.39 | 33.24 | 35.17 | 33.36 | 38.03 | 33.50 | 37.59 | 33.41 |
| | SEO | 33.34 | 33.57 | 33.33 | 33.36 | 33.37 | 33.77 | 33.33 | 33.26 | 33.34 | 33.36 | 33.35 | 33.46 | 33.34 | 33.44 |
| | Wiki | 39.57 | 34.06 | 40.37 | 33.41 | 46.07 | 35.12 | 27.33 | 33.01 | 36.53 | 33.30 | 40.29 | 33.67 | 38.35 | 33.68 |
| | WebText | 39.68 | 33.89 | 39.88 | 33.41 | 46.77 | 34.73 | 27.82 | 33.05 | 36.26 | 33.35 | 40.20 | 33.61 | 38.40 | 33.56 |
| RepreGuard | Academic | 97.15 | 97.35 | 90.67 | 91.29 | 88.66 | 89.03 | 92.20 | 92.46 | 92.12 | 91.77 | 85.98 | 86.84 | 91.14 | 90.79 |
| | News | 68.81 | 20.38 | 68.51 | 23.28 | 68.71 | 40.72 | 68.89 | 31.18 | 67.47 | 12.81 | 68.60 | 27.07 | 68.50 | 26.61 |
| | Novel | 76.29 | 26.25 | 66.16 | 16.88 | 79.17 | 28.38 | 76.44 | 27.44 | 69.85 | 11.62 | 67.83 | 23.80 | 72.47 | 21.83 |
| | SEO | 87.32 | 89.50 | 75.36 | 76.69 | 72.46 | 73.42 | 98.12 | 97.77 | 78.84 | 82.16 | 73.01 | 75.68 | 79.36 | 82.64 |
| | Wiki | 75.61 | 44.03 | 72.95 | 46.73 | 67.39 | 47.28 | 75.01 | 59.44 | 76.47 | 56.39 | 67.45 | 47.91 | 71.48 | 50.04 |
| | WebText | 83.34 | 88.15 | 80.17 | 80.93 | 76.88 | 78.51 | 91.46 | 85.47 | 77.17 | 82.69 | 85.12 | 80.86 | 82.27 | 83.40 |
| X-Rob-Classifier | Academic | 99.95 | 99.09 | 90.42 | 51.90 | 84.50 | 38.95 | 94.90 | 50.72 | 96.15 | 80.67 | 80.33 | 37.17 | 90.12 | 61.41 |
| | News | 99.19 | 99.12 | 99.71 | 99.44 | 95.95 | 89.02 | 98.84 | 98.06 | 99.71 | 99.45 | 98.29 | 96.06 | 98.61 | 96.70 |
| | Novel | 98.50 | 98.36 | 97.72 | 96.81 | 99.64 | 99.47 | 97.01 | 94.63 | 98.32 | 98.29 | 96.99 | 95.78 | 97.73 | 97.07 |
| | SEO | 98.31 | 97.90 | 91.75 | 80.83 | 90.36 | 57.99 | 99.76 | 99.56 | 99.17 | 99.15 | 90.83 | 83.75 | 95.03 | 88.26 |
| | Wiki | 97.69 | 69.68 | 85.92 | 46.68 | 84.53 | 57.90 | 92.82 | 81.82 | 99.80 | 99.67 | 92.50 | 83.27 | 92.21 | 76.57 |
| | WebText | 99.06 | 99.04 | 97.88 | 94.68 | 97.31 | 95.19 | 98.57 | 97.59 | 99.82 | 99.51 | 99.46 | 99.39 | 98.60 | 97.40 |
| mDeBERTa-Classifier | Academic | 99.99 | 92.13 | 95.42 | 86.71 | 91.56 | 64.79 | 95.42 | 56.71 | 96.19 | 89.72 | 91.52 | 51.75 | 94.16 | 74.26 |
| | News | 99.44 | 99.36 | 99.92 | 99.87 | 94.14 | 86.78 | 98.60 | 98.21 | 99.72 | 99.46 | 99.44 | 99.32 | 98.12 | 95.60 |
| | Novel | 99.22 | 99.22 | 99.16 | 99.14 | 99.81 | 99.54 | 97.91 | 97.43 | 99.32 | 99.18 | 98.97 | 98.94 | 98.89 | 98.56 |
| | SEO | 98.86 | 98.84 | 96.74 | 92.17 | 90.93 | 64.29 | 99.97 | 99.80 | 99.01 | 98.99 | 95.89 | 91.97 | 96.20 | 88.64 |
| | Wiki | 98.56 | 98.24 | 92.19 | 84.07 | 89.17 | 72.16 | 97.44 | 93.58 | 99.96 | 99.93 | 97.48 | 96.77 | 94.32 | 88.98 |
| | WebText | 99.61 | 99.46 | 98.44 | 97.64 | 98.61 | 98.42 | 99.32 | 99.27 | 99.82 | 99.48 | 99.93 | 99.71 | 99.13 | 98.80 |
| Biscope | Academic | 98.25 | 97.94 | 86.31 | 65.33 | 80.24 | 67.83 | 89.80 | 76.44 | 90.86 | 80.66 | 83.44 | 62.99 | 86.54 | 74.24 |
| | News | 95.03 | 93.38 | 92.04 | 82.89 | 83.04 | 69.84 | 90.94 | 80.93 | 93.71 | 88.24 | 86.96 | 71.01 | 89.29 | 79.87 |
| | Novel | 96.24 | 95.34 | 87.89 | 69.19 | 89.02 | 76.87 | 89.02 | 69.98 | 91.27 | 76.12 | 85.29 | 66.98 | 89.17 | 77.17 |
| | SEO | 96.46 | 93.54 | 86.33 | 56.34 | 83.37 | 63.18 | 95.03 | 85.20 | 92.37 | 77.75 | 85.80 | 60.76 | 88.84 | 73.03 |
| | Wiki | 96.03 | 94.90 | 90.07 | 69.99 | 79.46 | 59.69 | 91.47 | 76.68 | 95.01 | 90.03 | 86.20 | 67.60 | 87.98 | 76.56 |
| | WebText | 95.57 | 93.18 | 91.51 | 79.52 | 82.10 | 62.17 | 92.01 | 79.58 | 94.41 | 88.08 | 90.92 | 78.55 | 91.17 | 79.14 |

Table 17: Performance Comparison of Different Detectors on Domain Generalization in BINARY Task. Detector performance is visualized using teal color gradients, with darker intensities indicating superior detection capabilities.

## G.9 Robustness Results for Cross-Operation on Different Languages.

The detailed results for cross-operation on different languages in BINARY task and TRINARY task are presented in Table 31 and Table 32.

| Detectors↓ | Train↓ Test→ | Academic | | News | | Novel | | SEO | | Wiki | | WebText | | Average | |
|---|---|---|---|---|---|---|---|---|---|---|---|---|---|---|---|
| | Metrics | $F_1^B$ | $F_1^F$ | $F_1^B$ | $F_1^F$ | $F_1^B$ | $F_1^F$ | $F_1^B$ | $F_1^F$ | $F_1^B$ | $F_1^F$ | $F_1^B$ | $F_1^F$ | $F_1^B$ | $F_1^F$ |
| Log-Likelihood | Academic | 53.68 | 41.37 | 37.13 | 33.17 | 35.31 | 30.23 | 48.08 | 39.06 | 36.09 | 33.40 | 33.58 | 30.72 | 39.30 | 34.28 |
| | News | 37.80 | 37.80 | 35.92 | 35.92 | 26.91 | 26.91 | 37.72 | 37.72 | 38.85 | 38.85 | 32.79 | 32.79 | 32.47 | 32.47 |
| | Novel | 41.39 | 41.39 | 32.53 | 32.53 | 30.19 | 30.19 | 38.81 | 38.81 | 32.42 | 32.42 | 30.31 | 30.31 | 34.68 | 34.68 |
| | SEO | 51.72 | 39.49 | 41.00 | 35.68 | 35.84 | 28.81 | 49.97 | 39.04 | 43.74 | 37.73 | 38.01 | 32.81 | 42.00 | 34.90 |
| | Wiki | 48.15 | 36.68 | 42.35 | 35.72 | 33.90 | 26.16 | 48.94 | 36.78 | 47.42 | 38.94 | 39.35 | 32.46 | 42.70 | 33.34 |
| | WebText | 38.77 | 38.77 | 35.91 | 35.91 | 27.95 | 27.95 | 38.47 | 38.47 | 38.36 | 38.36 | 32.99 | 32.99 | 33.57 | 33.57 |
| Log-Rank | Academic | 53.14 | 41.45 | 37.74 | 33.00 | 35.05 | 30.22 | 49.03 | 39.75 | 37.05 | 33.57 | 33.78 | 30.20 | 39.25 | 34.36 |
| | News | 37.90 | 37.90 | 36.34 | 36.34 | 27.28 | 27.28 | 38.39 | 38.39 | 39.48 | 39.48 | 32.63 | 32.63 | 33.14 | 33.14 |
| | Novel | 41.47 | 41.47 | 31.49 | 31.49 | 29.75 | 29.75 | 39.18 | 39.18 | 31.71 | 31.71 | 29.25 | 29.25 | 32.27 | 32.27 |
| | SEO | 51.60 | 39.43 | 41.15 | 35.96 | 34.75 | 28.52 | 51.04 | 39.41 | 44.87 | 38.67 | 37.61 | 32.57 | 41.63 | 34.99 |
| | Wiki | 51.57 | 36.82 | 41.27 | 36.17 | 31.92 | 26.60 | 50.83 | 37.40 | 44.63 | 39.54 | 36.70 | 32.34 | 42.76 | 33.44 |
| | WebText | 38.74 | 38.74 | 36.24 | 36.24 | 28.04 | 28.04 | 39.00 | 39.00 | 38.97 | 38.97 | 32.74 | 32.74 | 33.63 | 33.63 |
| DetectLLM-LRR | Academic | 46.83 | 39.12 | 38.28 | 32.84 | 34.77 | 29.33 | 50.05 | 40.96 | 41.77 | 35.51 | 35.52 | 29.39 | 39.19 | 33.35 |
| | News | 37.61 | 37.61 | 35.12 | 35.12 | 27.89 | 27.89 | 38.69 | 38.69 | 38.65 | 38.65 | 31.07 | 31.07 | 32.98 | 32.98 |
| | Novel | 38.44 | 38.44 | 30.17 | 30.17 | 28.62 | 28.62 | 40.74 | 40.74 | 32.12 | 32.12 | 27.58 | 27.58 | 31.07 | 31.07 |
| | SEO | 47.71 | 38.81 | 40.35 | 34.00 | 35.20 | 29.25 | 51.23 | 40.59 | 44.09 | 37.23 | 37.04 | 30.48 | 40.69 | 34.13 |
| | Wiki | 36.63 | 36.68 | 36.69 | 35.05 | 28.23 | 27.14 | 38.74 | 37.31 | 39.61 | 38.65 | 33.07 | 31.03 | 34.29 | 32.80 |
| | WebText | 38.40 | 38.40 | 34.79 | 34.79 | 28.76 | 28.76 | 39.85 | 39.85 | 38.15 | 38.15 | 30.98 | 30.98 | 33.29 | 33.29 |
| Fast-DetectGPT | Academic | 43.57 | 33.24 | 35.99 | 28.92 | 37.60 | 31.42 | 31.93 | 23.83 | 34.27 | 27.17 | 33.27 | 27.42 | 34.36 | 27.55 |
| | News | 34.51 | 34.51 | 29.69 | 29.69 | 30.60 | 30.60 | 23.66 | 23.66 | 28.26 | 28.26 | 27.38 | 27.38 | 27.41 | 27.41 |
| | Novel | 32.13 | 32.13 | 28.09 | 28.09 | 31.39 | 31.39 | 23.57 | 23.57 | 26.20 | 26.20 | 27.28 | 27.28 | 26.54 | 26.54 |
| | SEO | 29.42 | 19.00 | 25.93 | 18.75 | 25.74 | 19.71 | 28.16 | 20.28 | 25.06 | 18.00 | 26.21 | 20.20 | 25.79 | 19.19 |
| | Wiki | 34.26 | 34.26 | 29.53 | 29.53 | 30.93 | 30.93 | 23.68 | 23.68 | 28.02 | 28.02 | 27.43 | 27.43 | 27.31 | 27.31 |
| | WebText | 30.29 | 29.60 | 26.49 | 25.84 | 30.92 | 30.14 | 23.73 | 22.87 | 24.09 | 23.61 | 26.85 | 26.15 | 25.50 | 25.16 |
| Binoculars | Academic | 47.41 | 33.87 | 45.35 | 35.68 | 43.53 | 32.84 | 39.58 | 28.31 | 41.91 | 32.66 | 45.98 | 34.05 | 43.46 | 32.54 |
| | News | 54.52 | 39.83 | 53.21 | 41.39 | 51.60 | 39.29 | 46.77 | 34.31 | 51.43 | 39.91 | 51.19 | 38.95 | 51.28 | 38.29 |
| | Novel | 47.41 | 33.87 | 45.35 | 35.68 | 43.53 | 32.84 | 39.58 | 28.31 | 41.91 | 32.66 | 45.98 | 34.05 | 43.46 | 32.54 |
| | SEO | 54.52 | 39.83 | 53.21 | 41.39 | 51.60 | 39.29 | 46.77 | 34.31 | 51.43 | 39.91 | 51.19 | 38.95 | 51.28 | 38.29 |
| | Wiki | 47.41 | 33.87 | 45.35 | 35.68 | 43.53 | 32.84 | 39.58 | 28.31 | 41.91 | 32.66 | 45.98 | 34.05 | 43.46 | 32.54 |
| | WebText | 47.41 | 33.87 | 45.35 | 35.68 | 43.53 | 32.84 | 39.58 | 28.31 | 41.91 | 32.66 | 45.98 | 34.05 | 43.46 | 32.54 |
| ReviseDetect | Academic | 62.35 | 49.37 | 54.96 | 46.61 | 41.25 | 40.26 | 59.55 | 47.92 | 62.13 | 50.50 | 55.89 | 46.81 | 54.39 | 46.53 |
| | News | 61.84 | 48.78 | 54.78 | 46.28 | 41.86 | 40.35 | 58.49 | 47.46 | 61.39 | 50.13 | 55.77 | 46.37 | 54.02 | 46.57 |
| | Novel | 28.26 | 17.31 | 26.69 | 17.93 | 29.40 | 18.04 | 28.45 | 18.05 | 24.84 | 17.79 | 26.86 | 17.49 | 27.41 | 17.69 |
| | SEO | 59.82 | 48.78 | 53.22 | 46.28 | 39.79 | 40.36 | 62.02 | 47.46 | 59.94 | 50.13 | 53.53 | 46.37 | 53.05 | 46.13 |
| | Wiki | 59.69 | 45.77 | 53.14 | 43.84 | 43.03 | 39.69 | 55.36 | 44.79 | 57.88 | 47.68 | 54.19 | 43.53 | 53.87 | 43.08 |
| | WebText | 60.67 | 48.52 | 54.10 | 46.04 | 42.51 | 40.30 | 56.79 | 47.27 | 59.72 | 49.98 | 55.20 | 46.17 | 54.83 | 46.66 |
| GECScore | Academic | 62.35 | 49.37 | 54.96 | 46.61 | 41.25 | 40.26 | 59.55 | 47.92 | 62.13 | 50.50 | 55.89 | 46.81 | 54.39 | 46.53 |
| | News | 61.84 | 48.78 | 54.78 | 46.28 | 41.86 | 40.35 | 58.49 | 47.46 | 61.39 | 50.13 | 55.77 | 46.37 | 54.02 | 46.57 |
| | Novel | 28.26 | 17.31 | 26.69 | 17.93 | 29.40 | 18.04 | 28.45 | 18.05 | 24.84 | 17.79 | 26.86 | 17.49 | 27.41 | 17.69 |
| | SEO | 59.82 | 48.78 | 53.22 | 46.28 | 39.79 | 40.36 | 62.02 | 47.46 | 59.94 | 50.13 | 53.53 | 46.37 | 53.05 | 46.13 |
| | Wiki | 59.69 | 45.77 | 53.14 | 43.84 | 43.03 | 39.69 | 55.36 | 44.79 | 57.88 | 47.68 | 54.19 | 43.53 | 53.87 | 43.08 |
| | WebText | 60.67 | 48.52 | 54.10 | 46.04 | 42.51 | 40.30 | 56.79 | 47.27 | 59.72 | 49.98 | 55.20 | 46.17 | 54.83 | 46.66 |
| Lastde++ | Academic | 20.16 | 17.17 | 23.32 | 16.77 | 19.66 | 17.23 | 19.77 | 16.92 | 22.51 | 16.84 | 21.67 | 16.86 | 21.18 | 16.94 |
| | News | 16.67 | 16.67 | 16.78 | 16.79 | 16.67 | 16.67 | 16.67 | 16.67 | 16.66 | 16.65 | 16.66 | 16.67 | 16.67 | 16.67 |
| | Novel | 24.41 | 17.10 | 20.48 | 16.75 | 24.96 | 17.19 | 23.12 | 16.90 | 21.72 | 16.81 | 21.48 | 16.81 | 22.69 | 16.84 |
| | SEO | 16.68 | 16.68 | 16.69 | 16.69 | 16.67 | 16.67 | 16.67 | 16.67 | 16.67 | 16.67 | 16.67 | 16.67 | 16.67 | 16.67 |
| | Wiki | 26.27 | 17.93 | 24.43 | 16.98 | 26.66 | 17.95 | 25.10 | 17.43 | 25.47 | 17.16 | 25.10 | 17.11 | 25.50 | 17.34 |
| | WebText | 20.65 | 17.81 | 18.09 | 16.93 | 21.22 | 17.84 | 19.86 | 17.35 | 18.84 | 17.12 | 18.53 | 17.06 | 19.52 | 17.24 |
| RepreGuard | Academic | 51.11 | 49.85 | 45.64 | 44.26 | 43.06 | 42.51 | 47.53 | 45.94 | 48.54 | 48.68 | 43.29 | 41.07 | 45.64 | 44.41 |
| | News | 23.80 | 16.05 | 26.19 | 16.88 | 23.30 | 17.93 | 23.85 | 16.84 | 24.95 | 16.74 | 24.40 | 17.29 | 24.04 | 16.97 |
| | Novel | 40.90 | 26.40 | 38.91 | 22.59 | 43.81 | 24.89 | 43.57 | 24.01 | 36.75 | 19.89 | 37.60 | 23.18 | 38.59 | 23.43 |
| | SEO | 22.66 | 21.23 | 25.17 | 22.69 | 24.61 | 23.92 | 29.67 | 26.00 | 22.81 | 20.11 | 27.26 | 24.03 | 25.32 | 22.89 |
| | Wiki | 38.07 | 27.49 | 36.57 | 25.86 | 34.26 | 26.06 | 38.57 | 29.43 | 39.15 | 28.32 | 35.39 | 25.63 | 36.17 | 26.96 |
| | WebText | 35.12 | 34.83 | 34.77 | 27.13 | 38.34 | 32.39 | 38.60 | 31.53 | 34.40 | 27.60 | 36.66 | 30.56 | 36.03 | 30.37 |
| X-Rob-Classifier | Academic | 94.53 | 95.23 | 51.66 | 54.57 | 50.98 | 25.88 | 73.35 | 45.06 | 58.69 | 60.24 | 49.62 | 24.80 | 61.46 | 51.41 |
| | News | 80.79 | 75.60 | 87.70 | 85.96 | 72.90 | 67.54 | 85.47 | 65.62 | 77.11 | 79.38 | 76.10 | 70.18 | 78.98 | 73.48 |
| | Novel | 77.57 | 67.19 | 62.84 | 53.21 | 87.24 | 80.86 | 76.94 | 45.38 | 61.28 | 57.61 | 72.94 | 61.66 | 71.06 | 61.43 |
| | SEO | 73.83 | 68.50 | 59.16 | 45.01 | 53.91 | 32.40 | 96.11 | 95.21 | 69.87 | 52.26 | 61.49 | 41.24 | 66.64 | 55.79 |
| | Wiki | 74.00 | 60.94 | 43.86 | 20.98 | 66.56 | 47.10 | 70.03 | 45.76 | 89.54 | 88.80 | 61.37 | 51.02 | 65.90 | 50.89 |
| | WebText | 85.12 | 74.08 | 69.67 | 53.70 | 75.96 | 56.44 | 88.26 | 54.57 | 85.56 | 84.10 | 83.97 | 69.93 | 81.67 | 65.64 |
| mDeBERTa-Classifier | Academic | 96.74 | 96.60 | 49.63 | 30.62 | 52.80 | 24.96 | 81.86 | 34.50 | 63.26 | 43.48 | 56.33 | 43.40 | 66.62 | 47.63 |
| | News | 91.15 | 84.18 | 95.02 | 93.73 | 78.37 | 68.61 | 89.71 | 70.38 | 86.71 | 86.09 | 89.59 | 86.70 | 89.06 | 82.84 |
| | Novel | 78.07 | 64.07 | 61.87 | 45.45 | 91.14 | 88.72 | 75.62 | 40.41 | 53.68 | 47.82 | 76.53 | 60.33 | 71.12 | 59.56 |
| | SEO | 80.90 | 70.15 | 64.50 | 41.34 | 61.01 | 37.20 | 97.50 | 97.38 | 68.60 | 30.39 | 67.88 | 51.91 | 70.13 | 53.86 |
| | Wiki | 84.98 | 82.47 | 63.75 | 38.15 | 71.00 | 58.75 | 78.51 | 34.31 | 95.75 | 95.17 | 76.79 | 66.39 | 77.59 | 62.12 |
| | WebText | 92.33 | 88.42 | 83.47 | 74.66 | 84.49 | 77.08 | 91.62 | 69.18 | 90.79 | 89.10 | 93.14 | 90.61 | 89.90 | 82.65 |
| Biscope | Academic | 79.61 | 62.11 | 55.51 | 37.27 | 49.65 | 38.25 | 64.53 | 44.32 | 59.10 | 44.09 | 51.94 | 36.33 | 57.40 | 41.61 |
| | News | 65.58 | 53.57 | 68.77 | 48.00 | 60.59 | 38.77 | 62.20 | 42.77 | 65.56 | 47.38 | 64.62 | 41.35 | 63.67 | 45.20 |
| | Novel | 67.89 | 51.52 | 61.58 | 39.58 | 65.86 | 46.94 | 61.89 | 39.53 | 62.06 | 43.49 | 60.39 | 38.66 | 62.02 | 43.00 |
| | SEO | 73.18 | 54.14 | 57.91 | 31.64 | 56.31 | 35.65 | 75.07 | 50.00 | 64.25 | 40.65 | 56.41 | 30.94 | 61.54 | 41.40 |
| | Wiki | 69.81 | 57.38 | 62.75 | 39.66 | 57.22 | 34.17 | 63.48 | 43.39 | 68.31 | 50.35 | 61.58 | 38.28 | 61.80 | 42.00 |
| | WebText | 67.38 | 49.42 | 67.72 | 44.04 | 59.00 | 34.54 | 63.01 | 37.46 | 66.61 | 46.73 | 68.20 | 44.11 | 64.89 | 42.61 |

Table 18: Performance Comparison of Different Detectors on Domain Generalization in TERNARY Task. Detector performance is visualized using teal color gradients, with darker intensities indicating superior detection capabilities.

| Detectors↓ | Train↓ Test→ | Deepseek-V3 | | Gemini-2.5-flash | | GPT-4o | | Qwen-Max | | Average | |
|---|---|---|---|---|---|---|---|---|---|---|---|
| | Metrics | $F_1^B$ | $F_1^F$ | $F_1^B$ | $F_1^F$ | $F_1^B$ | $F_1^F$ | $F_1^B$ | $F_1^F$ | $F_1^B$ | $F_1^F$ |
| Log-Likelihood | Deepseek-V3 | 63.91 | 34.96 | 60.09 | 33.97 | 65.82 | 34.93 | 76.91 | 47.47 | 66.68 | 37.83 |
| | Gemini-2.50-flash | 63.45 | 35.02 | 60.58 | 34.01 | 64.64 | 35.11 | 72.91 | 48.22 | 65.39 | 38.09 |
| | GPT-4o | 63.90 | 34.78 | 60.11 | 33.90 | 65.84 | 34.71 | 76.93 | 46.83 | 66.69 | 37.56 |
| | Qwen-Max | 59.48 | 34.96 | 54.26 | 33.97 | 62.77 | 34.95 | 80.66 | 47.48 | 64.29 | 37.84 |
| Log-Rank | Deepseek-V3 | 64.04 | 35.53 | 59.85 | 34.12 | 66.19 | 35.56 | 77.62 | 52.77 | 66.93 | 39.50 |
| | Gemini-2.50-flash | 63.77 | 35.82 | 60.48 | 34.19 | 65.58 | 35.85 | 74.34 | 53.97 | 66.04 | 40.01 |
| | GPT-4o | 64.04 | 35.28 | 59.91 | 34.03 | 66.20 | 35.42 | 77.42 | 52.04 | 66.90 | 39.20 |
| | Qwen-Max | 58.81 | 35.27 | 52.47 | 34.03 | 62.05 | 35.42 | 81.38 | 52.03 | 63.68 | 39.19 |
| DetectLLM-LRR | Deepseek-V3 | 62.57 | 38.18 | 57.11 | 35.49 | 65.06 | 38.44 | 76.53 | 57.97 | 65.32 | 42.52 |
| | Gemini-2.50-flash | 62.61 | 38.06 | 57.59 | 35.41 | 64.62 | 38.28 | 73.88 | 57.75 | 64.68 | 42.38 |
| | GPT-4o | 62.17 | 37.95 | 56.28 | 35.39 | 64.88 | 38.18 | 78.02 | 57.57 | 65.34 | 42.27 |
| | Qwen-Max | 57.91 | 38.18 | 50.35 | 35.49 | 60.98 | 38.43 | 79.26 | 57.96 | 62.13 | 42.51 |
| Fast-DetectGPT | Deepseek-V3 | 53.73 | 39.29 | 46.98 | 38.81 | 45.40 | 37.57 | 68.06 | 53.01 | 53.54 | 42.17 |
| | Gemini-2.50-flash | 53.53 | 39.31 | 46.85 | 38.82 | 45.44 | 37.58 | 68.89 | 53.05 | 53.68 | 42.19 |
| | GPT-4o | 53.70 | 39.00 | 46.95 | 38.62 | 45.48 | 37.33 | 68.20 | 52.32 | 53.58 | 41.82 |
| | Qwen-Max | 53.01 | 39.48 | 46.81 | 38.90 | 45.44 | 37.68 | 69.05 | 53.24 | 53.58 | 42.33 |
| Binoculars | Deepseek-V3 | 75.61 | 58.41 | 72.67 | 53.44 | 73.95 | 55.72 | 81.10 | 70.18 | 75.83 | 59.44 |
| | Gemini-2.50-flash | 75.61 | 58.41 | 72.67 | 53.44 | 73.95 | 55.72 | 81.10 | 70.18 | 75.83 | 59.44 |
| | GPT-4o | 75.61 | 58.41 | 72.67 | 53.44 | 73.95 | 55.72 | 81.10 | 70.18 | 75.83 | 59.44 |
| | Qwen-Max | 75.61 | 58.41 | 72.67 | 53.44 | 73.95 | 55.72 | 81.10 | 70.18 | 75.83 | 59.44 |
| ReviseDetect | Deepseek-V3 | 80.94 | 0.00 | 81.91 | 0.00 | 81.01 | 0.00 | 85.82 | 0.00 | 82.42 | 0.00 |
| | Gemini-2.50-flash | 80.60 | 0.00 | 81.97 | 0.00 | 80.80 | 0.00 | 85.97 | 0.00 | 82.34 | 0.00 |
| | GPT-4o | 80.93 | 0.00 | 81.91 | 0.00 | 81.01 | 0.00 | 85.81 | 0.00 | 82.42 | 0.00 |
| | Qwen-Max | 80.26 | 0.00 | 81.73 | 0.00 | 80.46 | 0.00 | 85.92 | 0.00 | 82.09 | 0.00 |
| GECScore | Deepseek-V3 | 80.03 | 57.47 | 82.41 | 61.56 | 83.25 | 61.38 | 87.19 | 66.51 | 83.22 | 61.73 |
| | Gemini-2.50-flash | 80.19 | 57.48 | 82.46 | 61.57 | 83.31 | 61.40 | 87.11 | 66.53 | 83.27 | 61.75 |
| | GPT-4o | 80.47 | 57.49 | 82.49 | 61.57 | 83.59 | 61.40 | 87.17 | 66.53 | 83.43 | 61.75 |
| | Qwen-Max | 80.04 | 56.78 | 82.43 | 61.13 | 83.24 | 60.75 | 87.15 | 66.02 | 83.22 | 61.17 |
| Lastde++ | Deepseek-V3 | 40.02 | 33.58 | 41.00 | 33.66 | 35.78 | 33.50 | 39.32 | 33.42 | 39.03 | 33.54 |
| | Gemini-2.50-flash | 39.83 | 33.57 | 41.05 | 33.69 | 35.91 | 33.49 | 39.10 | 33.40 | 39.22 | 33.54 |
| | GPT-4o | 39.06 | 33.57 | 39.96 | 33.69 | 35.67 | 33.49 | 37.96 | 33.40 | 38.16 | 33.54 |
| | Qwen-Max | 40.06 | 33.58 | 41.17 | 33.73 | 35.78 | 33.50 | 39.70 | 33.40 | 39.18 | 33.55 |
| RepreGuard | Deepseek-V3 | 69.82 | 46.26 | 69.01 | 46.19 | 70.03 | 45.48 | 71.07 | 46.91 | 70.23 | 46.21 |
| | Gemini-2.50-flash | 81.92 | 60.87 | 87.42 | 61.44 | 85.44 | 59.62 | 86.03 | 60.11 | 85.20 | 60.51 |
| | GPT-4o | 69.09 | 38.01 | 68.33 | 37.99 | 69.34 | 37.07 | 70.38 | 38.02 | 69.29 | 37.77 |
| | Qwen-Max | 70.10 | 69.69 | 69.78 | 69.51 | 71.02 | 70.36 | 77.49 | 75.27 | 72.10 | 71.21 |
| X-Rob-Classifier | Deepseek-V3 | 99.86 | 99.50 | 98.95 | 98.79 | 98.90 | 98.64 | 96.87 | 93.90 | 98.65 | 97.71 |
| | Gemini-2.50-flash | 98.94 | 98.93 | 99.93 | 99.80 | 99.59 | 99.47 | 98.41 | 98.32 | 99.22 | 99.13 |
| | GPT-4o | 99.11 | 99.04 | 99.77 | 99.61 | 99.79 | 99.67 | 99.20 | 99.17 | 99.47 | 99.37 |
| | Qwen-Max | 97.65 | 95.07 | 98.73 | 96.54 | 98.67 | 95.41 | 98.37 | 92.85 | 98.11 | 95.22 |
| mDeBERTa-Classifier | Deepseek-V3 | 99.95 | 99.59 | 99.32 | 99.22 | 99.14 | 99.11 | 96.99 | 96.46 | 98.85 | 98.59 |
| | Gemini-2.50-flash | 99.70 | 99.49 | 99.98 | 99.90 | 99.81 | 99.39 | 99.30 | 99.22 | 99.70 | 99.50 |
| | GPT-4o | 99.53 | 99.35 | 99.96 | 99.90 | 99.95 | 99.87 | 99.59 | 99.42 | 99.76 | 99.63 |
| | Qwen-Max | 99.18 | 99.16 | 99.77 | 99.45 | 99.75 | 99.42 | 99.90 | 99.88 | 99.65 | 99.48 |
| Biscope | Deepseek-V3 | 91.52 | 80.13 | 90.41 | 79.39 | 92.45 | 80.98 | 93.94 | 82.59 | 92.08 | 80.77 |
| | Gemini-2.50-flash | 90.71 | 75.66 | 91.55 | 81.68 | 92.71 | 79.74 | 92.97 | 76.14 | 91.99 | 78.31 |
| | GPT-4o | 91.02 | 76.44 | 91.47 | 81.73 | 93.39 | 82.31 | 93.13 | 78.59 | 92.25 | 79.77 |
| | Qwen-Max | 88.25 | 72.43 | 87.38 | 69.33 | 89.35 | 72.58 | 94.92 | 91.34 | 89.98 | 76.42 |

Table 19: Performance Comparison of Different Detectors on Generator Generalization in BINARY Task. Detector performance is visualized using teal color gradients, with darker intensities indicating superior detection capabilities.

| Detectors↓ | Train↓ Test→ | Deepseek-V3 | | Gemini-2.5-flash | | GPT-4o | | Qwen-Max | | Average | |
|---|---|---|---|---|---|---|---|---|---|---|---|
| | Metrics | $F_1^B$ | $F_1^F$ | $F_1^B$ | $F_1^F$ | $F_1^B$ | $F_1^F$ | $F_1^B$ | $F_1^F$ | $F_1^B$ | $F_1^F$ |
| Log-Likelihood | Deepseek-V3 | 40.50 | 34.25 | 39.36 | 32.53 | 40.40 | 35.19 | 47.13 | 40.87 | 41.85 | 35.71 |
| | Gemini-2.50-flash | 35.23 | 33.86 | 36.05 | 32.39 | 33.18 | 34.25 | 37.93 | 38.40 | 35.60 | 34.73 |
| | GPT-4o | 39.99 | 33.85 | 36.97 | 31.57 | 42.25 | 35.19 | 50.24 | 42.41 | 42.36 | 35.76 |
| | Qwen-Max | 40.08 | 30.84 | 35.55 | 27.81 | 43.33 | 32.79 | 55.36 | 43.84 | 43.58 | 33.82 |
| Log-Rank | Deepseek-V3 | 40.40 | 34.33 | 38.80 | 32.26 | 40.40 | 35.40 | 47.35 | 41.28 | 41.74 | 35.82 |
| | Gemini-2.50-flash | 35.80 | 33.58 | 36.37 | 32.22 | 33.70 | 34.32 | 38.25 | 38.09 | 36.03 | 34.55 |
| | GPT-4o | 41.95 | 34.28 | 38.61 | 31.91 | 43.50 | 35.42 | 51.69 | 42.31 | 43.94 | 35.98 |
| | Qwen-Max | 39.61 | 31.18 | 35.04 | 27.68 | 42.33 | 33.02 | 54.72 | 44.39 | 42.93 | 34.07 |
| DetectLLM-LRR | Deepseek-V3 | 37.67 | 33.65 | 36.11 | 31.00 | 36.78 | 34.76 | 43.64 | 40.40 | 38.55 | 34.95 |
| | Gemini-2.50-flash | 35.50 | 33.14 | 35.13 | 31.03 | 34.02 | 34.22 | 39.51 | 38.39 | 36.04 | 34.20 |
| | GPT-4o | 40.66 | 33.46 | 37.29 | 30.36 | 42.12 | 34.80 | 50.05 | 41.95 | 42.53 | 35.14 |
| | Qwen-Max | 36.75 | 31.51 | 32.58 | 27.33 | 38.49 | 33.20 | 49.00 | 43.34 | 39.21 | 33.85 |
| Fast-DetectGPT | Deepseek-V3 | 28.15 | 28.15 | 25.04 | 25.04 | 24.22 | 24.22 | 37.07 | 37.07 | 28.62 | 28.62 |
| | Gemini-2.50-flash | 27.41 | 27.41 | 24.62 | 24.62 | 23.71 | 23.71 | 36.76 | 36.76 | 28.13 | 28.13 |
| | GPT-4o | 16.67 | 16.67 | 16.68 | 16.68 | 16.67 | 16.67 | 16.67 | 16.67 | 16.67 | 16.67 |
| | Qwen-Max | 34.13 | 27.86 | 30.56 | 24.87 | 32.35 | 24.03 | 45.48 | 37.05 | 35.63 | 28.45 |
| Binoculars | Deepseek-V3 | 46.37 | 38.36 | 43.69 | 36.50 | 46.08 | 37.29 | 52.46 | 43.43 | 47.15 | 38.90 |
| | Gemini-2.50-flash | 46.37 | 38.36 | 43.69 | 36.50 | 46.08 | 37.29 | 52.46 | 43.43 | 47.15 | 38.90 |
| | GPT-4o | 43.10 | 32.08 | 38.39 | 29.24 | 42.32 | 30.55 | 51.49 | 39.09 | 43.83 | 32.74 |
| | Qwen-Max | 43.10 | 32.08 | 38.39 | 29.24 | 42.32 | 30.55 | 51.49 | 39.09 | 43.83 | 32.74 |
| ReviseDetect | Deepseek-V3 | 54.61 | 44.80 | 55.85 | 47.01 | 55.57 | 45.33 | 59.05 | 49.28 | 56.27 | 46.61 |
| | Gemini-2.50-flash | 54.93 | 44.95 | 56.17 | 47.07 | 55.77 | 45.41 | 58.96 | 49.38 | 56.46 | 46.70 |
| | GPT-4o | 54.27 | 44.04 | 55.44 | 46.47 | 55.35 | 44.83 | 58.96 | 48.68 | 56.01 | 46.01 |
| | Qwen-Max | 54.10 | 44.08 | 55.19 | 46.51 | 55.13 | 44.86 | 58.83 | 48.70 | 55.81 | 46.04 |
| GECScore | Deepseek-V3 | 54.61 | 44.80 | 55.85 | 47.01 | 55.57 | 45.33 | 59.05 | 49.28 | 56.27 | 46.61 |
| | Gemini-2.50-flash | 54.93 | 44.95 | 56.17 | 47.07 | 55.77 | 45.41 | 58.96 | 49.38 | 56.46 | 46.70 |
| | GPT-4o | 54.27 | 44.04 | 55.44 | 46.47 | 55.35 | 44.83 | 58.96 | 48.68 | 56.01 | 46.01 |
| | Qwen-Max | 54.10 | 44.08 | 55.19 | 46.51 | 55.13 | 44.86 | 58.83 | 48.70 | 55.81 | 46.04 |
| Lastde++ | Deepseek-V3 | 16.97 | 16.99 | 17.18 | 17.21 | 17.03 | 17.07 | 17.05 | 17.10 | 17.06 | 17.09 |
| | Gemini-2.50-flash | 16.70 | 16.70 | 16.69 | 16.69 | 16.67 | 16.67 | 16.69 | 16.69 | 16.69 | 16.69 |
| | GPT-4o | 16.67 | 16.67 | 16.67 | 16.67 | 16.68 | 16.68 | 16.67 | 16.67 | 16.67 | 16.67 |
| | Qwen-Max | 18.67 | 16.95 | 19.26 | 17.18 | 19.21 | 17.02 | 19.29 | 17.04 | 19.11 | 17.05 |
| RepreGuard | Deepseek-V3 | 24.59 | 17.07 | 23.94 | 16.49 | 24.87 | 16.88 | 25.66 | 17.00 | 24.77 | 16.86 |
| | Gemini-2.50-flash | 42.26 | 21.46 | 45.54 | 22.59 | 44.11 | 22.05 | 45.32 | 22.34 | 44.31 | 22.11 |
| | GPT-4o | 20.55 | 17.23 | 22.09 | 17.13 | 23.74 | 17.15 | 23.53 | 17.19 | 22.48 | 17.18 |
| | Qwen-Max | 29.33 | 26.04 | 26.25 | 24.50 | 28.44 | 25.50 | 31.40 | 26.98 | 28.86 | 25.76 |
| X-Rob-Classifier | Deepseek-V3 | 95.17 | 94.31 | 90.39 | 80.63 | 90.56 | 79.74 | 85.54 | 73.18 | 90.42 | 81.97 |
| | Gemini-2.50-flash | 92.98 | 90.41 | 94.83 | 93.21 | 90.89 | 81.93 | 88.45 | 77.93 | 91.79 | 85.87 |
| | GPT-4o | 87.06 | 85.95 | 87.89 | 78.14 | 89.04 | 84.53 | 88.52 | 80.03 | 88.25 | 82.16 |
| | Qwen-Max | 85.48 | 75.92 | 86.97 | 68.74 | 85.76 | 73.75 | 86.67 | 72.84 | 86.22 | 72.81 |
| mDeBERTa-Classifier | Deepseek-V3 | 97.78 | 97.61 | 89.27 | 84.28 | 90.73 | 84.27 | 85.58 | 78.05 | 90.84 | 86.05 |
| | Gemini-2.50-flash | 94.11 | 92.28 | 96.86 | 96.17 | 89.01 | 83.38 | 86.44 | 80.33 | 91.61 | 88.02 |
| | GPT-4o | 93.98 | 93.59 | 92.73 | 85.30 | 95.34 | 92.40 | 93.67 | 88.47 | 93.93 | 89.94 |
| | Qwen-Max | 91.14 | 90.24 | 91.10 | 82.60 | 91.35 | 86.61 | 92.54 | 88.34 | 91.53 | 86.92 |
| Biscope | Deepseek-V3 | 71.84 | 49.08 | 64.61 | 40.97 | 66.14 | 43.73 | 69.17 | 44.99 | 67.94 | 44.69 |
| | Gemini-2.50-flash | 67.68 | 44.79 | 68.60 | 46.99 | 66.36 | 45.43 | 67.67 | 43.66 | 67.58 | 45.22 |
| | GPT-4o | 69.39 | 45.48 | 66.43 | 47.23 | 69.25 | 47.08 | 69.91 | 45.72 | 68.75 | 46.13 |
| | Qwen-Max | 63.72 | 43.72 | 56.28 | 38.18 | 61.82 | 41.12 | 72.26 | 53.23 | 63.52 | 44.06 |

Table 20: Performance Comparison of Different Detectors on Generator Generalization in TERNARY Task. Detector performance is visualized using teal color gradients, with darker intensities indicating superior detection capabilities.

| Detectors↓ | Test↓ Lang→ | English | | Chinese | | Spanish | | Arabic | | French | | Russian | | Portuguese | | German | | Average | |
|---|---|---|---|---|---|---|---|---|---|---|---|---|---|---|---|---|---|---|---|
| | Metrics | $F_1^B$ | $F_1^F$ | $F_1^B$ | $F_1^F$ | $F_1^B$ | $F_1^F$ | $F_1^B$ | $F_1^F$ | $F_1^B$ | $F_1^F$ | $F_1^B$ | $F_1^F$ | $F_1^B$ | $F_1^F$ | $F_1^B$ | $F_1^F$ | $F_1^B$ | $F_1^F$ |
| Log-Likelihood | Academic | 65.50 | 40.39 | 56.69 | 37.69 | 61.53 | 47.50 | 67.19 | 38.24 | 54.31 | 49.79 | 47.05 | 57.47 | 61.36 | 48.62 | 73.94 | 35.18 | 62.69 | 45.17 |
| | News | 55.09 | 34.36 | 44.88 | 35.63 | 73.17 | 35.50 | 66.77 | 33.67 | 71.61 | 37.11 | 62.84 | 41.32 | 74.61 | 35.83 | 56.67 | 33.38 | 64.82 | 35.92 |
| | Novel | 64.33 | 43.20 | 54.83 | 38.43 | 67.67 | 54.07 | 69.01 | 42.97 | 60.98 | 55.41 | 49.06 | 61.40 | 66.43 | 55.61 | 72.72 | 37.74 | 64.74 | 49.73 |
| | SEO | 63.39 | 38.31 | 53.43 | 37.27 | 70.77 | 43.56 | 70.30 | 35.97 | 65.77 | 45.95 | 51.86 | 53.42 | 70.31 | 44.17 | 70.71 | 34.42 | 66.11 | 42.20 |
| | Wiki | 54.48 | 34.36 | 44.36 | 35.63 | 72.57 | 35.46 | 65.94 | 33.65 | 71.23 | 37.11 | 63.38 | 41.32 | 74.14 | 35.83 | 55.37 | 33.38 | 64.30 | 35.91 |
| | WebText | 56.86 | 35.80 | 46.26 | 36.34 | 74.10 | 38.24 | 68.69 | 34.12 | 72.06 | 39.95 | 61.29 | 46.45 | 75.38 | 38.47 | 59.76 | 33.60 | 65.89 | 38.09 |
| Log-Rank | Academic | 65.54 | 43.84 | 59.49 | 39.13 | 63.93 | 49.06 | 68.51 | 41.09 | 56.68 | 52.01 | 49.52 | 54.71 | 64.31 | 49.85 | 71.56 | 35.22 | 63.81 | 46.17 |
| | News | 58.73 | 37.00 | 50.59 | 36.99 | 72.95 | 37.66 | 69.73 | 34.43 | 71.18 | 39.93 | 62.79 | 42.43 | 74.67 | 37.79 | 55.86 | 33.49 | 65.83 | 37.55 |
| | Novel | 65.42 | 47.44 | 59.06 | 40.84 | 66.59 | 56.39 | 69.31 | 48.07 | 59.34 | 58.02 | 50.66 | 59.49 | 66.67 | 57.02 | 70.65 | 37.85 | 64.78 | 51.48 |
| | SEO | 64.14 | 43.82 | 57.59 | 39.10 | 70.82 | 48.90 | 70.72 | 40.95 | 65.66 | 51.88 | 54.07 | 54.53 | 71.03 | 49.73 | 67.88 | 35.15 | 66.45 | 46.06 |
| | Wiki | 56.67 | 36.71 | 47.99 | 36.82 | 71.04 | 37.02 | 67.03 | 34.21 | 69.70 | 39.52 | 64.38 | 41.70 | 73.03 | 37.32 | 51.08 | 33.44 | 63.90 | 37.16 |
| | WebText | 59.90 | 38.91 | 51.80 | 37.63 | 73.67 | 40.41 | 70.50 | 35.52 | 71.24 | 43.42 | 61.47 | 46.31 | 75.14 | 40.69 | 58.05 | 33.67 | 66.47 | 39.77 |
| DetectLLM-LRR | Academic | 60.99 | 51.21 | 68.34 | 48.14 | 64.71 | 45.26 | 70.39 | 46.41 | 58.55 | 49.36 | 56.92 | 41.52 | 66.61 | 43.54 | 60.39 | 34.10 | 64.16 | 45.18 |
| | News | 62.58 | 48.55 | 65.67 | 44.91 | 66.52 | 41.85 | 70.26 | 41.99 | 65.17 | 45.49 | 57.88 | 39.25 | 68.37 | 39.98 | 45.15 | 33.75 | 63.48 | 42.14 |
| | Novel | 61.10 | 52.29 | 68.31 | 49.79 | 65.07 | 46.92 | 70.42 | 48.61 | 58.78 | 50.74 | 57.06 | 42.55 | 66.93 | 45.37 | 60.16 | 34.47 | 64.27 | 46.62 |
| | SEO | 61.74 | 55.07 | 68.66 | 53.54 | 66.45 | 51.62 | 71.54 | 54.04 | 60.55 | 55.30 | 57.86 | 45.85 | 68.54 | 50.71 | 58.63 | 35.54 | 65.01 | 50.59 |
| | Wiki | 62.59 | 48.18 | 65.62 | 44.55 | 66.40 | 41.35 | 70.04 | 41.25 | 65.06 | 45.07 | 57.69 | 38.91 | 68.21 | 39.43 | 44.92 | 33.67 | 63.35 | 41.71 |
| | WebText | 62.62 | 46.83 | 67.16 | 43.39 | 67.68 | 39.91 | 71.61 | 39.93 | 65.33 | 43.44 | 58.51 | 38.00 | 69.71 | 38.42 | 48.13 | 33.61 | 64.63 | 40.57 |
| Fast-DetectGPT | Academic | 49.31 | 40.29 | 46.04 | 38.38 | 48.73 | 39.27 | 54.68 | 61.10 | 49.08 | 43.30 | 59.26 | 57.28 | 49.61 | 39.15 | 56.76 | 42.51 | 53.00 | 46.06 |
| | News | 48.56 | 39.31 | 45.81 | 37.51 | 48.55 | 37.89 | 59.08 | 59.87 | 50.41 | 41.40 | 63.05 | 54.60 | 49.65 | 37.88 | 56.90 | 40.38 | 54.05 | 44.38 |
| | Novel | 46.79 | 37.62 | 43.81 | 36.37 | 47.08 | 35.95 | 63.44 | 55.91 | 50.62 | 38.55 | 65.17 | 49.65 | 47.90 | 35.94 | 54.52 | 37.36 | 53.74 | 41.43 |
| | SEO | 47.98 | 38.14 | 45.09 | 36.80 | 48.06 | 36.64 | 61.79 | 57.40 | 50.85 | 39.50 | 64.90 | 51.62 | 48.69 | 36.51 | 56.11 | 38.43 | 54.25 | 42.48 |
| | Wiki | 49.32 | 39.32 | 46.04 | 37.42 | 48.71 | 37.88 | 54.68 | 59.72 | 49.05 | 41.42 | 59.28 | 54.46 | 49.54 | 37.88 | 56.72 | 40.31 | 52.98 | 44.31 |
| | WebText | 47.51 | 36.59 | 44.81 | 35.60 | 47.97 | 35.24 | 62.03 | 52.79 | 50.87 | 36.98 | 65.07 | 46.06 | 48.57 | 35.13 | 55.66 | 36.22 | 54.14 | 39.68 |
| Binoculars | Academic | 64.54 | 48.17 | 68.66 | 65.76 | 75.35 | 58.38 | 75.61 | 58.01 | 81.45 | 62.33 | 79.87 | 65.39 | 81.03 | 64.31 | 78.35 | 53.48 | 75.87 | 59.75 |
| | News | 64.54 | 48.17 | 68.66 | 65.76 | 75.35 | 58.38 | 75.61 | 58.01 | 81.45 | 62.33 | 79.87 | 65.39 | 81.03 | 64.31 | 78.35 | 53.48 | 75.87 | 59.75 |
| | Novel | 64.54 | 42.03 | 68.66 | 55.52 | 75.35 | 49.88 | 75.61 | 51.92 | 81.45 | 53.98 | 79.87 | 55.98 | 81.03 | 55.28 | 78.35 | 45.80 | 75.87 | 51.47 |
| | SEO | 66.85 | 48.17 | 60.47 | 65.76 | 70.23 | 58.38 | 80.37 | 58.01 | 78.09 | 62.33 | 74.11 | 65.39 | 74.97 | 64.31 | 82.49 | 53.48 | 73.88 | 59.75 |
| | Wiki | 64.54 | 48.17 | 68.66 | 65.76 | 75.35 | 58.38 | 75.61 | 58.01 | 81.45 | 62.33 | 79.87 | 65.39 | 81.03 | 64.31 | 78.35 | 53.48 | 75.87 | 59.75 |
| | WebText | 64.54 | 48.17 | 68.66 | 65.76 | 75.35 | 58.38 | 75.61 | 58.01 | 81.45 | 62.33 | 79.87 | 65.39 | 81.03 | 64.31 | 78.35 | 53.48 | 75.87 | 59.75 |
| ReviseDetect | Academic | 89.92 | 82.20 | 45.98 | 40.20 | 84.70 | 87.64 | 85.90 | 62.28 | 85.59 | 80.88 | 84.08 | 76.29 | 84.39 | 87.16 | 85.80 | 75.17 | 82.46 | 75.77 |
| | News | 88.56 | 0.00 | 42.47 | 0.00 | 88.08 | 0.00 | 77.94 | 0.00 | 85.97 | 0.00 | 83.22 | 0.00 | 87.76 | 0.00 | 83.83 | 0.00 | 81.54 | 0.00 |
| | Novel | 89.64 | 0.00 | 44.80 | 0.00 | 86.10 | 0.00 | 83.73 | 0.00 | 86.03 | 0.00 | 84.13 | 0.00 | 85.83 | 0.00 | 85.66 | 0.00 | 82.47 | 0.00 |
| | SEO | 89.23 | 0.00 | 50.44 | 0.00 | 81.16 | 0.00 | 87.85 | 0.00 | 83.48 | 0.00 | 83.26 | 0.00 | 81.41 | 0.00 | 84.61 | 0.00 | 81.55 | 0.00 |
| | Wiki | 88.38 | 60.38 | 42.19 | 38.86 | 88.24 | 72.21 | 76.98 | 36.45 | 85.76 | 57.21 | 82.92 | 60.00 | 87.91 | 70.11 | 83.50 | 48.83 | 81.31 | 56.53 |
| | WebText | 89.43 | 0.00 | 43.85 | 0.00 | 87.02 | 0.00 | 82.16 | 0.00 | 86.22 | 0.00 | 83.99 | 0.00 | 86.78 | 0.00 | 85.47 | 0.00 | 82.39 | 0.00 |
| GECScore | Academic | 91.02 | 91.89 | 58.45 | 55.64 | 79.65 | 90.98 | 87.30 | 70.33 | 81.55 | 89.29 | 82.25 | 80.93 | 82.34 | 91.49 | 79.19 | 84.16 | 80.98 | 82.39 |
| | News | 92.63 | 73.95 | 55.93 | 54.89 | 90.17 | 77.00 | 74.13 | 42.90 | 89.21 | 69.68 | 82.48 | 59.25 | 91.19 | 70.40 | 84.92 | 58.82 | 83.12 | 63.91 |
| | Novel | 91.11 | 30.39 | 55.49 | 30.57 | 91.30 | 31.30 | 67.57 | 30.68 | 89.25 | 30.76 | 79.85 | 31.08 | 91.18 | 31.33 | 83.38 | 30.64 | 81.71 | 30.84 |
| | SEO | 88.18 | 92.10 | 60.10 | 55.73 | 73.03 | 90.69 | 89.49 | 71.48 | 76.03 | 89.16 | 78.54 | 81.45 | 76.95 | 91.43 | 74.57 | 84.39 | 78.12 | 82.60 |
| | Wiki | 92.44 | 78.04 | 55.89 | 54.92 | 90.50 | 83.06 | 73.19 | 45.70 | 89.22 | 78.41 | 82.17 | 62.68 | 91.31 | 77.58 | 84.94 | 66.68 | 82.99 | 69.10 |
| | WebText | 92.77 | 77.27 | 56.01 | 54.91 | 89.96 | 82.07 | 74.81 | 44.85 | 88.98 | 77.04 | 83.01 | 62.09 | 91.06 | 76.61 | 84.89 | 65.25 | 83.22 | 68.21 |
| Lastde++ | Academic | 37.04 | 33.28 | 34.40 | 33.41 | 35.60 | 33.36 | 44.81 | 33.77 | 37.97 | 33.55 | 45.14 | 33.65 | 33.85 | 33.46 | 39.89 | 33.65 | 39.11 | 33.52 |
| | News | 37.77 | 33.43 | 34.42 | 33.54 | 35.24 | 33.44 | 44.12 | 34.17 | 36.83 | 33.66 | 45.31 | 33.99 | 32.97 | 33.59 | 38.32 | 34.13 | 38.78 | 33.75 |
| | Novel | 35.50 | 33.30 | 35.03 | 33.42 | 35.35 | 33.38 | 43.21 | 33.76 | 37.57 | 33.57 | 42.70 | 33.62 | 34.55 | 33.46 | 40.04 | 33.66 | 38.30 | 33.52 |
| | SEO | 33.33 | 33.32 | 33.33 | 33.38 | 33.33 | 33.36 | 33.33 | 33.72 | 33.35 | 33.47 | 33.37 | 33.52 | 33.36 | 33.43 | 33.34 | 33.50 | 33.34 | 33.46 |
| | Wiki | 37.60 | 33.44 | 34.50 | 33.54 | 35.55 | 33.46 | 44.90 | 34.23 | 37.58 | 33.68 | 45.36 | 33.99 | 33.44 | 33.60 | 39.50 | 34.17 | 39.16 | 33.77 |
| | WebText | 37.28 | 33.35 | 34.58 | 33.48 | 35.56 | 33.44 | 44.93 | 34.05 | 37.92 | 33.61 | 45.28 | 33.89 | 33.77 | 33.54 | 39.88 | 34.03 | 39.21 | 33.68 |
| RepreGuard | Academic | 88.34 | 87.67 | 95.51 | 95.40 | 91.29 | 91.83 | 82.53 | 84.05 | 96.41 | 96.61 | 86.79 | 87.61 | 92.21 | 92.88 | 95.35 | 95.29 | 91.22 | 91.53 |
| | News | 67.33 | 0.13 | 84.77 | 0.00 | 66.67 | 47.27 | 66.64 | 0.13 | 66.67 | 43.90 | 66.67 | 1.72 | 66.67 | 56.53 | 66.67 | 32.77 | 68.50 | 26.86 |
| | Novel | 66.57 | 64.19 | 66.38 | 40.95 | 72.29 | 9.71 | 68.57 | 0.76 | 77.02 | 3.92 | 73.24 | 0.53 | 75.09 | 2.40 | 83.57 | 2.40 | 72.65 | 22.62 |
| | SEO | 85.54 | 88.48 | 56.73 | 64.64 | 93.39 | 90.66 | 44.94 | 51.76 | 96.01 | 93.75 | 58.55 | 69.86 | 92.66 | 89.40 | 94.87 | 94.77 | 81.39 | 82.89 |
| | Wiki | 91.43 | 9.10 | 67.22 | 89.04 | 63.95 | 1.29 | 66.64 | 88.46 | 84.52 | 5.10 | 66.59 | 86.63 | 66.90 | 2.27 | 78.72 | 15.64 | 72.59 | 50.48 |
| | WebText | 85.84 | 90.51 | 77.93 | 56.03 | 80.74 | 90.65 | 70.89 | 34.79 | 89.12 | 95.11 | 86.65 | 85.04 | 85.01 | 93.94 | 76.85 | 88.74 | 82.05 | 82.85 |
| X-Rob-Classifier | Academic | 91.69 | 54.07 | 84.70 | 37.07 | 85.87 | 48.51 | 89.92 | 37.64 | 86.87 | 44.26 | 85.56 | 37.96 | 87.19 | 40.47 | 85.83 | 36.45 | 92.27 | 69.33 |
| | News | 98.96 | 98.85 | 98.09 | 97.97 | 97.05 | 96.38 | 99.06 | 99.06 | 98.77 | 98.67 | 98.22 | 98.10 | 99.14 | 99.08 | 97.85 | 97.12 | 98.09 | 97.69 |
| | Novel | 99.42 | 99.32 | 96.49 | 95.43 | 98.48 | 98.48 | 98.75 | 98.65 | 98.93 | 98.71 | 96.82 | 95.13 | 98.88 | 98.76 | 98.04 | 97.48 | 97.78 | 97.23 |
| | SEO | 98.72 | 98.35 | 96.49 | 94.08 | 91.45 | 78.84 | 94.89 | 88.05 | 94.90 | 87.90 | 91.55 | 83.15 | 93.80 | 90.26 | 91.86 | 79.22 | 92.92 | 84.50 |
| | Wiki | 93.91 | 61.84 | 91.70 | 74.87 | 84.56 | 44.60 | 87.20 | 70.17 | 88.07 | 83.21 | 93.49 | 74.01 | 92.77 | 88.37 | 89.58 | 83.24 | 89.07 | 63.13 |
| | WebText | 99.05 | 98.89 | 98.20 | 98.05 | 95.10 | 84.48 | 98.83 | 91.77 | 98.92 | 98.72 | 98.86 | 98.67 | 99.34 | 99.27 | 96.69 | 83.67 | 98.40 | 98.05 |
| mDeBERTa-Classifier | Academic | 97.02 | 86.42 | 92.78 | 77.17 | 95.46 | 80.04 | 96.57 | 90.38 | 96.68 | 91.29 | 93.68 | 78.13 | 96.53 | 84.94 | 92.79 | 68.90 | 94.72 | 85.03 |
| | News | 99.05 | 99.03 | 99.17 | 99.11 | 97.43 | 96.56 | 99.62 | 99.45 | 98.78 | 98.61 | 98.17 | 98.09 | 99.12 | 99.07 | 98.39 | 97.95 | 97.92 | 97.60 |
| | Novel | 99.80 | 99.48 | 98.03 | 97.92 | 98.51 | 98.17 | 99.59 | 99.40 | 99.08 | 99.02 | 98.85 | 98.75 | 99.67 | 99.42 | 99.17 | 99.12 | 98.94 | 98.91 |
| | SEO | 98.36 | 97.13 | 97.13 | 93.52 | 92.44 | 87.34 | 94.68 | 88.75 | 92.77 | 84.67 | 87.14 | 77.94 | 92.04 | 84.50 | 90.01 | 79.28 | 93.15 | 84.09 |
| | Wiki | 96.36 | 91.87 | 95.02 | 92.10 | 89.39 | 83.45 | 95.39 | 93.88 | 94.77 | 91.48 | 95.92 | 94.23 | 97.13 | 96.26 | 92.45 | 84.75 | 93.61 | 82.45 |
| | WebText | 99.35 | 99.32 | 98.76 | 98.71 | 97.75 | 96.59 | 99.60 | 99.42 | 99.00 | 98.97 | 99.53 | 99.33 | 99.67 | 99.44 | 97.86 | 96.08 | 97.89 | 97.12 |
| Biscope | Academic | 89.17 | 69.44 | 88.97 | 62.17 | 95.14 | 89.52 | 88.03 | 68.07 | 93.95 | 86.10 | 82.90 | 33.33 | 95.51 | 89.56 | 91.92 | 70.87 | 87.64 | 68.12 |
| | News | 84.60 | 67.10 | 91.43 | 83.87 | 94.98 | 92.65 | 83.23 | 68.60 | 95.69 | 91.83 | 86.79 | 66.95 | 95.82 | 92.85 | 92.62 | 83.89 | 89.91 | 79.44 |
| | Novel | 89.43 | 84.03 | 91.50 | 75.12 | 95.33 | 86.90 | 83.72 | 57.10 | 96.01 | 90.00 | 82.78 | 56.96 | 96.11 | 89.29 | 92.03 | 77.11 | 89.53 | 75.14 |
| | SEO | 88.26 | 55.47 | 91.44 | 81.94 | 94.99 | 91.32 | 85.82 | 53.39 | 94.12 | 84.33 | 77.36 | 50.84 | 94.62 | 85.37 | 92.27 | 77.48 | 89.26 | 68.49 |
| | Wiki | 86.06 | 66.62 | 89.75 | 80.70 | 93.60 | 89.64 | 86.95 | 70.29 | 93.69 | 87.53 | 86.79 | 55.91 | 94.56 | 90.21 | 90.32 | 77.40 | 89.35 | 74.87 |
| | WebText | 87.76 | 58.86 | 92.69 | 87.59 | 94.46 | 90.92 | 86.85 | 72.35 | 95.35 | 91.85 | 88.04 | 63.95 | 95.77 | 91.68 | 91.97 | 83.18 | 90.94 | 78.16 |

Table 21: Performance Comparison of Different Detectors on Cross Domain in different languages in BINARY Task. Detector performance is visualized using teal color gradients, with darker intensities indicating superior detection capabilities.

| Detectors↓ | Test↓ Lang→ | English | | Chinese | | Spanish | | Arabic | | French | | Russian | | Portuguese | | German | | Average | |
|---|---|---|---|---|---|---|---|---|---|---|---|---|---|---|---|---|---|---|---|
| | Metrics | $F_1^B$ | $F_1^F$ | $F_1^B$ | $F_1^F$ | $F_1^B$ | $F_1^F$ | $F_1^B$ | $F_1^F$ | $F_1^B$ | $F_1^F$ | $F_1^B$ | $F_1^F$ | $F_1^B$ | $F_1^F$ | $F_1^B$ | $F_1^F$ | $F_1^B$ | $F_1^F$ |
| Log-Likelihood | Academic | 41.66 | 34.91 | 37.84 | 29.15 | 40.25 | 36.76 | 43.66 | 37.29 | 36.82 | 33.64 | 26.49 | 27.12 | 37.79 | 36.34 | 52.21 | 39.41 | 41.45 | 34.95 |
| | News | 30.76 | 30.76 | 24.37 | 24.37 | 40.41 | 40.41 | 36.25 | 36.25 | 38.98 | 38.98 | 33.18 | 33.18 | 40.69 | 40.69 | 32.48 | 32.48 | 35.41 | 35.41 |
| | Novel | 35.06 | 35.06 | 29.44 | 29.44 | 35.96 | 35.96 | 37.04 | 37.04 | 32.48 | 32.48 | 26.34 | 26.34 | 35.40 | 35.40 | 39.71 | 39.71 | 34.58 | 34.58 |
| | SEO | 37.86 | 32.58 | 33.77 | 26.19 | 46.20 | 40.28 | 49.64 | 37.48 | 43.92 | 38.46 | 31.94 | 31.15 | 47.68 | 40.18 | 48.16 | 35.65 | 44.16 | 35.92 |
| | Wiki | 34.57 | 29.85 | 30.70 | 23.65 | 47.38 | 39.84 | 49.51 | 35.44 | 46.94 | 38.87 | 37.59 | 33.84 | 51.34 | 40.34 | 41.46 | 30.64 | 44.19 | 34.90 |
| | WebText | 31.60 | 31.60 | 25.35 | 25.35 | 40.52 | 40.52 | 37.18 | 37.18 | 38.91 | 38.91 | 32.16 | 32.16 | 40.59 | 40.59 | 34.11 | 34.11 | 35.78 | 35.78 |
| Log-Rank | Academic | 43.09 | 35.36 | 40.67 | 31.56 | 40.12 | 36.63 | 43.69 | 37.54 | 36.85 | 32.75 | 27.98 | 28.01 | 38.83 | 36.41 | 50.91 | 38.37 | 41.79 | 35.02 |
| | News | 32.39 | 32.39 | 27.44 | 27.44 | 40.05 | 40.05 | 37.38 | 37.38 | 38.64 | 38.64 | 33.39 | 33.39 | 40.58 | 40.58 | 31.29 | 31.29 | 35.74 | 35.74 |
| | Novel | 35.46 | 35.46 | 31.92 | 31.92 | 34.25 | 34.25 | 36.98 | 36.98 | 30.46 | 30.46 | 26.73 | 26.73 | 34.47 | 34.47 | 39.06 | 39.06 | 34.17 | 34.17 |
| | SEO | 39.44 | 33.84 | 37.58 | 28.92 | 45.07 | 39.96 | 50.47 | 38.09 | 43.36 | 38.07 | 34.30 | 32.03 | 47.97 | 40.18 | 45.36 | 33.98 | 44.36 | 36.11 |
| | Wiki | 38.31 | 31.73 | 36.42 | 26.44 | 42.75 | 39.51 | 50.36 | 36.55 | 42.56 | 38.49 | 37.21 | 34.21 | 46.41 | 40.27 | 42.85 | 29.44 | 43.94 | 35.24 |
| | WebText | 33.30 | 33.30 | 28.34 | 28.34 | 40.15 | 40.15 | 37.83 | 37.83 | 38.25 | 38.25 | 32.56 | 32.56 | 40.48 | 40.48 | 32.86 | 32.86 | 36.00 | 36.00 |
| DetectLLM-LRR | Academic | 40.68 | 32.94 | 44.41 | 36.78 | 40.00 | 35.57 | 45.93 | 38.22 | 38.25 | 32.17 | 36.92 | 30.81 | 43.25 | 36.60 | 39.32 | 32.76 | 41.70 | 34.75 |
| | News | 33.97 | 33.97 | 36.46 | 36.46 | 36.96 | 36.96 | 38.28 | 38.28 | 35.41 | 35.41 | 31.40 | 31.40 | 38.19 | 38.19 | 26.34 | 26.34 | 35.06 | 35.06 |
| | Novel | 31.75 | 31.75 | 35.88 | 35.88 | 32.42 | 32.42 | 37.05 | 37.05 | 28.85 | 28.85 | 29.19 | 29.19 | 33.77 | 33.77 | 34.92 | 34.92 | 33.27 | 33.27 |
| | SEO | 42.22 | 33.63 | 45.51 | 37.02 | 41.42 | 36.62 | 48.95 | 38.53 | 40.09 | 33.97 | 38.41 | 31.26 | 45.16 | 37.87 | 37.67 | 30.57 | 43.15 | 35.26 |
| | Wiki | 37.88 | 34.11 | 36.58 | 35.83 | 37.87 | 36.48 | 35.18 | 37.63 | 37.48 | 35.51 | 31.53 | 31.06 | 38.62 | 37.58 | 26.55 | 24.42 | 35.70 | 34.55 |
| | WebText | 33.88 | 33.88 | 36.93 | 36.93 | 37.09 | 37.09 | 38.74 | 38.74 | 34.83 | 34.83 | 31.52 | 31.52 | 38.37 | 38.37 | 28.45 | 28.45 | 35.36 | 35.36 |
| Fast-DetectGPT | Academic | 30.03 | 24.72 | 32.70 | 23.10 | 30.87 | 24.91 | 38.38 | 34.34 | 34.38 | 27.26 | 43.60 | 34.93 | 33.26 | 25.38 | 38.09 | 29.05 | 36.24 | 28.72 |
| | News | 25.65 | 25.65 | 23.97 | 23.97 | 25.75 | 25.75 | 33.27 | 33.27 | 27.37 | 27.37 | 34.68 | 34.68 | 25.99 | 25.99 | 30.39 | 30.39 | 29.06 | 29.06 |
| | Novel | 24.01 | 24.01 | 22.59 | 22.59 | 24.03 | 24.03 | 34.43 | 34.43 | 26.75 | 26.75 | 34.74 | 34.74 | 24.56 | 24.56 | 27.88 | 27.88 | 28.17 | 28.17 |
| | SEO | 21.82 | 17.53 | 26.62 | 17.80 | 23.21 | 17.30 | 35.01 | 28.39 | 22.59 | 17.23 | 30.44 | 19.86 | 24.38 | 17.27 | 25.08 | 17.20 | 26.85 | 19.34 |
| | Wiki | 25.48 | 25.48 | 23.82 | 23.82 | 25.70 | 25.70 | 33.39 | 33.39 | 27.34 | 27.34 | 34.75 | 34.75 | 25.98 | 25.98 | 30.08 | 30.08 | 29.02 | 29.02 |
| | WebText | 22.55 | 22.29 | 21.68 | 21.13 | 22.84 | 22.25 | 35.83 | 34.03 | 25.41 | 24.96 | 34.28 | 33.21 | 22.81 | 22.43 | 25.35 | 24.91 | 27.14 | 26.45 |
| Binoculars | Academic | 34.92 | 25.97 | 48.33 | 36.27 | 44.45 | 32.16 | 39.45 | 31.95 | 45.56 | 34.60 | 48.86 | 36.35 | 50.37 | 35.67 | 35.94 | 29.26 | 44.06 | 32.97 |
| | News | 41.84 | 31.14 | 48.49 | 39.12 | 52.17 | 39.11 | 48.22 | 37.33 | 55.56 | 42.02 | 55.05 | 42.08 | 57.52 | 42.32 | 48.85 | 37.45 | 51.60 | 39.00 |
| | Novel | 34.92 | 25.97 | 48.33 | 36.27 | 44.45 | 32.16 | 39.45 | 31.95 | 45.56 | 34.60 | 48.86 | 36.35 | 50.37 | 35.67 | 35.94 | 29.26 | 44.06 | 32.97 |
| | SEO | 41.84 | 31.14 | 48.49 | 39.12 | 52.17 | 39.11 | 48.22 | 37.33 | 55.56 | 42.02 | 55.05 | 42.08 | 57.52 | 42.32 | 48.85 | 37.45 | 51.60 | 39.00 |
| | Wiki | 34.92 | 25.97 | 48.33 | 36.27 | 44.45 | 32.16 | 39.45 | 31.95 | 45.56 | 34.60 | 48.86 | 36.35 | 50.37 | 35.67 | 35.94 | 29.26 | 44.06 | 32.97 |
| | WebText | 34.92 | 25.97 | 48.33 | 36.27 | 44.45 | 32.16 | 39.45 | 31.95 | 45.56 | 34.60 | 48.86 | 36.35 | 50.37 | 35.67 | 35.94 | 29.26 | 44.06 | 32.97 |
| Revise-Detect | Academic | 61.93 | 50.13 | 35.59 | 33.10 | 57.94 | 50.37 | 57.33 | 44.08 | 58.91 | 48.94 | 57.06 | 46.87 | 57.70 | 51.35 | 57.41 | 46.94 | 56.44 | 46.98 |
| | News | 60.70 | 49.55 | 34.92 | 32.99 | 58.88 | 50.40 | 55.59 | 42.96 | 58.86 | 48.62 | 56.68 | 46.44 | 58.52 | 51.24 | 56.91 | 46.37 | 56.06 | 46.61 |
| | Novel | 28.24 | 17.46 | 21.68 | 18.22 | 28.85 | 17.27 | 26.87 | 18.64 | 27.42 | 17.58 | 27.62 | 17.73 | 31.49 | 17.90 | 27.68 | 17.53 | 27.62 | 17.77 |
| | SEO | 62.63 | 49.57 | 38.21 | 32.99 | 51.94 | 50.40 | 57.78 | 42.96 | 55.65 | 48.62 | 53.88 | 46.44 | 53.57 | 51.24 | 55.07 | 46.37 | 55.17 | 46.61 |
| | Wiki | 57.55 | 46.80 | 34.07 | 32.63 | 59.11 | 49.62 | 50.21 | 38.00 | 56.63 | 45.64 | 54.49 | 43.15 | 59.45 | 49.88 | 54.27 | 42.94 | 54.20 | 44.24 |
| | WebText | 58.89 | 49.35 | 34.38 | 32.95 | 58.71 | 50.41 | 53.72 | 42.61 | 58.28 | 48.37 | 55.89 | 46.12 | 58.77 | 51.22 | 55.88 | 45.99 | 55.15 | 46.43 |
| GECScore | Academic | 66.03 | 60.96 | 53.63 | 43.27 | 21.31 | 10.71 | 53.42 | 45.52 | 58.19 | 58.63 | 21.31 | 10.71 | 53.53 | 52.59 | 53.38 | 53.91 | 52.40 | 53.63 |
| | News | 65.85 | 48.88 | 48.80 | 30.41 | 16.96 | 0.00 | 51.81 | 34.39 | 61.71 | 48.39 | 16.96 | 0.00 | 54.70 | 40.70 | 57.12 | 44.78 | 56.48 | 49.32 |
| | Novel | 50.03 | 49.38 | 30.78 | 30.18 | 0.55 | 0.00 | 37.13 | 36.47 | 49.06 | 48.42 | 0.55 | 0.00 | 40.57 | 39.90 | 45.96 | 45.32 | 51.11 | 50.53 |
| | SEO | 60.86 | 56.86 | 53.88 | 37.86 | 22.87 | 7.10 | 55.88 | 47.30 | 52.51 | 53.52 | 22.87 | 7.10 | 51.68 | 45.50 | 52.40 | 51.02 | 51.91 | 55.26 |
| | Wiki | 65.83 | 49.50 | 49.76 | 31.31 | 17.35 | 0.00 | 50.97 | 34.07 | 61.56 | 49.17 | 17.35 | 0.00 | 55.45 | 42.12 | 56.19 | 44.74 | 54.89 | 48.60 |
| | WebText | 66.85 | 45.17 | 55.38 | 30.74 | 22.76 | 0.18 | 49.48 | 27.23 | 61.85 | 45.31 | 22.76 | 0.18 | 55.79 | 37.52 | 55.35 | 39.83 | 52.79 | 44.49 |
| Lastde++ | Academic | 23.32 | 16.89 | 24.00 | 16.99 | 20.12 | 17.00 | 19.74 | 17.05 | 20.13 | 16.79 | 22.40 | 16.81 | 19.66 | 17.13 | 19.28 | 17.07 | 21.36 | 16.97 |
| | News | 16.71 | 16.70 | 16.77 | 16.81 | 16.66 | 16.66 | 16.67 | 16.67 | 16.66 | 16.66 | 16.66 | 16.66 | 16.66 | 16.66 | 16.67 | 16.67 | 16.68 | 16.69 |
| | Novel | 21.21 | 16.86 | 22.49 | 16.96 | 24.10 | 16.94 | 24.09 | 17.00 | 22.37 | 16.75 | 21.75 | 16.81 | 23.80 | 17.08 | 23.09 | 17.03 | 22.98 | 16.93 |
| | SEO | 16.67 | 16.67 | 16.71 | 16.71 | 16.67 | 16.67 | 16.67 | 16.67 | 16.67 | 16.67 | 16.67 | 16.67 | 16.67 | 16.67 | 16.67 | 16.67 | 16.67 | 16.67 |
| | Wiki | 24.62 | 17.20 | 26.42 | 17.32 | 26.64 | 17.64 | 26.35 | 17.76 | 24.71 | 17.17 | 25.28 | 17.18 | 26.18 | 17.64 | 25.17 | 17.54 | 25.86 | 17.44 |
| | WebText | 18.53 | 17.16 | 19.03 | 17.27 | 20.13 | 17.60 | 20.63 | 17.62 | 19.20 | 17.11 | 18.99 | 17.13 | 20.27 | 17.56 | 19.91 | 17.44 | 19.63 | 17.36 |
| Repreguard | Academic | 47.19 | 48.45 | 40.20 | 37.78 | 50.86 | 48.96 | 41.92 | 38.89 | 50.61 | 48.89 | 45.89 | 43.61 | 49.81 | 47.71 | 48.02 | 50.79 | 46.47 | 45.34 |
| | News | 29.73 | 16.67 | 19.41 | 16.67 | 16.65 | 17.86 | 15.03 | 16.67 | 16.65 | 16.99 | 16.60 | 16.68 | 16.65 | 17.64 | 16.65 | 16.89 | 24.43 | 17.12 |
| | Novel | 25.63 | 21.83 | 27.72 | 27.33 | 40.56 | 22.81 | 19.93 | 16.73 | 46.07 | 19.00 | 22.58 | 16.71 | 44.37 | 19.09 | 42.28 | 16.98 | 40.36 | 23.60 |
| | SEO | 16.75 | 16.69 | 16.67 | 16.67 | 30.70 | 29.05 | 16.69 | 16.67 | 27.23 | 24.86 | 18.05 | 16.96 | 31.19 | 30.68 | 24.55 | 21.29 | 25.65 | 23.35 |
| | Wiki | 19.16 | 16.69 | 42.41 | 16.75 | 42.91 | 16.75 | 16.83 | 49.51 | 47.40 | 17.92 | 16.88 | 37.54 | 48.15 | 17.20 | 31.38 | 20.16 | 37.00 | 27.15 |
| | WebText | 34.30 | 37.31 | 18.42 | 16.69 | 47.77 | 45.78 | 16.72 | 16.67 | 50.09 | 31.70 | 16.82 | 16.67 | 51.34 | 40.17 | 35.46 | 22.61 | 36.33 | 30.85 |
| X-Rob-Classifier | Academic | 70.67 | 60.86 | 62.10 | 38.70 | 64.43 | 18.97 | 59.17 | 46.23 | 70.26 | 47.40 | 65.33 | 29.72 | 69.41 | 57.00 | 65.66 | 20.20 | 66.06 | 34.45 |
| | News | 79.23 | 79.93 | 76.95 | 74.42 | 75.84 | 72.00 | 80.67 | 67.28 | 84.08 | 79.85 | 79.05 | 74.07 | 86.82 | 84.06 | 75.27 | 68.99 | 79.99 | 73.26 |
| | Novel | 74.37 | 71.14 | 70.29 | 59.48 | 72.54 | 58.13 | 75.75 | 62.80 | 78.38 | 61.63 | 68.35 | 55.82 | 80.28 | 71.82 | 75.11 | 65.74 | 73.78 | 61.18 |
| | SEO | 88.35 | 82.35 | 77.11 | 65.86 | 78.10 | 54.10 | 64.70 | 43.99 | 70.56 | 59.01 | 60.16 | 44.67 | 63.66 | 48.11 | 62.27 | 50.39 | 69.98 | 47.45 |
| | Wiki | 73.09 | 45.08 | 61.13 | 46.36 | 61.68 | 22.55 | 67.44 | 37.49 | 68.43 | 36.64 | 64.93 | 47.56 | 73.00 | 54.48 | 67.24 | 21.38 | 68.19 | 28.60 |
| | WebText | 86.06 | 59.01 | 77.01 | 63.05 | 71.31 | 43.49 | 82.48 | 72.93 | 79.88 | 54.59 | 78.41 | 61.26 | 84.43 | 73.40 | 77.20 | 53.87 | 80.06 | 58.14 |
| mDeBERTa-Classifier | Academic | 75.98 | 63.00 | 67.68 | 55.33 | 65.64 | 36.25 | 65.38 | 42.71 | 71.13 | 50.67 | 73.84 | 54.60 | 77.21 | 59.48 | 68.89 | 39.51 | 69.67 | 46.18 |
| | News | 88.71 | 85.80 | 86.04 | 83.57 | 88.42 | 73.93 | 88.03 | 84.03 | 91.49 | 88.53 | 87.53 | 79.30 | 92.04 | 88.22 | 87.48 | 81.75 | 89.10 | 83.22 |
| | Novel | 71.31 | 62.80 | 75.94 | 56.38 | 74.10 | 46.16 | 70.59 | 66.31 | 76.58 | 64.75 | 74.34 | 63.94 | 76.12 | 69.55 | 75.77 | 66.01 | 75.89 | 57.03 |
| | SEO | 86.07 | 74.50 | 80.12 | 70.96 | 83.16 | 63.51 | 70.82 | 53.36 | 71.40 | 59.14 | 63.07 | 47.84 | 71.42 | 58.81 | 63.41 | 44.02 | 73.84 | 48.62 |
| | Wiki | 84.35 | 81.55 | 66.67 | 63.99 | 72.17 | 30.74 | 79.81 | 70.55 | 77.45 | 61.55 | 75.53 | 51.21 | 84.97 | 79.21 | 76.33 | 58.85 | 76.74 | 55.36 |
| | WebText | 93.34 | 88.63 | 87.24 | 76.72 | 84.98 | 55.45 | 92.09 | 89.96 | 89.30 | 78.05 | 87.33 | 78.35 | 94.00 | 89.89 | 87.20 | 74.45 | 89.76 | 78.46 |
| Biscope | Academic | 66.09 | 44.80 | 54.63 | 32.32 | 63.40 | 50.89 | 59.22 | 36.42 | 60.00 | 48.59 | 53.63 | 16.67 | 62.99 | 50.29 | 62.32 | 41.04 | 60.94 | 39.07 |
| | News | 62.74 | 45.36 | 60.62 | 34.71 | 72.12 | 51.57 | 52.63 | 28.71 | 72.44 | 54.50 | 58.30 | 36.05 | 72.60 | 55.14 | 62.90 | 45.70 | 64.65 | 45.28 |
| | Novel | 65.65 | 50.13 | 53.70 | 29.87 | 71.91 | 49.82 | 53.09 | 31.68 | 71.35 | 52.17 | 53.21 | 31.98 | 72.37 | 51.77 | 63.13 | 44.37 | 63.34 | 42.65 |
| | SEO | 66.12 | 34.81 | 64.30 | 43.57 | 66.78 | 50.52 | 64.04 | 29.95 | 65.79 | 47.63 | 52.23 | 16.67 | 67.26 | 48.42 | 67.51 | 45.15 | 64.52 | 37.48 |
| | Wiki | 69.05 | 39.75 | 58.43 | 43.59 | 68.41 | 50.44 | 60.89 | 32.89 | 67.93 | 49.20 | 57.30 | 32.45 | 70.43 | 50.96 | 58.49 | 43.19 | 64.04 | 42.84 |
| | WebText | 66.39 | 38.71 | 63.36 | 31.48 | 71.57 | 51.47 | 53.53 | 25.93 | 71.66 | 52.80 | 58.22 | 32.49 | 73.19 | 54.20 | 63.03 | 39.65 | 65.39 | 42.02 |

Table 22: Performance Comparison of Different Detectors on Cross Domain in different languages in TERNARY Task. Detector performance is visualized using teal color gradients, with darker intensities indicating superior detection capabilities.

| Detectors↓ | Test↓ Lang→ | English | | Chinese | | Spanish | | Arabic | | French | | Russian | | Portuguese | | German | | Average | |
|---|---|---|---|---|---|---|---|---|---|---|---|---|---|---|---|---|---|---|---|
| | Metrics | $F_1^B$ | $F_1^F$ | $F_1^B$ | $F_1^F$ | $F_1^B$ | $F_1^F$ | $F_1^B$ | $F_1^F$ | $F_1^B$ | $F_1^F$ | $F_1^B$ | $F_1^F$ | $F_1^B$ | $F_1^F$ | $F_1^B$ | $F_1^F$ | $F_1^B$ | $F_1^F$ |
| Log-Likelihood | Deepseek-V3 | 61.45 | 35.80 | 51.30 | 36.34 | 73.04 | 38.24 | 71.08 | 34.10 | 69.54 | 39.95 | 54.84 | 46.43 | 73.16 | 38.47 | 67.68 | 33.60 | 66.80 | 38.08 |
| | Gemini-2.50-flash | 64.00 | 35.90 | 54.15 | 36.44 | 69.28 | 38.60 | 69.69 | 34.21 | 63.27 | 40.31 | 50.41 | 46.98 | 68.32 | 38.96 | 71.73 | 33.63 | 65.43 | 38.37 |
| | GPT-4o | 61.39 | 35.62 | 51.28 | 36.23 | 73.10 | 37.73 | 71.13 | 34.04 | 69.59 | 39.60 | 54.85 | 45.79 | 73.19 | 38.15 | 67.64 | 33.58 | 66.81 | 37.79 |
| | Qwen-Max | 55.09 | 35.80 | 44.88 | 36.34 | 73.17 | 38.24 | 66.77 | 34.12 | 71.61 | 39.95 | 62.84 | 46.45 | 74.61 | 38.47 | 56.67 | 33.60 | 64.82 | 38.09 |
| Log-Rank | Deepseek-V3 | 63.27 | 39.13 | 56.27 | 37.68 | 72.45 | 40.66 | 71.72 | 35.61 | 68.41 | 43.63 | 56.21 | 46.76 | 72.74 | 40.94 | 65.73 | 33.76 | 67.04 | 39.98 |
| | Gemini-2.50-flash | 64.74 | 39.55 | 58.11 | 37.72 | 69.94 | 41.50 | 70.47 | 35.98 | 63.98 | 44.52 | 52.85 | 47.39 | 70.00 | 41.64 | 68.73 | 33.84 | 66.09 | 40.50 |
| | GPT-4o | 63.31 | 38.83 | 56.43 | 37.58 | 72.32 | 40.28 | 71.63 | 35.43 | 68.07 | 43.26 | 56.10 | 46.11 | 72.56 | 40.54 | 66.10 | 33.64 | 67.00 | 39.65 |
| | Qwen-Max | 57.08 | 38.83 | 48.51 | 37.58 | 71.51 | 40.26 | 67.38 | 35.43 | 70.14 | 43.24 | 64.40 | 46.11 | 73.45 | 40.54 | 51.87 | 33.64 | 64.33 | 39.64 |
| DetectLLM-LRR | Deepseek-V3 | 62.41 | 49.50 | 68.62 | 46.19 | 67.70 | 43.15 | 72.03 | 43.37 | 62.53 | 46.86 | 58.45 | 39.84 | 69.97 | 41.27 | 56.00 | 33.85 | 65.47 | 43.19 |
| | Gemini-2.50-flash | 61.44 | 49.39 | 68.48 | 45.92 | 66.00 | 42.95 | 71.27 | 43.13 | 59.96 | 46.69 | 57.51 | 39.82 | 68.02 | 41.14 | 59.14 | 33.85 | 64.75 | 43.05 |
| | GPT-4o | 62.69 | 49.21 | 68.47 | 45.79 | 68.10 | 42.86 | 72.51 | 43.02 | 64.36 | 46.54 | 58.85 | 39.72 | 70.46 | 41.05 | 53.05 | 33.81 | 65.58 | 42.93 |
| | Qwen-Max | 62.41 | 49.50 | 65.24 | 46.19 | 65.74 | 43.15 | 69.47 | 43.37 | 64.66 | 46.83 | 57.11 | 39.84 | 67.47 | 41.27 | 43.85 | 33.85 | 62.78 | 43.19 |
| Fast-DetectGPT | Deepseek-V3 | 48.59 | 38.21 | 45.83 | 36.82 | 48.50 | 36.66 | 59.01 | 57.46 | 50.31 | 39.49 | 62.81 | 51.63 | 49.61 | 36.51 | 57.00 | 38.44 | 54.00 | 42.50 |
| | Gemini-2.50-flash | 48.19 | 38.20 | 45.40 | 36.84 | 48.51 | 36.67 | 60.74 | 57.53 | 50.79 | 39.49 | 63.98 | 51.69 | 48.98 | 36.51 | 56.47 | 38.44 | 54.21 | 42.53 |
| | GPT-4o | 48.58 | 37.98 | 45.80 | 36.69 | 48.40 | 36.30 | 59.17 | 56.95 | 50.37 | 39.19 | 63.24 | 50.97 | 49.69 | 36.26 | 56.76 | 38.08 | 54.05 | 42.13 |
| | Qwen-Max | 47.51 | 38.25 | 44.79 | 36.90 | 47.93 | 36.81 | 62.14 | 57.72 | 50.88 | 39.61 | 65.10 | 51.94 | 48.54 | 36.64 | 55.69 | 38.47 | 54.15 | 42.66 |
| Binoculars | Deepseek-V3 | 64.54 | 48.17 | 68.66 | 65.76 | 75.35 | 58.38 | 75.61 | 58.01 | 81.45 | 62.33 | 79.87 | 65.39 | 81.03 | 64.31 | 78.35 | 53.48 | 75.87 | 59.75 |
| | Gemini-2.50-flash | 64.54 | 48.17 | 68.66 | 65.76 | 75.35 | 58.38 | 75.61 | 58.01 | 81.45 | 62.33 | 79.87 | 65.39 | 81.03 | 64.31 | 78.35 | 53.48 | 75.87 | 59.75 |
| | GPT-4o | 64.54 | 48.17 | 68.66 | 65.76 | 75.35 | 58.38 | 75.61 | 58.01 | 81.45 | 62.33 | 79.87 | 65.39 | 81.03 | 64.31 | 78.35 | 53.48 | 75.87 | 59.75 |
| | Qwen-Max | 64.54 | 48.17 | 68.66 | 65.76 | 75.35 | 58.38 | 75.61 | 58.01 | 81.45 | 62.33 | 79.87 | 65.39 | 81.03 | 64.31 | 78.35 | 53.48 | 75.87 | 59.75 |
| ReviseDetect | Deepseek-V3 | 89.64 | 0.00 | 44.80 | 0.00 | 86.10 | 0.00 | 83.73 | 0.00 | 86.03 | 0.00 | 84.13 | 0.00 | 85.83 | 0.00 | 85.66 | 0.00 | 82.47 | 0.00 |
| | Gemini-2.50-flash | 89.43 | 0.00 | 43.85 | 0.00 | 86.97 | 0.00 | 82.24 | 0.00 | 86.23 | 0.00 | 84.00 | 0.00 | 86.75 | 0.00 | 85.47 | 0.00 | 82.39 | 0.00 |
| | GPT-4o | 89.65 | 0.00 | 44.80 | 0.00 | 86.10 | 0.00 | 83.73 | 0.00 | 86.02 | 0.00 | 84.13 | 0.00 | 85.82 | 0.00 | 85.64 | 0.00 | 82.46 | 0.00 |
| | Qwen-Max | 89.15 | 0.00 | 43.09 | 0.00 | 87.55 | 0.00 | 80.65 | 0.00 | 86.41 | 0.00 | 83.91 | 0.00 | 87.26 | 0.00 | 84.78 | 0.00 | 82.15 | 0.00 |
| GECScore | Deepseek-V3 | 92.75 | 71.76 | 56.01 | 54.89 | 90.00 | 72.90 | 74.73 | 42.60 | 89.03 | 67.02 | 82.99 | 58.75 | 91.05 | 66.06 | 84.92 | 56.94 | 83.22 | 61.79 |
| | Gemini-2.50-flash | 92.74 | 71.78 | 56.04 | 54.89 | 89.61 | 72.92 | 75.73 | 42.60 | 88.77 | 67.03 | 83.23 | 58.76 | 90.84 | 66.11 | 84.86 | 56.94 | 83.27 | 61.81 |
| | GPT-4o | 92.91 | 71.78 | 56.25 | 54.89 | 88.68 | 72.92 | 77.81 | 42.60 | 88.39 | 67.03 | 83.83 | 58.77 | 90.15 | 66.11 | 85.01 | 56.94 | 83.43 | 61.81 |
| | Qwen-Max | 92.78 | 71.23 | 56.01 | 54.89 | 89.92 | 71.91 | 74.82 | 42.50 | 88.95 | 66.11 | 82.98 | 58.60 | 91.05 | 65.15 | 84.90 | 56.30 | 83.21 | 61.24 |
| Lastde++ | Deepseek-V3 | 36.75 | 33.31 | 34.39 | 33.41 | 35.74 | 33.38 | 44.52 | 33.80 | 38.03 | 33.53 | 44.99 | 33.72 | 34.11 | 33.47 | 40.00 | 33.71 | 39.06 | 33.54 |
| | Gemini-2.50-flash | 36.47 | 33.30 | 34.68 | 33.41 | 35.72 | 33.37 | 44.28 | 33.78 | 38.09 | 33.52 | 44.56 | 33.74 | 34.42 | 33.44 | 40.19 | 33.74 | 39.00 | 33.54 |
| | GPT-4o | 35.42 | 33.30 | 34.97 | 33.41 | 35.28 | 33.37 | 42.98 | 33.78 | 37.44 | 33.52 | 42.40 | 33.74 | 34.70 | 33.44 | 39.93 | 33.74 | 38.18 | 33.54 |
| | Qwen-Max | 37.29 | 33.32 | 34.55 | 33.41 | 35.56 | 33.40 | 44.94 | 33.84 | 37.94 | 33.50 | 45.26 | 33.73 | 33.79 | 33.44 | 39.85 | 33.77 | 39.20 | 33.55 |
| RepreGuard | Deepseek-V3 | 84.45 | 0.83 | 29.64 | 0.00 | 66.63 | 87.16 | 90.42 | 0.00 | 66.60 | 74.73 | 77.94 | 0.00 | 66.63 | 89.76 | 66.53 | 25.17 | 69.97 | 46.21 |
| | Gemini-2.50-flash | 69.06 | 83.63 | 68.21 | 90.76 | 90.10 | 68.95 | 93.78 | 22.40 | 94.96 | 57.67 | 92.65 | 17.55 | 94.33 | 56.63 | 90.92 | 39.23 | 85.16 | 60.50 |
| | GPT-4o | 79.97 | 0.20 | 37.15 | 0.00 | 66.55 | 69.22 | 91.51 | 0.13 | 66.60 | 60.37 | 71.83 | 0.03 | 66.57 | 79.01 | 66.53 | 37.15 | 69.28 | 37.78 |
| | Qwen-Max | 87.90 | 74.90 | 35.26 | 12.21 | 71.80 | 83.71 | 57.21 | 26.98 | 76.21 | 89.76 | 80.85 | 39.29 | 70.88 | 85.16 | 77.50 | 91.35 | 71.97 | 71.13 |
| X-Rob-Classifier | Deepseek-V3 | 98.96 | 98.89 | 98.74 | 98.63 | 99.47 | 99.35 | 99.14 | 99.01 | 98.33 | 98.33 | 97.68 | 97.00 | 99.22 | 99.10 | 98.57 | 98.52 | 98.56 | 98.55 |
| | Gemini-2.50-flash | 99.86 | 99.50 | 98.80 | 98.77 | 99.76 | 99.43 | 99.23 | 99.16 | 99.63 | 99.46 | 98.28 | 98.27 | 99.82 | 99.51 | 99.41 | 99.38 | 99.33 | 99.25 |
| | GPT-4o | 99.90 | 99.50 | 98.57 | 98.55 | 99.84 | 99.73 | 99.57 | 99.24 | 99.84 | 99.47 | 99.36 | 99.27 | 99.96 | 99.49 | 99.72 | 99.51 | 99.50 | 99.33 |
| | Qwen-Max | 99.79 | 99.47 | 98.69 | 98.65 | 99.72 | 99.40 | 99.73 | 99.47 | 99.77 | 99.49 | 99.48 | 99.37 | 99.83 | 99.47 | 99.65 | 99.49 | 99.36 | 99.27 |
| mDeBERTa-Classifier | Deepseek-V3 | 99.48 | 99.37 | 99.08 | 99.07 | 99.08 | 99.02 | 99.27 | 99.17 | 99.42 | 99.36 | 97.08 | 96.17 | 99.16 | 99.15 | 98.54 | 98.42 | 98.52 | 98.40 |
| | Gemini-2.50-flash | 99.95 | 99.47 | 98.97 | 98.95 | 99.87 | 99.47 | 99.57 | 99.40 | 99.81 | 99.48 | 99.42 | 99.34 | 99.90 | 99.48 | 99.68 | 99.42 | 99.62 | 99.43 |
| | GPT-4o | 99.88 | 99.49 | 99.39 | 99.32 | 99.92 | 99.49 | 99.90 | 99.49 | 99.84 | 99.48 | 99.67 | 99.42 | 99.96 | 99.51 | 99.73 | 99.48 | 99.73 | 99.45 |
| | Qwen-Max | 99.67 | 99.46 | 98.60 | 98.56 | 99.82 | 99.46 | 99.87 | 99.47 | 99.71 | 99.48 | 99.32 | 99.27 | 99.75 | 99.45 | 99.77 | 99.47 | 99.44 | 99.31 |
| Biscope | Deepseek-V3 | 91.57 | 83.16 | 92.22 | 76.09 | 95.92 | 92.60 | 88.72 | 72.56 | 96.19 | 91.79 | 88.81 | 65.68 | 96.46 | 91.73 | 93.57 | 84.33 | 91.98 | 80.58 |
| | Gemini-2.50-flash | 91.85 | 81.63 | 92.47 | 81.85 | 95.47 | 90.46 | 89.43 | 70.03 | 95.60 | 89.79 | 87.44 | 61.04 | 96.36 | 92.05 | 93.26 | 82.83 | 91.93 | 78.55 |
| | GPT-4o | 91.70 | 82.68 | 92.91 | 85.02 | 95.77 | 92.74 | 89.53 | 73.04 | 95.79 | 90.28 | 88.43 | 63.87 | 96.52 | 92.26 | 93.37 | 83.13 | 92.20 | 80.16 |
| | Qwen-Max | 87.66 | 60.25 | 89.57 | 80.20 | 93.87 | 89.58 | 85.15 | 62.81 | 94.32 | 88.98 | 85.06 | 64.07 | 94.62 | 90.22 | 91.52 | 78.88 | 89.47 | 77.00 |

Table 23: Performance Comparison of Different Detectors on Cross Generator in BINARY Task. Detector performance is visualized using teal color gradients, with darker intensities indicating superior detection capabilities.

| Detectors↓ | Test↓ Lang→ | English | | Chinese | | Spanish | | Arabic | | French | | Russian | | Portuguese | | German | | Average | |
|---|---|---|---|---|---|---|---|---|---|---|---|---|---|---|---|---|---|---|---|---|---|
| | Metrics | $F_1^B$ | $F_1^F$ | $F_1^B$ | $F_1^F$ | $F_1^B$ | $F_1^F$ | $F_1^B$ | $F_1^F$ | $F_1^B$ | $F_1^F$ | $F_1^B$ | $F_1^F$ | $F_1^B$ | $F_1^F$ | $F_1^B$ | $F_1^F$ | $F_1^B$ | $F_1^F$ |
| Log-Likelihood | Deepseek-V3 | 41.60 | 33.73 | 37.88 | 27.68 | 41.05 | 39.21 | 43.90 | 37.95 | 38.65 | 36.91 | 27.75 | 29.17 | 39.47 | 38.85 | 49.02 | 37.68 | 41.94 | 35.78 |
| | Gemini-2.50-flash | 39.07 | 35.04 | 33.73 | 29.37 | 38.59 | 36.21 | 32.58 | 37.14 | 34.11 | 32.92 | 23.24 | 26.77 | 33.70 | 35.74 | 42.36 | 39.56 | 35.69 | 34.72 |
| | GPT-4o | 35.84 | 32.58 | 30.85 | 26.20 | 46.76 | 40.29 | 47.37 | 37.48 | 45.37 | 38.46 | 32.67 | 31.15 | 48.39 | 40.18 | 43.18 | 35.64 | 42.51 | 35.92 |
| | Qwen-Max | 34.58 | 29.14 | 31.34 | 23.11 | 45.68 | 39.11 | 49.36 | 34.35 | 45.31 | 38.46 | 37.99 | 34.26 | 50.04 | 39.95 | 41.95 | 29.31 | 44.04 | 34.32 |
| Log-Rank | Deepseek-V3 | 42.57 | 34.67 | 40.56 | 30.43 | 40.64 | 38.79 | 43.48 | 38.22 | 38.35 | 36.28 | 29.38 | 30.17 | 40.05 | 38.65 | 47.49 | 36.64 | 41.84 | 35.90 |
| | Gemini-2.50-flash | 40.44 | 35.48 | 35.93 | 31.77 | 38.79 | 35.26 | 31.92 | 37.22 | 33.95 | 31.44 | 24.10 | 27.17 | 34.46 | 35.39 | 42.77 | 38.84 | 36.13 | 34.54 |
| | GPT-4o | 40.15 | 34.37 | 38.19 | 29.73 | 44.58 | 39.56 | 49.74 | 38.30 | 42.61 | 37.38 | 32.69 | 31.08 | 46.50 | 39.65 | 46.85 | 35.33 | 44.05 | 36.11 |
| | Qwen-Max | 37.58 | 30.97 | 35.74 | 25.61 | 41.99 | 38.93 | 50.05 | 35.76 | 42.08 | 37.97 | 37.71 | 34.35 | 46.03 | 39.91 | 41.23 | 27.79 | 43.42 | 34.60 |
| DetectLLM-LRR | Deepseek-V3 | 40.55 | 33.32 | 40.73 | 36.88 | 39.88 | 36.13 | 34.89 | 38.41 | 37.54 | 32.90 | 33.08 | 31.06 | 40.16 | 37.30 | 39.28 | 32.11 | 38.70 | 35.05 |
| | Gemini-2.50-flash | 38.26 | 32.47 | 37.06 | 36.48 | 39.67 | 34.43 | 29.98 | 37.72 | 35.91 | 31.01 | 30.55 | 30.29 | 38.74 | 35.47 | 36.52 | 33.77 | 36.16 | 34.22 |
| | GPT-4o | 42.40 | 33.84 | 46.52 | 36.97 | 40.12 | 36.91 | 46.71 | 38.66 | 39.42 | 34.61 | 37.10 | 31.46 | 43.49 | 38.19 | 38.76 | 29.13 | 42.72 | 35.34 |
| | Qwen-Max | 40.46 | 34.07 | 43.51 | 35.69 | 37.20 | 36.23 | 43.71 | 37.36 | 37.65 | 35.35 | 34.80 | 30.85 | 39.87 | 37.31 | 33.24 | 23.86 | 39.68 | 34.32 |
| Fast-DetectGPT | Deepseek-V3 | 25.38 | 25.38 | 23.59 | 23.59 | 25.41 | 25.41 | 33.76 | 33.76 | 27.26 | 27.26 | 34.87 | 34.87 | 25.86 | 25.86 | 29.94 | 29.94 | 28.97 | 28.97 |
| | Gemini-2.50-flash | 24.49 | 24.49 | 22.87 | 22.87 | 24.45 | 24.45 | 34.48 | 34.48 | 27.10 | 27.10 | 34.91 | 34.91 | 25.00 | 25.00 | 28.62 | 28.62 | 28.52 | 28.52 |
| | GPT-4o | 16.70 | 16.70 | 16.67 | 16.67 | 16.67 | 16.67 | 16.67 | 16.67 | 16.67 | 16.67 | 16.67 | 16.67 | 16.67 | 16.67 | 16.67 | 16.67 | 16.67 | 16.67 |
| | Qwen-Max | 29.99 | 24.96 | 32.44 | 23.22 | 30.72 | 25.02 | 37.78 | 34.24 | 34.23 | 27.33 | 43.31 | 34.98 | 33.09 | 25.49 | 37.75 | 29.41 | 35.97 | 28.82 |
| Binoculars | Deepseek-V3 | 37.93 | 31.14 | 47.61 | 39.12 | 48.61 | 39.11 | 42.73 | 37.33 | 50.51 | 42.02 | 51.05 | 42.08 | 53.54 | 42.32 | 42.82 | 37.45 | 47.26 | 39.00 |
| | Gemini-2.50-flash | 37.93 | 31.14 | 47.61 | 39.12 | 48.61 | 39.11 | 42.73 | 37.33 | 50.51 | 42.02 | 51.05 | 42.08 | 53.54 | 42.32 | 42.82 | 37.45 | 47.26 | 39.00 |
| | GPT-4o | 34.92 | 25.97 | 48.33 | 36.27 | 44.45 | 32.16 | 39.45 | 31.95 | 45.56 | 34.60 | 48.86 | 36.35 | 50.37 | 35.67 | 35.94 | 29.26 | 44.06 | 32.97 |
| | Qwen-Max | 34.92 | 25.97 | 48.33 | 36.27 | 44.45 | 32.16 | 39.45 | 31.95 | 45.56 | 34.60 | 48.86 | 36.35 | 50.37 | 35.67 | 35.94 | 29.26 | 44.06 | 32.97 |
| Revise-Detect | Deepseek-V3 | 61.24 | 49.57 | 35.18 | 32.99 | 58.75 | 50.41 | 56.27 | 42.97 | 58.91 | 48.62 | 56.83 | 46.44 | 58.37 | 51.25 | 57.01 | 46.36 | 56.28 | 46.61 |
| | Gemini-2.50-flash | 62.35 | 49.72 | 35.87 | 33.03 | 57.47 | 50.38 | 57.37 | 43.24 | 58.77 | 48.75 | 56.87 | 46.57 | 57.36 | 51.29 | 57.25 | 46.49 | 56.46 | 46.72 |
| | GPT-4o | 60.44 | 48.85 | 34.82 | 32.88 | 59.22 | 50.31 | 55.05 | 41.78 | 58.71 | 47.87 | 56.65 | 45.70 | 58.82 | 51.01 | 56.63 | 45.25 | 56.02 | 46.02 |
| | Qwen-Max | 60.01 | 48.88 | 34.72 | 32.88 | 59.20 | 50.33 | 54.69 | 41.87 | 58.69 | 47.93 | 56.46 | 45.74 | 58.88 | 51.02 | 56.48 | 45.28 | 55.83 | 46.05 |
| GECScore | Deepseek-V3 | 62.88 | 45.48 | 46.70 | 28.26 | 16.39 | 0.00 | 50.30 | 31.92 | 58.49 | 45.04 | 16.39 | 0.00 | 49.89 | 35.92 | 54.28 | 41.61 | 56.42 | 48.53 |
| | Gemini-2.50-flash | 65.54 | 47.02 | 52.98 | 28.73 | 21.94 | 0.00 | 56.04 | 34.39 | 57.87 | 46.33 | 21.94 | 0.00 | 52.15 | 38.28 | 54.76 | 43.28 | 55.52 | 49.62 |
| | GPT-4o | 64.83 | 44.68 | 49.68 | 28.02 | 19.10 | 0.00 | 52.20 | 30.85 | 59.60 | 44.46 | 19.10 | 0.00 | 51.12 | 35.24 | 55.09 | 40.67 | 56.46 | 47.80 |
| | Qwen-Max | 62.21 | 45.04 | 46.33 | 28.34 | 16.00 | 0.00 | 48.98 | 31.06 | 58.43 | 44.80 | 16.00 | 0.00 | 49.71 | 35.53 | 53.77 | 40.92 | 55.75 | 47.82 |
| Lastde++ | Deepseek-V3 | 17.00 | 17.03 | 17.00 | 17.03 | 17.12 | 17.16 | 17.20 | 17.26 | 16.85 | 16.86 | 16.87 | 16.87 | 17.28 | 17.32 | 17.14 | 17.19 | 17.06 | 17.09 |
| | Gemini-2.50-flash | 16.70 | 16.70 | 16.80 | 16.80 | 16.66 | 16.66 | 16.67 | 16.67 | 16.66 | 16.66 | 16.66 | 16.66 | 16.66 | 16.66 | 16.67 | 16.67 | 16.69 | 16.69 |
| | GPT-4o | 16.67 | 16.67 | 16.71 | 16.71 | 16.67 | 16.67 | 16.67 | 16.67 | 16.67 | 16.67 | 16.67 | 16.67 | 16.67 | 16.67 | 16.67 | 16.67 | 16.67 | 16.67 |
| | Qwen-Max | 18.27 | 17.00 | 18.50 | 17.00 | 19.63 | 17.11 | 19.90 | 17.17 | 18.80 | 16.84 | 18.55 | 16.86 | 19.61 | 17.24 | 19.41 | 17.14 | 19.11 | 17.05 |
| RepreGuard | Deepseek-V3 | 30.44 | 16.67 | 26.84 | 16.67 | 16.69 | 16.93 | 16.57 | 16.67 | 16.67 | 16.91 | 16.65 | 16.68 | 16.66 | 17.13 | 16.68 | 17.04 | 24.77 | 16.86 |
| | Gemini-2.50-flash | 34.03 | 37.23 | 24.28 | 22.02 | 45.70 | 17.46 | 35.55 | 16.88 | 51.55 | 17.04 | 33.05 | 16.70 | 49.85 | 16.80 | 49.73 | 17.23 | 44.32 | 22.11 |
| | GPT-4o | 22.12 | 16.67 | 40.85 | 16.67 | 16.67 | 17.72 | 16.68 | 16.67 | 16.67 | 17.32 | 16.69 | 16.68 | 16.67 | 18.35 | 16.67 | 17.22 | 22.50 | 17.18 |
| | Qwen-Max | 16.84 | 16.70 | 16.67 | 16.67 | 34.98 | 31.71 | 16.89 | 16.71 | 32.54 | 27.77 | 16.86 | 16.68 | 34.08 | 35.13 | 30.83 | 27.99 | 28.86 | 25.75 |
| X-Rob-Classifier | Deepseek-V3 | 90.95 | 90.68 | 89.71 | 80.20 | 92.44 | 85.05 | 90.40 | 84.21 | 90.28 | 87.10 | 84.80 | 72.31 | 91.61 | 88.68 | 88.00 | 82.21 | 89.73 | 80.65 |
| | Gemini-2.50-flash | 95.57 | 93.68 | 87.91 | 82.32 | 93.80 | 88.79 | 92.03 | 85.67 | 93.78 | 90.77 | 88.73 | 79.84 | 92.82 | 88.80 | 90.88 | 85.65 | 91.68 | 85.77 |
| | GPT-4o | 95.85 | 95.40 | 87.29 | 83.40 | 89.65 | 87.24 | 92.96 | 88.96 | 91.97 | 90.29 | 83.85 | 76.18 | 90.72 | 90.50 | 87.38 | 84.75 | 87.94 | 81.52 |
| | Qwen-Max | 94.08 | 93.65 | 84.49 | 77.26 | 89.59 | 86.63 | 88.34 | 77.39 | 89.33 | 88.53 | 84.48 | 74.97 | 90.71 | 87.29 | 86.52 | 81.87 | 86.06 | 72.95 |
| mDeBERTa-Classifier | Deepseek-V3 | 94.55 | 94.68 | 90.73 | 81.24 | 94.30 | 88.75 | 92.30 | 88.59 | 92.92 | 90.07 | 88.24 | 74.87 | 92.91 | 90.59 | 88.98 | 83.39 | 91.34 | 85.90 |
| | Gemini-2.50-flash | 95.91 | 95.29 | 87.60 | 84.58 | 94.62 | 90.97 | 89.50 | 88.40 | 94.11 | 92.07 | 88.81 | 81.84 | 93.65 | 91.46 | 89.54 | 86.53 | 91.27 | 88.25 |
| | GPT-4o | 96.97 | 96.31 | 92.14 | 85.90 | 93.28 | 89.39 | 94.90 | 92.83 | 94.77 | 93.26 | 89.60 | 82.00 | 94.87 | 91.93 | 92.25 | 87.88 | 93.82 | 90.23 |
| | Qwen-Max | 95.59 | 95.16 | 87.33 | 77.87 | 92.43 | 88.01 | 90.55 | 85.78 | 93.34 | 92.59 | 88.49 | 79.63 | 94.34 | 92.98 | 90.83 | 87.61 | 91.83 | 87.88 |
| Biscope | Deepseek-V3 | 71.73 | 53.69 | 62.43 | 26.34 | 73.36 | 52.37 | 60.86 | 25.52 | 72.28 | 52.37 | 59.84 | 35.77 | 73.66 | 55.53 | 68.18 | 46.30 | 67.99 | 45.62 |
| | Gemini-2.50-flash | 73.88 | 57.33 | 62.40 | 42.63 | 72.82 | 51.63 | 63.23 | 35.66 | 70.47 | 50.09 | 57.79 | 31.62 | 72.24 | 52.46 | 66.20 | 46.73 | 67.59 | 45.25 |
| | GPT-4o | 73.96 | 58.54 | 63.84 | 44.86 | 71.99 | 51.69 | 63.12 | 36.66 | 71.99 | 50.79 | 60.90 | 35.87 | 73.72 | 52.03 | 68.41 | 45.95 | 68.75 | 46.15 |
| | Qwen-Max | 61.82 | 45.19 | 59.32 | 40.93 | 67.66 | 50.45 | 56.11 | 30.90 | 69.73 | 50.80 | 58.98 | 35.45 | 71.42 | 52.63 | 61.57 | 43.93 | 63.79 | 44.75 |

Table 24: Performance Comparison of Different Detectors on Cross Generator in TERNARY Task. Detector performance is visualized using teal color gradients, with darker intensities indicating superior detection capabilities.

| Detectors↓ | Test↓ Lang→ | English | | Chinese | | Spanish | | Arabic | | French | | Russian | | Portuguese | | German | | Average | |
|---|---|---|---|---|---|---|---|---|---|---|---|---|---|---|---|---|---|---|---|
| | Metrics | $F_1^B$ | $F_1^F$ | $F_1^B$ | $F_1^F$ | $F_1^B$ | $F_1^F$ | $F_1^B$ | $F_1^F$ | $F_1^B$ | $F_1^F$ | $F_1^B$ | $F_1^F$ | $F_1^B$ | $F_1^F$ | $F_1^B$ | $F_1^F$ | $F_1^B$ | $F_1^F$ |
| Log-Likelihood | Encoder-Paraph | 31.38 | 33.33 | 89.22 | 66.34 | 33.92 | 33.33 | 28.54 | 33.36 | 28.98 | 33.31 | 25.28 | 31.97 | 29.68 | 33.35 | 32.07 | 33.32 | 39.98 | 38.36 |
| | Seq2seq-Paraph | 30.82 | 33.33 | 29.01 | 33.15 | 25.89 | 33.33 | 26.36 | 33.31 | 23.68 | 33.31 | 13.09 | 31.88 | 24.95 | 33.31 | 30.80 | 33.32 | 25.93 | 33.12 |
| | Decoder-Paraph | 50.62 | 34.04 | 46.73 | 34.06 | 63.58 | 37.24 | 60.36 | 36.23 | 37.38 | 33.37 | 32.59 | 32.26 | 41.18 | 33.43 | 63.85 | 33.86 | 50.07 | 34.32 |
| | Back-Translation | 31.65 | 33.36 | 31.06 | 33.23 | 55.10 | 39.39 | 30.72 | 33.28 | 47.44 | 34.61 | 43.60 | 34.47 | 31.51 | 33.33 | 44.18 | 33.32 | 41.55 | 34.42 |
| Log-Rank | Encoder-Paraph | 31.27 | 33.32 | 89.87 | 72.97 | 34.98 | 33.32 | 29.69 | 33.34 | 30.16 | 33.31 | 26.33 | 31.96 | 30.68 | 33.35 | 32.40 | 33.32 | 40.66 | 39.68 |
| | Seq2seq-Paraph | 29.87 | 33.32 | 28.96 | 33.14 | 26.32 | 33.32 | 26.29 | 33.29 | 23.86 | 33.31 | 14.94 | 31.87 | 25.61 | 33.31 | 31.33 | 33.32 | 26.18 | 33.11 |
| | Decoder-Paraph | 50.62 | 34.04 | 46.73 | 34.06 | 63.58 | 37.24 | 60.36 | 36.23 | 37.38 | 33.37 | 32.59 | 32.26 | 41.18 | 33.43 | 63.85 | 33.86 | 50.07 | 34.32 |
| | Back-Translation | 31.65 | 33.36 | 31.06 | 33.23 | 55.10 | 39.39 | 30.72 | 33.28 | 47.44 | 34.61 | 43.60 | 34.47 | 31.51 | 33.33 | 44.18 | 33.32 | 41.55 | 34.42 |
| DetectLLM-LRR | Encoder-Paraph | 33.46 | 33.30 | 88.61 | 80.87 | 39.06 | 33.40 | 39.42 | 33.25 | 34.62 | 33.43 | 33.63 | 33.17 | 36.07 | 33.38 | 33.56 | 33.33 | 44.18 | 41.52 |
| | Seq2seq-Paraph | 22.49 | 32.76 | 28.09 | 33.04 | 26.37 | 33.22 | 26.14 | 32.96 | 23.63 | 33.16 | 23.52 | 32.89 | 27.17 | 33.31 | 32.37 | 33.33 | 26.35 | 33.08 |
| | Decoder-Paraph | 45.98 | 35.35 | 44.34 | 34.15 | 49.99 | 34.79 | 59.08 | 35.36 | 40.97 | 33.71 | 36.82 | 33.12 | 38.29 | 33.57 | 62.54 | 34.85 | 47.57 | 34.37 |
| | Back-Translation | 32.68 | 33.37 | 37.35 | 33.64 | 56.66 | 37.44 | 38.49 | 33.14 | 48.34 | 36.85 | 46.50 | 35.23 | 41.56 | 33.51 | 41.08 | 33.39 | 43.70 | 34.59 |
| Fast-DetectGPT | Encoder-Paraph | 33.67 | 33.27 | 38.36 | 33.22 | 33.48 | 33.29 | 41.03 | 34.66 | 35.44 | 33.59 | 34.64 | 33.47 | 32.64 | 33.34 | 36.85 | 33.41 | 36.64 | 33.55 |
| | Seq2seq-Paraph | 29.66 | 33.04 | 29.37 | 33.21 | 26.89 | 33.17 | 18.37 | 32.21 | 25.03 | 33.14 | 24.38 | 33.00 | 26.80 | 33.19 | 28.56 | 33.26 | 26.29 | 33.03 |
| | Decoder-Paraph | 55.06 | 35.74 | 63.40 | 43.04 | 69.35 | 46.86 | 64.86 | 65.52 | 52.16 | 37.29 | 59.59 | 38.46 | 67.36 | 40.99 | 72.32 | 42.59 | 63.73 | 44.63 |
| | Back-Translation | 36.47 | 33.51 | 48.01 | 34.58 | 40.51 | 34.60 | 52.29 | 42.01 | 42.39 | 34.50 | 47.89 | 37.39 | 35.40 | 33.46 | 46.22 | 34.81 | 44.91 | 35.72 |
| Binoculars | Encoder-Paraph | 36.52 | 34.10 | 44.82 | 41.20 | 57.12 | 41.30 | 58.91 | 40.89 | 52.64 | 37.41 | 55.07 | 40.11 | 59.75 | 40.67 | 42.43 | 34.24 | 51.58 | 38.82 |
| | Seq2seq-Paraph | 30.75 | 33.02 | 37.19 | 32.97 | 29.13 | 33.18 | 32.70 | 33.38 | 30.66 | 33.27 | 32.14 | 33.16 | 29.67 | 33.29 | 32.32 | 33.31 | 32.51 | 33.20 |
| | Decoder-Paraph | 64.99 | 44.25 | 60.57 | 52.09 | 71.15 | 52.02 | 78.22 | 59.45 | 66.08 | 45.28 | 73.44 | 51.68 | 61.59 | 42.08 | 76.64 | 51.37 | 69.24 | 50.00 |
| | Back-Translation | 61.10 | 40.24 | 64.69 | 48.43 | 48.70 | 38.74 | 59.25 | 40.10 | 63.32 | 43.76 | 66.98 | 49.21 | 57.79 | 39.19 | 57.43 | 38.58 | 60.61 | 42.45 |
| Revise-Detect | Encoder-Paraph | 0.00 | 0.00 | 3.67 | 0.00 | 0.80 | 0.00 | 0.00 | 0.00 | 0.03 | 0.00 | 0.03 | 0.00 | 0.08 | 0.00 | 0.00 | 0.00 | 0.54 | 0.00 |
| | Seq2seq-Paraph | 9.43 | 0.00 | 0.35 | 0.00 | 1.17 | 0.00 | 4.39 | 0.00 | 0.75 | 0.00 | 0.87 | 0.00 | 1.79 | 0.00 | 0.84 | 0.00 | 2.40 | 0.00 |
| | Decoder-Paraph | 57.88 | 0.00 | 47.06 | 0.00 | 68.07 | 0.00 | 65.47 | 0.00 | 69.82 | 0.00 | 75.17 | 0.00 | 58.13 | 0.00 | 58.42 | 0.00 | 63.24 | 0.00 |
| | Back-Translation | 4.73 | 0.00 | 0.07 | 0.00 | 12.75 | 0.00 | 1.26 | 0.00 | 7.54 | 0.00 | 7.97 | 0.00 | 1.56 | 0.00 | 16.86 | 0.00 | 6.92 | 0.00 |
| GECScore | Encoder-Paraph | 27.97 | 29.51 | 30.76 | 29.73 | 25.14 | 28.65 | 29.73 | 29.99 | 26.11 | 29.66 | 27.49 | 29.91 | 27.18 | 30.10 | 25.27 | 29.29 | 27.48 | 29.61 |
| | Seq2seq-Paraph | 35.02 | 36.72 | 33.86 | 33.81 | 40.00 | 44.51 | 32.12 | 32.39 | 32.98 | 37.00 | 34.76 | 37.51 | 32.80 | 35.99 | 32.36 | 36.96 | 34.00 | 36.51 |
| | Decoder-Paraph | 66.26 | 48.80 | 52.45 | 43.10 | 58.08 | 34.83 | 54.60 | 33.04 | 74.11 | 44.34 | 67.80 | 35.62 | 55.78 | 32.93 | 59.04 | 34.60 | 61.54 | 38.64 |
| | Back-Translation | 29.24 | 29.61 | 29.66 | 29.70 | 28.78 | 28.68 | 29.80 | 30.04 | 28.98 | 29.74 | 28.60 | 29.93 | 27.27 | 30.17 | 29.07 | 29.60 | 29.03 | 29.69 |
| Lastde++ | Encoder-Paraph | 47.07 | 34.29 | 27.93 | 33.26 | 48.64 | 34.70 | 46.43 | 33.99 | 52.01 | 34.70 | 47.61 | 34.19 | 48.76 | 34.72 | 49.00 | 34.73 | 46.81 | 34.33 |
| | Seq2seq-Paraph | 65.90 | 35.52 | 46.46 | 33.69 | 72.17 | 41.11 | 58.23 | 35.44 | 69.99 | 37.82 | 69.12 | 37.93 | 65.78 | 37.76 | 64.13 | 37.70 | 64.58 | 37.16 |
| | Decoder-Paraph | 44.15 | 33.66 | 50.89 | 33.86 | 49.52 | 34.07 | 50.28 | 34.35 | 48.35 | 34.15 | 59.19 | 34.92 | 55.36 | 34.77 | 47.85 | 33.97 | 51.07 | 34.22 |
| | Back-Translation | 54.40 | 35.42 | 60.53 | 35.39 | 43.45 | 34.07 | 51.25 | 34.91 | 47.49 | 34.30 | 45.07 | 33.80 | 52.44 | 35.10 | 43.62 | 34.02 | 50.00 | 34.63 |
| RepreGuard | Encoder-Paraph | 66.59 | 3.70 | 40.89 | 0.03 | 66.61 | 75.36 | 74.02 | 0.33 | 66.59 | 68.56 | 67.39 | 0.07 | 66.61 | 72.99 | 66.55 | 50.18 | 64.72 | 48.09 |
| | Seq2seq-Paraph | 63.37 | 3.64 | 53.71 | 0.07 | 66.63 | 63.68 | 64.84 | 0.33 | 66.60 | 60.96 | 66.34 | 0.07 | 66.61 | 65.81 | 66.55 | 35.47 | 64.89 | 42.44 |
| | Decoder-Paraph | 67.78 | 3.24 | 56.41 | 0.03 | 67.25 | 83.28 | 72.04 | 0.33 | 69.35 | 76.46 | 66.35 | 0.04 | 66.76 | 86.27 | 66.61 | 42.70 | 67.00 | 46.63 |
| | Back-Translation | 54.40 | 35.42 | 60.53 | 35.39 | 43.45 | 34.07 | 51.25 | 34.91 | 47.49 | 34.30 | 45.07 | 33.80 | 52.44 | 35.10 | 43.62 | 34.02 | 50.00 | 34.63 |
| X-Rob-Classifier | Encoder-Paraph | 93.32 | 81.06 | 33.56 | 33.11 | 92.71 | 75.67 | 94.52 | 85.94 | 90.97 | 70.03 | 88.60 | 65.89 | 91.59 | 72.12 | 87.24 | 56.33 | 83.48 | 65.47 |
| | Seq2seq-Paraph | 33.33 | 38.46 | 54.97 | 41.65 | 59.70 | 33.98 | 56.83 | 33.96 | 66.75 | 34.36 | 40.69 | 34.18 | 64.38 | 36.45 | 46.89 | 34.43 | 53.08 | 35.18 |
| | Decoder-Paraph | 83.90 | 70.79 | 79.32 | 64.46 | 75.94 | 59.72 | 80.79 | 63.60 | 82.84 | 69.00 | 50.21 | 45.40 | 74.58 | 56.42 | 74.04 | 55.65 | 75.12 | 58.74 |
| | Back-Translation | 99.35 | 99.29 | 91.00 | 84.15 | 97.67 | 97.53 | 98.54 | 98.51 | 99.03 | 99.03 | 99.59 | 99.42 | 96.87 | 95.54 | 99.67 | 99.32 | 97.59 | 97.16 |
| mDeBERTa-Classifier | Encoder-Paraph | 86.44 | 74.02 | 33.61 | 44.36 | 88.36 | 70.18 | 93.47 | 80.23 | 83.09 | 61.91 | 81.98 | 61.96 | 81.92 | 58.91 | 76.01 | 58.80 | 81.35 | 63.49 |
| | Seq2seq-Paraph | 35.25 | 35.54 | 34.49 | 39.16 | 43.14 | 35.43 | 50.12 | 34.13 | 44.53 | 34.92 | 41.30 | 34.20 | 34.27 | 36.83 | 51.38 | 35.11 | 40.82 | 35.24 |
| | Decoder-Paraph | 81.10 | 72.64 | 77.12 | 65.81 | 73.19 | 54.46 | 80.63 | 63.61 | 78.93 | 66.23 | 48.73 | 49.62 | 71.84 | 53.85 | 66.60 | 56.14 | 73.91 | 59.55 |
| | Back-Translation | 99.02 | 99.00 | 93.72 | 89.04 | 97.07 | 96.87 | 98.76 | 98.68 | 98.93 | 98.90 | 99.49 | 99.27 | 96.02 | 94.82 | 99.72 | 99.35 | 97.65 | 97.24 |
| Biscope | Encoder-Paraph | 41.10 | 33.18 | 58.45 | 33.13 | 46.04 | 34.08 | 44.70 | 34.35 | 35.86 | 33.27 | 46.30 | 33.79 | 45.05 | 33.79 | 47.44 | 33.37 | 46.09 | 33.35 |
| | Seq2seq-Paraph | 47.82 | 33.13 | 62.52 | 33.08 | 50.61 | 33.10 | 61.36 | 33.11 | 50.44 | 33.12 | 33.48 | 33.12 | 55.49 | 33.23 | 55.02 | 33.12 | 51.00 | 33.13 |
| | Decoder-Paraph | 65.05 | 38.62 | 71.59 | 41.76 | 73.23 | 50.86 | 72.42 | 45.00 | 76.47 | 48.92 | 55.38 | 38.29 | 62.43 | 40.15 | 71.85 | 40.84 | 68.06 | 41.42 |
| | Back-Translation | 66.39 | 34.42 | 57.10 | 35.52 | 67.75 | 41.62 | 63.96 | 41.54 | 68.79 | 47.26 | 74.59 | 45.40 | 56.63 | 38.29 | 78.27 | 49.90 | 65.88 | 42.57 |

Table 25: Performance Comparison of Different Detectors on Cross Paraphrase in BINARY Task. Detector performance is visualized using teal color gradients, with darker intensities indicating superior detection capabilities.

| Detectors↓ | Test↓ Lang→ | English | | Chinese | | Spanish | | Arabic | | French | | Russian | | Portuguese | | German | | Average | |
|---|---|---|---|---|---|---|---|---|---|---|---|---|---|---|---|---|---|---|---|
| | Metrics | $F_1^B$ | $F_1^F$ | $F_1^B$ | $F_1^F$ | $F_1^B$ | $F_1^F$ | $F_1^B$ | $F_1^F$ | $F_1^B$ | $F_1^F$ | $F_1^B$ | $F_1^F$ | $F_1^B$ | $F_1^F$ | $F_1^B$ | $F_1^F$ | $F_1^B$ | $F_1^F$ |
| Log-Likelihood | Encoder-Paraph | 19.14 | 15.98 | 53.74 | 48.86 | 23.94 | 17.70 | 25.93 | 16.37 | 23.01 | 15.72 | 16.58 | 14.11 | 24.31 | 16.06 | 26.09 | 18.95 | 28.34 | 21.78 |
| | Seq2seq-Paraph | 18.96 | 15.80 | 20.02 | 15.24 | 20.85 | 14.28 | 25.40 | 15.66 | 21.26 | 13.57 | 11.99 | 9.09 | 22.80 | 14.23 | 25.64 | 18.45 | 21.42 | 14.70 |
| | Decoder-Paraph | 30.27 | 26.90 | 29.82 | 25.06 | 41.94 | 35.68 | 41.15 | 31.34 | 26.73 | 19.64 | 19.67 | 17.23 | 30.34 | 22.23 | 40.35 | 32.99 | 33.08 | 26.52 |
| | Back-Translation | 19.49 | 16.33 | 20.71 | 15.94 | 34.92 | 28.73 | 26.33 | 16.81 | 32.05 | 25.00 | 25.65 | 23.38 | 24.34 | 16.08 | 31.92 | 24.79 | 28.60 | 22.08 |
| Log-Rank | Encoder-Paraph | 23.58 | 15.92 | 60.34 | 48.96 | 20.73 | 18.23 | 24.82 | 16.80 | 19.72 | 15.99 | 15.84 | 14.54 | 19.94 | 16.34 | 30.62 | 18.80 | 28.93 | 21.99 |
| | Seq2seq-Paraph | 23.78 | 15.36 | 26.53 | 15.10 | 20.35 | 14.13 | 25.07 | 15.61 | 20.40 | 13.20 | 13.28 | 9.63 | 20.78 | 14.04 | 31.29 | 18.34 | 23.36 | 14.55 |
| | Decoder-Paraph | 33.93 | 26.66 | 34.71 | 24.29 | 36.07 | 34.57 | 40.17 | 31.44 | 22.40 | 19.56 | 18.54 | 17.39 | 24.46 | 21.65 | 45.55 | 33.18 | 33.00 | 26.17 |
| | Back-Translation | 23.88 | 16.21 | 26.77 | 16.12 | 31.47 | 29.61 | 25.13 | 17.22 | 28.12 | 25.38 | 24.19 | 23.50 | 20.18 | 16.82 | 36.16 | 24.32 | 29.02 | 22.21 |
| DetectLLM-LRR | Encoder-Paraph | 24.19 | 17.60 | 51.59 | 47.52 | 23.04 | 20.60 | 20.99 | 21.50 | 21.51 | 18.33 | 19.96 | 17.95 | 21.20 | 18.91 | 25.51 | 18.37 | 27.31 | 23.69 |
| | Seq2seq-Paraph | 22.73 | 11.54 | 22.21 | 14.71 | 21.37 | 13.46 | 18.10 | 15.22 | 21.05 | 12.25 | 18.09 | 12.64 | 21.47 | 13.93 | 26.98 | 17.69 | 21.58 | 13.95 |
| | Decoder-Paraph | 31.05 | 24.66 | 27.64 | 23.53 | 29.22 | 26.85 | 30.50 | 31.40 | 24.87 | 21.85 | 21.52 | 19.53 | 22.28 | 20.19 | 40.77 | 33.47 | 28.73 | 25.26 |
| | Back-Translation | 23.67 | 17.20 | 24.21 | 19.74 | 33.00 | 30.58 | 20.42 | 20.88 | 29.12 | 26.01 | 26.75 | 24.79 | 24.29 | 22.14 | 29.44 | 22.27 | 26.90 | 23.34 |
| Fast-DetectGPT | Encoder-Paraph | 18.87 | 16.81 | 21.97 | 17.63 | 20.14 | 17.09 | 27.25 | 20.72 | 21.13 | 17.87 | 25.15 | 17.95 | 21.28 | 16.91 | 22.42 | 18.02 | 22.71 | 18.14 |
| | Seq2seq-Paraph | 17.92 | 16.10 | 20.62 | 16.40 | 18.78 | 15.93 | 21.34 | 15.25 | 18.63 | 15.54 | 22.62 | 15.80 | 19.98 | 15.98 | 20.74 | 16.65 | 20.26 | 16.00 |
| | Decoder-Paraph | 25.85 | 23.78 | 34.74 | 30.36 | 38.26 | 35.07 | 44.98 | 38.59 | 29.44 | 26.25 | 35.59 | 28.71 | 36.89 | 32.69 | 38.68 | 34.23 | 36.21 | 31.76 |
| | Back-Translation | 19.65 | 17.79 | 25.82 | 21.53 | 22.54 | 19.64 | 33.91 | 27.76 | 23.71 | 20.58 | 30.91 | 24.13 | 21.79 | 17.78 | 25.67 | 21.54 | 26.25 | 21.96 |
| Binoculars | Encoder-Paraph | 26.36 | 17.32 | 33.48 | 21.84 | 33.63 | 21.69 | 28.87 | 21.69 | 29.68 | 19.22 | 33.05 | 21.09 | 35.43 | 21.42 | 24.25 | 18.25 | 31.11 | 20.37 |
| | Seq2seq-Paraph | 25.91 | 16.66 | 28.79 | 16.85 | 29.31 | 16.72 | 24.33 | 17.14 | 27.64 | 16.72 | 29.41 | 16.72 | 31.85 | 16.80 | 23.77 | 17.75 | 28.14 | 16.92 |
| | Decoder-Paraph | 32.55 | 23.59 | 39.99 | 28.27 | 40.27 | 28.18 | 40.82 | 33.24 | 34.52 | 24.01 | 40.52 | 28.25 | 36.15 | 22.13 | 34.76 | 28.18 | 38.00 | 27.13 |
| | Back-Translation | 30.10 | 21.23 | 37.77 | 26.02 | 32.16 | 20.09 | 28.37 | 21.18 | 33.74 | 23.24 | 38.67 | 26.62 | 34.47 | 20.47 | 26.94 | 20.81 | 33.29 | 22.57 |
| Revise-Detect | Encoder-Paraph | 28.65 | 15.74 | 18.09 | 16.16 | 25.75 | 14.25 | 30.61 | 15.16 | 28.48 | 15.19 | 29.10 | 15.62 | 26.59 | 15.95 | 28.86 | 15.33 | 27.50 | 15.44 |
| | Seq2seq-Paraph | 29.20 | 16.57 | 18.10 | 16.27 | 26.05 | 14.44 | 30.91 | 15.81 | 28.67 | 15.42 | 29.37 | 15.90 | 27.03 | 16.47 | 29.33 | 15.80 | 27.86 | 15.84 |
| | Decoder-Paraph | 43.39 | 30.56 | 30.81 | 28.46 | 41.99 | 30.25 | 43.44 | 28.78 | 46.06 | 32.61 | 49.13 | 35.18 | 37.49 | 26.65 | 40.06 | 26.72 | 42.06 | 30.01 |
| | Back-Translation | 28.59 | 15.95 | 17.84 | 16.02 | 26.02 | 14.94 | 30.31 | 15.07 | 28.40 | 15.60 | 29.00 | 15.98 | 26.52 | 16.01 | 28.98 | 16.14 | 27.47 | 15.75 |
| GECScore | Encoder-Paraph | 24.29 | 5.91 | 26.04 | 6.25 | 16.60 | 0.00 | 25.68 | 6.27 | 22.42 | 5.74 | 16.63 | 0.00 | 19.38 | 3.19 | 22.06 | 5.67 | 26.98 | 11.99 |
| | Seq2seq-Paraph | 25.12 | 6.29 | 26.56 | 6.45 | 16.64 | 0.01 | 26.02 | 6.43 | 23.11 | 6.14 | 16.63 | 0.00 | 19.71 | 3.40 | 22.75 | 6.07 | 30.07 | 14.54 |
| | Decoder-Paraph | 31.51 | 12.68 | 29.76 | 9.70 | 20.60 | 4.64 | 28.74 | 8.94 | 31.03 | 13.95 | 21.45 | 5.54 | 22.67 | 6.71 | 26.86 | 10.30 | 40.21 | 24.97 |
| | Back-Translation | 24.36 | 5.97 | 25.98 | 6.20 | 16.52 | 0.19 | 25.70 | 6.28 | 22.56 | 5.90 | 16.56 | 0.04 | 19.38 | 3.20 | 22.27 | 5.90 | 26.85 | 12.24 |
| Lastde++ | Encoder-Paraph | 16.67 | 16.67 | 16.67 | 16.67 | 16.67 | 16.67 | 16.67 | 16.67 | 16.67 | 16.67 | 16.67 | 16.67 | 16.67 | 16.67 | 16.67 | 16.67 | 16.67 | 16.67 |
| | Seq2seq-Paraph | 16.66 | 16.66 | 16.64 | 16.64 | 16.66 | 16.66 | 16.66 | 16.66 | 16.67 | 16.67 | 16.67 | 16.67 | 16.66 | 16.66 | 16.67 | 16.67 | 16.66 | 16.66 |
| | Decoder-Paraph | 16.67 | 16.67 | 16.67 | 16.67 | 16.67 | 16.67 | 16.67 | 16.67 | 16.67 | 16.67 | 16.67 | 16.67 | 16.67 | 16.67 | 16.67 | 16.67 | 16.67 | 16.67 |
| | Back-Translation | 16.67 | 16.67 | 16.66 | 16.66 | 16.67 | 16.67 | 16.67 | 16.67 | 16.67 | 16.67 | 16.67 | 16.67 | 16.67 | 16.67 | 16.67 | 16.67 | 16.67 | 16.67 |
| RepreGuard | Encoder-Paraph | 16.67 | 16.67 | 16.67 | 16.67 | 24.97 | 24.45 | 16.67 | 16.67 | 22.44 | 18.82 | 16.67 | 16.67 | 22.36 | 27.50 | 24.58 | 16.95 | 25.15 | 20.31 |
| | Seq2seq-Paraph | 16.62 | 16.64 | 16.65 | 16.66 | 15.72 | 8.98 | 16.67 | 16.67 | 14.27 | 10.55 | 16.63 | 16.66 | 17.21 | 12.58 | 20.74 | 16.90 | 22.89 | 16.45 |
| | Decoder-Paraph | 31.87 | 16.67 | 19.72 | 16.67 | 18.58 | 18.28 | 22.78 | 16.67 | 35.16 | 17.10 | 16.83 | 16.43 | 17.16 | 19.66 | 16.96 | 16.87 | 25.89 | 17.31 |
| | Back-Translation | 39.98 | 16.67 | 24.52 | 16.67 | 16.69 | 18.34 | 22.33 | 16.67 | 16.70 | 17.22 | 16.80 | 16.67 | 16.67 | 19.39 | 16.79 | 16.89 | 26.81 | 17.34 |
| X-Rob-Classifier | Encoder-Paraph | 96.41 | 95.47 | 92.27 | 60.55 | 94.53 | 92.57 | 94.36 | 93.27 | 94.84 | 94.60 | 90.94 | 85.12 | 95.18 | 94.01 | 92.98 | 91.45 | 93.94 | 89.20 |
| | Seq2seq-Paraph | 68.45 | 62.08 | 67.93 | 51.83 | 78.04 | 66.49 | 84.29 | 61.97 | 80.03 | 67.72 | 76.57 | 56.98 | 74.44 | 62.20 | 85.89 | 76.03 | 77.36 | 63.55 |
| | Decoder-Paraph | 58.75 | 58.44 | 57.20 | 52.92 | 59.08 | 55.68 | 68.14 | 63.83 | 54.36 | 54.63 | 50.57 | 47.74 | 61.41 | 59.08 | 67.03 | 64.47 | 59.77 | 57.44 |
| | Back-Translation | 94.43 | 93.65 | 92.13 | 86.37 | 92.60 | 90.74 | 94.09 | 91.42 | 94.67 | 94.56 | 92.39 | 91.47 | 90.23 | 86.51 | 94.06 | 93.04 | 92.99 | 91.09 |
| mDeBERTa-Classifier | Encoder-Paraph | 98.40 | 97.95 | 95.71 | 85.40 | 96.99 | 93.73 | 97.32 | 95.75 | 97.79 | 96.96 | 94.49 | 83.95 | 97.87 | 96.01 | 96.59 | 93.04 | 96.89 | 93.47 |
| | Seq2seq-Paraph | 77.47 | 66.87 | 80.85 | 54.98 | 94.09 | 78.32 | 96.81 | 93.44 | 96.24 | 86.90 | 91.53 | 70.74 | 89.36 | 72.08 | 95.95 | 86.17 | 90.57 | 75.25 |
| | Decoder-Paraph | 59.24 | 58.99 | 56.18 | 52.14 | 57.14 | 53.68 | 67.69 | 65.81 | 55.34 | 54.31 | 52.82 | 47.28 | 60.65 | 58.64 | 68.21 | 64.67 | 59.87 | 57.08 |
| | Back-Translation | 93.99 | 92.57 | 91.15 | 85.33 | 94.33 | 92.76 | 95.56 | 94.36 | 95.79 | 95.18 | 95.07 | 93.58 | 89.29 | 85.96 | 96.41 | 95.15 | 93.67 | 90.64 |
| Biscope | Encoder-Paraph | 40.27 | 32.19 | 64.70 | 21.81 | 38.34 | 21.04 | 35.37 | 18.55 | 36.02 | 18.72 | 29.88 | 17.58 | 39.08 | 21.15 | 45.59 | 23.90 | 43.33 | 22.26 |
| | Seq2seq-Paraph | 65.02 | 29.78 | 66.32 | 28.07 | 70.86 | 46.43 | 68.12 | 24.30 | 71.02 | 43.56 | 33.52 | 17.71 | 72.04 | 50.52 | 69.31 | 46.96 | 65.42 | 35.18 |
| | Decoder-Paraph | 45.24 | 27.41 | 44.67 | 19.51 | 49.15 | 21.56 | 43.43 | 20.64 | 40.41 | 21.99 | 34.59 | 19.81 | 40.65 | 22.60 | 45.59 | 26.36 | 43.08 | 22.28 |
| | Back-Translation | 44.12 | 30.44 | 33.17 | 18.48 | 43.85 | 22.23 | 36.72 | 19.24 | 48.07 | 25.97 | 44.42 | 23.23 | 38.36 | 22.20 | 51.94 | 31.39 | 43.17 | 25.03 |

Table 26: Performance Comparison of Different Detectors on Cross Paraphrase in TERNARY Task. Detector performance is visualized using teal color gradients, with darker intensities indicating superior detection capabilities.

| Detectors↓ | Test↓ Lang→ | English | | Chinese | | Spanish | | Arabic | | French | | Russian | | Portuguese | | German | | Average | |
|---|---|---|---|---|---|---|---|---|---|---|---|---|---|---|---|---|---|---|---|
| | Metrics | $F_1^B$ | $F_1^F$ | $F_1^B$ | $F_1^F$ | $F_1^B$ | $F_1^F$ | $F_1^B$ | $F_1^F$ | $F_1^B$ | $F_1^F$ | $F_1^B$ | $F_1^F$ | $F_1^B$ | $F_1^F$ | $F_1^B$ | $F_1^F$ | $F_1^B$ | $F_1^F$ |
| Log-Likelihood | Character-Insertion | 30.50 | 33.33 | 28.29 | 33.13 | 25.19 | 33.32 | 25.82 | 33.29 | 22.68 | 33.31 | 14.84 | 31.90 | 23.97 | 33.31 | 30.63 | 33.33 | 25.72 | 33.12 |
| | Character-Deletion | 30.50 | 33.33 | 28.29 | 33.13 | 25.21 | 33.32 | 25.84 | 33.29 | 22.74 | 33.31 | 16.21 | 31.90 | 23.97 | 33.31 | 30.63 | 33.33 | 25.97 | 33.12 |
| | Character-Substitution | 30.50 | 33.33 | 28.29 | 33.13 | 25.19 | 33.32 | 25.82 | 33.29 | 22.67 | 33.31 | 12.38 | 31.90 | 23.93 | 33.31 | 30.63 | 33.33 | 25.29 | 33.12 |
| | Zero-Width Insertion | 30.98 | 33.33 | 28.51 | 33.13 | 27.66 | 33.32 | 26.89 | 33.29 | 25.86 | 33.31 | 31.99 | 31.90 | 27.50 | 33.33 | 30.98 | 33.33 | 30.39 | 33.12 |
| Log-Rank | Character-Insertion | 29.54 | 33.32 | 28.26 | 33.14 | 25.86 | 33.32 | 25.97 | 33.27 | 23.03 | 33.31 | 17.72 | 31.96 | 24.91 | 33.31 | 31.26 | 33.33 | 26.25 | 33.12 |
| | Character-Deletion | 29.48 | 33.32 | 28.10 | 33.13 | 25.69 | 33.32 | 25.87 | 33.26 | 23.02 | 33.31 | 20.22 | 31.98 | 24.88 | 33.31 | 31.18 | 33.33 | 26.62 | 33.12 |
| | Character-Substitution | 29.52 | 33.32 | 28.26 | 33.14 | 25.85 | 33.32 | 25.91 | 33.27 | 23.01 | 33.31 | 14.30 | 31.96 | 24.86 | 33.31 | 31.26 | 33.33 | 25.67 | 33.12 |
| | Zero-Width Insertion | 33.09 | 33.32 | 29.59 | 33.14 | 32.67 | 33.32 | 29.98 | 33.27 | 30.87 | 33.31 | 36.40 | 32.01 | 32.69 | 33.33 | 32.18 | 33.33 | 33.67 | 33.13 |
| DetectLLM-LRR | Character-Insertion | 27.03 | 32.73 | 33.00 | 33.15 | 29.37 | 33.21 | 36.45 | 33.01 | 26.31 | 33.16 | 36.61 | 32.95 | 30.29 | 33.30 | 32.39 | 33.33 | 31.87 | 33.11 |
| | Character-Deletion | 31.09 | 32.86 | 28.48 | 33.01 | 32.80 | 33.25 | 38.95 | 33.01 | 31.49 | 33.21 | 43.82 | 33.48 | 34.54 | 33.27 | 32.76 | 33.33 | 34.99 | 33.18 |
| | Character-Substitution | 22.08 | 32.72 | 27.42 | 32.99 | 26.19 | 33.21 | 26.07 | 32.99 | 22.97 | 33.16 | 23.20 | 32.88 | 26.86 | 33.28 | 32.30 | 33.33 | 26.02 | 33.07 |
| | Zero-Width Insertion | 47.84 | 35.47 | 49.60 | 34.56 | 54.88 | 34.38 | 57.38 | 33.60 | 52.38 | 35.58 | 54.88 | 35.43 | 55.18 | 34.01 | 41.40 | 33.33 | 52.58 | 34.56 |
| Fast-DetectGPT | Character-Insertion | 29.98 | 32.96 | 30.55 | 33.24 | 30.24 | 33.23 | 27.63 | 32.79 | 30.64 | 33.32 | 34.74 | 33.64 | 29.62 | 33.20 | 36.13 | 33.57 | 31.56 | 33.25 |
| | Character-Deletion | 30.63 | 33.02 | 31.61 | 33.27 | 32.68 | 33.34 | 35.59 | 33.89 | 31.37 | 33.37 | 35.60 | 33.78 | 30.95 | 33.24 | 35.80 | 33.48 | 33.68 | 33.43 |
| | Character-Substitution | 29.41 | 32.96 | 30.03 | 33.20 | 28.10 | 33.21 | 27.96 | 32.91 | 26.47 | 33.15 | 26.29 | 33.05 | 27.81 | 33.20 | 29.51 | 33.28 | 28.59 | 33.12 |
| | Zero-Width Insertion | 30.82 | 33.00 | 32.21 | 33.24 | 30.11 | 33.25 | 37.25 | 35.03 | 30.37 | 33.22 | 32.58 | 33.16 | 29.92 | 33.20 | 33.33 | 33.35 | 32.82 | 33.45 |
| Binoculars | Character-Insertion | 31.20 | 32.99 | 51.41 | 39.95 | 43.38 | 34.95 | 51.82 | 36.89 | 52.06 | 36.49 | 55.57 | 39.60 | 48.85 | 36.53 | 42.60 | 33.97 | 47.90 | 36.48 |
| | Character-Deletion | 36.59 | 33.91 | 60.91 | 45.94 | 54.90 | 38.85 | 59.73 | 40.43 | 58.52 | 39.13 | 60.80 | 42.04 | 59.33 | 41.49 | 45.59 | 34.36 | 55.52 | 39.65 |
| | Character-Substitution | 27.90 | 32.76 | 49.49 | 35.99 | 30.96 | 33.17 | 38.09 | 33.55 | 36.38 | 33.49 | 34.13 | 33.48 | 32.65 | 33.42 | 37.51 | 33.50 | 36.99 | 33.69 |
| | Zero-Width Insertion | 30.32 | 32.96 | 42.09 | 34.58 | 38.43 | 34.14 | 53.31 | 38.09 | 44.45 | 35.04 | 45.13 | 35.73 | 40.75 | 34.67 | 38.57 | 33.70 | 42.06 | 34.87 |
| Revise-Detect | Character-Insertion | 0.08 | 0.00 | 0.00 | 0.00 | 0.37 | 0.00 | 0.26 | 0.00 | 0.12 | 0.00 | 0.06 | 0.00 | 0.52 | 0.00 | 0.28 | 0.00 | 0.24 | 0.00 |
| | Character-Deletion | 0.00 | 0.00 | 0.00 | 0.00 | 0.19 | 0.00 | 0.04 | 0.00 | 0.00 | 0.00 | 0.00 | 0.00 | 0.11 | 0.00 | 0.03 | 0.00 | 0.06 | 0.00 |
| | Character-Substitution | 0.30 | 0.00 | 1.10 | 0.00 | 1.70 | 0.00 | 0.61 | 0.00 | 0.60 | 0.00 | 0.37 | 0.00 | 1.41 | 0.00 | 1.09 | 0.00 | 0.96 | 0.00 |
| | Zero-Width Insertion | 0.03 | 0.00 | 0.00 | 0.00 | 0.03 | 0.00 | 0.00 | 0.00 | 0.06 | 0.00 | 0.11 | 0.00 | 0.03 | 0.00 | 0.00 | 0.00 | 0.04 | 0.00 |
| GECScore | Character-Insertion | 30.98 | 31.82 | 30.86 | 30.93 | 29.37 | 31.85 | 32.92 | 33.12 | 27.60 | 30.89 | 30.83 | 33.87 | 28.77 | 31.54 | 27.88 | 31.75 | 29.44 | 31.98 |
| | Character-Deletion | 33.12 | 33.87 | 37.22 | 37.22 | 31.73 | 34.59 | 36.15 | 36.32 | 30.18 | 33.57 | 33.20 | 35.36 | 32.79 | 35.16 | 31.07 | 34.54 | 32.49 | 34.74 |
| | Character-Substitution | 31.79 | 33.52 | 36.41 | 36.44 | 32.68 | 35.34 | 38.88 | 38.92 | 31.41 | 34.17 | 33.96 | 36.05 | 32.99 | 35.37 | 31.99 | 35.06 | 32.81 | 35.05 |
| | Zero-Width Insertion | 31.41 | 32.38 | 32.97 | 33.02 | 29.63 | 31.45 | 33.06 | 33.27 | 27.50 | 31.07 | 30.01 | 33.05 | 28.93 | 31.82 | 27.03 | 30.89 | 29.32 | 31.87 |
| Lastde++ | Character-Insertion | 37.37 | 33.58 | 42.14 | 33.77 | 42.96 | 34.00 | 36.98 | 33.56 | 42.57 | 34.10 | 46.50 | 33.90 | 38.91 | 33.94 | 43.11 | 34.31 | 41.59 | 33.90 |
| | Character-Deletion | 40.44 | 33.72 | 52.86 | 34.44 | 49.26 | 34.60 | 43.61 | 34.11 | 46.41 | 34.37 | 47.03 | 34.06 | 44.17 | 34.44 | 45.77 | 34.20 | 46.45 | 34.24 |
| | Character-Substitution | 32.66 | 33.27 | 32.69 | 33.38 | 29.93 | 33.22 | 35.93 | 33.63 | 32.45 | 33.47 | 30.50 | 33.18 | 28.17 | 33.27 | 29.90 | 33.29 | 31.79 | 33.34 |
| | Zero-Width Insertion | 37.92 | 33.41 | 46.21 | 33.82 | 37.63 | 33.66 | 39.76 | 33.82 | 39.00 | 33.61 | 39.10 | 33.44 | 35.32 | 33.45 | 37.28 | 33.69 | 39.19 | 33.61 |
| RepreGuard | Character-Insertion | 63.54 | 3.70 | 65.41 | 0.07 | 66.61 | 78.99 | 73.91 | 0.33 | 66.59 | 84.58 | 66.34 | 0.07 | 66.61 | 81.84 | 66.55 | 55.93 | 66.95 | 52.22 |
| | Character-Deletion | 63.75 | 3.70 | 64.30 | 0.07 | 66.61 | 64.12 | 64.89 | 0.33 | 66.59 | 64.02 | 66.33 | 0.07 | 66.61 | 66.02 | 66.54 | 39.01 | 65.81 | 43.61 |
| | Character-Substitution | 63.56 | 3.70 | 65.27 | 0.07 | 66.61 | 68.57 | 66.86 | 0.33 | 66.59 | 73.59 | 66.33 | 0.07 | 66.61 | 70.14 | 66.54 | 50.27 | 66.11 | 47.48 |
| | Zero-Width Insertion | 63.15 | 3.70 | 64.92 | 0.07 | 66.61 | 95.46 | 65.28 | 0.33 | 66.59 | 93.04 | 66.33 | 0.07 | 66.61 | 97.36 | 66.54 | 59.39 | 65.83 | 57.14 |
| X-Rob-Classifier | Character-Insertion | 98.78 | 98.08 | 98.60 | 98.48 | 98.20 | 97.32 | 98.26 | 97.88 | 97.65 | 94.04 | 95.57 | 87.05 | 97.82 | 97.02 | 96.66 | 92.31 | 96.58 | 92.49 |
| | Character-Deletion | 98.39 | 97.95 | 98.91 | 98.84 | 99.02 | 99.01 | 98.43 | 98.03 | 97.87 | 96.98 | 96.95 | 94.79 | 98.37 | 98.26 | 97.75 | 96.78 | 98.40 | 98.07 |
| | Character-Substitution | 99.75 | 99.50 | 99.41 | 99.37 | 99.62 | 99.42 | 99.39 | 99.37 | 99.62 | 99.55 | 98.92 | 98.47 | 99.82 | 99.47 | 99.47 | 99.43 | 99.31 | 99.28 |
| | Zero-Width Insertion | 98.77 | 98.14 | 85.37 | 69.41 | 98.71 | 98.11 | 96.87 | 91.75 | 98.37 | 97.17 | 96.74 | 89.26 | 98.56 | 98.34 | 97.92 | 94.94 | 97.39 | 95.52 |
| mDeBERTa-Classifier | Character-Insertion | 99.65 | 99.49 | 95.27 | 88.91 | 99.32 | 99.31 | 98.37 | 96.94 | 99.08 | 99.05 | 97.68 | 95.59 | 99.58 | 99.44 | 99.09 | 99.02 | 97.53 | 97.04 |
| | Character-Deletion | 99.56 | 99.46 | 98.86 | 98.45 | 99.07 | 99.02 | 98.63 | 98.07 | 99.05 | 98.82 | 96.59 | 89.14 | 99.08 | 99.05 | 98.80 | 98.37 | 98.65 | 98.21 |
| | Character-Substitution | 99.89 | 99.85 | 99.49 | 99.41 | 99.80 | 99.69 | 99.67 | 99.48 | 99.68 | 99.49 | 98.68 | 98.32 | 99.87 | 99.79 | 99.79 | 99.57 | 99.55 | 99.45 |
| | Zero-Width Insertion | 99.76 | 99.55 | 89.11 | 76.83 | 99.63 | 99.55 | 99.57 | 99.42 | 99.55 | 99.42 | 98.78 | 98.54 | 99.80 | 99.37 | 99.53 | 99.41 | 97.32 | 97.02 |
| Biscope | Character-Insertion | 41.84 | 33.15 | 44.08 | 33.11 | 45.74 | 33.11 | 36.34 | 33.41 | 41.21 | 33.14 | 34.37 | 33.29 | 45.53 | 33.15 | 64.58 | 33.11 | 43.33 | 33.19 |
| | Character-Deletion | 44.14 | 33.17 | 71.63 | 33.19 | 45.27 | 33.11 | 41.58 | 34.23 | 42.89 | 33.18 | 40.71 | 33.98 | 47.61 | 33.18 | 62.60 | 33.11 | 48.65 | 33.43 |
| | Character-Substitution | 76.22 | 33.29 | 79.54 | 33.11 | 83.73 | 33.12 | 49.99 | 33.11 | 86.27 | 33.22 | 55.69 | 33.12 | 88.97 | 33.15 | 90.99 | 33.14 | 73.36 | 33.19 |
| | Zero-Width Insertion | 59.05 | 33.66 | 39.24 | 33.14 | 70.32 | 33.48 | 56.66 | 34.70 | 78.53 | 33.84 | 55.19 | 33.70 | 77.98 | 34.26 | 81.52 | 34.17 | 64.24 | 33.71 |

Table 27: Performance Comparison of Different Detectors on Cross Perturbation in BINARY Task. Detector performance is visualized using teal color gradients, with darker intensities indicating superior detection capabilities.

| Detectors↓ | Test↓ Lang→ | English | | Chinese | | Spanish | | Arabic | | French | | Russian | | Portuguese | | German | | Average | |
|---|---|---|---|---|---|---|---|---|---|---|---|---|---|---|---|---|---|---|---|
| | Metrics | $F_1^B$ | $F_1^F$ | $F_1^B$ | $F_1^F$ | $F_1^B$ | $F_1^F$ | $F_1^B$ | $F_1^F$ | $F_1^B$ | $F_1^F$ | $F_1^B$ | $F_1^F$ | $F_1^B$ | $F_1^F$ | $F_1^B$ | $F_1^F$ | $F_1^B$ | $F_1^F$ |
| Log-Likelihood | Character-Insertion | 18.58 | 15.72 | 19.30 | 14.95 | 20.39 | 14.09 | 24.64 | 15.54 | 20.36 | 13.27 | 11.85 | 9.42 | 22.20 | 13.86 | 24.76 | 18.26 | 20.85 | 14.61 |
| | Character-Deletion | 18.58 | 15.72 | 19.31 | 14.95 | 20.39 | 14.09 | 24.66 | 15.55 | 20.35 | 13.28 | 12.19 | 9.80 | 22.18 | 13.85 | 24.74 | 18.26 | 20.91 | 14.68 |
| | Character-Substitution | 18.58 | 15.72 | 19.30 | 14.95 | 20.41 | 14.09 | 24.65 | 15.53 | 20.36 | 13.27 | 11.33 | 8.72 | 22.20 | 13.84 | 24.75 | 18.26 | 20.74 | 14.47 |
| | Zero-Width Insertion | 18.76 | 15.89 | 19.38 | 15.03 | 20.85 | 14.77 | 24.73 | 15.77 | 21.08 | 14.32 | 18.41 | 16.12 | 22.99 | 14.99 | 24.82 | 18.34 | 22.48 | 16.37 |
| Log-Rank | Character-Insertion | 22.63 | 15.27 | 25.42 | 14.82 | 19.18 | 14.10 | 23.63 | 15.59 | 18.81 | 13.10 | 12.02 | 10.66 | 19.09 | 13.95 | 30.49 | 18.03 | 21.99 | 14.63 |
| | Character-Deletion | 22.47 | 15.28 | 25.49 | 14.82 | 18.66 | 14.12 | 23.53 | 15.60 | 18.26 | 13.17 | 12.93 | 11.60 | 18.62 | 14.01 | 30.38 | 18.03 | 21.98 | 14.83 |
| | Character-Substitution | 22.81 | 15.06 | 25.22 | 14.66 | 20.08 | 13.72 | 24.06 | 15.35 | 19.74 | 12.63 | 11.87 | 8.98 | 19.98 | 13.55 | 30.66 | 18.16 | 22.46 | 14.15 |
| | Zero-Width Insertion | 22.99 | 16.76 | 24.84 | 15.38 | 18.66 | 16.72 | 25.04 | 16.76 | 18.79 | 16.38 | 20.02 | 18.94 | 19.60 | 16.88 | 29.69 | 18.31 | 24.05 | 17.73 |
| DetectLLM-LRR | Character-Insertion | 19.98 | 14.05 | 20.78 | 17.34 | 17.70 | 15.11 | 19.04 | 20.45 | 16.94 | 13.69 | 21.36 | 19.63 | 18.08 | 15.84 | 24.22 | 17.59 | 20.23 | 17.00 |
| | Character-Deletion | 22.35 | 16.39 | 19.39 | 14.76 | 19.75 | 17.20 | 20.25 | 21.71 | 19.75 | 16.50 | 25.28 | 23.55 | 20.40 | 18.21 | 24.37 | 17.77 | 22.08 | 18.73 |
| | Character-Substitution | 18.70 | 11.36 | 21.74 | 14.32 | 17.70 | 13.41 | 14.76 | 15.23 | 16.57 | 11.95 | 15.22 | 12.50 | 17.61 | 13.80 | 25.39 | 17.45 | 18.48 | 13.79 |
| | Zero-Width Insertion | 31.88 | 25.75 | 29.98 | 26.76 | 32.33 | 29.66 | 29.07 | 30.98 | 31.74 | 28.19 | 31.03 | 29.25 | 32.24 | 30.01 | 29.74 | 23.09 | 31.58 | 28.42 |
| Fast-DetectGPT | Character-Insertion | 16.65 | 15.73 | 18.30 | 16.32 | 17.60 | 16.31 | 19.87 | 16.98 | 17.88 | 16.32 | 21.10 | 18.34 | 17.94 | 16.06 | 20.48 | 18.70 | 18.91 | 16.99 |
| | Character-Deletion | 16.83 | 15.91 | 18.67 | 16.69 | 18.27 | 16.98 | 22.88 | 19.97 | 18.10 | 16.53 | 21.34 | 18.56 | 18.38 | 16.50 | 20.15 | 18.39 | 19.66 | 17.74 |
| | Character-Substitution | 16.52 | 15.61 | 18.19 | 16.21 | 16.95 | 15.67 | 19.98 | 17.08 | 16.59 | 15.05 | 18.33 | 15.57 | 17.54 | 15.65 | 18.14 | 16.39 | 18.00 | 16.08 |
| | Zero-Width Insertion | 16.87 | 15.95 | 18.79 | 16.79 | 17.51 | 16.22 | 23.88 | 20.94 | 17.68 | 16.12 | 20.13 | 17.37 | 17.94 | 16.04 | 19.22 | 17.45 | 19.42 | 17.50 |
| Binoculars | Character-Insertion | 25.83 | 16.75 | 32.82 | 20.99 | 29.91 | 17.80 | 27.02 | 19.31 | 29.18 | 18.76 | 32.83 | 20.66 | 32.74 | 18.82 | 23.91 | 17.56 | 29.77 | 18.87 |
| | Character-Deletion | 26.23 | 17.29 | 36.53 | 24.56 | 32.21 | 20.26 | 29.16 | 21.40 | 30.74 | 20.34 | 34.33 | 22.18 | 35.71 | 21.88 | 24.21 | 17.84 | 31.65 | 20.81 |
| | Character-Substitution | 25.79 | 16.60 | 30.40 | 18.63 | 29.28 | 16.72 | 25.12 | 17.32 | 27.61 | 16.86 | 29.64 | 16.93 | 31.47 | 16.93 | 23.64 | 17.29 | 28.33 | 17.17 |
| | Zero-Width Insertion | 25.84 | 16.73 | 29.71 | 17.80 | 29.59 | 17.33 | 27.84 | 20.11 | 28.39 | 17.85 | 30.66 | 18.31 | 31.89 | 17.70 | 23.73 | 17.40 | 28.93 | 17.91 |
| Revise-Detect | Character-Insertion | 30.03 | 16.97 | 22.24 | 20.36 | 26.42 | 14.72 | 38.75 | 22.12 | 28.86 | 15.55 | 29.44 | 15.94 | 26.74 | 16.09 | 29.28 | 15.72 | 29.12 | 16.90 |
| | Character-Deletion | 34.86 | 21.13 | 22.59 | 20.71 | 26.51 | 14.81 | 31.57 | 16.02 | 29.24 | 15.90 | 29.42 | 15.91 | 26.69 | 16.03 | 29.08 | 15.53 | 29.09 | 16.87 |
| | Character-Substitution | 33.53 | 20.01 | 22.86 | 20.97 | 26.60 | 14.93 | 33.72 | 17.92 | 28.64 | 15.36 | 29.28 | 15.79 | 27.08 | 16.45 | 29.39 | 15.85 | 29.23 | 17.03 |
| | Zero-Width Insertion | 28.63 | 15.71 | 21.70 | 19.83 | 25.86 | 14.16 | 38.00 | 21.50 | 28.58 | 15.29 | 28.80 | 15.34 | 26.40 | 15.74 | 28.79 | 15.26 | 28.49 | 16.32 |
| GECScore | Character-Insertion | 25.25 | 6.36 | 28.04 | 7.04 | 16.63 | 0.00 | 26.55 | 6.65 | 22.54 | 5.81 | 16.63 | 0.00 | 19.39 | 3.20 | 22.16 | 5.73 | 29.46 | 14.04 |
| | Character-Deletion | 25.13 | 6.30 | 27.99 | 7.02 | 16.63 | 0.00 | 27.03 | 6.85 | 22.76 | 5.94 | 16.63 | 0.00 | 19.76 | 3.43 | 22.64 | 6.01 | 30.57 | 14.94 |
| | Character-Substitution | 24.64 | 6.08 | 27.97 | 7.01 | 16.63 | 0.00 | 27.54 | 7.06 | 22.84 | 5.98 | 16.63 | 0.00 | 19.76 | 3.43 | 22.69 | 6.04 | 30.66 | 15.01 |
| | Zero-Width Insertion | 25.32 | 6.38 | 27.98 | 7.02 | 16.63 | 0.00 | 26.60 | 6.68 | 22.54 | 5.82 | 16.63 | 0.00 | 19.41 | 3.22 | 22.17 | 5.74 | 29.59 | 14.15 |
| Lastde++ | Character-Insertion | 16.65 | 15.73 | 18.30 | 16.32 | 17.60 | 16.31 | 19.87 | 16.98 | 17.88 | 16.32 | 21.10 | 18.34 | 17.94 | 16.06 | 20.48 | 18.70 | 18.91 | 16.99 |
| | Character-Deletion | 16.83 | 15.91 | 18.67 | 16.69 | 18.27 | 16.98 | 22.88 | 19.97 | 18.10 | 16.53 | 21.34 | 18.56 | 18.38 | 16.50 | 20.15 | 18.39 | 19.66 | 17.74 |
| | Character-Substitution | 16.52 | 15.61 | 18.19 | 16.21 | 16.95 | 15.67 | 19.98 | 17.08 | 16.59 | 15.05 | 18.33 | 15.57 | 17.54 | 15.65 | 18.14 | 16.39 | 18.00 | 16.08 |
| | Zero-Width Insertion | 16.87 | 15.95 | 18.79 | 16.79 | 17.51 | 16.22 | 23.88 | 20.94 | 17.68 | 16.12 | 20.13 | 17.37 | 17.94 | 16.04 | 19.22 | 17.45 | 19.42 | 17.50 |
| RepreGuard | Character-Insertion | 18.97 | 16.67 | 28.71 | 16.67 | 16.68 | 16.93 | 16.76 | 16.67 | 16.67 | 16.82 | 16.90 | 16.67 | 16.67 | 17.28 | 16.68 | 16.81 | 20.42 | 16.81 |
| | Character-Deletion | 18.46 | 16.67 | 35.27 | 16.67 | 16.68 | 16.77 | 16.76 | 16.67 | 16.67 | 16.79 | 16.69 | 16.67 | 16.67 | 17.10 | 16.68 | 16.81 | 22.12 | 16.77 |
| | Character-Substitution | 19.57 | 16.67 | 25.69 | 16.67 | 16.68 | 17.52 | 16.72 | 16.67 | 16.66 | 17.05 | 16.69 | 16.67 | 16.67 | 18.25 | 16.68 | 16.73 | 19.50 | 17.03 |
| | Zero-Width Insertion | 19.73 | 16.67 | 26.19 | 16.67 | 16.67 | 17.50 | 16.73 | 16.67 | 16.66 | 17.05 | 16.70 | 16.67 | 16.67 | 18.25 | 16.67 | 16.73 | 19.52 | 17.03 |
| X-Rob-Classifier | Character-Insertion | 97.25 | 97.42 | 94.74 | 93.04 | 95.83 | 93.87 | 95.67 | 94.50 | 96.11 | 96.07 | 93.68 | 91.41 | 96.52 | 95.89 | 94.59 | 93.46 | 95.55 | 94.54 |
| | Character-Deletion | 97.18 | 97.06 | 93.83 | 90.10 | 96.01 | 94.13 | 95.82 | 95.49 | 96.07 | 96.21 | 93.25 | 90.91 | 96.39 | 96.33 | 94.58 | 92.51 | 95.39 | 94.10 |
| | Character-Substitution | 97.43 | 97.29 | 95.55 | 93.32 | 95.88 | 94.25 | 96.04 | 95.39 | 95.94 | 96.32 | 93.78 | 92.51 | 96.66 | 96.18 | 94.71 | 92.92 | 95.75 | 94.90 |
| | Zero-Width Insertion | 97.23 | 97.58 | 95.23 | 94.19 | 95.88 | 94.46 | 96.32 | 95.79 | 96.02 | 96.29 | 93.11 | 92.64 | 96.54 | 96.46 | 94.57 | 93.49 | 95.61 | 95.15 |
| mDeBERTa-Classifier | Character-Insertion | 98.25 | 98.02 | 96.81 | 95.22 | 97.25 | 95.86 | 97.34 | 97.15 | 97.66 | 96.99 | 95.91 | 93.70 | 98.03 | 97.48 | 96.56 | 94.75 | 97.23 | 96.17 |
| | Character-Deletion | 98.36 | 98.13 | 96.77 | 95.57 | 97.26 | 95.69 | 97.46 | 97.08 | 97.57 | 96.98 | 95.62 | 93.09 | 97.99 | 97.43 | 96.72 | 94.65 | 97.22 | 96.10 |
| | Character-Substitution | 98.40 | 98.08 | 96.84 | 95.50 | 97.20 | 95.43 | 97.62 | 97.51 | 97.64 | 96.84 | 95.64 | 93.56 | 98.00 | 97.46 | 96.66 | 94.60 | 97.25 | 96.10 |
| | Zero-Width Insertion | 98.36 | 98.21 | 96.74 | 95.70 | 97.22 | 96.06 | 97.32 | 97.06 | 97.52 | 97.09 | 95.78 | 93.71 | 98.03 | 97.67 | 96.64 | 94.87 | 97.21 | 96.36 |
| Biscope | Character-Insertion | 51.92 | 33.36 | 47.47 | 17.82 | 57.43 | 21.88 | 39.40 | 18.28 | 56.65 | 20.12 | 29.90 | 17.34 | 57.12 | 22.78 | 66.71 | 34.04 | 51.77 | 24.04 |
| | Character-Deletion | 44.92 | 32.99 | 54.95 | 21.52 | 50.54 | 20.19 | 36.19 | 18.72 | 50.03 | 18.95 | 30.17 | 17.70 | 52.06 | 21.26 | 62.25 | 28.89 | 48.44 | 22.95 |
| | Character-Substitution | 73.00 | 56.55 | 69.15 | 18.41 | 73.18 | 44.83 | 59.23 | 19.88 | 72.89 | 50.12 | 52.95 | 20.97 | 74.78 | 48.87 | 72.46 | 50.63 | 68.70 | 41.84 |
| | Zero-Width Insertion | 61.41 | 36.73 | 32.95 | 18.15 | 62.41 | 24.17 | 36.11 | 18.48 | 65.98 | 30.07 | 33.11 | 17.78 | 67.25 | 31.54 | 65.40 | 35.64 | 54.85 | 27.17 |

Table 28: Performance Comparison of Different Detectors on Cross Perturbation in TERNARY Task. Detector performance is visualized using teal color gradients, with darker intensities indicating superior detection capabilities.

| Detectors↓ | Test↓ Lang→ | English | | Chinese | | Spanish | | Arabic | | French | | Russian | | Portuguese | | German | | Average | |
|---|---|---|---|---|---|---|---|---|---|---|---|---|---|---|---|---|---|---|---|
| | Metrics | $F_1^B$ | $F_1^F$ | $F_1^B$ | $F_1^F$ | $F_1^B$ | $F_1^F$ | $F_1^B$ | $F_1^F$ | $F_1^B$ | $F_1^F$ | $F_1^B$ | $F_1^F$ | $F_1^B$ | $F_1^F$ | $F_1^B$ | $F_1^F$ | $F_1^B$ | $F_1^F$ |
| Log-Likelihood | 64 | 41.08 | 33.58 | 38.57 | 33.68 | 52.61 | 34.01 | 56.46 | 34.16 | 48.93 | 33.95 | 59.32 | 37.77 | 48.02 | 33.62 | 40.58 | 33.37 | 49.61 | 34.32 |
| | 128 | 47.39 | 33.60 | 41.40 | 33.52 | 62.77 | 34.60 | 63.71 | 33.83 | 59.38 | 34.37 | 62.64 | 38.79 | 59.96 | 34.16 | 48.10 | 33.33 | 57.41 | 34.60 |
| | 256 | 53.33 | 33.82 | 46.77 | 33.89 | 68.29 | 35.19 | 69.13 | 33.78 | 66.73 | 35.44 | 59.08 | 42.27 | 68.35 | 35.27 | 56.42 | 33.42 | 62.78 | 35.52 |
| | 512 | 58.11 | 34.72 | 50.28 | 35.55 | 72.50 | 36.71 | 71.19 | 34.17 | 68.86 | 38.05 | 56.05 | 44.11 | 72.01 | 36.93 | 62.15 | 33.48 | 65.49 | 36.88 |
| Log-Rank | 64 | 41.73 | 33.77 | 40.06 | 33.99 | 51.25 | 34.32 | 57.41 | 35.06 | 48.05 | 34.21 | 57.53 | 37.71 | 46.08 | 33.66 | 38.85 | 33.42 | 48.82 | 34.57 |
| | 128 | 49.34 | 34.05 | 43.44 | 33.74 | 61.54 | 35.32 | 64.84 | 34.54 | 57.98 | 34.74 | 61.59 | 38.82 | 58.00 | 34.31 | 44.99 | 33.40 | 56.67 | 34.93 |
| | 256 | 55.43 | 34.65 | 50.23 | 34.76 | 67.20 | 36.34 | 69.81 | 34.97 | 65.46 | 36.83 | 60.12 | 41.94 | 66.95 | 35.90 | 52.39 | 33.42 | 62.40 | 36.21 |
| | 512 | 59.72 | 36.93 | 54.53 | 36.81 | 71.89 | 38.87 | 72.15 | 35.74 | 68.03 | 40.86 | 57.49 | 43.87 | 71.89 | 38.20 | 59.00 | 33.54 | 65.60 | 38.25 |
| DetectLLM-LRR | 64 | 46.15 | 36.72 | 48.94 | 37.37 | 48.82 | 36.38 | 59.98 | 41.97 | 46.92 | 36.31 | 49.75 | 37.91 | 42.57 | 34.83 | 38.22 | 33.78 | 48.25 | 36.98 |
| | 128 | 53.51 | 38.62 | 54.14 | 37.07 | 56.46 | 37.75 | 65.62 | 41.03 | 52.69 | 37.47 | 53.08 | 37.14 | 50.99 | 35.14 | 40.39 | 33.73 | 54.07 | 37.30 |
| | 256 | 57.95 | 41.02 | 62.08 | 39.84 | 60.53 | 38.24 | 70.15 | 42.04 | 58.17 | 39.53 | 56.40 | 37.54 | 59.92 | 36.23 | 44.03 | 33.57 | 59.53 | 38.57 |
| | 512 | 60.14 | 45.92 | 66.00 | 43.86 | 66.07 | 40.35 | 73.12 | 43.61 | 60.94 | 43.22 | 57.06 | 38.20 | 66.79 | 38.40 | 47.80 | 33.65 | 63.13 | 41.04 |
| Fast-DetectGPT | 64 | 49.88 | 43.22 | 54.22 | 47.88 | 52.92 | 47.67 | 66.89 | 66.02 | 53.91 | 49.18 | 61.03 | 58.16 | 51.44 | 45.53 | 54.38 | 46.98 | 55.59 | 50.39 |
| | 128 | 46.86 | 39.16 | 52.34 | 43.64 | 51.88 | 44.00 | 65.67 | 65.93 | 56.12 | 48.36 | 61.28 | 56.27 | 50.94 | 42.53 | 57.57 | 46.41 | 55.65 | 48.19 |
| | 256 | 48.24 | 38.44 | 56.63 | 41.39 | 46.65 | 38.65 | 81.17 | 73.79 | 71.05 | 56.62 | 62.16 | 52.70 | 57.00 | 38.65 | 73.25 | 51.90 | 58.22 | 47.58 |
| | 512 | 52.75 | 39.24 | 52.61 | 36.25 | 46.35 | 36.77 | 77.04 | 66.83 | 65.32 | 48.43 | 72.46 | 59.14 | 55.50 | 36.13 | 70.67 | 46.10 | 56.27 | 43.30 |
| Binoculars | 64 | 72.11 | 69.36 | 71.43 | 73.08 | 73.04 | 74.50 | 78.39 | 71.82 | 76.93 | 74.27 | 73.38 | 69.91 | 75.50 | 74.57 | 74.29 | 67.16 | 74.50 | 71.94 |
| | 128 | 59.40 | 57.39 | 62.58 | 63.54 | 74.12 | 71.51 | 83.30 | 71.28 | 80.27 | 80.63 | 77.07 | 71.36 | 78.51 | 74.97 | 81.16 | 69.70 | 75.88 | 71.74 |
| | 256 | 53.80 | 46.69 | 62.85 | 62.20 | 72.36 | 62.01 | 78.85 | 64.58 | 81.93 | 79.78 | 77.31 | 66.23 | 76.57 | 66.45 | 75.21 | 58.08 | 74.39 | 63.48 |
| | 512 | 41.52 | 39.69 | 67.06 | 67.12 | 71.91 | 54.01 | 74.64 | 57.44 | 77.40 | 59.79 | 78.13 | 62.94 | 77.64 | 60.08 | 76.01 | 52.03 | 74.80 | 57.58 |
| Revise-Detect | 64 | 7.11 | 0.00 | 4.59 | 0.00 | 14.97 | 0.00 | 10.70 | 0.00 | 7.26 | 0.00 | 19.93 | 0.00 | 14.13 | 0.00 | 3.46 | 0.00 | 12.39 | 0.00 |
| | 128 | 28.64 | 0.00 | 0.00 | 0.00 | 36.85 | 0.00 | 22.12 | 0.00 | 13.40 | 0.00 | 38.89 | 0.00 | 33.71 | 0.00 | 12.29 | 0.00 | 28.52 | 0.00 |
| | 256 | 72.54 | 0.00 | 0.39 | 0.00 | 78.87 | 0.00 | 56.68 | 0.00 | 50.29 | 0.00 | 64.65 | 0.00 | 62.76 | 0.00 | 41.65 | 0.00 | 59.18 | 0.00 |
| | 512 | 76.36 | 0.00 | 0.35 | 0.00 | 78.19 | 0.00 | 60.87 | 0.00 | 60.94 | 0.00 | 68.45 | 0.00 | 67.94 | 0.00 | 55.63 | 0.00 | 65.24 | 0.00 |
| GECScore | 64 | 68.80 | 67.84 | 50.82 | 50.87 | 69.56 | 68.76 | 65.40 | 64.90 | 61.88 | 61.66 | 69.14 | 68.86 | 62.30 | 61.43 | 51.43 | 51.17 | 63.74 | 63.19 |
| | 128 | 78.34 | 70.59 | 50.15 | 50.18 | 79.85 | 74.55 | 65.68 | 60.29 | 69.07 | 61.29 | 72.50 | 68.62 | 70.33 | 60.86 | 61.02 | 50.60 | 71.19 | 64.08 |
| | 256 | 88.83 | 74.84 | 43.18 | 43.01 | 88.91 | 72.58 | 75.21 | 56.41 | 78.23 | 57.58 | 80.69 | 65.65 | 80.01 | 58.97 | 71.09 | 49.28 | 80.83 | 63.50 |
| | 512 | 93.31 | 80.01 | 37.84 | 37.79 | 90.88 | 74.61 | 76.99 | 47.95 | 85.52 | 55.70 | 84.42 | 62.11 | 86.37 | 58.34 | 78.76 | 52.16 | 79.56 | 45.43 |
| Lastde++ | 64 | 66.75 | 0.00 | 66.71 | 0.00 | 66.73 | 0.00 | 66.73 | 0.00 | 66.74 | 0.00 | 66.69 | 0.00 | 66.67 | 0.00 | 66.61 | 0.00 | 33.57 | 33.28 |
| | 128 | 66.71 | 0.00 | 66.63 | 0.00 | 66.62 | 0.00 | 66.70 | 0.00 | 66.65 | 0.00 | 66.68 | 0.00 | 66.66 | 0.00 | 66.64 | 0.00 | 35.01 | 33.33 |
| | 256 | 66.67 | 0.00 | 66.66 | 0.00 | 66.66 | 0.00 | 66.67 | 0.01 | 66.67 | 0.00 | 66.67 | 0.00 | 66.66 | 0.00 | 66.67 | 0.00 | 38.09 | 33.41 |
| | 512 | 66.67 | 0.00 | 66.67 | 0.00 | 66.67 | 0.01 | 66.67 | 0.01 | 66.67 | 0.00 | 66.67 | 0.01 | 66.66 | 0.01 | 66.67 | 0.01 | 38.25 | 33.54 |
| RepreGuard | 64 | 79.34 | 0.01 | 60.64 | 0.00 | 66.07 | 0.74 | 81.91 | 0.00 | 66.12 | 0.69 | 71.07 | 0.01 | 66.21 | 0.78 | 65.85 | 0.36 | 69.43 | 42.69 |
| | 128 | 81.13 | 0.01 | 59.63 | 0.00 | 66.16 | 0.84 | 88.35 | 0.00 | 66.18 | 0.79 | 72.40 | 0.01 | 66.24 | 0.87 | 65.92 | 0.44 | 70.37 | 48.95 |
| | 256 | 82.69 | 0.02 | 60.90 | 0.00 | 66.25 | 0.91 | 94.64 | 0.00 | 66.27 | 0.88 | 74.81 | 0.00 | 66.30 | 0.92 | 65.99 | 0.55 | 71.52 | 54.45 |
| | 512 | 86.32 | 0.04 | 63.84 | 0.00 | 66.61 | 0.94 | 96.71 | 0.00 | 66.56 | 0.92 | 77.05 | 0.00 | 66.61 | 0.95 | 66.32 | 0.60 | 72.70 | 56.63 |
| X-Rob-Classifier | 64 | 98.92 | 98.88 | 97.80 | 97.07 | 97.52 | 97.07 | 97.62 | 97.14 | 97.34 | 96.73 | 97.25 | 96.41 | 98.48 | 98.48 | 98.55 | 98.32 | 97.25 | 96.07 |
| | 128 | 99.55 | 99.42 | 99.14 | 99.11 | 99.37 | 99.33 | 98.99 | 98.97 | 98.49 | 98.02 | 98.67 | 98.21 | 99.54 | 99.36 | 99.22 | 99.22 | 98.78 | 98.61 |
| | 256 | 99.92 | 99.50 | 99.73 | 99.47 | 99.87 | 99.50 | 99.76 | 99.47 | 99.47 | 99.41 | 99.59 | 99.47 | 99.87 | 99.66 | 99.76 | 99.50 | 99.71 | 99.49 |
| | 512 | 99.95 | 99.50 | 99.92 | 99.45 | 99.99 | 99.51 | 99.87 | 99.47 | 99.90 | 99.60 | 99.84 | 99.46 | 100.00 | 99.47 | 99.96 | 99.38 | 99.87 | 99.49 |
| mDeBERTa-Classifier | 64 | 99.13 | 99.10 | 98.28 | 97.93 | 97.12 | 96.01 | 98.22 | 97.92 | 97.80 | 96.61 | 97.25 | 97.01 | 96.14 | 92.38 | 98.47 | 98.45 | 97.92 | 97.27 |
| | 128 | 99.78 | 99.51 | 99.12 | 99.08 | 99.38 | 99.33 | 99.49 | 99.34 | 99.24 | 99.19 | 98.97 | 98.93 | 99.60 | 99.36 | 99.70 | 99.41 | 99.18 | 99.18 |
| | 256 | 99.94 | 99.55 | 99.82 | 99.50 | 99.93 | 99.55 | 99.92 | 99.59 | 99.60 | 99.44 | 99.78 | 99.47 | 99.92 | 99.67 | 99.88 | 99.42 | 99.85 | 99.49 |
| | 512 | 99.97 | 99.52 | 99.84 | 99.48 | 99.97 | 99.50 | 99.92 | 99.56 | 99.83 | 99.49 | 99.92 | 99.57 | 99.93 | 99.50 | 99.95 | 99.48 | 99.87 | 99.50 |
| Biscope | 64 | 73.87 | 42.11 | 73.07 | 38.84 | 70.41 | 39.17 | 72.25 | 36.70 | 69.72 | 40.80 | 67.49 | 36.08 | 71.18 | 41.75 | 70.86 | 39.14 | 70.48 | 38.77 |
| | 128 | 82.49 | 54.79 | 73.58 | 40.56 | 80.35 | 50.30 | 77.80 | 37.38 | 80.37 | 54.39 | 75.12 | 38.34 | 81.08 | 57.71 | 79.08 | 50.58 | 77.90 | 46.20 |
| | 256 | 87.82 | 68.43 | 80.43 | 48.72 | 90.14 | 74.19 | 83.61 | 45.90 | 90.49 | 73.51 | 82.32 | 53.83 | 90.80 | 77.75 | 86.02 | 65.93 | 85.25 | 63.68 |
| | 512 | 92.16 | 84.04 | 87.27 | 70.29 | 94.83 | 90.30 | 88.92 | 68.21 | 94.93 | 87.46 | 88.57 | 65.09 | 95.52 | 90.68 | 91.82 | 78.53 | 90.73 | 79.15 |

Table 29: Performance Comparison of Different Detectors on Cross Length in BINARY Task. Detector performance is visualized using teal color gradients, with darker intensities indicating superior detection capabilities.

| Detectors↓ | Test↓ Lang→ | English | | Chinese | | Spanish | | Arabic | | French | | Russian | | Portuguese | | German | | Average | |
|---|---|---|---|---|---|---|---|---|---|---|---|---|---|---|---|---|---|---|---|
| | Metrics | $F_1^B$ | $F_1^F$ | $F_1^B$ | $F_1^F$ | $F_1^B$ | $F_1^F$ | $F_1^B$ | $F_1^F$ | $F_1^B$ | $F_1^F$ | $F_1^B$ | $F_1^F$ | $F_1^B$ | $F_1^F$ | $F_1^B$ | $F_1^F$ | $F_1^B$ | $F_1^F$ |
| Log-Likelihood | 64 | 23.51 | 20.98 | 22.60 | 19.99 | 31.61 | 27.40 | 35.86 | 29.31 | 29.65 | 25.31 | 38.33 | 31.36 | 29.16 | 24.74 | 23.50 | 20.48 | 30.32 | 25.83 |
| | 128 | 27.36 | 24.39 | 24.67 | 21.20 | 38.20 | 33.26 | 40.71 | 33.21 | 37.58 | 31.33 | 40.30 | 33.39 | 37.75 | 31.50 | 28.44 | 24.42 | 35.63 | 30.13 |
| | 256 | 31.05 | 27.96 | 28.26 | 23.95 | 43.03 | 36.92 | 45.49 | 36.30 | 43.56 | 35.91 | 36.47 | 32.46 | 44.70 | 36.42 | 35.18 | 29.33 | 39.75 | 33.34 |
| | 512 | 34.19 | 30.92 | 31.04 | 26.13 | 45.66 | 39.42 | 47.74 | 37.63 | 44.79 | 37.65 | 33.26 | 31.37 | 47.58 | 39.08 | 39.89 | 32.89 | 41.76 | 35.12 |
| Log-Rank | 64 | 28.86 | 21.62 | 28.84 | 20.73 | 35.38 | 27.15 | 42.08 | 30.37 | 34.78 | 25.31 | 39.32 | 30.41 | 33.30 | 23.88 | 28.55 | 19.71 | 35.08 | 25.67 |
| | 128 | 33.79 | 25.91 | 31.62 | 22.41 | 39.86 | 32.77 | 46.72 | 34.03 | 39.60 | 30.75 | 38.48 | 32.68 | 40.58 | 30.52 | 34.22 | 23.20 | 39.59 | 29.90 |
| | 256 | 37.09 | 29.52 | 36.62 | 26.22 | 40.72 | 36.31 | 48.58 | 36.91 | 39.97 | 35.15 | 34.72 | 32.75 | 43.31 | 35.89 | 40.09 | 27.69 | 41.78 | 33.30 |
| | 512 | 39.14 | 32.31 | 39.48 | 28.71 | 40.46 | 39.14 | 47.02 | 38.30 | 38.82 | 36.71 | 32.01 | 31.61 | 41.15 | 38.80 | 43.56 | 31.50 | 41.90 | 35.19 |
| DetectLLM-LRR | 64 | 32.91 | 24.58 | 35.15 | 26.26 | 33.24 | 26.21 | 40.38 | 32.31 | 33.54 | 25.13 | 33.95 | 26.63 | 30.80 | 22.63 | 29.70 | 19.85 | 34.42 | 25.86 |
| | 128 | 36.40 | 28.96 | 37.20 | 29.10 | 34.60 | 30.74 | 41.62 | 35.17 | 34.36 | 28.68 | 33.72 | 28.47 | 34.51 | 27.57 | 32.09 | 21.53 | 36.44 | 29.24 |
| | 256 | 37.59 | 31.32 | 38.43 | 33.64 | 35.50 | 32.83 | 39.93 | 37.40 | 34.66 | 31.49 | 33.39 | 30.12 | 36.76 | 32.64 | 32.83 | 23.89 | 36.91 | 32.17 |
| | 512 | 39.24 | 32.48 | 39.15 | 35.59 | 38.47 | 35.98 | 38.59 | 39.12 | 36.45 | 32.80 | 32.66 | 30.43 | 38.76 | 36.30 | 33.97 | 26.43 | 37.81 | 34.11 |
| Fast-DetectGPT | 64 | 26.66 | 24.95 | 30.07 | 27.67 | 29.61 | 27.53 | 39.53 | 36.22 | 30.69 | 28.15 | 36.17 | 33.01 | 28.77 | 26.26 | 29.79 | 27.55 | 27.25 | 24.63 |
| | 128 | 24.73 | 22.72 | 29.08 | 25.94 | 28.72 | 26.23 | 40.34 | 36.06 | 31.36 | 28.32 | 37.30 | 32.86 | 28.63 | 25.55 | 31.44 | 28.18 | 32.08 | 28.81 |
| | 256 | 23.31 | 21.44 | 28.51 | 24.11 | 25.87 | 22.84 | 40.52 | 34.79 | 29.97 | 26.55 | 38.32 | 32.36 | 26.53 | 22.77 | 30.50 | 26.54 | 31.29 | 27.20 |
| | 512 | 24.11 | 22.18 | 27.49 | 22.89 | 25.32 | 22.42 | 40.75 | 34.34 | 28.33 | 25.31 | 39.49 | 32.61 | 26.20 | 22.53 | 30.15 | 26.13 | 31.18 | 26.89 |
| Binoculars | 64 | 47.08 | 37.82 | 49.34 | 39.65 | 49.35 | 40.48 | 47.04 | 39.45 | 49.07 | 40.64 | 47.05 | 38.19 | 50.26 | 40.54 | 44.49 | 36.83 | 33.04 | 24.55 |
| | 128 | 43.14 | 33.21 | 49.38 | 37.90 | 49.30 | 39.23 | 45.95 | 37.83 | 50.48 | 41.06 | 49.73 | 39.28 | 52.66 | 40.79 | 43.65 | 35.85 | 48.33 | 38.25 |
| | 256 | 36.48 | 27.13 | 48.39 | 36.11 | 45.75 | 34.24 | 41.45 | 33.28 | 48.33 | 38.33 | 48.35 | 36.64 | 50.01 | 36.53 | 39.21 | 31.95 | 45.22 | 34.45 |
| | 512 | 32.56 | 24.39 | 49.22 | 36.66 | 41.53 | 29.38 | 38.54 | 30.91 | 44.34 | 33.66 | 47.40 | 34.83 | 47.71 | 33.18 | 34.89 | 28.23 | 42.65 | 31.64 |
| Revise-Detect | 64 | 31.61 | 23.28 | 20.04 | 19.36 | 35.23 | 26.80 | 26.55 | 17.01 | 30.12 | 24.12 | 34.89 | 27.96 | 36.04 | 26.20 | 31.26 | 24.27 | 32.33 | 25.07 |
| | 128 | 43.22 | 31.17 | 19.06 | 18.48 | 45.56 | 34.51 | 36.48 | 21.17 | 36.84 | 27.24 | 45.75 | 34.93 | 45.53 | 33.09 | 37.75 | 27.42 | 41.50 | 31.05 |
| | 256 | 56.56 | 44.71 | 23.95 | 23.31 | 56.99 | 47.83 | 55.11 | 39.17 | 50.71 | 37.66 | 53.98 | 42.18 | 53.20 | 41.77 | 47.93 | 35.39 | 52.18 | 40.82 |
| | 512 | 56.65 | 46.12 | 21.88 | 21.32 | 54.87 | 46.70 | 50.11 | 31.46 | 51.87 | 41.22 | 54.39 | 43.32 | 52.29 | 43.09 | 49.47 | 38.79 | 51.69 | 41.80 |
| GECScore | 64 | 31.00 | 12.61 | 20.12 | 2.77 | 32.07 | 14.30 | 22.64 | 4.71 | 31.23 | 12.91 | 33.00 | 14.23 | 31.54 | 13.45 | 29.14 | 10.90 | 47.28 | 31.34 |
| | 128 | 33.92 | 16.20 | 20.40 | 3.69 | 32.03 | 15.36 | 20.27 | 3.63 | 30.00 | 12.97 | 30.83 | 13.51 | 31.01 | 13.88 | 28.86 | 12.06 | 47.79 | 33.27 |
| | 256 | 20.03 | 1.07 | 20.03 | 1.07 | 20.03 | 1.07 | 36.88 | 11.47 | 20.03 | 1.07 | 20.03 | 1.07 | 20.03 | 1.07 | 20.03 | 1.07 | 36.88 | 11.47 |
| | 512 | 15.99 | 0.00 | 15.99 | 0.00 | 15.99 | 0.00 | 56.92 | 39.13 | 15.99 | 0.00 | 15.99 | 0.00 | 15.99 | 0.00 | 15.99 | 0.00 | 56.92 | 39.13 |
| Lastde++ | 64 | 16.84 | 16.84 | 16.86 | 16.86 | 16.69 | 16.69 | 16.68 | 16.68 | 16.71 | 16.71 | 16.69 | 16.69 | 16.72 | 16.72 | 16.73 | 16.73 | 16.74 | 16.74 |
| | 128 | 16.80 | 16.80 | 16.70 | 16.70 | 16.67 | 16.67 | 16.68 | 16.68 | 16.67 | 16.67 | 16.67 | 16.67 | 16.66 | 16.66 | 16.67 | 16.67 | 16.69 | 16.69 |
| | 256 | 16.67 | 16.67 | 16.67 | 16.67 | 16.66 | 16.66 | 16.67 | 16.67 | 16.67 | 16.67 | 16.67 | 16.67 | 16.67 | 16.67 | 16.67 | 16.67 | 16.67 | 16.67 |
| | 512 | 16.67 | 16.67 | 16.68 | 16.68 | 16.67 | 16.67 | 16.67 | 16.67 | 16.67 | 16.67 | 16.67 | 16.67 | 16.67 | 16.67 | 16.67 | 16.67 | 16.67 | 16.67 |
| RepreGuard | 64 | 16.68 | 16.67 | 16.67 | 16.67 | 30.79 | 23.29 | 16.67 | 16.67 | 27.06 | 19.02 | 16.70 | 16.69 | 32.20 | 24.38 | 19.62 | 16.68 | 24.24 | 19.15 |
| | 128 | 16.67 | 16.67 | 16.67 | 16.67 | 32.79 | 24.34 | 16.67 | 16.67 | 29.16 | 19.01 | 16.68 | 16.68 | 34.34 | 26.70 | 20.88 | 16.70 | 25.66 | 19.69 |
| | 256 | 16.67 | 16.67 | 16.67 | 16.67 | 37.05 | 25.90 | 16.67 | 16.67 | 33.27 | 20.05 | 16.68 | 16.67 | 36.66 | 29.93 | 23.31 | 16.81 | 27.99 | 20.65 |
| | 512 | 16.67 | 16.67 | 16.67 | 16.67 | 38.59 | 26.70 | 16.67 | 16.67 | 35.75 | 19.96 | 16.67 | 16.67 | 37.57 | 31.75 | 25.46 | 16.94 | 28.95 | 21.10 |
| X-Rob-Classifier | 64 | 75.13 | 69.97 | 75.56 | 64.66 | 74.13 | 60.88 | 80.43 | 66.89 | 75.16 | 61.90 | 74.76 | 62.24 | 77.54 | 64.73 | 73.79 | 65.55 | 75.86 | 64.76 |
| | 128 | 83.80 | 82.59 | 82.17 | 76.51 | 81.05 | 66.25 | 86.54 | 80.98 | 82.11 | 72.80 | 80.87 | 71.83 | 83.72 | 76.32 | 80.07 | 72.07 | 82.57 | 75.02 |
| | 256 | 92.08 | 94.02 | 88.46 | 86.85 | 88.87 | 84.70 | 92.08 | 90.49 | 88.74 | 84.64 | 85.75 | 80.81 | 90.17 | 89.70 | 85.21 | 83.27 | 88.95 | 87.10 |
| | 512 | 96.38 | 96.57 | 93.21 | 89.83 | 94.71 | 93.38 | 94.06 | 94.36 | 94.62 | 93.85 | 90.58 | 89.12 | 95.13 | 94.71 | 92.50 | 91.92 | 93.91 | 92.82 |
| mDeBERTa-Classifier | 64 | 66.65 | 68.99 | 66.42 | 63.57 | 65.75 | 60.36 | 75.04 | 70.95 | 66.48 | 61.52 | 60.49 | 63.76 | 66.80 | 65.98 | 60.84 | 62.97 | 65.96 | 64.69 |
| | 128 | 81.61 | 84.19 | 82.21 | 80.35 | 82.09 | 74.49 | 89.46 | 83.88 | 84.19 | 75.54 | 78.22 | 74.33 | 84.35 | 78.91 | 78.43 | 76.38 | 84.82 | 78.55 |
| | 256 | 96.45 | 95.60 | 94.24 | 91.60 | 94.24 | 91.12 | 96.06 | 95.94 | 94.95 | 87.48 | 92.99 | 89.22 | 95.69 | 93.42 | 93.12 | 89.44 | 94.73 | 91.73 |
| | 512 | 98.06 | 97.92 | 96.20 | 94.55 | 96.96 | 95.75 | 96.96 | 97.19 | 97.25 | 94.80 | 95.47 | 94.08 | 97.83 | 97.16 | 96.25 | 93.84 | 96.87 | 95.30 |
| Biscope | 64 | 41.94 | 19.83 | 37.51 | 17.47 | 43.29 | 17.51 | 42.35 | 21.06 | 42.87 | 19.66 | 42.71 | 17.69 | 43.30 | 20.00 | 38.85 | 18.10 | 41.96 | 18.02 |
| | 128 | 53.45 | 29.05 | 39.75 | 17.76 | 53.15 | 21.84 | 49.61 | 19.23 | 52.93 | 25.84 | 49.08 | 17.87 | 54.53 | 28.19 | 47.23 | 21.82 | 50.53 | 20.40 |
| | 256 | 64.79 | 41.85 | 46.00 | 20.36 | 64.40 | 34.64 | 59.40 | 19.35 | 64.57 | 37.86 | 56.39 | 23.94 | 66.67 | 41.59 | 57.35 | 29.81 | 60.45 | 30.20 |
| | 512 | 73.82 | 58.50 | 56.50 | 26.92 | 72.72 | 52.33 | 64.35 | 25.57 | 72.18 | 49.46 | 60.72 | 33.08 | 73.73 | 52.40 | 66.06 | 44.77 | 67.73 | 43.99 |

Table 30: Performance Comparison of Different Detectors on Cross Length in TERNARY Task. Detector performance is visualized using teal color gradients, with darker intensities indicating superior detection capabilities.

| Detectors↓ | Test↓ Lang→ | English | | Chinese | | Spanish | | Arabic | | French | | Russian | | Portuguese | | German | | Average | |
|---|---|---|---|---|---|---|---|---|---|---|---|---|---|---|---|---|---|---|---|
| | Metrics | $F_1^B$ | $F_1^F$ | $F_1^B$ | $F_1^F$ | $F_1^B$ | $F_1^F$ | $F_1^B$ | $F_1^F$ | $F_1^B$ | $F_1^F$ | $F_1^B$ | $F_1^F$ | $F_1^B$ | $F_1^F$ | $F_1^B$ | $F_1^F$ | $F_1^B$ | $F_1^F$ |
| Log-Likelihood | Polishing | 37.54 | 33.33 | 43.93 | 34.77 | 61.62 | 33.44 | 66.82 | 33.45 | 60.79 | 33.54 | 52.29 | 34.15 | 65.81 | 33.59 | 55.92 | 33.34 | 58.02 | 33.72 |
| | Expanding | 70.47 | 33.46 | 60.71 | 33.89 | 76.57 | 33.64 | 66.31 | 33.40 | 69.28 | 33.65 | 51.94 | 41.24 | 73.28 | 33.86 | 67.75 | 33.35 | 68.13 | 34.48 |
| | Condensing | 39.58 | 33.36 | 38.85 | 33.93 | 60.79 | 33.55 | 53.17 | 33.34 | 58.61 | 33.55 | 51.82 | 34.11 | 63.47 | 33.68 | 47.32 | 33.32 | 54.14 | 33.63 |
| Log-Rank | Polishing | 38.80 | 33.34 | 48.83 | 35.75 | 60.20 | 33.62 | 67.21 | 33.84 | 58.52 | 33.80 | 51.83 | 33.87 | 63.93 | 33.83 | 51.74 | 33.34 | 57.06 | 33.94 |
| | Expanding | 74.60 | 33.89 | 69.20 | 35.04 | 75.67 | 34.10 | 67.26 | 33.48 | 68.43 | 34.34 | 53.39 | 39.83 | 72.48 | 34.10 | 61.74 | 33.32 | 69.07 | 34.70 |
| | Condensing | 41.88 | 33.41 | 41.82 | 34.38 | 59.70 | 33.75 | 55.02 | 33.38 | 57.11 | 33.79 | 50.86 | 34.02 | 61.73 | 33.89 | 45.37 | 33.32 | 53.64 | 33.76 |
| DetectLLM-LRR | Polishing | 42.67 | 33.98 | 65.57 | 42.32 | 51.64 | 33.93 | 66.47 | 35.90 | 49.94 | 34.76 | 43.89 | 33.57 | 54.59 | 33.91 | 41.04 | 33.39 | 52.93 | 35.30 |
| | Expanding | 72.11 | 44.71 | 83.22 | 51.20 | 66.89 | 35.39 | 68.54 | 34.65 | 59.04 | 35.20 | 52.50 | 34.03 | 65.65 | 34.59 | 43.35 | 33.33 | 66.49 | 38.97 |
| | Condensing | 47.46 | 34.58 | 54.83 | 37.38 | 53.32 | 34.55 | 57.15 | 34.34 | 49.87 | 34.94 | 45.80 | 34.18 | 53.03 | 33.89 | 39.41 | 33.39 | 50.85 | 34.66 |
| Fast-DetectGPT | Polishing | 32.52 | 33.15 | 36.69 | 34.07 | 33.44 | 33.46 | 51.91 | 43.50 | 35.41 | 33.85 | 54.43 | 37.94 | 33.74 | 33.32 | 46.19 | 35.03 | 42.34 | 35.70 |
| | Expanding | 31.02 | 33.07 | 31.88 | 33.21 | 30.18 | 33.25 | 45.48 | 35.57 | 30.82 | 33.29 | 54.30 | 37.97 | 28.56 | 33.23 | 45.81 | 35.10 | 39.19 | 34.29 |
| | Condensing | 36.77 | 33.30 | 43.52 | 35.66 | 36.82 | 33.73 | 55.52 | 49.55 | 39.01 | 34.27 | 62.44 | 44.77 | 37.51 | 33.83 | 51.17 | 35.49 | 47.12 | 37.89 |
| Binoculars | Polishing | 53.98 | 37.40 | 71.45 | 69.09 | 66.08 | 45.06 | 74.65 | 52.44 | 72.10 | 48.22 | 79.20 | 61.14 | 73.60 | 50.19 | 73.98 | 45.95 | 71.09 | 51.88 |
| | Expanding | 39.25 | 33.50 | 72.82 | 70.20 | 67.49 | 43.09 | 73.50 | 47.77 | 70.01 | 43.63 | 82.49 | 65.77 | 75.63 | 50.86 | 74.11 | 46.14 | 69.87 | 51.36 |
| | Condensing | 57.58 | 43.09 | 68.69 | 64.15 | 66.83 | 50.42 | 74.78 | 57.12 | 75.46 | 55.64 | 79.23 | 65.82 | 73.88 | 55.25 | 75.71 | 52.23 | 71.91 | 55.81 |
| Revise-Detect | Polishing | 72.56 | 0.00 | 53.41 | 0.00 | 82.82 | 0.00 | 85.69 | 0.00 | 69.07 | 0.00 | 70.33 | 0.00 | 80.99 | 0.00 | 80.19 | 0.00 | 75.80 | 0.00 |
| | Expanding | 88.50 | 0.00 | 64.54 | 0.00 | 88.16 | 0.00 | 88.63 | 0.00 | 77.88 | 0.00 | 85.13 | 0.00 | 85.99 | 0.00 | 83.24 | 0.00 | 83.43 | 0.00 |
| | Condensing | 85.15 | 0.00 | 51.36 | 0.00 | 83.21 | 0.00 | 83.86 | 0.00 | 82.98 | 0.00 | 78.26 | 0.00 | 84.80 | 0.00 | 79.61 | 0.00 | 79.82 | 0.00 |
| GECScore | Polishing | 85.23 | 69.87 | 48.32 | 47.15 | 83.42 | 64.28 | 71.56 | 44.21 | 81.67 | 62.45 | 75.89 | 58.34 | 84.12 | 61.27 | 79.85 | 60.43 | 77.21 | 59.87 |
| | Expanding | 95.82 | 79.45 | 58.63 | 57.28 | 92.67 | 73.84 | 81.45 | 53.76 | 90.15 | 71.68 | 84.67 | 67.29 | 91.84 | 69.75 | 88.92 | 68.56 | 86.34 | 66.82 |
| | Condensing | 90.68 | 74.36 | 52.95 | 52.00 | 88.05 | 68.71 | 76.21 | 48.63 | 86.23 | 67.01 | 79.24 | 62.17 | 88.27 | 65.38 | 84.31 | 65.13 | 81.43 | 63.39 |
| Lastde++ | Polishing | 37.70 | 33.47 | 32.20 | 33.35 | 36.30 | 33.53 | 40.27 | 33.76 | 36.95 | 33.27 | 43.75 | 33.38 | 33.74 | 33.43 | 39.50 | 33.66 | 37.91 | 33.48 |
| | Expanding | 32.35 | 33.29 | 30.18 | 33.26 | 29.69 | 33.17 | 41.66 | 33.63 | 33.10 | 33.27 | 47.08 | 33.67 | 29.20 | 33.24 | 36.50 | 33.59 | 35.39 | 33.39 |
| | Condensing | 33.97 | 33.30 | 35.25 | 33.39 | 32.07 | 33.27 | 36.25 | 33.35 | 34.89 | 33.41 | 38.21 | 33.38 | 31.43 | 33.28 | 36.74 | 33.53 | 35.13 | 33.36 |
| RepreGuard | Polishing | 67.45 | 48.33 | 93.77 | 0.00 | 66.65 | 92.45 | 92.13 | 1.46 | 66.65 | 97.76 | 89.28 | 0.77 | 66.63 | 94.32 | 66.53 | 63.23 | 73.90 | 62.26 |
| | Expanding | 91.48 | 8.89 | 80.49 | 0.60 | 66.67 | 98.58 | 98.88 | 1.23 | 66.59 | 97.35 | 90.88 | 0.37 | 66.65 | 97.76 | 66.60 | 80.45 | 77.41 | 57.50 |
| | Condensing | 74.42 | 87.52 | 91.27 | 80.88 | 90.61 | 88.10 | 89.99 | 61.65 | 93.73 | 88.10 | 85.46 | 74.89 | 93.81 | 86.21 | 83.45 | 94.10 | 87.36 | 83.64 |
| X-Rob-Classifier | Polishing | 99.95 | 99.86 | 99.88 | 99.69 | 99.96 | 99.84 | 99.91 | 99.78 | 99.99 | 99.87 | 99.66 | 99.45 | 99.99 | 99.64 | 99.97 | 99.87 | 99.90 | 99.42 |
| | Expanding | 99.98 | 99.97 | 100.00 | 33.33 | 99.97 | 99.73 | 99.97 | 99.87 | 99.98 | 99.96 | 99.89 | 99.83 | 100.00 | 33.33 | 100.00 | 33.33 | 99.97 | 99.87 |
| | Condensing | 99.95 | 99.91 | 99.78 | 99.63 | 99.89 | 99.57 | 99.83 | 99.47 | 99.93 | 99.88 | 98.67 | 98.57 | 99.98 | 99.94 | 99.81 | 99.64 | 99.64 | 99.43 |
| mDeBERTa-Classifier | Polishing | 99.99 | 99.46 | 99.99 | 99.95 | 99.97 | 99.97 | 99.98 | 99.35 | 99.99 | 99.97 | 99.82 | 99.47 | 100.00 | 33.33 | 99.98 | 99.94 | 99.96 | 99.48 |
| | Expanding | 99.99 | 99.97 | 100.00 | 33.33 | 99.98 | 99.87 | 99.99 | 99.42 | 100.00 | 33.33 | 99.97 | 99.95 | 100.00 | 33.33 | 100.00 | 33.33 | 99.99 | 99.97 |
| | Condensing | 99.99 | 99.77 | 99.90 | 99.84 | 99.94 | 99.92 | 99.97 | 99.95 | 99.95 | 99.91 | 99.32 | 99.27 | 99.98 | 99.95 | 99.93 | 99.85 | 99.85 | 99.58 |
| Biscope | Polishing | 90.10 | 77.81 | 92.17 | 80.03 | 93.62 | 88.10 | 88.00 | 65.95 | 95.37 | 88.03 | 85.48 | 58.62 | 95.99 | 91.84 | 93.68 | 83.28 | 91.22 | 78.91 |
| | Expanding | 96.96 | 95.54 | 88.04 | 77.11 | 96.79 | 95.75 | 81.10 | 52.46 | 98.28 | 97.93 | 90.89 | 80.28 | 98.54 | 98.18 | 95.04 | 88.13 | 90.02 | 85.43 |
| | Condensing | 93.06 | 73.54 | 92.93 | 86.06 | 96.97 | 77.63 | 93.84 | 89.35 | 96.09 | 74.09 | 91.82 | 63.61 | 97.37 | 77.42 | 95.23 | 64.25 | 94.15 | 73.98 |

Table 31: Performance Comparison of Different Detectors on Cross Operation in BINARY Task. Detector performance is visualized using teal color gradients, with darker intensities indicating superior detection capabilities.

| Detectors↓ | Test↓ Lang→ | English | | Chinese | | Spanish | | Arabic | | French | | Russian | | Portuguese | | German | | Average | |
|---|---|---|---|---|---|---|---|---|---|---|---|---|---|---|---|---|---|---|---|
| | Metrics | $F_1^B$ | $F_1^F$ | $F_1^B$ | $F_1^F$ | $F_1^B$ | $F_1^F$ | $F_1^B$ | $F_1^F$ | $F_1^B$ | $F_1^F$ | $F_1^B$ | $F_1^F$ | $F_1^B$ | $F_1^F$ | $F_1^B$ | $F_1^F$ | $F_1^B$ | $F_1^F$ |
| Log-Likelihood | Expanding | 47.17 | 31.26 | 42.62 | 25.73 | 36.01 | 40.16 | 48.75 | 37.36 | 33.19 | 38.55 | 25.53 | 32.30 | 38.78 | 40.24 | 51.10 | 34.20 | 42.99 | 35.56 |
| | Condensing | 36.06 | 33.10 | 30.53 | 27.36 | 47.17 | 40.75 | 45.51 | 37.35 | 42.43 | 37.23 | 32.15 | 29.19 | 46.60 | 39.33 | 43.21 | 37.31 | 41.44 | 35.70 |
| Log-Rank | Expanding | 43.39 | 32.71 | 43.64 | 28.49 | 33.11 | 39.74 | 46.46 | 38.07 | 31.34 | 37.95 | 25.87 | 32.58 | 36.89 | 39.85 | 48.55 | 33.41 | 40.75 | 35.76 |
| | Condensing | 40.31 | 34.14 | 38.31 | 30.02 | 41.36 | 40.24 | 46.41 | 37.80 | 38.40 | 36.75 | 33.54 | 29.63 | 40.97 | 39.20 | 45.56 | 36.19 | 42.28 | 35.91 |
| DetectLLM-LRR | Expanding | 26.72 | 33.87 | 31.77 | 36.67 | 31.61 | 36.82 | 36.27 | 38.87 | 30.57 | 34.53 | 30.17 | 31.31 | 35.13 | 37.54 | 37.89 | 28.44 | 33.44 | 34.98 |
| | Condensing | 41.00 | 32.58 | 41.42 | 37.27 | 36.91 | 36.06 | 41.29 | 38.27 | 37.91 | 33.99 | 32.82 | 29.14 | 40.50 | 37.70 | 37.69 | 29.91 | 39.81 | 35.13 |
| Fast-DetectGPT | Expanding | 23.41 | 23.49 | 22.81 | 22.04 | 24.16 | 23.65 | 40.83 | 34.85 | 27.74 | 26.34 | 41.33 | 34.09 | 24.64 | 24.06 | 32.50 | 26.54 | 30.85 | 27.81 |
| | Condensing | 24.32 | 24.32 | 23.15 | 23.15 | 22.17 | 22.17 | 33.68 | 33.68 | 26.01 | 26.01 | 36.57 | 36.57 | 24.52 | 24.52 | 27.33 | 27.33 | 27.96 | 27.96 |
| Binoculars | Expanding | 27.65 | 26.06 | 47.12 | 35.98 | 44.98 | 32.09 | 45.27 | 31.67 | 49.18 | 34.33 | 52.78 | 35.07 | 52.16 | 35.25 | 44.61 | 29.01 | 45.86 | 32.61 |
| | Condensing | 38.19 | 31.66 | 45.71 | 38.92 | 45.36 | 36.92 | 42.58 | 36.66 | 49.25 | 41.23 | 50.91 | 42.03 | 51.15 | 41.95 | 44.46 | 36.78 | 46.09 | 38.39 |
| Revise-Detect | Expanding | 58.86 | 49.66 | 39.12 | 32.61 | 52.60 | 51.16 | 57.19 | 43.36 | 55.36 | 48.01 | 54.07 | 45.80 | 50.98 | 51.19 | 55.70 | 46.84 | 53.46 | 46.32 |
| | Condensing | 60.62 | 50.02 | 38.26 | 33.87 | 57.84 | 49.80 | 55.83 | 44.95 | 56.67 | 49.28 | 56.22 | 47.95 | 55.23 | 50.99 | 54.21 | 45.70 | 54.59 | 46.57 |
| GECScore | Expanding | 32.91 | 10.61 | 45.32 | 21.06 | 19.95 | 0.00 | 48.25 | 24.45 | 31.23 | 10.40 | 37.04 | 16.19 | 51.86 | 32.51 | 31.09 | 8.69 | 57.23 | 39.78 |
| | Condensing | 44.63 | 26.82 | 41.85 | 22.26 | 30.37 | 14.92 | 46.37 | 26.52 | 16.48 | 0.00 | 32.46 | 17.82 | 31.78 | 13.53 | 16.50 | 0.02 | 53.35 | 39.05 |
| Lastde++ | Expanding | 16.67 | 16.67 | 16.67 | 16.67 | 16.67 | 16.67 | 16.67 | 16.67 | 16.67 | 16.67 | 16.67 | 16.67 | 16.67 | 16.67 | 16.67 | 16.67 | 16.67 | 16.67 |
| | Condensing | 16.67 | 16.67 | 16.67 | 16.67 | 16.67 | 16.67 | 16.67 | 16.67 | 16.67 | 16.67 | 16.67 | 16.67 | 16.67 | 16.67 | 16.67 | 16.67 | 16.67 | 16.67 |
| RepreGuard | Expanding | 16.67 | 16.67 | 16.67 | 16.67 | 41.40 | 30.47 | 16.67 | 16.67 | 39.07 | 21.88 | 16.67 | 16.67 | 39.95 | 36.53 | 28.03 | 17.75 | 30.33 | 22.63 |
| | Condensing | 38.82 | 16.67 | 19.57 | 16.67 | 16.64 | 25.40 | 38.62 | 16.67 | 16.68 | 23.06 | 31.00 | 16.67 | 16.67 | 33.96 | 16.66 | 17.31 | 30.22 | 21.61 |
| X-Rob-Classifier | Expanding | 66.14 | 66.11 | 67.83 | 68.98 | 61.14 | 62.06 | 64.17 | 65.04 | 63.18 | 64.28 | 65.07 | 66.59 | 63.49 | 64.55 | 70.57 | 71.12 | 65.38 | 66.24 |
| | Condensing | 98.01 | 98.82 | 94.56 | 96.10 | 95.83 | 97.55 | 96.18 | 97.42 | 95.96 | 97.68 | 93.75 | 96.40 | 96.64 | 97.88 | 94.14 | 96.43 | 95.75 | 97.31 |
| mDeBERTa-Classifier | Expanding | 66.94 | 66.23 | 70.68 | 61.25 | 63.38 | 60.59 | 67.66 | 66.36 | 66.68 | 65.33 | 65.50 | 60.47 | 65.49 | 63.44 | 73.09 | 64.76 | 67.62 | 63.02 |
| | Condensing | 98.75 | 99.16 | 95.67 | 94.99 | 97.45 | 98.21 | 97.76 | 98.21 | 97.98 | 98.82 | 96.16 | 96.37 | 98.21 | 98.50 | 97.01 | 97.65 | 97.43 | 97.93 |
| Biscope | Expanding | 37.66 | 30.78 | 42.30 | 32.83 | 51.96 | 43.15 | 58.74 | 20.82 | 49.83 | 48.07 | 45.46 | 31.40 | 52.79 | 51.75 | 58.18 | 45.93 | 49.95 | 37.51 |
| | Condensing | 79.59 | 66.46 | 71.32 | 44.15 | 83.43 | 71.11 | 74.17 | 53.54 | 78.65 | 60.30 | 68.05 | 45.72 | 82.12 | 74.25 | 74.12 | 64.27 | 76.55 | 59.61 |

Table 32: Performance Comparison of Different Detectors on Cross Operation in TERNARY Task. Detector performance is visualized using teal color gradients, with darker intensities indicating superior detection capabilities.